\documentclass[doctor,english,final]{kaist-ucs}

\usepackage[utf8]{inputenc} 
\usepackage[T1]{fontenc}    
\usepackage{hyperref}       
\usepackage{url}            
\usepackage{booktabs}       
\usepackage{amsfonts}       
\usepackage{nicefrac}       
\usepackage[utf8]{inputenc}
\usepackage{multirow}
\usepackage{amssymb, amsthm, mathalpha, mathtools} 
\usepackage{bbold, dsfont} 
\usepackage{kotex, xcolor}
\usepackage[table,svgnames,dvipsnames]{xcolor}
\usepackage[ruled,vlined]{algorithm2e}
\usepackage{algorithmic}
\usepackage[pdftex]{graphicx}
\usepackage{subcaption}
\usepackage{wrapfig}
\usepackage{enumitem}
\usepackage{pifont}

\usepackage{array}
\usepackage{eqparbox}
\usepackage{etoolbox}  
\usepackage[skins]{tcolorbox}
\usepackage{arydshln}
\usepackage{tabulary}
\usepackage{makecell}
\usepackage{natbib}

\definecolor{myred}{RGB}{204, 4, 67}
\definecolor{custom_green}{rgb}{0.0, 0.5, 0.0}
\definecolor{custom_red}{rgb}{1.0, 0.01, 0.24}
\definecolor{custom_blue}{HTML}{C9DAF7}
\definecolor{custom_purple}{HTML}{D9D1E9}
\definecolor{title_blue}{HTML}{204899} 
\definecolor{cite_blue}{HTML}{044dc1}  
\definecolor{cite_purple}{HTML}{7406a7}  

\newcommand{\cmark}{\ding{51}}%
\newcommand{\xmark}{\ding{55}}%

\usepackage{amsmath,amsfonts,bm}

\def\eqref#1{equation~\ref{#1}}

\def\1{\bm{1}}

\DeclareMathAlphabet{\mathsfit}{\encodingdefault}{\sfdefault}{m}{sl}
\SetMathAlphabet{\mathsfit}{bold}{\encodingdefault}{\sfdefault}{bx}{n}

\DeclareMathOperator*{\argmin}{arg\,min}

\title[korean] {조기 종료 알고리즘을 활용한 초거대\linebreak 언어 모델 추론 가속화}
\title[english]{Accelerating Large Language Model Inference\linebreak via Early-Exiting Algorithms}

\author[korean] {배}{상 민}
\author[korean2]{배}{상민}    
\author[chinese]{裵}{祥 珉}
\author[english]{Bae}{Sangmin}

\advisor[major]{윤 세 영}{Se-Young Yun}{signed}
\advisor[major2]{윤세영}{Se-Young Yun}{signed}  
\advisorinfo{Professor of Kim Jaechul Graduate School of AI} 
\department{AI}{engineering}{a}

\referee[1]{윤 세 영}
\referee[2]{김 기 응}
\referee[3]{송 환 준}
\referee[4]{신 진 우}
\referee[5]{양 은 호}

\approvaldate{2025}{6}{2}

\refereedate{2025}{6}{2}

\gradyear{2025}

\begin{document}


   \thesisinfo
    \begin{summary}      
    거대 언어 모델은 괄목할 만한 성능을 달성했지만, 막대한 연산 비용으로 인해 실용적인 배포에 어려움을 겪고 있다. 조기 종료와 같은 적응형 연산 방법론은 이러한 비용을 절감할 수 있는 유망한 접근법이지만, 연산을 절약하려는 토큰 단위의 동적임이 배치 추론 환경에서는 오히려 시스템 수준의 병목 현상을 일으켜 역설적으로 전체 처리량을 감소시킬 수 있다. 본 학위 논문은 적응형 알고리즘과 모델 구조를 함께 설계하여 동적임과 효율성 사이의 최적의 균형을 맞춤으로써 이 문제를 해결했다. 이를 위해, 본 연구는 먼저 효율적인 병렬 디코딩 알고리즘을 제안하여 기존의 조기 종료 방식이 갖는 주요 추론 병목 원인들을 해결하였다. 그다음, 파라미터 공유 메커니즘을 통해 소형의 효율적인 모델을 만들 뿐만 아니라, 동적 추론에서 발생하는 중대한 동기화 문제를 본질적으로 완화하는 새로운 모델 구조를 제안하였다. 마지막으로, 본 연구는 라우터가 각 토큰에 최적의 순환 깊이를 동적으로 할당하도록 사전 훈련하는 통합된 프레임워크를 제시하였다. 이 접근법은 단일 모델 내에서 적응형 연산과 파라미터 효율성을 효과적으로 동시에 최적화하여, 비용 대비 성능을 극대화하는 새로운 모델 설계 방향을 제시하였다.
    \end{summary}
   
    \begin{Korkeyword}
    조기 종료, 적응형 연산, 거대 언어 모델, 추론 효율성, 병렬 디코딩, 파라미터 공유, 라우팅 메커니즘.
    \end{Korkeyword}

    \begin{abstract}
    Large language models have achieved remarkable capabilities, but their practical deployment is hindered by significant computational costs. While adaptive computation methods like early-exiting promise to reduce these costs, they introduce a fundamental conflict: the per-token dynamism intended to save computation often creates system-level bottlenecks that can paradoxically reduce throughput in batched inference. This dissertation resolves this conflict by co-designing adaptive algorithms and model architectures to strike an optimal balance between dynamism and efficiency. To this end, our work first addresses critical sources of overhead in conventional early-exiting by proposing an efficient parallel decoding mechanism. We then show that deep parameter sharing provides an architectural foundation that not only yields compact, parameter-efficient models but also inherently mitigates the critical synchronization issues affecting dynamic inference. Finally, this work presents a unified framework where lightweight routers are pretrained to dynamically assign an optimal recursion depth for each token. This approach establishes a new Pareto frontier between efficiency and performance by effectively optimizing for both adaptive computation and parameter efficiency within a single model.
    \end{abstract} 
     
    \begin{Engkeyword}
    Early-exiting, adaptive computation, large language models, inference efficiency, parallel decoding, parameter sharing, routing mechanism.
    \end{Engkeyword}

    \addtocounter{pagemarker}{1}                 
    \newpage

    \hypersetup{linkcolor=black, citecolor=black, urlcolor=black}
    \tableofcontents

    \listoftables

    \listoffigures



\hypersetup{
    linkcolor=red,
    citecolor=RoyalBlue, 
    filecolor=cyan,
    urlcolor=Magenta
}

\chapter{Introduction}

Large language models (LLMs) have demonstrated capabilities comparable to, or even superior to, human performance across a wide range of benchmarks such as code generation and logical reasoning. Nevertheless, achieving this level of performance comes at a significant computational cost due to their vast number of parameters and the inherent inefficiency of autoregressive decoding. This sequential process, where the model generates a single token at a time, is inefficient as it requires the entire model's computational graph to be activated at every step, leading to a major bottleneck in inference.

\begin{figure*}[h]
    \centering
    \begin{subfigure}[b]{0.45\linewidth}
    \centering
    \includegraphics[width=\linewidth]{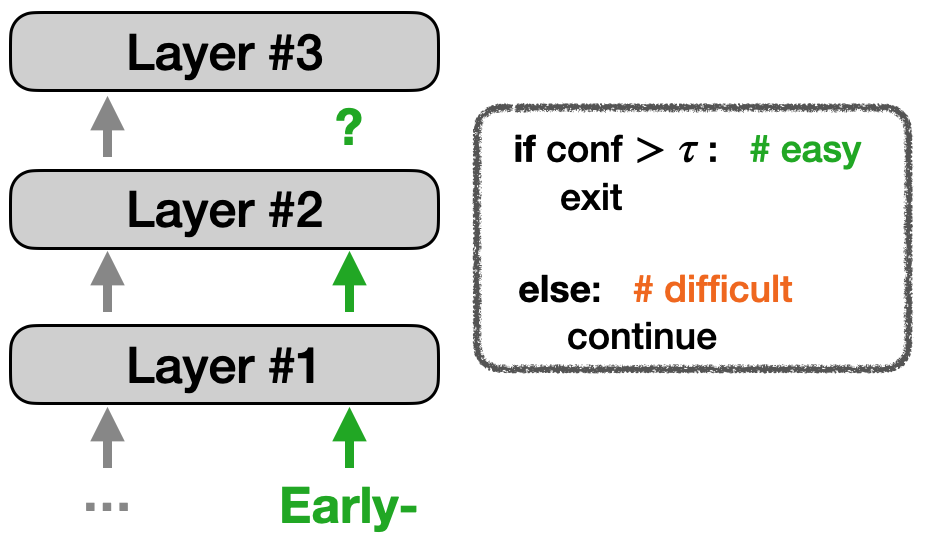}
    \caption{Early-Exiting mechanism}
    \label{fig_intro:early_exit}
    \end{subfigure}
    \hfill
    \begin{subfigure}[b]{0.30\linewidth}
    \centering
    \includegraphics[width=\linewidth]{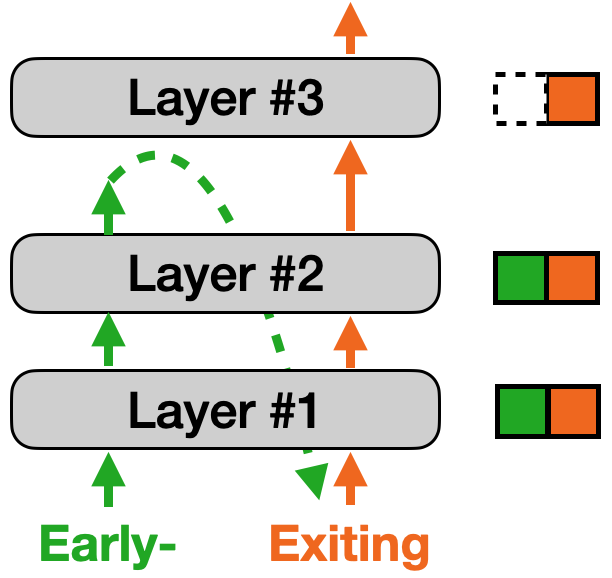}
    \caption{Missing cache issue}
    \label{fig_intro:problem_1}
    \end{subfigure}
    \hfill
    \begin{subfigure}[b]{0.215\linewidth}
    \centering
    \includegraphics[width=\linewidth]{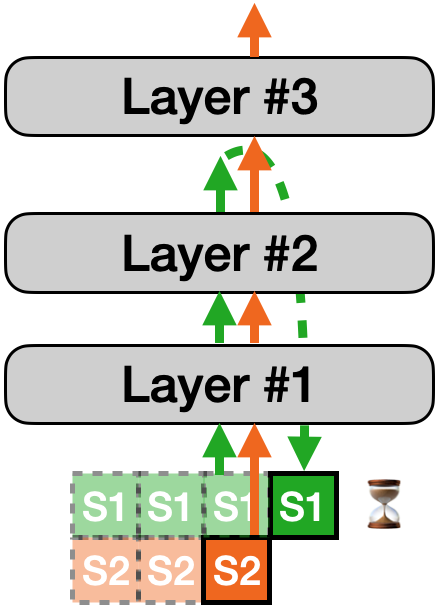}
    \vspace{-20pt}
    \caption{Batching issue}
    \label{fig_intro:problem_2}
    \end{subfigure}
    \caption{
    (a) The inference process dynamically decides whether to exit or continue by comparing a confidence measure against a threshold $\tau$. (b) The missing cache issue arises when skipped layers fail to update key-value (KV) caches needed for future tokens. (c) The batching inefficiency occurs when early-exited samples must wait for other samples to finish, leading to computational idle time.
    }
    \label{fig_intro:early_exit_overview}
\end{figure*}

To enhance the inference efficiency of LLMs, prior work has leveraged a critical insight into their computational redundancy: not all parameters are necessary for generating every token. In light of this, \textit{adaptive computation} has emerged as a promising approach that dynamically allocates computational resources based on the needs of each input. The introduction of \textit{dynamism} into the computational path is the cornerstone of all adaptive approaches. Among these methods, the \textbf{early-exiting} framework is being actively explored for its simple yet effective principle: if the model is confident in its prediction at an intermediate layer, it can terminate the process and output the result earlier without passing through the remaining layers (refer to Figure~\ref{fig_intro:early_exit}). This framework enhances computational efficiency by creating input-dependent computational paths of varying depths.

Ironically, while dynamic computational paths are intended to optimize inference, their excessive dynamism and naive implementation can complicate batched inference, sometimes even reducing actual throughput. This issue stems from \textit{two} fundamental challenges that hinder the deployment of early-exiting algorithms in real-world scenarios. \textit{First} is the missing key-value (KV) cache problem (Figure~\ref{fig_intro:problem_1}); early-exited tokens skip computing their KV pairs in remained deeper layers, which can be necessary for the decoding of future tokens. Approximating these missing values often degrades performance, while their exact computation can decrease the efficiency gains of adaptiveness. \textit{Second} is the batch synchronization issue (Figure~\ref{fig_intro:problem_2}); tokens that have exited are forced to idle while waiting for other tokens in the same batch to complete their forward passes through deeper layers, as tokens at different network depths cannot be normally batched together.

\clearpage

\begin{figure*}[h]
    \centering
    \includegraphics[width=\linewidth]{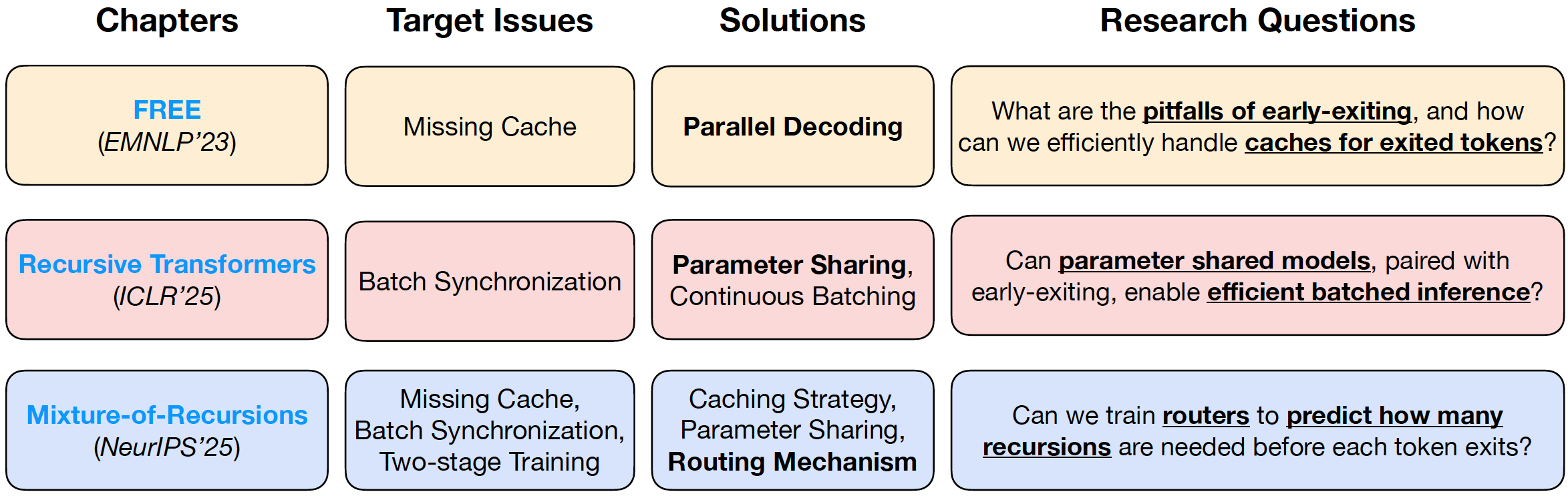}
    \caption{
    Overview of the dissertation structure. This figure outlines the three core chapters, detailing the specific challenges in early-exiting frameworks, the proposed methodologies, and the corresponding research questions addressed in each study.
    }
    \label{fig_intro:thesis_overview}
\end{figure*}

Accordingly, this dissertation \textbf{co-designs adaptive algorithms and model architectures} to strike an optimal balance between dynamism and efficiency, thereby resolving these two major inference challenges without compromising the original model's performance. We believe the principles presented herein will serve as a cornerstone for the wider adoption of early exiting in real-world deployments.

In \textbf{Chapter \ref{chapter:free}}, we first address the missing KV cache problem within a single batch setting via \textit{parallel decoding} mechanism. This approach accumulates hidden states of consecutive early-exited tokens and processes them in a single, parallelized forward pass alongside the next non-exiting token. It exploits the memory-bound nature of Transformer decoding, where latency is dominated by memory access rather than computation. Consequently, this mechanism incurs virtually no overhead, as it avoids additional memory accesses while introducing only a marginal computational cost—negligible compared to recent GPU's FLOPs performance. As a result, our framework accelerates inference by allowing exited tokens to skip the memory retrieval and computation in deeper layers, while guaranteeing lossless generation through the efficient computation of their exact KV pairs in deeper layers.

Meanwhile, \textbf{Chapter \ref{chapter:rrt}} mitigates the batch synchronization issue by exploring and optimizing \textit{parameter sharing} in LLMs. This enables \textit{continuous depth-wise batching}, a novel inference paradigm where tokens at different model stages can be processed in a single batch because they utilize the same parameter block. While we demonstrate significant hypothetical speedup by resolving the synchronization bottleneck, this approach introduces its own challenges. Specifically, it relies on a sensitive two-stage training process—one for parameter sharing and another for early-exiting. Furthermore, it does not resolve the missing KV cache problem, still necessitating the integration of parallel decoding, which may result in a complex implementation.

Finally, in \textbf{Chapter \ref{chapter:mor}}, we present a unified framework that resolves both the missing KV cache and batch synchronization challenges inherent to early exiting. This is achieved through \textit{routers}, which are trained end-to-end to assign a dynamic recursion depth to each token, thereby allocating computation only where needed. This eliminates the complex two-stage training of prior work while preserving model performance and proving its scalability. Furthermore, we introduce a recursion-wise KV caching mechanism that selectively stores KV pairs only from tokens actively processed at each recursion depth. This not only solves the missing cache problem but also significantly reduces memory sizes. By systematically addressing the limitations of previous chapters, this unified architecture establishes a new Pareto frontier in the trade-off between efficiency and performance.

\chapter{Fast and Robust Early-Exiting Framework\\for Autoregressive Language Models with\\Synchronized Parallel Decoding}\label{chapter:free}

\begin{tcolorbox}[colback=gray!10, colframe=black, arc=3mm, boxrule=1pt]
    \textbf{Publication Note:} This chapter is based on the paper accepted to the Conference on Empirical Methods in Natural Language Processing (EMNLP 2023)~\citep{DBLP:conf/emnlp/BaeKSY23}.
    \\
    
    \textbf{Abstract:} To tackle the high inference latency exhibited by autoregressive language models, previous studies have proposed an early-exiting framework that allocates adaptive computation paths for each token based on the complexity of generating the subsequent token. However, we observed several shortcomings, including performance degradation caused by a state copying mechanism or numerous exit paths, and sensitivity to exit confidence thresholds. Consequently, we propose a \textbf{Fast and Robust Early-Exiting (FREE) framework}, which incorporates a shallow-deep module and a synchronized parallel decoding. Our framework enables faster inference by synchronizing the decoding process of the current token with previously stacked early-exited tokens. Furthermore, as parallel decoding allows us to observe predictions from both shallow and deep models, we present a novel adaptive threshold estimator that exploits a Beta mixture model to determine suitable confidence thresholds. We empirically demonstrated the superiority of our proposed framework on extensive generation tasks.\looseness=-1
\end{tcolorbox}

\section{Introduction}

\begin{figure*}[h]
    \centering
    \begin{subfigure}[b]{0.318\linewidth}
    \centering
    \includegraphics[width=\linewidth]{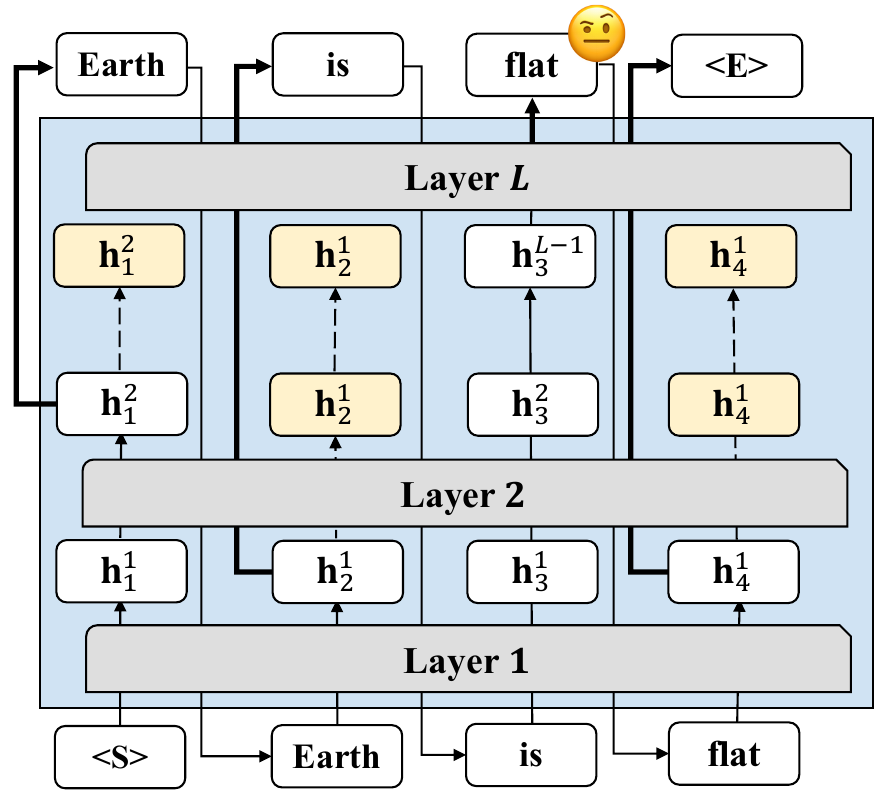}
    \caption{Conventional Early-Exiting}
    \label{fig_free:concept_left}
    \end{subfigure}
    \begin{subfigure}[b]{0.66\linewidth}
    \centering
    \includegraphics[width=\linewidth]{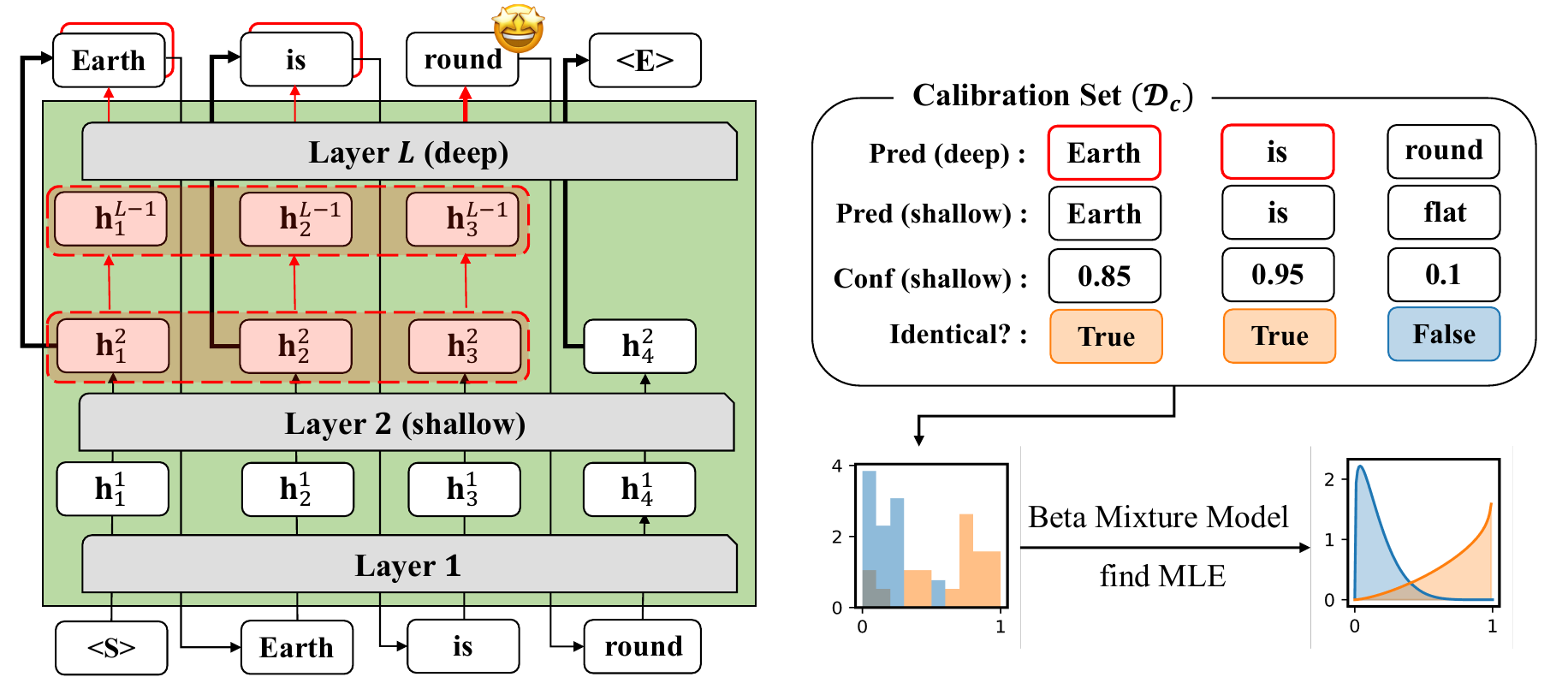}
    \caption{Fast and Robust Early-Exiting (FREE)}
    \label{fig_free:concept_right}
    \end{subfigure}
    \caption{Overview of our FREE framework compared to the conventional early-exiting framework. FREE exhibits three key differences: (1) FREE employs a shallow-deep module that utilizes two exit points instead of employing all layers as exit points, (2) FREE replaces the state copying mechanism (yellow colored) with synchronized parallel decoding (red colored) to prevent performance degradation while accelerating inference speed, and (3) FREE utilizes an adaptive threshold estimator to determine the appropriate threshold values for each dataset during inference.}
    \label{fig_free:concept}
\end{figure*}

\clearpage
Recent advancements in autoregressive language models\,\cite{brown2020language, raffel2020exploring, hoffmann2022an, touvron2023llama} have revolutionized the quality of language generation in various generative tasks, including question answering\,\cite{rajpurkar2016squad}, summarization\,\cite{nallapati2016abstractive, fabbri2019multi}, and machine translation\,\cite{cettolo2017overview}. Nevertheless, these large transformer models have shown high inference latency due to the considerable number of layers and the autoregressive decoding step. As the multiple stacks of transformer layers have to be computed sequentially for each individual token, the inference process poses significant computational burdens and hinders their real-time adaptability~\cite{jiao-etal-2020-tinybert}.

In light of the necessity to expedite inference latency, the \textit{early-exiting} framework\,\cite{elbayad2019depth, liu2021faster, schuster2022confident} emerges as a promising approach that dynamically allocates computation paths based on the complexity of generation for each token. 
As illustrated in Figure~\ref{fig_free:concept_left}, tokens that are relatively easy to predict the subsequent token yield consistent predictions with only a few layer computations, while those with higher difficulty require computations across a larger number of layers to generate accurate predictions.
In an ideal scenario, the early-exiting method empowers models to achieve notable acceleration in inference without compromising the generation quality when compared to that of a full model.

However, our extensive analysis identified {four} challenges in the early-exiting framework.
\emph{Firstly}, despite the potential to exit at earlier layers, key and value states for remaining layers are still required for processing subsequent tokens. 
While previous works have proposed the state copying mechanism~\cite{elbayad2019depth, schuster2022confident} to efficiently compute these states by reusing hidden states from the early-exited layer, our findings reveal that this method performs poorly with larger models and longer output sequences (see Section~\ref{sec_free:3_1}).
\emph{Additionally}, setting all layers as possible exit positions does not guarantee faster inference due to (1) the defective performance of earlier layers that can generate abnormally long sequence outputs, and (2) the computational overhead from confidence measurement at every layer (see Section~\ref{sec_free:3_2} and \ref{sec_free:3_3}).
\emph{Lastly}, achieving the desired level of latency and accuracy with early-exiting heavily depends on selecting the appropriate confidence threshold for the target task. This often entails significant efforts and additional computational overhead (see Section~\ref{sec_free:4_4}).
Hence, these challenges call for a new approach that consistently demonstrates high performance and low latency across diverse language models and datasets.\looseness=-1

In this paper, we introduce a \textbf{F}ast and \textbf{R}obust \textbf{E}arly-\textbf{E}xiting (\textbf{FREE}) framework that incorporates a shallow-deep module and synchronized parallel decoding. Our framework not only offers consistent speedup and performance even for larger models and longer output sequences, but also eliminates the need for the computationally expensive process of finding the appropriate exit threshold.

Specifically, the \emph{shallow-deep} module bifurcates the computation paths into a shallow model (with a specified number of early layers) and a deep model (including all layers).
Our \emph{synchronized parallel decoding} accumulates consecutive early-exited tokens that only pass through the shallow model until a non-exiting token is encountered. Thereby, we synchronize the decoding process of the current non-exiting token with the previously stacked tokens, as shown in the left of Figure~\ref{fig_free:concept_right}.
This prevents performance degradation by utilizing actual attention computed key and value instead of approximated states through state copying, while it also achieves a more efficient approach compared to decoding each token autoregressively.
Furthermore, we devise a novel \emph{adaptive threshold estimator}, as shown in the right of Figure~\ref{fig_free:concept_right}, by leveraging the fact that parallel decoding outputs predictions even for early-exited tokens from the deep model. This estimator uses a Beta mixture model\,(BMM) to capture the correlation between confidence scores and prediction alignment of two models, determining the proper confidence threshold for each dataset.
In practice, we demonstrate the efficiency of our FREE framework on extensive generation tasks.
\section{Related Work}

\subsection{Early-exiting Framework} 

As the size of language models has significantly increased, there have been numerous efforts to develop efficient decoding methods that reduce the computational cost of language generation tasks. Motivated by prior literature~\citep{teerapittayanon2016branchynet, graves2016adaptive, zhang2019your}, Elbayad
et al. (2020)~\citep{elbayad2019depth} introduced an early-exiting framework for faster inference, which dynamically adjusts the depth of the decoder for each token generation by making predictions at an intermediate layer.
To achieve the better trade-off between speed and accuracy, Schuster et al. (2022)~\citep{schuster2022confident} recently explored confidence thresholding methodologies, including various confidence measures, a decaying threshold function, and a calibration method. 

However, their experiments were primarily conducted on small-sized decoder models, necessitating further validation on larger models.
In addition, their approaches require additional training time for statistical tests on the extra calibration sets, which prevents them from real deployment scenarios.

\subsection{Parallel Decoding} 

The non-autoregressive decoding, which generates multiple output tokens in parallel, was initially proposed by Gu et al. (2018)~\citep{gu2018nonautoregressive}. Several works~\cite{marjan2019mask, gu-kong-2021-fully, savinov2022stepunrolled, santilli2023accelerating} have since focused on enhancing generation quality in machine translation tasks.
Subsequently, Leviathan et al. (2023)~\citep{leviathan2022fast} introduced speculative decoding for sequence generation tasks. In this approach, an approximation model (small size) predicts outputs autoregressively, while a target model (large size) runs in parallel to verify the acceptance of predictions made by the approximation model. With only accepted tokens, they resample the next token from an adjusted distribution. Related approaches have been proposed by Chen et al. (2023)~\cite{chen2023accelerating} and Kim et al.
(2023)~\citep{kim2023big}, where they also utilize two models of varying depth and focus on refining the small model through speculative sampling or a rollback policy in a non-autoregressive manner.\looseness=-1

Our approach is notably distinct from the aforementioned works as we focus on early-exiting framework by introducing synchronized parallel decoding within a \textit{single} network, which incorporates a shallow-deep module. While we also leverage the advantage of simultaneously obtaining predictions from models of different depth, we rather aim to develop a novel and effective estimation methodology to adaptively determine the optimal threshold for each dataset. It is worth noting that their refining strategies may result in unbounded latency increase as they restart from incorrect predictions.
\section{Preliminary}

A Transformer network~\cite{vaswani2017attention} is composed of $L$ layers, where each layer consists of two sublayers, a multi-head attention (MHA) layer and a feed-forward network (FFN) layer. The computation for hidden states at time step $t+1$ via stacked Transformer blocks is as follows:
\begin{align*}
    \mathbf{h}^{\ell}_{t+1} = \text{Transformer}_{\ell} (\mathbf{h}^{\ell - 1}_{t+1}), \; \ell \in [1, L],
\end{align*}
where $\mathbf{h}^{0}_{t+1}$ is the embedding layer outputs of $y_{t}$ that represents the generated token at time step $t$.

After $L^{\text{th}}$ layer of the decoder network, the predicted token $\hat{y}_{t+1}$, is determined by the probability output from a softmax classifier $\mathbf{W}_{L}$:
\begin{equation}
    p(y_{t+1} \vert \mathbf{h}_{t+1}^{L}) = \texttt{softmax}(\mathbf{W}_{L}^{\intercal} \mathbf{h}_{t+1}^{L}) \nonumber
\end{equation}
However, unlike the standard LMs, the early-exiting framework enables the generation of a subsequent token in earlier layers by using $p(y_{t+1} \vert \mathbf{h}_{t+1}^{\ell})$. 
If the confidence score $c^{\ell}$ is larger than the predefined threshold, we can make a prediction at time step $t+1$ as $\arg\max p(y_{t+1} \vert \mathbf{h}_{t+1}^{\ell})$.
While classifiers can be parameterized independently or shared across the $L$ layers, most early-exiting methods~\cite{elbayad2019depth, liu2021faster, schuster2022confident} utilize the shared classifier due to its large number of parameters caused by enormous vocabulary size. 

After the current token is early-exited at the $\ell^{\text{th}}$ layer, we need to calculate the key and value states for all deeper blocks in order to perform the self-attention for the subsequent tokens that pass through deeper blocks. For a more efficient approach of caching key and value states, the early-exiting frameworks employ the state copying mechanism. It duplicates the hidden states of the early-exited layer (\textit{i.e.}, $\mathbf{h}^{i}_{t+1} = \mathbf{h}^{\ell}_{t+1}, \forall i \in [\ell+1, L]$), allowing us to compute the approximate key and value states required for the self-attention of Transformer networks. Schuster et al. (2022)~\cite{schuster2022confident} have verified that state copying from lower layers does not have a detrimental effect on performance in the case of small-sized T5 models~\cite{raffel2020exploring}.\looseness=-1

\section{Re-evaluating Early-exit Framework}
\label{sec_free:re-evaluate}

In this section, we present \emph{four} new findings from our re-evaluation of the early-existing framework. We utilized different model sizes of T5~\cite{raffel2020exploring} on SAMSum~\cite{gliwa-etal-2019-samsum} and CNN/DailyMail~\cite{see-etal-2017-get}, and LongT5-base~\cite{guo-etal-2022-longt5} architectures on Multi-News~\cite{fabbri-etal-2019-multi} and BIGPATENT~\cite{sharma-etal-2019-bigpatent}.

\begin{table*}[t]
    \centering
    \small
    \caption{(\textit{Left}) Comparison of ROUGE-L scores between a full model, fine-tuned using all layer outputs, and \textit{oracle}-exiting. We also measured cosine similarity between hidden states of the last layer and oracle-exited layer. (\textit{Right}) The optimal confidence threshold to achieve desired performance. We chose the best values among threshold value from 0 to 1 with step size of 0.1. The numbers sequentially represent the selected threshold and corresponding performance (gray colored).}
    \begin{minipage}[h]{0.495\textwidth}
        \centering
        \renewcommand{\arraystretch}{1.25}
        \resizebox{\columnwidth}{!}{
        \addtolength{\tabcolsep}{-1pt}
        \begin{tabular}{l|l|cc|c}
        \toprule
        Dataset                 & Model       & Full\,M. & Oracle & Sim. \\ \midrule
        \multirow{2}{*}{SAMSum} & T5-small    & 44.84 & 44.17 (\textcolor{myred}{-0.67}) & 0.913 \\
                                & T5-large    & 48.82 & 47.58 (\textcolor{myred}{-1.24}) & 0.809 \\ \hline
        \multirow{2}{*}{CNN/DM} & T5-small    & 37.82 & 37.60 (\textcolor{myred}{-0.22}) & 0.902 \\
                                & T5-large    & 41.15 & 40.15 (\textcolor{myred}{-1.00}) & 0.792 \\ \hline
        Multi-News              & LongT5-base & 37.62 & 29.63 (\textcolor{myred}{-7.99}) & 0.724 \\ \hline
        BIGPATENT               & LongT5-base & 49.68 & 44.99 (\textcolor{myred}{-4.69}) & 0.686 \\
        \bottomrule
        \end{tabular}
        }
        \label{tab_free:obs1}
    \end{minipage}
    \hfill
    \centering
    \small
    \begin{minipage}[h]{0.495\textwidth}
        \centering
        \renewcommand{\arraystretch}{0.98}
        \resizebox{\columnwidth}{!}{
        \addtolength{\tabcolsep}{-1pt}
        \begin{tabular}{c|l|ccc}
        \toprule
         &  & \multicolumn{3}{c}{Performance Drop} \\ 
         \cmidrule(l{2pt}r{2pt}){3-5}
        Task & Dataset & $\sim$1\% & $\sim$5\% & $\sim$10\% \\\midrule
        \multirow{4}{*}{SUM} & SAMSum & 1.0~\textcolor{gray}{(48.8)} & 0.7~\textcolor{gray}{(46.8)} & 0.5~\textcolor{gray}{(45.0)} \\
          & CNN/DM & 1.0~\textcolor{gray}{(41.2)} & 0.5~\textcolor{gray}{(39.2)} & 0.3~\textcolor{gray}{(37.3)} \\
          & Multi-News & 0.8~\textcolor{gray}{(37.3)} & 0.5~\textcolor{gray}{(35.9)} & 0.4~\textcolor{gray}{(34.9)} \\
          & BIGPATENT & 1.0~\textcolor{gray}{(49.7)} & 0.8~\textcolor{gray}{(47.3)} & 0.6~\textcolor{gray}{(45.2)} \\ \hline
        QA & SQuAD & 0.1~\textcolor{gray}{(90.1)} & 0.0~\textcolor{gray}{(88.3)} & 0.0~\textcolor{gray}{(88.3)} \\ \hline
        MT & IWSLT & 1.0~\textcolor{gray}{(39.4)} & 1.0~\textcolor{gray}{(39.4)} & 1.0~\textcolor{gray}{(39.4)} \\
        \bottomrule
        \end{tabular}
        }\label{tab_free:obs4}
    \end{minipage}
\end{table*}

\subsection{Lack of Robustness to Model Size and Output Sequence Length}
\label{sec_free:3_1}

We first re-evaluate the state copying mechanism which is an essential component of the early-exiting framework. Following Schuster et al. (2022)~\cite{schuster2022confident}, we use an \textit{oracle} confidence measure that enables tokens to exit at the earliest layer, such that their predictions are identical to those of the final layer. Notably, as observed in the left of Table~\ref{tab_free:obs1}, \emph{the degradation of the generation quality with the state copying gets severe on larger models and datasets with the longer sequence} ($\vartriangleright$~\textbf{Obs.~1}). For instance, when considering the oracle-exiting results, the T5-small model demonstrates the degradation of only 0.67 on the SAMSum dataset, whereas the T5-large model experiences a much larger drop of 1.24. Similarly, on datasets such as Multi-News and BIGPATENT, which consist of relatively long output sequences, the oracle-exiting results exhibit a decrease of 7.99 and 4.69, respectively.

To strengthen the supporting evidence, we further discover the substantial variation in the distribution of hidden states across different layers. In the left of Table~\ref{tab_free:obs1}, we also reported cosine similarity between the hidden states of the final layer and the oracle-exited layer. 
Even though the hidden states of the final and oracle-exited layers yield the same predictions, the cosine similarity between them decreases significantly as the decoder network gets larger and the output sequences become longer.

\begin{figure*}[t]
    \centering
    \small
    \begin{minipage}[h]{0.54\textwidth}
        \centering
        \includegraphics[width=\linewidth]{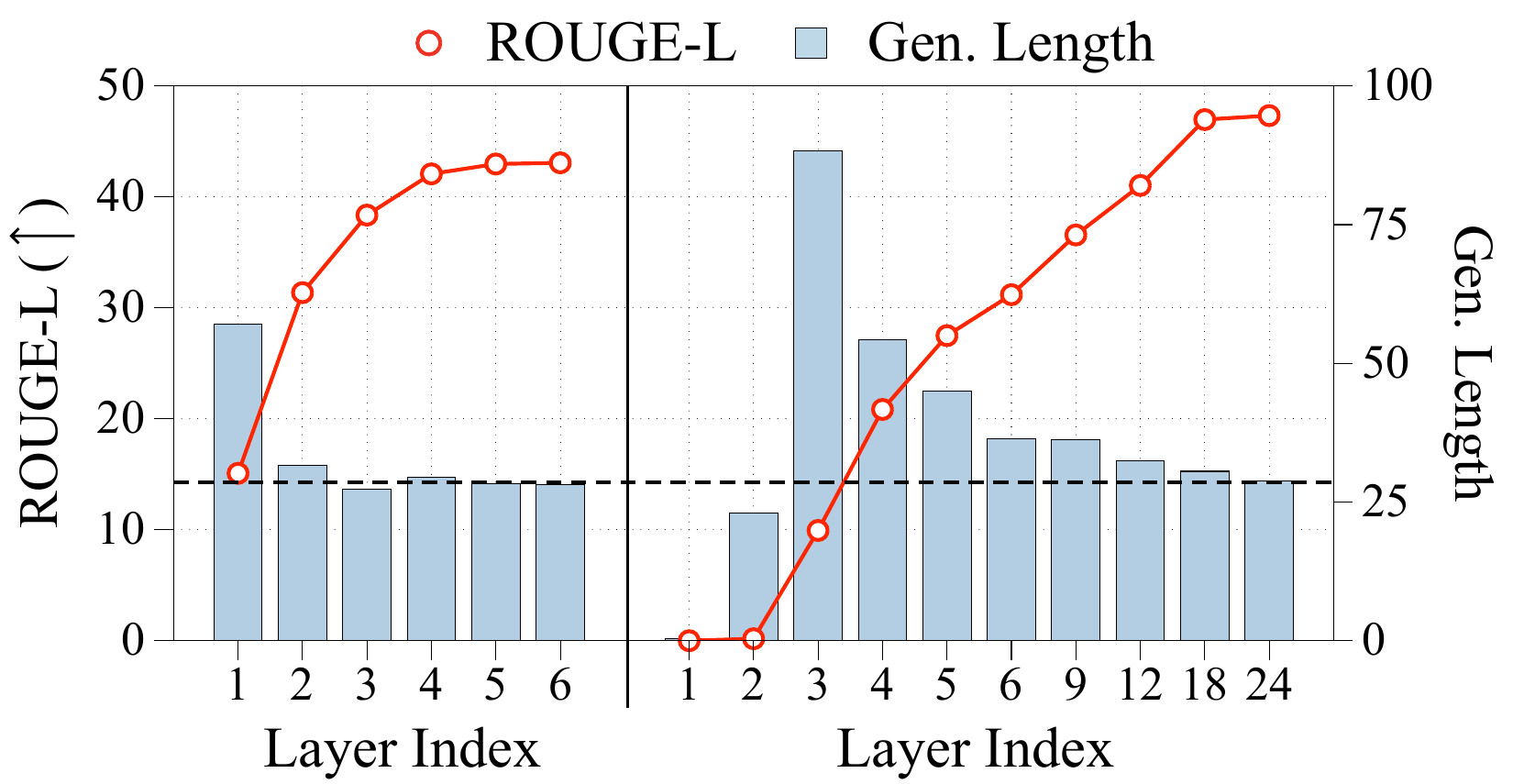}
    \end{minipage}
    \hfill
    \begin{minipage}[h]{0.445\textwidth}        
        \centering
        \includegraphics[width=\linewidth]{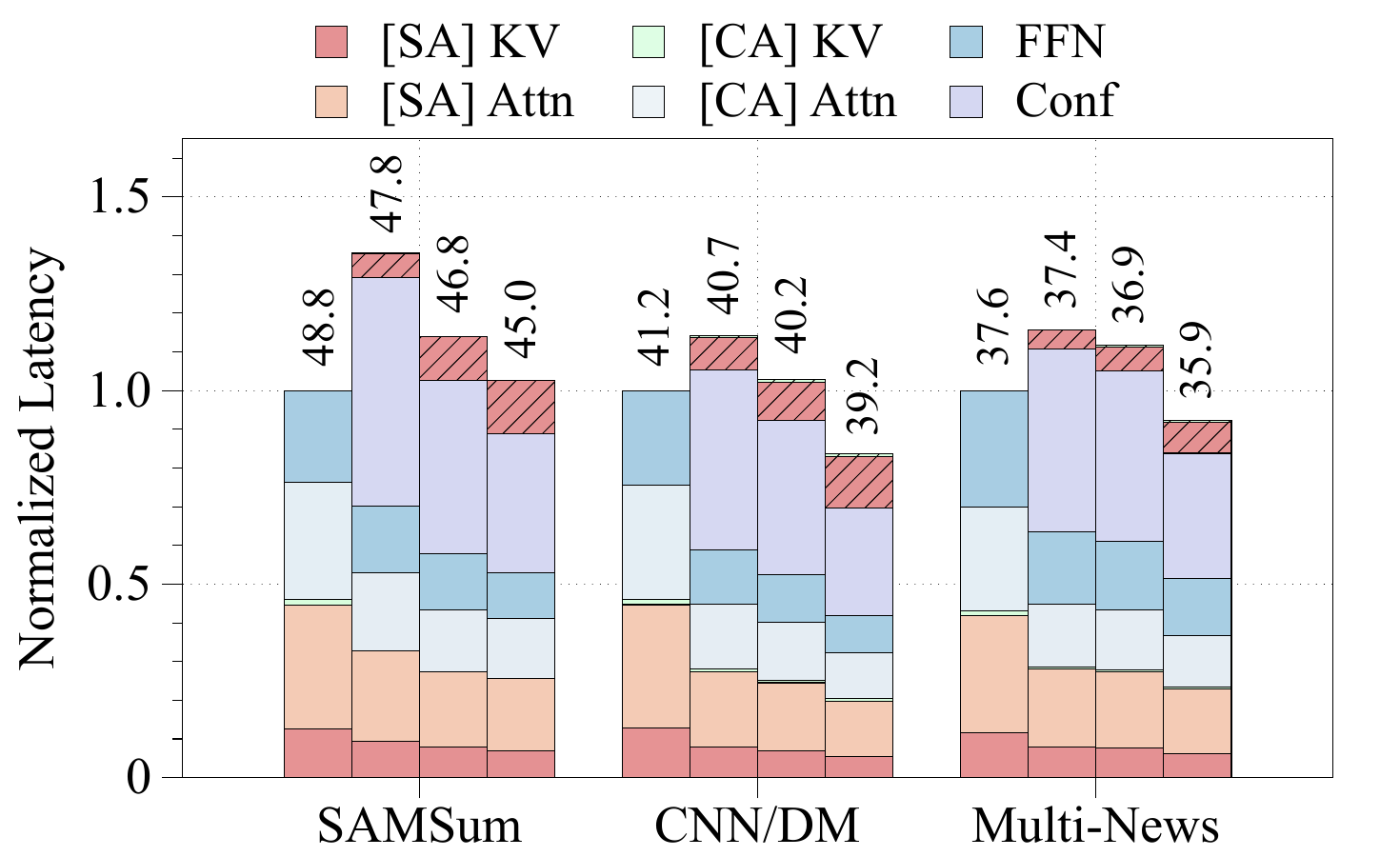}
    \end{minipage}
    \caption{(\textit{Left}) Illustration of the ROUGE-L scores and generated sequence length from the \textit{static}-exiting approach in T5-small (left) and T5-large (right) on the SAMSum dataset. The horizontal dashed line represents the average sequence length of the ground truth. (\textit{Right}) Component-wise computational cost on three datasets. Four bars correspond to full model and early-exiting with thresholds of 0.9, 0.7, and 0.5. The hatched color denotes the elapsed time after the token exits, related to the state copying mechanism. The numbers above the bars represent the ROUGE-L scores. SA and CA denote self- and cross-attention, respectively.
    }
    \label{fig_free:obs2_obs3}
\end{figure*}

\subsection{Performance Drop by Exit Position}
\label{sec_free:3_2}

To facilitate early-exiting for all decoder layers, the training objectives need to be a combination of the training objectives for each individual layer. We can present as follows:
\begin{equation}
    \mathcal{L} = \sum_{i=1}^{L} \alpha_{i} \mathcal{L}_{i} \text{\; where \;} \sum_{i} \alpha_{i} = 1,
    \label{eq:wce}
\end{equation}
$\mathcal{L}_{i}$ and $\alpha_{i}$ is negative log-likelihood loss function and weight coefficient for $i^{\text{th}}$ layer, respectively. Especially, previous work set $\alpha_{i}$ as $1/L$~(unweighted average~\cite{elbayad2019depth}) or $i/\sum_{i} i$~(weighted average~\cite{schuster2022confident}). They demonstrated that these weighting rules effectively facilitate learning in earlier layers without compromising the overall performance of the full model on small-sized decoder models.

However, as shown in Figure~\ref{fig_free:obs2_obs3}, \emph{we observed a notable decrease in the performance of \textit{static}-exiting, which utilizes the same number of layers for all tokens, when utilizing only a small portion of the early layers from the T5-large model.} ($\vartriangleright$~\textbf{Obs.~2}). For instance, if all tokens are exited in the first or second layers, the model achieved nearly zero ROUGE-L scores.
Furthermore, when we apply the early-exiting framework to these models during inference, we verified that the T5-large model generates abnormally long sentences, actually consuming more inference time. 
Based on these results, in the subsequent experiments, we have excluded the first two or four layers from the candidates for early-exiting layers of base and large models, respectively.

\subsection{Non-negligible Computational Cost}
\label{sec_free:3_3}

During our analysis, we observed that the conventional early-exiting framework not only presents performance disadvantages but also poses challenges for inference latency. In the right of Figure~\ref{fig_free:obs2_obs3}, we conducted a breakdown of the computational costs associated with a decoder model across three summarization datasets.
Surprisingly, early-exiting has often shown an unexpected increase in total decoding time when compared to the baseline model without using early-exiting. 

This can be attributed to \emph{the non-negligible computational cost involved in measuring confidence at each layer}, particularly due to the softmax operations with the large vocabulary size. In addition, although the state copying method aims to reduce computation time in the MHA and FFN layers of the remaining layers, \emph{the computation of key and value states using duplicated hidden states incurs additional non-negligible overhead} ($\vartriangleright$~\textbf{Obs.~3}).

\subsection{Disparate Optimal Confidence Threshold}
\label{sec_free:4_4}

Determining the appropriate threshold for exit confidence is a crucial challenge in the early-exiting framework as it directly impacts the trade-off between performance and latency~\cite{zhang2019scan, schuster2022confident}. 
As summarized in the right of Table~\ref{tab_free:obs4}, our observations indicate that \textit{the optimal confidence thresholds for achieving the lowest latency in the same performance significantly vary across datasets} ($\vartriangleright$~\textbf{Obs.~4}).
For instance, SQuAD and CNN/DailyMail datasets can maintain performance with relatively lower exit thresholds, whereas higher threshold values are required in the case of the IWSLT dataset.
Previous work~\cite{schuster2022confident} has leveraged distribution-free risk control techniques for confident generations. However, these methods require additional training time for statistical tests on the extra calibration set before the deployment, where time can be also influenced by the size of the threshold candidate sets.
\section{Novel Early-Exiting Framework: FREE}

Building upon the discoveries in Section~\ref{sec_free:re-evaluate}, we introduce a Fast and Robust Early-Exiting framework named FREE, leveraging a shallow-deep module and capitalizing on the structure of parallel decoding.  Furthermore, we present a confidence estimation algorithm designed to enhance the robustness of early-exiting within the FREE framework.

\subsection{Shallow-Deep Module}

We present an effective shallow-deep module, which strategically assigns a predetermined number of early layers ($L_S$) as a shallow model, while all the layers as a deep model. This module tackles the performance degradation associated with co-training numerous exiting layers in the conventional early-exiting framework. 
 
To enhance the performance of the shallow model, we exploit layerwise knowledge distillation (KD) as an additive loss term to Eq.~(\ref{eq:wce}) with $\alpha_{L_{s}}=L_{s}/(L + L_{s})$ and $\alpha_{L}=L/(L + L_{s})$:
\begin{equation}
    \mathcal{L}_{\text{KD}} = \frac{1}{|L_S|} \sum_{i=1}^{L_S} \text{MSE} (\mathbf{H}_{S}^{i}, \mathbf{H}_{D}^{m(i)}),
\end{equation}
where $m(i)$ indicates the layer in the deep model that extracts knowledge into the corresponding layer $i$ of the shallow model. $\mathbf{H}_S$ and $\mathbf{H}_D$ are hidden states from shallow and deep models.

We have experimented with the distillation from the last layer (KD-last~\cite{wang2020minilm, ko-etal-2023-revisiting}), from fixed uniform mapped layers (KD-unif~\cite{jiao-etal-2020-tinybert, park-etal-2021-distilling}), and from dynamically mapped layers (KD-dyna~\cite{xia-etal-2022-structured}). Especially, dynamic mapping function allows us to align each deep model layer with its closest counterpart in the shallow model:
\begin{equation}
    m(i) = \argmin_{j} \text{MSE} (\mathbf{H}_{S}^{i}, \mathbf{H}_{D}^{j})
\end{equation}
where $j$ denotes the layer indices of the deep model selected by the total number of $L_S$, and the condition of $m(1) \leq \dots \leq m(L_S)$ should be satisfied. Based on the consistently superior performance of KD-dyna loss (see Figure~\ref{fig_free:distill_loss}), we utilized it for all experiments with the shallow-deep module.


\begin{figure}[t]
    \centering
    \small
    \includegraphics[width=0.6\linewidth]{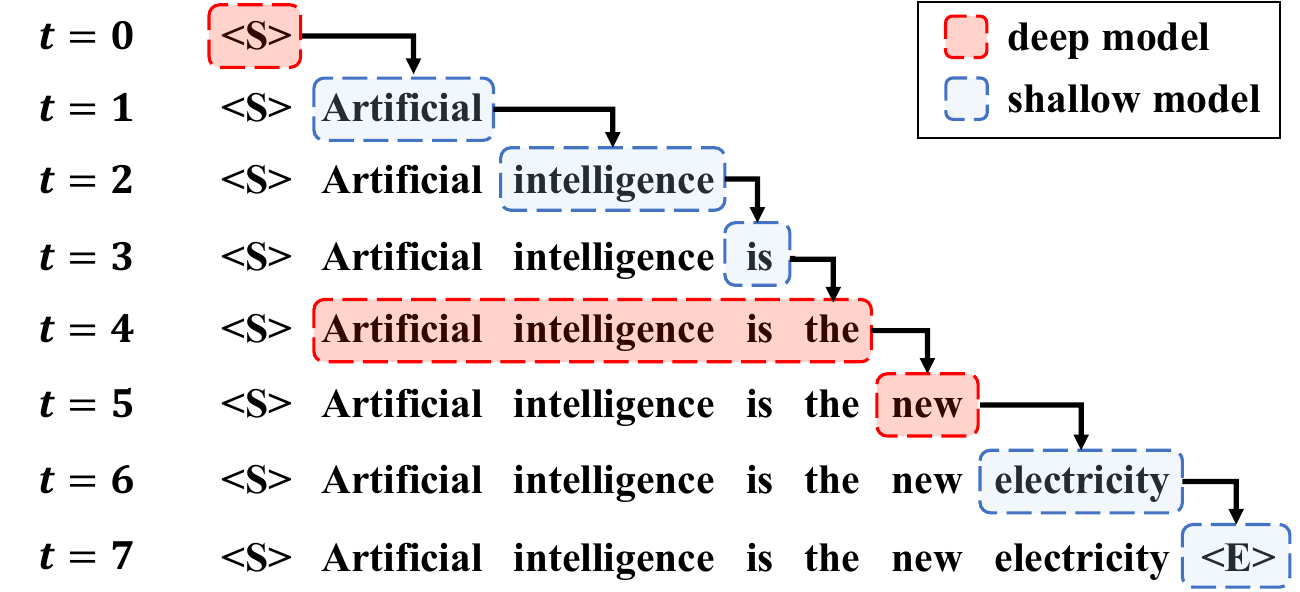}
    \caption{Overview of synchronized parallel decoding.
    We colored the tokens used to generate the next token based on the model that they forward.
    }
    \label{fig_free:parallel_example}
\end{figure}

\subsection{Synchronized Parallel Decoding}

We present synchronized parallel decoding as an alternative to the state copying mechanism, which is a key component of the conventional early-exiting framework but can lead to a significant performance decrease, as demonstrated in Section~\ref{sec_free:3_1}. In contrast to traditional approaches that have multiple exit points, our method incorporates the shallow-deep module, enabling us to stack consecutive early-exited tokens in the shallow model until a non-exiting token is encountered. When decoding the token with the deep model, we enhance efficiency and effectiveness through parallel decoding, synchronously computing the key and value states of previously stacked tokens. The example of the parallel decoding process is depicted in Figure~\ref{fig_free:parallel_example}.

The underlying principle of this approach is to leverage the enhanced parallelism offered by modern hardware accelerators. This allows for efficient computations to be carried out simultaneously on the large number of sequences. Thus, by employing synchronized parallel decoding, we can directly compute multiple hidden states similar to a single token processing time. Besides, this can eliminate the potential performance degradation that may arise from inaccurate approximations of hidden states resulting from the state copying mechanism.

\subsection{Adaptive Threshold Estimation}

We propose a novel adaptive threshold estimation method that updates the threshold to be retailed for different datasets. Unlike the previous methods that utilize extra calibration sets~\cite{schuster2022confident}, we quickly adapt the threshold by using the information of early-stage instances, regardless of the initial threshold values. Especially, during parallel decoding, we collect samples to evaluate the correspondence between the confidence scores of the shallow model and the prediction alignment between shallow and deep models.\looseness=-1

As depicted in Figure~\ref{fig_free:concept_right}, we observe that when the predictions of the deep and shallow models are identical, the confidence tends to skew towards one, otherwise it skews towards zero. To model this skewed distribution over $[0, 1]$, we utilize a \emph{beta mixture} model~(BMM~\cite{ma2011bayesian}) due to its flexibility and the appropriate support set of the beta distribution. The probability density function of beta distribution over $x \in [0, 1]$ is defined as:
\begin{equation}
    p(x \vert \alpha, \beta) = \frac{\Gamma(\alpha + \beta)}{\Gamma(\alpha)\Gamma(\beta)} x^{\alpha-1} (1-x)^{\beta-1}
\end{equation}
The parameters of the BMM are updated using the maximum likelihood estimator~(MLE~\cite{norden1972survey}) with observed data points.
\begin{equation}
    \alpha_{k} = \bar{c}_{k} \left( \frac{\bar{c}_{k} (1-\bar{c}_{k})}{s_{k}^{2}} - 1 \right), \beta_{k} = \frac{\alpha_{k} (1 - \bar{c}_{k})}{\bar{c}_{k}},
\end{equation}
where $\bar{c}_{k}$ being a average of the confidence $\{c_{i}^{L_{s}}\}_{i=1}^{\vert \mathcal{D}_{c} \vert}$ for corresponding $k$. $k$ is set to $1$ if the predictions of the two models are identical, and $0$ otherwise. Similarly, $s_{k}$ is the standard deviation of confidence of related $k$.
\begin{align}
    \bar{c}_{k} = \frac{\sum_{i=1}^{N} \gamma_{i} c_{i}^{L_{s}}}{\sum_{i=1}^{N} \gamma_{i}}, \bar{s}_{k}^{2} = \frac{\sum_{i=1}^{N} \gamma_{i} (c_{i}^{L_{s}} - \bar{c}_{k})^{2}}{\sum_{i=1}^{N} \gamma_{i}},
\end{align}
where $\gamma_{i} \coloneqq \text{I}(\hat{y}_{i}^{L_{s}} = \hat{y}_{i}^{L})$ denote whether the prediction of two models are same.

After updating the BMM, we find an appropriate threshold for future tokens by identifying the point at which the posterior probability, defined as below, reaches $\zeta$:
\begin{equation}
    p(k=1 \vert \lambda_{c}) = \frac{p(k=1) p(\lambda_{c} \vert \alpha_{1}, \beta_{1})}{\sum_{j \in \{ 0, 1 \}}p(k=j) p(\lambda_{c} \vert \alpha_{j}, \beta_{j})}.
\end{equation}
Here, as we observe the severe imbalance between the case of $k=0$ and $1$, we restrict the prior value of each class to $0.5$ for the balance between two cases~(\textit{i.e.,} $p(k=j)=0.5 \text{\,\,} \forall j$). As this restriction makes us to use a smaller value of $\zeta$, we na\"ively set it as 0.4. A detailed algorithm can be found in Algorithm~\ref{alg_free:threshold}.

\begin{algorithm}[t]
    \caption{Adaptive Threshold Estimation}
    \label{alg_free:threshold}
    \textbf{Input}: empty calibration dataset $\mathcal{D}_{c}$, initial confidence threshold $\lambda_{c}^{0}$, posterior condition $\zeta$, update number $T$ \\
    \textbf{Output}: updated confidence threshold $\lambda_{c}$
    \begin{algorithmic}[1]
    \STATE initialize $t \leftarrow 0$, $\lambda_{c} \leftarrow \lambda_{c}^{0}$
    \WHILE{$t \leq T$}
    \STATE Generate $t^{\text{th}}$ sentence with $N_{t}$ tokens 
    \STATE \textcolor{gray}{/* Update $\mathcal{D}_{c}$ */}
    \STATE $\mathcal{D}_{c} \leftarrow \mathcal{D}_{c} \cup \{c_{i}^{L_{s}}, \text{I}(\hat{y}_{i}^{L_{s}} = \hat{y}_{i}^{L})\}_{i = 1}^{N_{t}}$
    \STATE \textcolor{gray}{/* Find threshold with Eq.(2)-(4)*/}
    \STATE $\alpha_{k}, \beta_{k} \leftarrow \text{MLE}_{\text{BMM}}$($\mathcal{D}_{c}$) for $k \in \{0, 1\}$
    \STATE $\lambda_{c} \leftarrow \argmin_{\lambda: p(k=1 \vert \lambda) \geq \zeta} \lambda$
    \STATE update $t \leftarrow t + 1$
    \ENDWHILE
    \end{algorithmic}
\end{algorithm}
\section{Experiments}

\begin{table*}{t}
    \centering
    \caption{Optimized hyperparameters for training shallow-deep T5 models. The column labeled `\#\,Batch' indicates the product of the batch size per GPU and the number of GPUs. `In\,len.' and `Out\,len.' represent the maximum length of the input and output, respectively.}
    \renewcommand{\arraystretch}{1.1}
    \resizebox{0.65\columnwidth}{!}{
    \begin{tabular}{ll|cccc}
    \toprule[0.1em]
         Dataset & Model & \#\,Batch & Epochs & In\,len. & Out\,len. \\ \midrule
         SAMSum & T5-large & 4$\times$2 & 20 & 512 & 128 \\
         CNN/DM & T5-large & 4$\times$4 & 3 & 512 & 128 \\
         Multi-News & LongT5-base & 2$\times$2 & 3 & 2048 & 512 \\
         BIGPATENT & LongT5-base & 2$\times$2 & 3 & 2048 & 512 \\ 
         SQuAD & T5-large & 4$\times$2 & 10 & 512 & 30 \\ 
         IWSLT\,2017 & mT5-large & 4$\times$4 & 2 & 1024 & 128 \\
    \bottomrule[0.1em]
    \end{tabular}%
    }
    \label{tab_free:setup}
\end{table*}

\subsection{Experimental Setup}

\paragraph{Dataset description.}
We conducted experiments on various sequence modeling tasks, including text summarization tasks using SAMSum, CNN/DailyMail, Multi-News, and BIGPATENT datasets, question answering (SQuAD~\cite{rajpurkar-etal-2016-squad}), and machine translation (IWSLT 2017 En-De~\cite{cettolo-etal-2017-overview}). The LongT5-base model was used for the Multi-News and BIGPATENT datasets, while the T5-large model was used for the other datasets. We provide detailed descriptions of the datasets used.

\begin{itemize}[leftmargin=4.5mm]
    \item \textbf{SAMSum}: SAMSum~\cite{gliwa-etal-2019-samsum} consists of 16K messenger-like conversations that are annotated with a summary for providing a concise overview of the conversation's content in the third person.\looseness=-1
    \item \textbf{CNN/DailyMail}: CNN/ DailyMail~\cite{see-etal-2017-get} consists of over 300K English news articles that were originally designed for machine-reading and comprehension as well as abstractive question answering, but it now also supports extractive and abstractive summarization.
    \item \textbf{Multi-News}: Multi-News~\cite{fabbri-etal-2019-multi} comprises 45K news articles and corresponding summaries, where each summary is professionally crafted and provides links to the original articles referenced.\looseness=-1
    \item \textbf{BIGPATENT}: BIGPATENT~\cite{sharma-etal-2019-bigpatent} contains 1.3M records of U.S. patent documents, each accompanied by abstractive summaries written by humans. In our work, we specifically focus on the Fixed Constructions category, which is one of the nine classification categories available in the dataset.
    \item \textbf{SQuAD}: The Stanford Question Answering (SQuAD~\cite{rajpurkar-etal-2016-squad}) is a collection of 87.6K reading comprehension tasks. It includes questions generated by crowd workers based on a set of Wikipedia articles.\looseness=-1
    \item \textbf{IWSLT 2017}: IWSLT 2017~\cite{cettolo-etal-2017-overview} addresses text translation, using a single machine translation (MT) system for multiple language directions such as English and German. Here, we specifically focus on a German-to-English translation task.
\end{itemize}

\paragraph{Training settings.} 

All implementations are based on PyTorch using \texttt{Huggingface}~\cite{wolf-etal-2020-transformers, Lhoest2021huggingface}. We utilize the NVIDIA RTX 3090 GPUs for training the language models, and we summarize the training configuration in Table~\ref{tab_free:setup}. For all dataset, we use AdaFactor~\cite{shazeer2018adafactor} optimizer with the learning rate of 1e-4. For the adaptive threshold estimation, we set the initial threshold value $\lambda_{c}^{0}$ as 0.9, $\zeta$ as 0.4, $T$ as 3\% of total sample number (refer to Algorithm~\ref{alg_free:threshold}).

\paragraph{Performance metrics.} 

To numerically measure the output quality of our method, we utilize the F1 score for SQuAD, BLEU score~\cite{papineni-etal-2002-bleu} for IWSLT2017, and ROUGE score~\cite{lin-2004-rouge} for four summarization tasks.

\paragraph{Inference latency evaluation.}

For measuring inference speed, we execute 500 inference predictions for each dataset under each examined configuration in PyTorch~\cite{paszke2019pytorch} compiled function in a single server with a single NVIDIA GeForce RTX 3039 GPU and 12th Gen Intel(R) Core(TM) i7-12700K CPU. For each inference prediction, we use batch size 1, which is a common use case for online serving~\cite{schuster2022confident}. Also, we use to generate output sequences through greedy sampling with a beam size of 1. We measure the time including all decoding steps until completion.

\begin{figure*}[t]
    \centering
    \includegraphics[width=\linewidth]{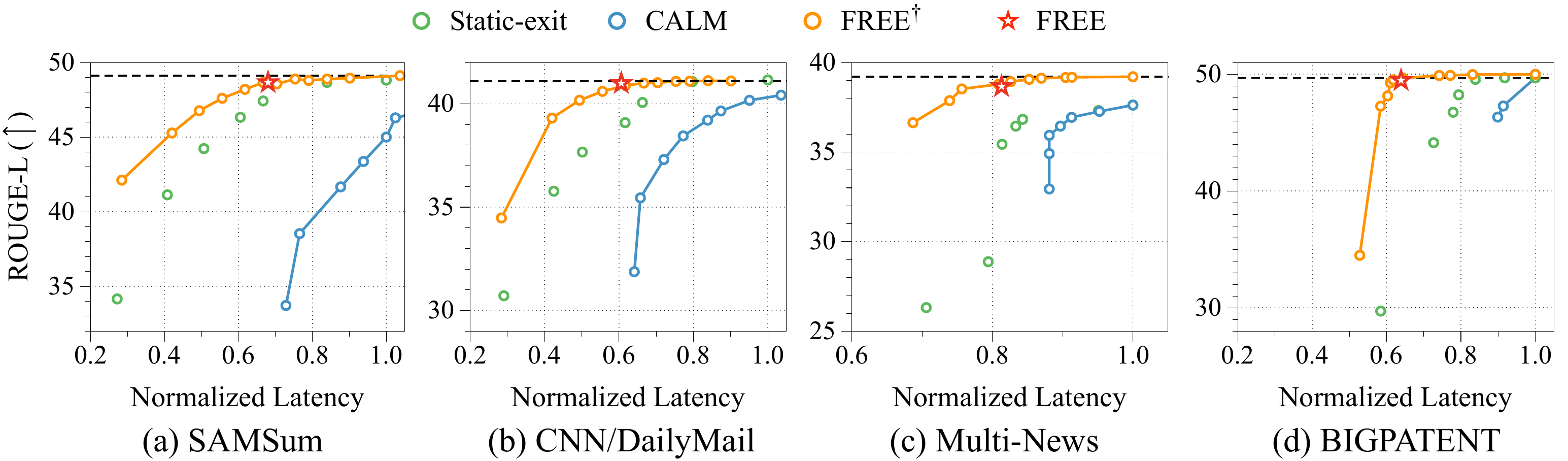}    
    \caption{The trade-off between the generated output quality and normalized latency under different exit conditions. We varied the exit threshold values between 0 and 1 for both CALM and FREE$^{\dagger}$ and the number of exit layers for the static-exiting framework. We exclude the inner point of the Pareto curve, and the dashed line represents the ROUGE-L score of the full model, which is the fine-tuned shallow-deep module.}
    \label{fig_free:main_result}
\end{figure*}

\subsection{Experimental Results}
\label{sec_free:6_2}

In order to investigate the effect of the individual component of our proposed framework, we evaluate both FREE without and with an adaptive threshold estimator, denoted as FREE$^{\dagger}$ and FREE.

\paragraph{Overall performance.} 

In Figure~\ref{fig_free:main_result}, we present a comparison of the quality of generated output (ROUGE-L) and the inference latency between the FREE framework and baselines, including static-exiting and the conventional early-exiting method (CALM~\cite{schuster2022confident}). CALM method exhibited poorer performance compared to a simple static-exiting approach on all summarization datasets, likely due to the state copying mechanism and the presence of numerous exit positions, as observed in Section~\ref{sec_free:re-evaluate}. In contrast, FREE$^{\dagger}$ demonstrated robust performance and the larger AUC~(area under the curve) across datasets by adjusting exit thresholds.

\begin{figure*}[t]
    \centering
    \small
    \begin{minipage}[h]{0.505\textwidth}
        \includegraphics[width=\linewidth]{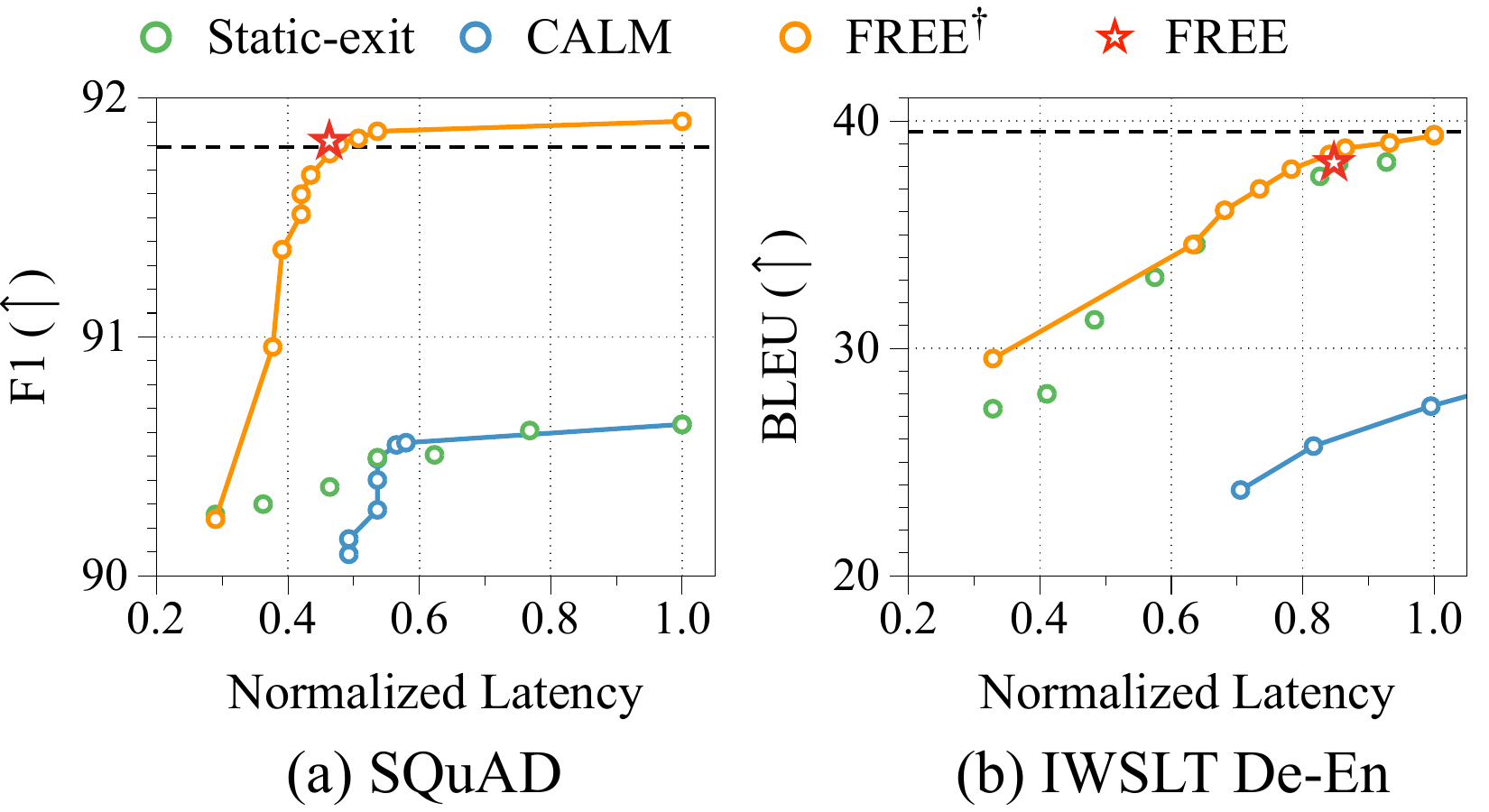}
    \end{minipage}
    \hfill
    \begin{minipage}[h]{0.485\textwidth}
        \centering
        \includegraphics[width=\linewidth]{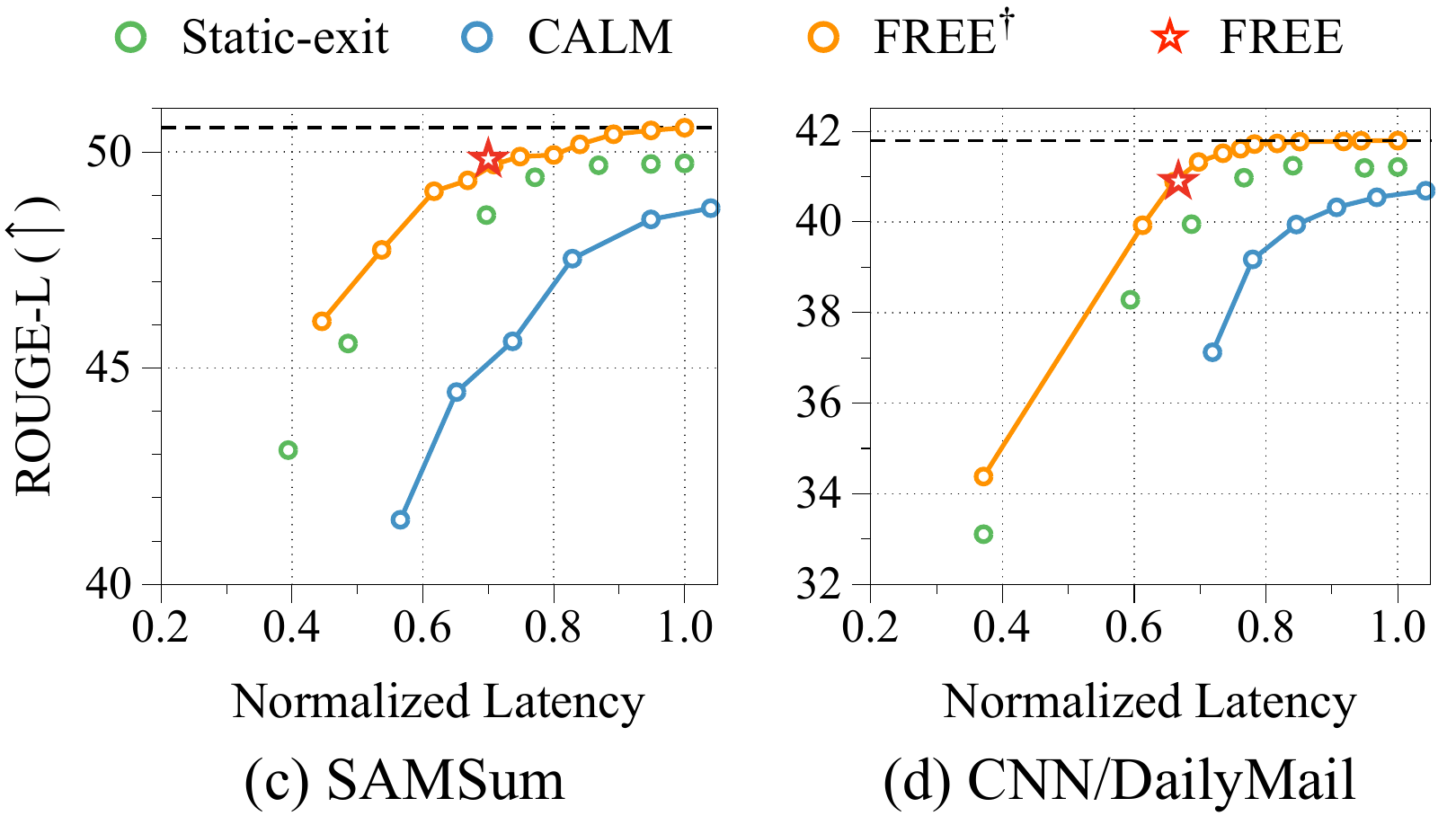}
    \end{minipage}
    \caption{(\textit{Left}) The trade-off between the generated output quality and normalized latency under different exit conditions. The dashed line represents the F1 and BLEU scores of the full model, which is the fine-tuned shallow-deep module, respectively. Similar to Figure~\ref{fig_free:main_result}, we exclude the inner point of the Pareto curve. (\textit{Right}) The trade-off between the generated output quality and normalized latency on T5-3B models.}
    \label{fig_free:main_qa_mt_3b}
\end{figure*}

We also present a comparison of the quality of the generated output (F1 or BLEU) and the inference latency on the SQuAD and IWSLT 2017 datasets. The left of Figure~\ref{fig_free:main_qa_mt_3b} illustrates that both FREE$^{\dagger}$ and FREE consistently outperform the CALM and static-exiting baselines in the SQuAD dataset, which aligns with our previous findings. 

However, their performance advantages in the IWSLT dataset are slightly reduced compared to other datasets. This can be attributed to the larger vocabulary size of mT5 compared to T5, resulting in longer processing times for the confidence measurement. The CALM approach, which also utilizes large linear classifiers, exhibits much lower effectiveness in this dataset as well. We believe that this challenge, regarding the large vocabulary size, can be mitigated by employing a vocabulary size-independent confidence measure that proposed in previous work~\cite{schuster2022confident}. Nonetheless, our proposed algorithm still outperforms the other baselines on various datasets.

\paragraph{Performance on large language models.}

Recently, various studies~\cite{dettmers2022llm, guangxuan2022smoothquant, leviathan2022fast, zichang2023dejavu} have aimed at boosting the inference speed of large language models (LLMs). To validate the applicability of the FREE framework on LLMs, we conducted experiments utilizing the T5-3B model~\cite{raffel2020exploring} on the SAMSum and CNN/DailyMail datasets. Due to substantial computational overhead, we utilized the LoRA adapter~\cite{edward2022lora}, targeting both self-attention and feed-forward layers with a rank of 64. The right of Figure~\ref{fig_free:main_qa_mt_3b} summarized a comprehensive comparison of early-exiting methods. Our method maintained superiority over the baselines in terms of latency and ROUGE-L scores, showing the consistent performance trend observed in the T5-large model. Thus, we are confident that our proposed framework would demonstrate consistent level of inference acceleration, even with larger language models.

\begin{table*}[t]
\centering
\caption{Comparison between early-exiting frameworks on various datasets. For CALM and FREE$^{\dagger}$, we reported the performance using the smallest threshold value that achieves 99\% performance of the full model, fine-tuned by weighted average or KD-dyna losses, respectively. The parentheses denote relative speedup based on the first row.}
\renewcommand{\arraystretch}{1.1}
\resizebox{1.0\textwidth}{!}{
\addtolength{\tabcolsep}{2.5pt}
\begin{tabular}{l|cccc|c|c}
\toprule[0.1em]
        & \multicolumn{4}{c|}{SUM} & \multicolumn{1}{c|}{QA} & \multicolumn{1}{c}{MT} \\ 
        \cmidrule(l{2pt}r{2pt}){2-5} \cmidrule(l{2pt}r{2pt}){6-6} \cmidrule(l{2pt}r{2pt}){7-7} Method & SAMSum & CNN/DailyMail & Multi-News & BIGPATENT & SQuAD & IWSLT\,De-En \\ \midrule
        Full Model & 48.82 ($\times$1.00) & 41.15 ($\times$1.00) & 37.62 ($\times$1.00) & 49.68 ($\times$1.00) & 90.63 ($\times$1.00) & 39.19 ($\times$1.00) \\
        CALM & 48.37 ($\times$0.72) & 40.78 ($\times$0.86) & 37.27 ($\times$0.85) & 49.21 ($\times$0.65) & 90.09 ($\times$2.03) & 39.19 ($\times$1.00) \\ \hline  
        Full Model & 49.11 ($\times$1.00) & 41.09 ($\times$1.00) & 39.20 ($\times$1.00) & 49.68 ($\times$1.00) & 91.90 ($\times$1.00) & 39.39 ($\times$1.00) \\
        FREE$^{\dagger}$ & 48.65 ($\times$1.50) & 40.89 ($\times$1.80) & 38.93 ($\times$1.07) & 49.51 ($\times$1.62) & 91.31 ($\times$2.76) & 39.04 ($\times$1.07) \\
        FREE & 48.66 ($\times$1.47) & 40.99 ($\times$1.65) & 38.66 ($\times$1.23) & 49.47 ($\times$1.58) & 91.82 ($\times$2.16) & 38.17 ($\times$1.18) \\
        \bottomrule[0.1em]
\end{tabular}
}\label{tab_free:main_exp}
\end{table*}

\paragraph{Adaptive threshold evaluation.}

In the early-exiting framework, choosing the appropriate confidence threshold is crucial for achieving the best trade-off between generation quality and latency. Unlike previous calibration methods \cite{schuster2022confident} that require an extra calibration set and training time, 
our methodology effectively addresses this challenge by leveraging the byproduct of parallel decoding.
As summarized in Table~\ref{tab_free:main_exp}, FREE with adaptive threshold estimation successfully achieved significant speedup, by up to $\times$2.16, when preserving the 99\% of full model performance. 
Furthermore, in Figure~\ref{fig_free:main_result}, the estimated threshold demonstrated nearly the maximum achievable speed improvement without sacrificing performance, represented by red stars.

\subsection{Ablation Study}

\begin{table*}[t]
    \centering
    \small
    \caption{(\textit{Left}) The comparison of ROUGE-L and speedup based on different numbers of layers for shallow model and confidence thresholds. (\textit{Right}) The comparison between synchronized parallel decoding (SPC) and state copying (SC). The shallow-deep module is utilized in both decoding methods.}
    \begin{minipage}[h]{0.495\textwidth}
        \centering
        \renewcommand{\arraystretch}{1.}
        \resizebox{\columnwidth}{!}{
        \addtolength{\tabcolsep}{-1pt}
        \begin{tabular}{l|c|ccc}
        \toprule
         & & \multicolumn{3}{c}{Threshold}\\ 
        \cmidrule(l{2pt}r{2pt}){3-5}
        Dataset & $L_S$ & 0.7 & 0.5 & 0.3 \\ \midrule
        \multirow{4}{*}{SAMSum} 
         & 4 & 48.27\,($\times$1.04) & 46.95\,($\times$1.09) & 44.72\,($\times$1.15) \\
         & 6 & 48.89\,($\times$\textbf{1.32}) & 48.65\,($\times$\textbf{1.50}) & 47.60\,($\times$\textbf{1.80}) \\
         & 8 & 48.74\,($\times$1.11) & 47.97\,($\times$1.17) & 47.09\,($\times$1.31) \\
         & 12 & \textbf{48.97}\,($\times$1.21) & \textbf{48.74}\,($\times$1.28) & \textbf{48.10}\,($\times$1.37) \\\hline
        \multirow{4}{*}{CNN/DM} 
         & 4 & 41.03\,($\times$1.45) & 40.59\,($\times$1.68) & 39.88\,($\times$1.86) \\
         & 6 & 41.08\,($\times$\textbf{1.53}) & 41.00\,($\times$\textbf{1.69}) & 40.60\,($\times$\textbf{2.07}) \\
         & 8 & \textbf{41.19}\,($\times$1.44) & \textbf{41.15}\,($\times$1.64) & 40.95\,($\times$1.69) \\
         & 12 & 41.11\,($\times$1.33) & 41.09\,($\times$1.47) & \textbf{40.95}\,($\times$1.55) \\
        \bottomrule
        \end{tabular}
        }\label{tab_free:depth}
    \end{minipage}
    \hfill
    \begin{minipage}[h]{0.495\textwidth}
        \centering
        \renewcommand{\arraystretch}{1.27}
        \resizebox{\columnwidth}{!}{
        \addtolength{\tabcolsep}{-1pt}
        \begin{tabular}{l|cc|ccccc}
        \toprule
         & \multicolumn{2}{c|}{Method} & \multicolumn{5}{c}{Threshold} \\ 
        \cmidrule(l{2pt}r{2pt}){2-8}
        Dataset & SC & SPD & 0.9 & 0.7 & 0.5 & 0.3 & 0.1\\\midrule
        \multirow{2}{*}{SAMSum} & \cmark & \xmark & 46.35 & 44.59 & 43.92 & 42.36 & 41.27 \\
         & \xmark & \cmark & \textbf{48.89} & \textbf{48.89} & \textbf{48.65} & \textbf{47.60} & \textbf{45.27} \\\hline
        \multirow{2}{*}{CNN/DM} & \cmark & \xmark & 40.92 & 40.92 & 40.71 & 39.99 & 38.17 \\
         & \xmark & \cmark & \textbf{41.12} & \textbf{41.08} & \textbf{41.00} & \textbf{40.60} & \textbf{39.30} \\ \hline
        \multirow{2}{*}{Multi-News} & \cmark & \xmark & 38.43 & 37.61 & 36.55 & 33.99 & 29.34 \\
         & \xmark & \cmark & \textbf{39.16} & \textbf{39.06} & \textbf{38.78} & \textbf{37.87} & \textbf{33.98} \\
        \bottomrule
        \end{tabular}
        }\label{tab_free:copying}
    \end{minipage}
\end{table*}

\paragraph{Different depth of shallow model.} 

In the left of Table~\ref{tab_free:depth}, we also ablate on the number of layers for the shallow model to observe the trade-offs. While our method demonstrated a trend towards higher speedup gains as the depth of the shallow model decreases, we experienced some decreases in performance and speed gain when the depth of the model is reduced too much~(\emph{e.g.,} four layers). We assumed that this is due to incorrect and redundant output sentences, similarly observed in the conventional early-exiting framework. Consequently, with enough depth~(\emph{e.g.,} six layers), FREE consistently showed robust performance and inference speedup.

\paragraph{Robustness of parallel decoding.} 

In order to verify the robustness of our decoding mechanism, we conducted a comparative analysis between synchronized parallel decoding (SPD) and state copying (SC), both implemented with the shallow-deep module. Synchronized parallel decoding consistently outperformed state copying across all three datasets by much higher ROUGE-L metrics, as summarized in the right of Table~\ref{tab_free:copying}. This improvement can be attributed to the updated hidden states that are obtained through the accurate computation of Transformer layers during parallel decoding. These findings suggest that our efficient decoding method for early-exited tokens can enhance the overall performance of the early-exiting framework as well.

\vspace{-3pt}
\paragraph{Dependency on size of calibration set.} 

By using the early-stage instances as the calibration set, we iteratively update the adaptive confidence threshold to converge to the appropriate value. 
Here, we have observed the sample efficiency of the adaptive threshold estimator by varying the sizes of this calibration set.
Interestingly, even with only 3\% of the total samples, our estimator can approximate the threshold, measured by the full sample set, as shown in the left of Table~\ref{tab_free:calibration}. This ensures minimal additional computation time required for threshold estimation.

\begin{table*}[t]
    \centering
    \small
    \caption{(\textit{Left}) The experimental results of FREE framework based on different sizes of the calibration set. (\textit{Right}) The evaluation of refinement methods in the FREE framework. Refining thresholds control the level of acceptance for predictions from the shallow model.}
    \begin{minipage}[h]{0.58\textwidth}
        \centering
        \renewcommand{\arraystretch}{1.5}
        \resizebox{\columnwidth}{!}{
        \addtolength{\tabcolsep}{-2.5pt}
        \begin{tabular}{l|ccc|ccc|c}
        \toprule
         & \multicolumn{3}{c|}{3\%} & \multicolumn{3}{c|}{10\%} & 100\% \\ 
        \cmidrule(l{2pt}r{2pt}){2-4}  \cmidrule(l{2pt}r{2pt}){5-7} \cmidrule(l{2pt}r{2pt}){8-8}
        Dataset & Thr. & Perf. & Speed & Thr. & Perf. & Speed & Thr. \\ \midrule
        SAMSum & 0.51 & 48.66 & $\times$1.47 & 0.49 & 48.69 & $\times$1.51 & 0.48 \\
        BIGPATENT & 0.54 & 49.47 & $\times$1.58 & 0.54 & 49.39 & $\times$1.63 & 0.54 \\
        \bottomrule
        \end{tabular}
        }\label{tab_free:calibration}
    \end{minipage}
    \hfill
    \begin{minipage}[h]{0.41\textwidth}
        \centering
        \renewcommand{\arraystretch}{0.93}
        \resizebox{\columnwidth}{!}{
        \addtolength{\tabcolsep}{-3pt}
        \begin{tabular}{l|cc|cc|cc}
        \toprule
         &  &  & \multicolumn{2}{c|}{Thr.~0.7} & \multicolumn{2}{c}{Thr.~0.3} \\ 
         \cmidrule(l{2pt}r{2pt}){4-5}  \cmidrule(l{2pt}r{2pt}){6-7} 
        Dataset & Ref. & Thr. & Perf. & Speed & Perf. & Speed \\\midrule
        \multirow{3}{*}{SAMSum} & \xmark & - & 48.89 & \textbf{$\times$1.33} & 47.60 & $\times$\textbf{1.80} \\
         & \cmark & 1.0 & \textbf{49.08} & $\times$1.26 & \textbf{48.31} & $\times$1.50 \\
         & \cmark & 0.1 & 49.06 & $\times$1.17 & 48.27 & $\times$1.12 \\\hline
        \multirow{3}{*}{CNN/DM} & \xmark & - & \textbf{41.08} & \textbf{$\times$1.53} & 40.60 & $\times$\textbf{2.07} \\
         & \cmark & 1.0 & 40.86 & $\times$1.51 & \textbf{40.78} & $\times$1.67 \\
         & \cmark & 0.1 & 40.85 & $\times$1.35 & 40.75 & $\times$1.21 \\
        \bottomrule
        \end{tabular}
        }\label{tab_free:rollback}
    \end{minipage}
\end{table*}

\vspace{-3pt}
\paragraph{Refining shallow model predictions.} 

Prior works~\cite{leviathan2022fast, chen2023accelerating, kim2023big} have proposed refinement methods to correct erroneous outputs from an approximation model.
Specifically, when a wrong token is detected in previous sequences, they remove all subsequently generated tokens and restart the generation process from that point.
In the right of Table~\ref{tab_free:rollback}, we conducted experiments in order to evaluate the effects of this refinement method~\cite{kim2023big} in our early-exiting framework.
We observed that when the refinement threshold is set low, allowing for more correction by the deep model, the performance improvement is minimal compared to the increase in latency.
Our findings suggest that these approaches that cannot guarantee an upper bound on latency increase may not be well-suited for integration into the early-exiting framework.\looseness=-1

\begin{figure*}[t]
    \centering
    \small
    \begin{minipage}[h]{0.495\textwidth}
        \begin{tcolorbox}
        [width=\linewidth, sharp corners=all, colback=gray!10, boxrule=0.2mm]
        Evaluate the quality of summaries written for a news article. Rate each summary on four dimensions: \textcolor{blue}{\{Dimension\_1\}}, \textcolor{blue}{\{Dimension\_2\}}, \textcolor{blue}{\{Dimension\_3\}}, and \textcolor{blue}{\{Dimension\_4\}}. You should rate on a scale from 1 (worst) to 5 (best). \\
        
        Article: \textcolor{blue}{\{Article\}} \\
        Summary: \textcolor{blue}{\{Summary\}}
        \end{tcolorbox}
    \end{minipage}
    \hfill
    \begin{minipage}[h]{0.495\textwidth}
        \begin{tcolorbox}
        [width=\linewidth, sharp corners=all, colback=gray!10, boxrule=0.2mm]
        Given a new article, which summary is better? Answer "Summary 0" or "Summary 1". You do not need to explain the reason. \\
        \\
        
        Article: \textcolor{blue}{\{Article\}} \\
        Summary 0: \textcolor{blue}{\{Summary\_0\}} \\
        Summary 1: \textcolor{blue}{\{Summary\_1\}}
        \end{tcolorbox}
    \end{minipage}
    \caption{(\textit{Left}) The template for Likert scale scoring. The four dimensions are relevance, informativeness, fluency, and coherence. (\textit{Right}) The template for pairwise comparison. We measured twice by changing the order of summaries for a fair comparison.}
    \label{fig_free:likert_scale_pairwise_comp}
\end{figure*}

\paragraph{Human-like summarization evaluation.}

Recent studies~\cite{gao2023human, liu2023gpteval, zhang2023wider} have argued that existing summarization evaluation metrics like ROUGE-L do not accurately represent the true summarization capabilities. Instead, they explored the human-like evaluation using LLMs based on their strong correlation with human judgment. Thereby, we conducted two human-like evaluation methods, Likert scale scoring and pairwise comparison~\cite{gao2023human}, using ChatGPT API (gpt-3.5-turbo-0613). We compared a full model and our FREE framework on 100 instances, randomly drawn from the CNN/DailyMail dataset. Figure~\ref{fig_free:likert_scale_pairwise_comp} provide the templates used for each evaluation task. For the full model, we observed scores of [4.73, 3.83, 3.87, 3.77], while our FREE method returned scores of [4.68, 3.84, 3.84, 3.72] across the four dimensions. Besides, the win counts for each method were 101 and 99, respectively. Given ROUGE-L scores of 41.09 ($\times$1.00) for the full model and 40.99 ($\times$1.65) for the FREE method, our method is certainly capable of yielding predictions of similar quality, while notably reducing computational overhead.

\begin{wrapfigure}{r}{0.5\textwidth}
    \centering
    \vspace{-13pt}
    \includegraphics[width=\linewidth]{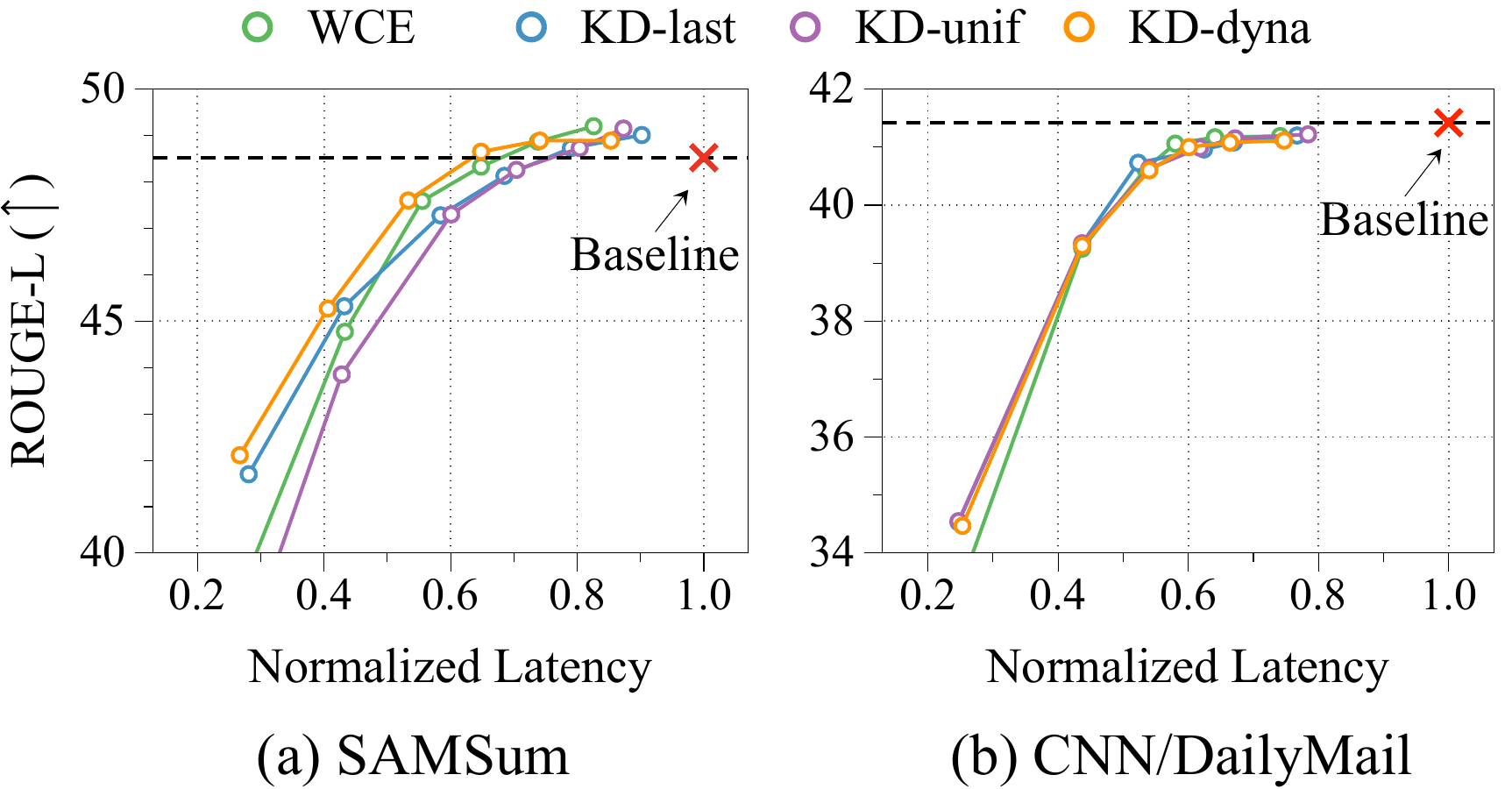}
    \caption{The trade-off between performance and normalized latency per sentence. We varied the exit thresholds in the range of $\{$0.0, 0.1, 0.3, 0.5, 0.7, 0.9$\}$. The latency values are normalized by the latency of a baseline, which is a simple fine-tuned full model.}
    \label{fig_free:distill_loss}
\end{wrapfigure}

\paragraph{Layerwise knowledge distillation.}
Given the only two exit positions in our shallow-deep module, since their performance significantly impacts the overall robustness of the early-exiting approach, we carefully design the loss function for training. In Figure~\ref{fig_free:distill_loss}, we observed the performance trends of four different loss functions as we varied the exit thresholds. While the differences are not significant, the KD-dyna loss demonstrates better trade-offs compared to a weighted average or other KD-based losses. Specifically, the lower performance of KD-unif on the SAMSum dataset suggests that dynamically determining the layer mapping can facilitate more effective knowledge transfer between the deep and shallow models. Consequently, we trained our shallow-deep module using the KD-dyna loss for all experiments, and left the exploration of additional loss functions, such as contrastive distillation losses~\cite{tian2019contrastive, bae2021self}, for future work.\looseness=-1

\begin{table*}[t]
    \centering
    \small
    \caption{(\textit{Left}) Comparison between FREE with T5-large and directly trained small-sized T5-base. We apply threshold values of FREE$^\dagger$as 0.1 for SQuAD and 0.2 for CNN/DailyMail. (\textit{Right}) Comparison between early-exiting frameworks on SAMSum with different decoding strategies.}
    \begin{minipage}[h]{0.495\textwidth}
        \centering
        \renewcommand{\arraystretch}{1.15}
        \resizebox{\columnwidth}{!}{
        \begin{tabular}{ll|cc|cc}
        \toprule[0.1em]
        &        & \multicolumn{2}{c|}{SQuAD} & \multicolumn{2}{c}{CNN/DailyMail} \\ 
         \cmidrule(l{2pt}r{2pt}){3-4}  \cmidrule(l{2pt}r{2pt}){5-6} 
        Method    & Model            & F1 & Speedup & ROUGE-L & Speedup \\ \midrule
        Full Model & T5-large   & 91.82 & $\times$ 1.00 & 41.09 & $\times$ 1.00 \\
        Full Model & T5-base   & 90.50 & $\times$ 1.86 & 40.22 & $\times$ 2.06 \\
        FREE$^\dagger$ & T5-large         & 90.95 & $\times$ 2.76 & 40.17 & $\times$ 2.07 \\
        \bottomrule[0.1em]
        \end{tabular}
        }
        \label{tab_free:small_sized}
    \end{minipage}
    \hfill
    \begin{minipage}[h]{0.495\textwidth}
        \centering
        \renewcommand{\arraystretch}{1.02}
        \resizebox{\columnwidth}{!}{
        \begin{tabular}{l|cc|cc}
        \toprule[0.1em]
        & \multicolumn{2}{c|}{\textit{top-k} ($k=50$)} & \multicolumn{2}{c}{nucleus ($p=0.92$)} \\ 
         \cmidrule(l{2pt}r{2pt}){2-3}  \cmidrule(l{2pt}r{2pt}){4-5} 
        Method                  & ROUGE-L & Speedup & ROUGE-L & Speedup \\ \midrule
        Full Model              & 44.34 & $\times$ 1.00 & 45.84 & $\times$ 1.00 \\
        CALM                    & 42.35 & $\times$ 0.78 & 44.48 & $\times$ 0.82 \\
        FREE                    & 43.58 & $\times$ 1.30 & 45.78 & $\times$ 1.31 \\
        \bottomrule[0.1em]
        \end{tabular}
        }
        \label{tab_free:sampling}
    \end{minipage}
\end{table*}

\paragraph{Comparison with small-sized models.}

We conducted a comparison between the inference speed of FREE using T5-large model and a directly trained T5-base model. To ensure a fair comparison, we manually selected the appropriate confidence threshold for FREE$^\dagger$ (without relying on an adaptive threshold estimator) to align its performance closely with that of T5-base. The results, presented in the left of Table~\ref{tab_free:small_sized}, demonstrate that our proposed method exhibited a competitive speedup in inference performance on the CNN/DailyMail dataset. Moreover, it demonstrated a superior F1 score and significantly higher speedup on the SQuAD dataset. We believe that the variance in speedup across the datasets can be attributed to the performance achievable by a directly trained smaller model, as well as a shallow model within the FREE framework. In the case of SQuAD, the T5-base model (12 layers) achieved a ROUGE-L score of 90.50, whereas a shallow model (6 layers) of our FREE framework yielded a similar score of 90.24. Our method effectively leverage these inherent benefits, thereby facilitating the inference speedup through exiting at lower layers.

\paragraph{Various decoding strategies.}

To evaluate the applicability of FREE on various decoding methods, we conducted experiments with \textit{top-k} sampling~\cite{radford2019language} and nucleus sampling~(\textit{top-p} sampling~\cite{Holtzman2020The}). \textit{top-k} sampling samples the next word from the top $k$ most probable choices, instead of aiming to decode text that maximizes likelihood. On the other hand, nucleus sampling chooses from the smallest possible set of words whose cumulative probability exceeds the probability $p$. As detailed in the right of Table~\ref{tab_free:sampling}, FREE method exhibited consistent and robust performance while achieving a larger speedup compared to CALM. These results affirm that our FREE framework can be widely applied, irrespective of the chosen decoding method.
\section{Conclusion}

We proposed FREE framework to address the challenges of conventional early-exiting frameworks for autoregressive language models. Our approach incorporates three key components: (1) shallow-deep module, (2) synchronized parallel decoding, and (3) adaptive threshold estimation. Through extensive experiments on various generation tasks, we empirically demonstrated the superior performance of FREE framework, achieving significant acceleration in latency without compromising the quality of the generated output.

\paragraph{Limitations.}

Our work addressed a fast and robust existing framework that can be efficiently utilized without concerns about performance degradation. However, our approach does have a few limitations which we discuss below: (1) Our method requires additional computational resources to fine-tune the shallow model. However, as we have demonstrated, parameter-efficient fine-tuning methods would be a promising solution to overcome this limitation. (2) While our work demonstrates robustness in the depth of the shallow model, further investigation is required to determine the appropriate depth for various language models. This aspect remains an area for additional research.

\chapter{Relaxed Recursive Transformers:\\Effective Parameter Sharing with Layer-wise LoRA}
\label{chapter:rrt}

\begin{tcolorbox}[colback=gray!10, colframe=black, arc=3mm, boxrule=1pt]
    \textbf{Publication Note:} This chapter is based on the paper accepted to the International Conference on Learning Representation (ICLR 2025)~\citep{bae2024relaxed}.
    \\
    
    \textbf{Abstract:} Large language models (LLMs) are expensive to deploy. Parameter sharing offers a possible path towards reducing their size and cost, but its effectiveness in modern LLMs remains fairly limited. In this work, we revisit ``layer tying'' as form of parameter sharing in Transformers, and introduce novel methods for converting existing LLMs into smaller ``\textbf{Recursive Transformers}'' that share parameters across layers, with minimal loss of performance. Here, our Recursive Transformers are efficiently initialized from standard pretrained Transformers, but only use a single block of unique layers that is then repeated multiple times in a loop. We further improve  performance by introducing Relaxed Recursive Transformers that add flexibility to the layer tying constraint via depth-wise low-rank adaptation (LoRA) modules, yet still preserve the compactness of the overall model. We show that our recursive models (e.g., recursive Gemma 1B) outperform both similar-sized vanilla pretrained models (such as TinyLlama 1.1B and Pythia 1B) and knowledge distillation baselines---and can even recover most of the performance of the original ``full-size'' model (e.g., Gemma 2B with no shared parameters). Finally, we propose Continuous Depth-wise Batching, a promising new inference paradigm enabled by the Recursive Transformer that becomes more effective when paired with early exiting. This approach directly mitigates a synchronization issue in early-exiting frameworks: the overhead incurred when exited tokens are forced to idle, as they cannot be batched with tokens being processed at different depths.\looseness=-1
\end{tcolorbox}

\section{Introduction}
\label{sec_rrt:introduction}

\begin{figure}[h]
    \centering
    \includegraphics[width=0.98\textwidth]{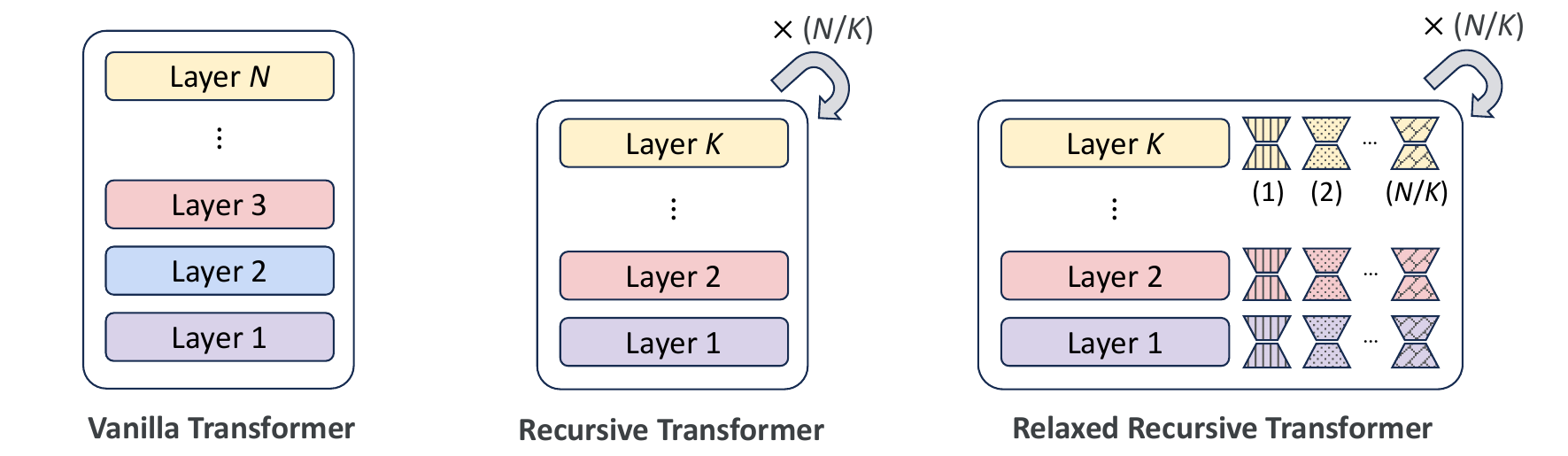}
    \caption{
    Overview of the conversion from a vanilla \textit{N}-layer Transformer to a Recursive Transformer with $N / K$ blocks of \textit{K} shared layers. The Recursive Transformer is obtained by repeating a single block of \textit{K} layers multiple times, resulting in a looped architecture. The Recursive Transformer can  also be converted into a Relaxed Recursive Transformer by adding  layer-specific LoRA modules. This  preserves many of the advantages of weight sharing, but also allows for better performance.
    }
    \label{fig_rrt:overview}
\end{figure}

Efficient deployment of large language models (LLMs) demands a balance between performance and resources~\citep{DBLP:journals/corr/abs-2404-02258, DBLP:conf/icml/LeviathanKM23, DBLP:journals/corr/abs-2408-00118, DBLP:journals/tmlr/Wan0LA0LQYZZC024, DBLP:journals/corr/abs-2404-14294}. While larger models with more parameters consistently demonstrate superior performance~\citep{rosenfeld2020a,rae2021scaling,hoffmann2022trainingcomputeoptimallargelanguage}, their substantial memory and computational demands are expensive~\citep{MLSYS2023_c4be71ab}. Parameter sharing approaches~\cite[e.g.][]{DBLP:conf/iclr/DehghaniGVUK19, DBLP:conf/aaai/XiaHTTHQ19, DBLP:conf/iclr/LanCGGSS20, DBLP:conf/sustainlp/TakaseK23}, wherein weights are reused across model layers, can lower these costs by reducing memory footprint, and thereby allow for the use of fewer (or lower-grade) accelerators, or larger batch sizes for better throughput. While parameter sharing has shown encouraging capabilities in previous work~\citep{DBLP:conf/iclr/LanCGGSS20,DBLP:conf/icml/GiannouRS0LP23}, its application to modern LLMs has yielded limited reported success.\looseness=-1

In this work, we revisit parameter sharing for LLMs, and propose novel methodologies to \emph{convert} existing, unshared models into smaller, and more efficient,  Recursive Transformers. These models use a single block of unique layers that are recursively reused across multiple loops, yet still achieve impressive performance relative to their reduced size. To mitigate the potential performance degradation associated with parameter sharing, we first initialize the shared block of layers based on the original model's pre-trained parameters, and then finetune the resulting recursive model for a limited number of ``uptraining'' steps. Importantly, we show that our initialization strategies allow us to achieve strong performance with minimal training time. This is aligned with observations that model compression techniques such as layer skipping~\citep{DBLP:conf/acl/Zhang00S0CM24, DBLP:journals/corr/abs-2311-15436, DBLP:conf/iclr/FanGJ20, DBLP:conf/acl/ElhoushiSLHWL0A24}, pruning~\citep{DBLP:conf/iclr/FrankleC19, DBLP:conf/cvpr/RamanujanWKFR20} or nesting~\citep{devvrit2023matformernestedtransformerelastic} can preserve surprisingly high performance---further motivating our approach of compressing models to more compact yet performant architectures (here, repeated layers with low-rank adapters).

As depicted in Figure\,\ref{fig_rrt:overview}, we further propose the Relaxed Recursive Transformer, an extension of the Recursive Transformer in which the  weight tying across repeated layer blocks is slightly relaxed through the incorporation of multiple layer-specific, low-rank adaptation\,(LoRA) modules~\citep{DBLP:conf/iclr/HuSWALWWC22}. Despite its simplicity, this strategy offers several non-trivial advantages. First, it allows for low-rank deltas between shared layers, while only adding minimal  overhead. Second, the rank of the LoRA matrices can be adjusted to control the degree of relaxation, which directly influences model capacity. Furthermore, since the relaxed model has the same overall shape as the original Transformer, we can efficiently initialize LoRA modules via truncated Singular Value Decomposition~\citep{hansen1987truncated} on the residual matrices between the original layer weights and the shared layer weights. Hence, the rank values serve as a pivotal hyperparameter, enabling the Relaxed Recursive Transformer to seamlessly transition between the two extremes of the vanilla and Recursive Transformer architectures.

While the primary focus of this paper lies in how to formulate and train Recursive Transformers, we also highlight their potential to achieve significant throughput gains via a new batched inference paradigm, Continuous Depth-wise Batching, that their recursive nature enables. Prior work introduced continuous sequence-wise batching~\citep{DBLP:conf/osdi/YuJKKC22, DBLP:conf/sosp/KwonLZ0ZY0ZS23}, which leverages the fact that the computation performed to compute a new token is functionally the same (and uses the same model parameters) regardless of the token position within the sequence. This allows new requests to be continuously scheduled when slots within a batch become available. For example, when one response is completed, the start of the next response to be formed can immediately take the finished response's place in the batch, without waiting for the rest of the batch responses that might be longer. In our Recursive Transformer, parameter sharing occurs not only across different timesteps, but also across different depths (loop iterations). This enables an extra dimension of dynamic grouping: jointly computing different iterations of the looped layer blocks per individual responses within the same batch.\looseness=-1

\clearpage
Our key contributions are as follows:
\begin{itemize}[leftmargin=*]
\item 
We introduce a framework for initializing and training Relaxed Recursive Transformers and demonstrate strong performance compared to non-recursive models of comparable size. For example, when we uptrained a recursive Gemma 1B model converted from a pretrained Gemma 2B~\citep{team2024gemma}, we observed up to 13.5 absolute accuracy improvement (22\% error reduction) on few-shot tasks compared to a non-recursive Gemma 1B model (pretrained from scratch). Furthermore, we show that by incorporating knowledge distillation~\citep{DBLP:journals/corr/HintonVD15, DBLP:conf/emnlp/KimR16}, our recursive Gemma model, uptrained on 60 billion tokens, achieves performance on par with the full-size Gemma model trained on a massive 3 trillion token corpus (see \textsection\ref{sec_rrt:main_results} for  details).\looseness=-1

\item
Based on our Relaxed Recursive Transformer, we also evaluate a key use case for continuous depth-wise batching with early-exiting~\citep{DBLP:conf/emnlp/BaeKSY23, DBLP:conf/nips/SchusterFG0B0TM22, DBLP:conf/iclr/ElbayadGGA20, DBLP:journals/corr/Graves16}, which opportunistically makes predictions for samples with high confidence at earlier stages. From our simulation, Early Exits reveal a substantial throughput improvement of up to 2-3$\times$ compared to a vanilla Transformer with the same architecture. Notably, the recursive Gemma model, which outperforms the vanilla Pythia model, can theoretically achieve a nearly 4$\times$ increase in throughput (see \textsection\ref{exp_rrt:hypothetical_generation_speedup} for details).
\end{itemize}

\section{Related Work}

Cross-layer parameter sharing has proven to be an effective method for achieving parameter efficiency in deep learning models such as RNNs~\citep{DBLP:journals/corr/abs-1808-03314, graves2016adaptive}, CNNs~\citep{DBLP:journals/corr/EigenRFL13, DBLP:conf/iclr/SavareseM19, DBLP:conf/cvpr/GuoYWLQY19, DBLP:conf/eccv/ShenLX22}, and the popular Transformer architecture. The Universal Transformer~\citep{DBLP:conf/iclr/DehghaniGVUK19}, a recurrent self-attentive model, demonstrated superior performance to non-recursive counterparts with significantly fewer parameters. This cross-layer parameter sharing approach has subsequently been explored in various tasks, including language understanding ~\citep{DBLP:conf/iclr/LanCGGSS20}, language modeling~\citep{DBLP:conf/nips/BaiKK19, mohtashami2023cotformer, DBLP:conf/icml/Liu0ILTFXCSKLC24, csordas2024moeut, DBLP:journals/corr/abs-2405-16712}, and machine translation~\citep{DBLP:conf/aaai/DabreF19, milbauer-etal-2023-lait, DBLP:conf/aaai/XiaHTTHQ19, DBLP:conf/sustainlp/TakaseK23, DBLP:conf/emnlp/GeCW22}. These methods often claim to achieve comparable performance with more compact models and increased computational speed, while also setting the ground for effective adaptive compute solutions~\citep{DBLP:conf/iclr/DehghaniGVUK19,graves2016adaptive,schuster-etal-2021-consistent}.

Concurrently, there has been growing interest in exploiting recurrent architectures for algorithmic or logical reasoning tasks~\citep{saunshi2024inductive}. Prior research~\citep{DBLP:conf/nips/SchwarzschildBG21, mcleish2022re} has shown that recurrent networks can extrapolate reasoning strategies learned on simple problems to harder, larger problems through additional recurrences during inference. The looped Transformer structure has also been employed to emulate basic computing blocks for program simulation~\citep{DBLP:conf/icml/GiannouRS0LP23}, to learn iterative algorithms for data-fitting problems~\citep{DBLP:conf/iclr/Yang0NP24}, to achieve length generalization in algorithmic tasks~\citep{fan2024looped}, and promising theoretical potential for few-shot learning~\citep{gatmiry2024looped}.

However, previous work has predominantly focused on relatively small Transformer models, trained from scratch without leveraging pretrained model weights. Our work distinguishes itself by investigating parameter sharing in the context of LLMs and proposing effective initialization strategies that leverage the knowledge embedded within existing LLMs. 
To the best of our knowledge, we are the first to propose a generalized framework for parameter-shared models, enabling relaxation in weight tying constraints through layer-specific modules.

In this paper, we also discuss how Recursive Transformers can be well suited for early-exiting techniques to accelerate decoding in LLMs. The inherent recursive structure readily enables early-exiting for individual responses within a large serving batch, which is often a practical limitation of such techniques. Vanilla Transformers encounter a synchronization issue with early-exiting, where the model must forward all layers if even a single token in a batch requires full processing (exited tokens must wait for them). Several approaches attempt to exploit this idle time by computing missing KV caches for exited tokens in later layers, which are essential for subsequent sequence generation. These techniques include state propagation~\citep{DBLP:conf/nips/SchusterFG0B0TM22, DBLP:conf/iclr/ElbayadGGA20}, SkipDecode~\citep{del2023skipdecode}, and parallel decoding (which can be combined with Speculative Decoding)~\citep{DBLP:conf/emnlp/BaeKSY23, DBLP:conf/acl/ElhoushiSLHWL0A24, liu2024kangaroo, DBLP:conf/icml/ChenPLDZ24, DBLP:conf/naacl/TangZLAMM24}. Nevertheless, the heterogeneous parameters across varying model depths still hinder the efficient progression of exited tokens to subsequent sequences. In contrast, our Recursive Transformers enable parallel computation for tokens at different depths and sequences (in a continuous depth-wise batching paradigm)---also allow for parallel computation of missing KV caches with minimal overhead during the memory-bounded decoding phase.

\section{Effective Model Compression with Recursive Patterns}
\label{sec_rrt:methods}

In this section, we present the main details of our method for converting a vanilla Transformer model into a parameter-shared model that outperforms models of equivalent size. We first provide a short overview of the Transformer architecture ($\S$\ref{sec_rrt:methods:background}). Then, we introduce the Recursive Transformer and present effective techniques to initialize its looped layers by leveraging the weights of the original pretrained model ($\S$\ref{sec_rrt:methods:recursive}). In $\S$\ref{sec_rrt:methods:relaxed}, we relax the parameter-sharing constraint in the model design, and add a limited set of layer-specific parameters to further improve the model's accuracy while maintaining compact representations. Finally, we show how, beyond reduced memory, Recursive Transformers readily support further throughput optimizations via a novel inference paradigm ($\S$\ref{sec_rrt:methods:early_exit}).

\subsection{Basic Transformer Architecture}
\label{sec_rrt:methods:background}

Large language models~\citep{DBLP:journals/corr/abs-2408-00118, reid2024gemini, DBLP:journals/corr/abs-2303-08774, grattafiori2024llama} typically leverage the Transformer architecture \citep{vaswani2017attention}. A Transformer consists of \textit{L} layers, where the hidden states at each time step \textit{t} are computed by running through the series of layers:\looseness=-1
\begin{align}
    \mathbf{h}^{\ell}_{t} = f(\mathbf{h}^{\ell - 1}_{t}; \,\Phi_\ell), \,\,\ell \in [1, L],
    \label{eq_rrt:full_size_model}
\end{align}
with $\mathbf{h}^{0}_{t}$ representing the embedding of the token $y_{t-1}$ from the previous time step, and $\Phi_\ell$ denoting the trainable parameters of the $\ell$-th layer.

The Transformer block consists of two core components: a multi-head attention (MHA) mechanism and a feed-forward network (FFN). MHA utilizes multiple attention heads to capture diverse relationships within the input sequence. The computation within each attention head is formulated as:
\begin{align}
\text{Attention}(\mathbf{Q}, \mathbf{K}, \mathbf{V}) = \text{softmax} \left( \frac{\mathbf{Q}\mathbf{K}^T}{\sqrt{d_k}} \right) \mathbf{V},
\end{align}
where $\mathbf{Q}$, $\mathbf{K}$, and $\mathbf{V}$ are linear projections of the input, parameterized by learned weight matrices $\mathbf{W}_\ell^Q$, $\mathbf{W}_\ell^K$, and $\mathbf{W}_\ell^V$, respectively. The outputs from each head of the multi-head attention are concatenated and then projected back to the original hidden size using a learned weight matrix $\mathbf{W}_\ell^{out}$.

While the FFN structure typically consists of two linear transformations, in the Gemma model, it deviates from this standard architecture as follows:
\begin{align}
\text{FFN}(\mathbf{x}) = \mathbf{W}_\ell^{down} (\text{GELU}(\mathbf{x} \mathbf{W}_\ell^{gate}) * \mathbf{x} \mathbf{W}_\ell^{up})
\end{align}
with three learned linear weight matrices and a GeGLU activation~\citep{DBLP:journals/corr/abs-2002-05202}.

\begin{figure}[t!]
    \centering
    \begin{subfigure}[t]{0.322\textwidth}
        \includegraphics[width=\textwidth]{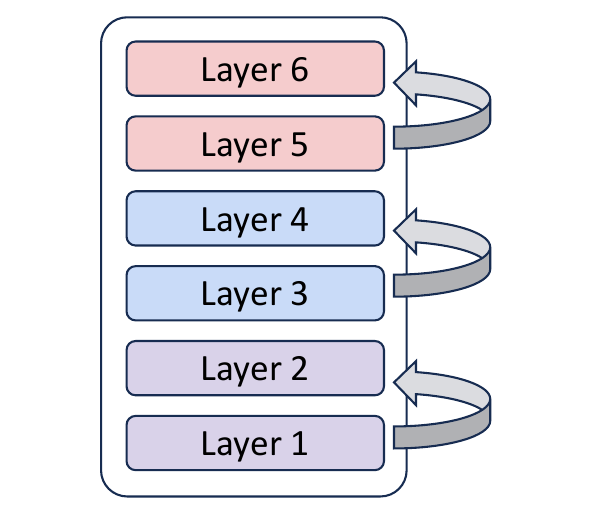}
        \subcaption{SEQUENCE}
    \end{subfigure}
    \centering
    \begin{subfigure}[t]{0.322\textwidth}
        \includegraphics[width=\textwidth]{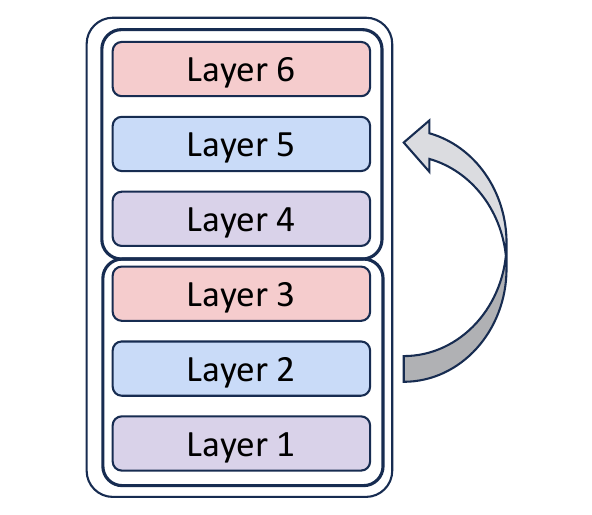}
        \subcaption{CYCLE}
    \end{subfigure}
    \centering
    \begin{subfigure}[t]{0.322\textwidth}
        \includegraphics[width=\textwidth]{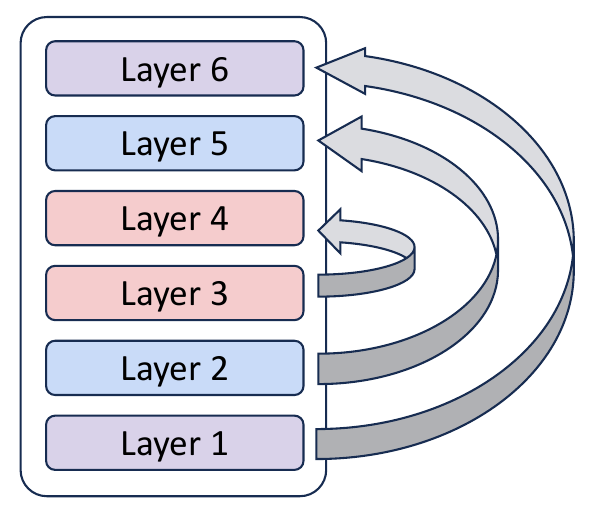}
        \subcaption{CYCLE\,(REV)}
    \end{subfigure}
    \caption{
    Three strategies for parameter sharing \citep{DBLP:conf/sustainlp/TakaseK23}. The examples utilize models with six layers, where identical colors represent shared weights. 
    }
    \label{fig_rrt:sharing_strategy}
\end{figure}

\subsection{Recursive Transformer: Looped Layer Tying}
\label{sec_rrt:methods:recursive}

In this work, we revisit parameter sharing in the context of LLMs and propose the Recursive Transformer architecture. Takase and Kiyono~(2023)~\cite{DBLP:conf/sustainlp/TakaseK23} discuss three strategies for partial layer tying in Transformer models, as depicted in Figure\,\ref{fig_rrt:sharing_strategy}. The SEQUENE strategy is the simplest, assigning the same parameters to consecutive layers. The CYCLE strategy repeatedly stacks a single block of unique layers to achieve the desired depth. Meanwhile, the CYCLE (REV) strategy stacks the lower layers in reverse order for the remaining layers.

In the comparative analysis of SEQUENCE and CYCLE strategies~\cite{DBLP:conf/icml/Liu0ILTFXCSKLC24}, CYCLE demonstrated marginally superior zero-shot performance. Although the SEQUENCE approach, which caches shared weights (the capacity of SRAM is typically sufficient to hold a single transformer block) and computes them iteratively, has the potential to mitigate the weight transfer bottleneck between SRAM and DRAM, we prioritized compatibility with early-exiting.

Consequently, we specifically adopt the CYCLE strategy, which enables continuous depth-wise batching and thereby maximizes the throughput of Recursive Transformers---wherein a single block of unique layers is recursively reused. This inherent design aligns seamlessly with early-exiting mechanisms, potentially offering substantial speedup. The model's hidden states are computed as:
\begin{align}
    \textbf{h}_t^\ell = f(\textbf{h}_t^{\ell-1}; \,\Phi'_{((\ell-1) \bmod L/B) + 1}), \,\,\ell \in [1, L],
    \label{eq_rrt:recursive_transformer}
\end{align}
where the parameter-shared model is parameterized by $\Phi'$, and \textit{B} denotes the number of looping blocks (we restrict \textit{B} to be a factor of \textit{L}).
For example, Gemma 2B~\cite{team2024gemma} with 18 layers can be converted to a recursive variant with 2 blocks by storing weights for only the first 9 layers. The forward pass will loop twice through these 9 layers.
We tie all trainable parameters, including the weights of the linear layers in the Transformer blocks and the weights of the RMSNorm~\cite{DBLP:conf/nips/ZhangS19a}.\looseness=-1

\begin{figure}[t!]
    \centering
        \includegraphics[width=\textwidth]{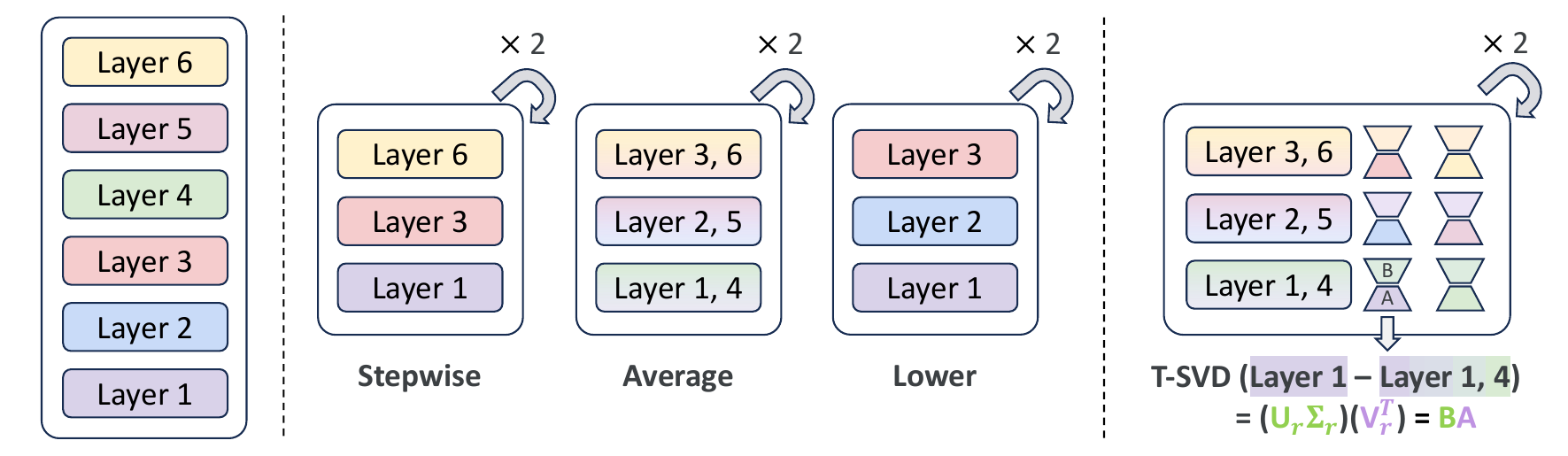}
    \caption{
    (\textit{Left}) An example of unshared, full-size model with 6 layers. (\textit{Middle}) Three proposed methodologies for initializing looped layers in a Recursive Transformer. Each layer number indicates the source layer in the full-size model used for initialization. (\textit{Right}) Example of a Relaxed Recursive Transformer initialized by SVD method. Here, looped layers are initialized using the Average method.
    }
    \label{fig_rrt:looping_init_overview}
\end{figure}

\paragraph{Initialization techniques for looped layers.}
To mitigate the potential performance drop associated with reduced capacity in parameter-shared models, we propose several novel initialization methodologies to facilitate effective knowledge transfer from unshared, pretrained models to Recursive Transformers. Figure~\ref{fig_rrt:looping_init_overview} illustrates three such techniques. The Stepwise method selects intermediate layers at specific intervals while keeping the first and last layer fixed. This is motivated by prior work~\citep{DBLP:conf/icml/LiuWDZY0S0TRC23, DBLP:conf/acl/Zhang00S0CM24, DBLP:journals/corr/abs-2311-15436, DBLP:conf/iclr/FanGJ20} showing minimal impact on generation quality when skipping a few layers in LLMs. The Average method initializes the shared weights among tied layers by averaging their weight matrices, whereas the Lower method directly uses weights from the first \textit{K} layers of the unshared model. We conducted a brief uptraining on 15 billion tokens to investigate the extent of performance recovery in these initialized models ($\S$\ref{sec_rrt:exp_initialization}) and found the Stepwise approach to perform best for Recursive Transformers. However, we found the Average method to perform best for Relaxed Recursive Transformers, discussed next.

\subsection{Relaxed Recursive Transformer: Multi-LoRA Layers}
\label{sec_rrt:methods:relaxed}

While full layer-tying is effective for compressing the model's size while maintaining strong capabilities, it has two noticeable limitations: (1) the set of possible model sizes is limited to scaling the number of layers, and (2) each model layer ends up having to serve multiple roles associated with different depths of the model. To address this, we introduce Relaxed Recursive Transformers in which we incorporate independent adapter modules~\citep{DBLP:conf/iclr/HuSWALWWC22, DBLP:conf/icml/HoulsbyGJMLGAG19} for each layer, relaxing the strict parameter sharing. While we experiment with various approaches like layer-specific prefixes~\citep{DBLP:journals/corr/abs-2110-07602}, we find low-rank adaptation (LoRA) modules~\citep{DBLP:conf/iclr/HuSWALWWC22} to efficiently capture the subtle variations between tied layers. Specifically, we modify Eq.\,\ref{eq_rrt:recursive_transformer} to:
\begin{align}
    \textbf{h}_t^\ell = f(\textbf{h}_t^{\ell-1}; \,\Phi'_{((\ell-1) \bmod L/B) + 1}, \Delta \Phi'_\ell), \,\,\ell \in [1, L],
    \label{eq_rrt:relaxed_recursive_transformer}
\end{align}
where $\Delta \Phi'$ is the (small) set of parameters for the LoRA modules. 

In this relaxed model, each looped layer is augmented with multiple LoRA modules. For example, a recursive model with two loop iterations has a single block of shared layers, and two different LoRA modules are attached to each layer within this block. The first and second LoRA modules per layer are used during the first and second loop iterations, respectively. 
Functionally, these LoRA modules introduce low-rank deltas to all of the shared, linear weight matrices. More concretely, for a base transformation $\textbf{h} = \textbf{W}'\textbf{x}$, our modified forward pass yields $\textbf{h} = \textbf{W}'\textbf{x} + \Delta \textbf{W}' \textbf{x} = \textbf{W}'\textbf{x} + \textbf{B}\textbf{A}\textbf{x}$, where $ \textbf {A}\in \mathbb{R}^{(r \times k)}$ and $\textbf {B} \in \mathbb{R}^{(d \times r)}$ denote the weight matrices of LoRA with rank \textit{r}.

\paragraph{LoRA initialization via truncated SVD.}

Unlike typical LoRA finetuning setups that train only the LoRA parameters, here we train all model parameters to let the shared parameters learn an optimal centroid for all of the layer depths that they support. Therefore, instead of following standard zero initialization for adaptation to the frozen base model, we propose novel initialization methods, especially designed for Relaxed Recursive Transformers. To effectively match the performance of the original full-size model after initializing the tied weights as described in $\S$\ref{sec_rrt:methods:recursive}, we aim for the sum of the tied weights ($\Phi'$) and LoRA weights ($\Delta \Phi'$) to approximately recover the full-size model's weights ($\Phi$). 
We exploit truncated Singular Value Decomposition (SVD)~\citep{hansen1987truncated} on residual matrices between original weights and tied weights:
\begin{align}
    \mathbf{U}_r^\ell, \mathbf{\Sigma}_r^\ell, \mathbf{V}_r^\ell = \text{Truncated SVD}(\mathbf{W}_\ell - \mathbf{W'}_{((\ell-1) \bmod L/B) + 1}; \,\,r), \,\,\ell \in [1, L],
    \label{eq_rrt:truncated_svd}
\end{align}
where outputs retain the first \textit{r} columns corresponding to the \textit{r} largest singular values. \textit{\textbf{W}} denotes the weight matrices of the full-size model, and $\text{\textit{\textbf{W}}}'$ denotes those of the Recursive Transformer.
We initialize the LoRA's weights with principal components in Eq.\,\ref{eq_rrt:truncated_svd}: \textit{\textbf{B}} as the product of $\text{\textit{\textbf{U}}}_r$ and $\mathbf{\Sigma}_r$, and \textit{\textbf{A}} as the transpose of the right singular vectors $\text{\textit{\textbf{V}}}_r$ (see the right of Figure\,\ref{fig_rrt:looping_init_overview}).

\newtheorem*{remark*}{Remark}
\begin{remark*}
    By initializing LoRA weights through the proposed truncated SVD methodology, the rank of the LoRA modules serves as a pivotal hyperparameter, enabling the Relaxed Recursive Transformer to seamlessly transition between the two extremes of the vanilla and Recursive Transformer architectures. 
\end{remark*}

With sufficiently large ranks, our Relaxed Recursive Transformer (Eq.\,\ref{eq_rrt:relaxed_recursive_transformer})  approximates the full-size vanilla model (Eq.\,\ref{eq_rrt:full_size_model}):
\begin{align}
    \mathbf{W}\textbf{x} \approx \mathbf{W}'\textbf{x} + (\mathbf{U}_r \mathbf{\Sigma}_r)(\mathbf{V}^\top_r)\textbf{x} = \mathbf{W}'\textbf{x} + \mathbf{B}\mathbf{A}\textbf{x} = \mathbf{W}'\textbf{x} + \Delta \mathbf{W}'\textbf{x},
    \label{eq_rrt:relaxed_to_vanilla}
\end{align}
Meanwhile, setting the rank to zero reduces the model to a Recursive Transformer, as the LoRA modules contribute no additional parameters, highlighting the flexibility of this relaxation approach.

\begin{figure}[t!]
    \centering
        \includegraphics[width=\textwidth]{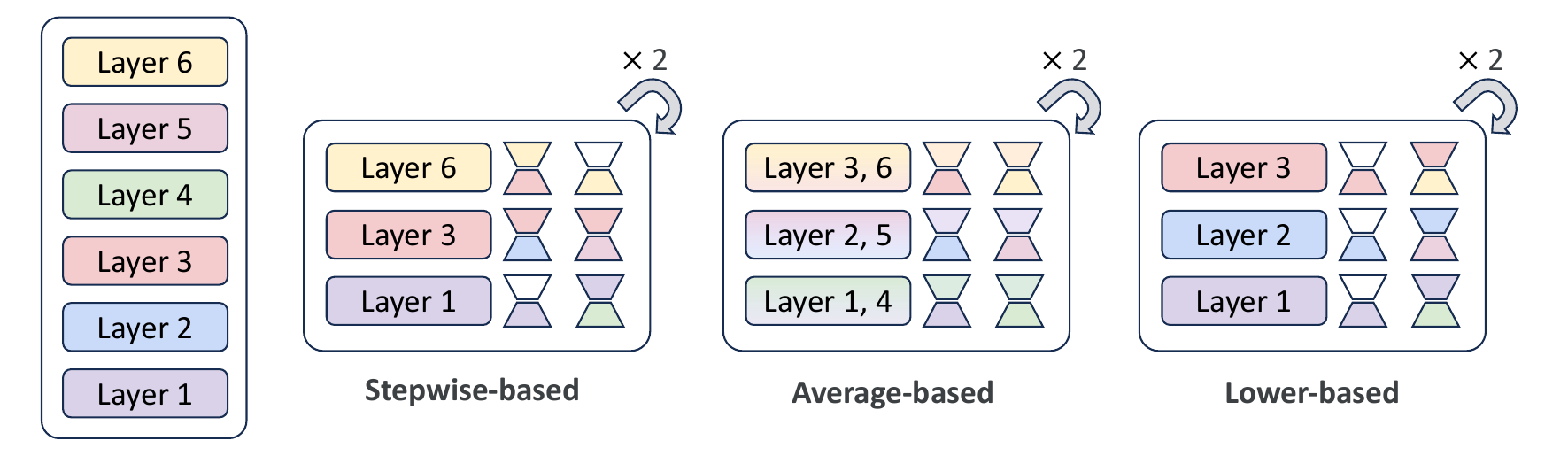}
    \caption{
    We visualize LoRA modules to show which residual matrices they target for initialization under three different looping initialization methods, assuming a full-size model with six layers and two looping blocks. For ease of understanding, \textit{{\textbf{A}}} matrices are colored according to the full-size model weights at the corresponding depth, while {\textit{\textbf{B}}} matrices are colored based on the looped layer weights. White \textit{{\textbf{B}}} matrices indicate cases where the full-size model and looped model weights are identical, resulting in standard zero initialization. 
    }
    \label{fig_rrt:lora_init_overview}
    \vspace{3pt}
\end{figure}

Figure\,\ref{fig_rrt:lora_init_overview} illustrates an overview of how the LoRA module is initialized under three different initialization techniques (Stepwise, Average, and Lower) for looped layers. One crucial point is that if the initialized looped layer's weights match those of the original pretrained model, its corresponding LoRA module undergoes standard zero initialization: random Gaussian for matrix \textit{\textbf{A}} and zero for \textit{\textbf{B}}. For example, with the Stepwise method, the first loop's LoRA module receives standard zero initialization, while the second loop's LoRA is initialized using our proposed initialization.

\vspace{6pt}

\subsection{Continuous Depth-wise Batching and Early-Exiting}
\label{sec_rrt:methods:early_exit}

\vspace{3pt}

In real-world deployments, user requests arrive sequentially and asynchronously. Recent research has introduced continuous sequence-wise batching~\citep{DBLP:conf/osdi/YuJKKC22, DBLP:conf/sosp/KwonLZ0ZY0ZS23}, a serving strategy that allows new requests to immediately replace completed (terminated) sequence within a batch. This approach exploits the fact that the computation performed for a new token is functionally the same and utilize the same model parameters. By continuously scheduling requests in this manner, models can operate at their maximum batch capacity, thereby enhancing serving efficiency.

\begin{figure}[t!]
    \centering
    \begin{subfigure}[t]{0.35\textwidth}
        \includegraphics[width=\textwidth]{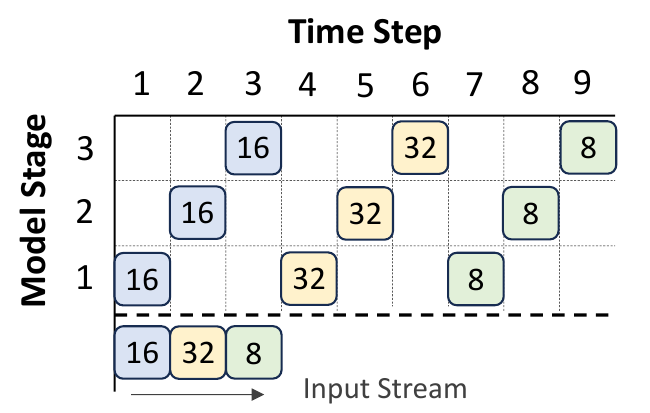}
        \subcaption{Vanilla Batching}
    \end{subfigure}
    \centering
    \begin{subfigure}[t]{0.30\textwidth}
        \includegraphics[width=\textwidth]{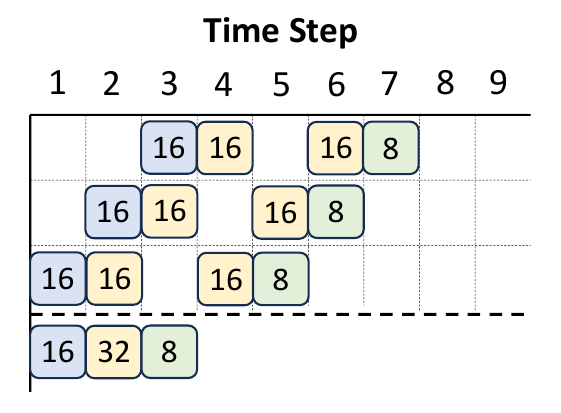}
        \subcaption{Depth-wise Batching}
    \end{subfigure}
    \centering
    \begin{subfigure}[t]{0.30\textwidth}
        \includegraphics[width=\textwidth]{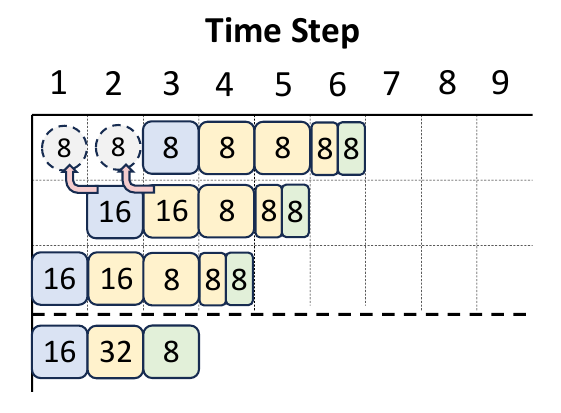}
        \subcaption{With Early-exiting}
        \label{fig_rrt:cdb_early_exiting}
    \end{subfigure}
    \caption{
    An illustrative example of a continuous depth-wise batching strategy together with early-exiting. We assume a maximum batch size of $32$, three model ``stages'' (e.g., layer blocks), and a stream of batched inputs that arrive sequentially in time. In {(a)},  all three model stages must complete for the first (non-maximal) batch of $16$ before the second batch of $32$ examples that arrives next can be started. In {(b)}, however, half of second batch of $32$ examples can share computation with the first batch of $16$ that is still finishing. Finally, {(c)} demonstrates a situation where some examples within each batch can early-exit after stage 2; their vacant slots in the batch are then immediately filled.
    }
    \label{fig_rrt:continuous_depthwise_batching}
\end{figure}

The repetitive structure of Recursive Transformers allows for the same function to be applied not just across sequences, but also across depths (loop iterations). This introduces a new dimension for continuous batching, which we call Continuous Depth-wise Batching. This technique enables the simultaneous computation of different iterations of the looped layer block for different samples (See Figure\,\ref{fig_rrt:continuous_depthwise_batching} for an example with a single forward pass; this easily extends to multiple decode iterations per request.) With a maximum batch size of 32, a standard Transformer must wait for all model stages to complete before processing new requests. In contrast, our Recursive Transformer, because it shares layer functions across all stages, can immediately schedule new incoming requests at timestep 2, maximizing batch size utilization. This strategy can yield a substantial speedup in generation and reduce the time to first token~\citep{DBLP:journals/corr/abs-2407-14057, DBLP:journals/corr/abs-2312-15234} through faster scheduling.

Throughput improvements from depth-wise batching are further amplified when combined with early-exiting~\citep{DBLP:conf/emnlp/BaeKSY23, DBLP:conf/nips/SchusterFG0B0TM22, DBLP:conf/iclr/ElbayadGGA20}. As depicted in Figure\,\ref{fig_rrt:cdb_early_exiting}, once some samples exit after certain looping iterations, queued requests can then be immediately scheduled. While Recursive Transformers leverage the speedup from early-exiting, they also inherently address a key challenge of batched inference in early-exiting approaches: the synchronization issue when serving large batches, as early-exited tokens might wait for others to complete processing through the entire model. We demonstrate that Recursive Transformers, equipped with this dynamic sample scheduling at various depths, can theoretically allow up to 2-3$\times$ speedup on evaluated LLMs. 
\section{Experiments}
\label{sec_rrt:experiments}

\subsection{Experimental Setup}
\label{sec_rrt:exp_setup}

We evaluate our method on three popular pretrained LLMs: Gemma 2B\,\citep{team2024gemma}, TinyLlama 1.1B\,\citep{zhang2024tinyllama}, and Pythia 1B\,\citep{DBLP:conf/icml/BidermanSABOHKP23}. Table\,\ref{tab_rrt:model_arch} summarizes each model's architecture and pretraining recipes, and their few-shot performance is summarized in Table\,\ref{tab_rrt:pretrained_model_performance}. Unless stated otherwise, the number of looping blocks\,(\textit{B}) is set to 2 for all experiments. The results for Gemma with $\text{\textit{B}}=3$ is provided in the appendix.

\begin{table}[t!]
    \small
    \centering
    \resizebox{\textwidth}{!}{
    \setlength{\tabcolsep}{5pt}
    \begin{tabular}{l|cccccccc|ccc}
    \toprule
      &  \multicolumn{8}{c|}{\textbf{Model Architecture}} & \multicolumn{3}{c}{\textbf{Pretraining}} \\
    \cmidrule(l{2pt}r{2pt}){2-9} \cmidrule(l{2pt}r{2pt}){10-12}
     \textbf{Models} & N-emb & Emb & $N_L$ & $d_{model}$ & $N_{head}$ & $N_{KV}$ & $d_{head}$ & Vocab & Dataset & $N_{tok}$ & $L_{ctx}$  \\
    \midrule
    Gemma\,2B & 1.98B & 0.52B & 18 & 2048 & 8 & 1 & 256 & 256K & Unreleased & 3T & 8K \\[3pt]
    \multirow{2}{*}{TinyLlama\,1.1B} & \multirow{2}{*}{0.97B} & \multirow{2}{*}{0.13B} & \multirow{2}{*}{22} & \multirow{2}{*}{2048} & \multirow{2}{*}{32} & \multirow{2}{*}{4} & \multirow{2}{*}{64} & \multirow{2}{*}{32K} & SlimPajama\,+ & 73B$^\ast$ & \multirow{2}{*}{2K}  \\
     &  &  &  &  &  &  &  &  & Starcoderdata  & 32B  &  \\[3pt]
    Pythia\,1B & 0.81B & 0.21B & 16 & 2048 & 8 & 8 & 256 & 50K & Pile & 300B & 2K \\
    \bottomrule
    \end{tabular}
    }
    \caption{
    Key parameters and pretraining details of three models. The sizes of each model refer to the number of embedding parameters (embedding matrices and classifier heads), and all other non-embedding parameters. Gemma and TinyLlama utilize Multi-Query~\citep{DBLP:journals/corr/abs-1911-02150} and Grouped-Query~\citep{DBLP:conf/emnlp/AinslieLJZLS23} attention mechanisms, which leads to a reduced number of key-value heads. $^\ast$We take an early TinyLlama checkpoint to study recursive conversions on top of an under-trained model on SlimPajama. The vanilla performance with longer pretraining is reported in Table~\ref{tab_rrt:pretrained_model_performance}.
    }
    \label{tab_rrt:model_arch}
\end{table}

\begin{table}[t]
    \small
    \centering
    \resizebox{\textwidth}{!}{
    \setlength{\tabcolsep}{5pt}
    \begin{tabular}{l|rcr|ccccccc|c}
    \toprule
      &  &  & & \multicolumn{8}{c}{\textbf{Few-shot Accuracy\,$\uparrow$}} \\
    \cmidrule(l{2pt}r{2pt}){5-12} 
     \textbf{Models} & N-emb & Dataset & $N_{token}$ & LD & HS & PQ & WG & ARC-e & ARC-c & OB & Avg  \\
    \midrule
    Gemma\,2B & 1.99B & Unreleased & 3T & {63.13} & {71.38} & {78.13} & {65.04} & {72.26} & {41.89} & 40.20 & {61.72}  \\
     \midrule
    \multirow{5}{*}{TinyLlama\,1.1B} & \multirow{5}{*}{0.97B} & \multirow{4}{*}{SlimPajama\,+} & 105B & 43.26 & 42.23 & 66.81 & 53.35 & 44.74 & 23.21 & 29.20 & 43.26  \\
    &  & \multirow{4}{*}{Starcoderdata} & 503B & 48.92 & 49.56 & 69.42 & 55.80 & 48.32 & 26.54 & 31.40 & 47.14  \\
    & & & 1T & 53.00 & 52.52 & 69.91 & 55.96 & 52.36 & 27.82 & 33.40 & 49.28  \\
    & & & 2T & 53.33 & 54.63 & 70.67 & 56.83 & 54.67 & 28.07 & 33.40 & 50.23  \\
    & & & 3T & 58.82 & 59.20 & 73.29 & 59.12 & 55.35 & 30.12 & 36.00 & 53.13  \\
     \midrule
    Pythia\,1B & 0.81B & Pile & 300B &  57.52 & 49.10 & 70.40 & 52.80 & {51.89} & 26.71 & {33.40} & {48.83}  \\
    \bottomrule
    \end{tabular}
    }
    \caption{
    Few-shot performance of three pretrained models. Few-shot accuracy is measured on the LAMBADA, HellaSwag, PIQA, WinoGrande, ARC-easy, ARC-challenge, and OpenBookQA benchmarks. We evaluated intermediate checkpoints up to the fully trained checkpoint for TinyLlama 1.1B. Among these, we utilized the 105B intermediate checkpoint to study an under-trained model.
    }
    \label{tab_rrt:pretrained_model_performance}
\end{table}

\paragraph{Uptraining setting.}
To convert vanilla Transformers into Recursive Transformers, we conducted further uptraining on either 15 billion or 60 billion tokens from the SlimPajama dataset\,\citep{cerebras2023slimpajama}. SlimPajama is an open-source dataset designed for training large language models, which is created by cleaning and deduplicating the RedPajama dataset\,\citep{together2023redpajama}. The source data primarily consists of web-crawled data, along with data from Github, books, Arxiv, Wikipedia, and StackExchange. We employed the HuggingFace framework\,\citep{wolf2020transformers} and enhanced memory efficiency through the Zero Redundancy Optimizer (ZeRO)\,\citep{rajbhandari2020zero} from the DeepSpeed\,\citep{rasley2020deepspeed}, along with mixed precision training. The context length was set to 2048, and the batch size was approximately 2 million tokens. We used the AdamW optimizer\,\citep{DBLP:conf/iclr/LoshchilovH19} with a learning rate of 2e-4, utilizing a cosine annealing learning rate scheduler~\citep{DBLP:conf/iclr/LoshchilovH17}. Additionally, we set warmup steps to 200 and 800 for 15 billion and 60 billion token training. Eight H100 GPUs were used for the training.\looseness=-1

\paragraph{Early-exit training setting.}
Similar to the uptraining process, we used the SlimPajama dataset to enable models to predict next tokens at intermediate loops. Models with two looping blocks underwent additional training on a total of two exit points, whereas models with three blocks were trained on three exit points. We explored various strategies, but by default, we continued training on an additional 15 billion tokens (SlimPajama dataset), starting from the uptrained Recursive Transformers. We also utilized eight H100 GPUs and maintained consistent configurations with the uptraining settings, including batch size, context length, and learning rates.

\paragraph{Evaluation setting.}
We evaluated perplexity on test sets from three language modeling datasets: SlimPajama, RedPajama, and PG19\,\citep{rae2019compressive}. Additionally, we used the Language Model Evaluation Harness framework\,\citep{eval-harness} to evaluate accuracy on seven few-shot tasks: LAMBADA\,\citep{paperno2016lambada}, HellaSwag\,\citep{zellers2019hellaswag}, PIQA\,\citep{bisk2020piqa}, WinoGrande\,\citep{DBLP:conf/aaai/SakaguchiBBC20}, ARC-easy and ARC-challenge\,\citep{clark2018think}, and OpenBookQA\,\citep{DBLP:conf/emnlp/MihaylovCKS18}. We adhered to the standard number of shots specified by the evaluation framework for each dataset. For few-shot datasets, excluding LAMBADA and WinoGrande, we normalized accuracy by the byte length of the target string. All evaluation performance measurements were conducted using a single H100 GPU.

\paragraph{Throughput measurement settings.}
To present the hypothetical generation speeds of our Recursive Transformers, we prepared two key elements: per-token generation time and exit trajectory datasets. Firstly, we measured the generation time under various model configurations using dummy weights and inputs. We measured the time for each component, such as embedding matrices, Transformer blocks, and the classifier head (final throughput comparisons were based solely on the time spent within Transformer blocks.) We tested two settings of prefix and decoding lengths (512\,/\,2048 and 64\,/\,256), calculating the per-token time by dividing the total elapsed time by the decoding length. Using a single A100 40GB GPU, we recorded these decoding times across different batch sizes, until an out-of-memory error occurred or under a specific memory constraint was reached.

To obtain exit trajectory data, we assumed an oracle-exiting approach, where all tokens could exit at intermediate loops if intermediate predictions matched the final loop's prediction. Since our models are not finetuned on any specific downstream tasks, we did simulation with three language modeling datasets (SlimPajama, RedPajama, and PG19) as if they were generated by our models. For simplicity, we assumed a queue of 20K samples, rather than considering their arrival in static or dynamic time intervals. We then recorded the exit loop of each token in these samples using the oracle-exiting algorithm. With these two measurement (per-token generation time and exit trajectories), we present the hypothetical throughput of Recursive Transformers under various simulation scenarios.

\begin{table}[t!]
    \small
    \centering
    \resizebox{\textwidth}{!}{
    \setlength{\tabcolsep}{3pt}
    \begin{tabular}{l|c|cc|rrr|ccccccc|c}
    \toprule
      &  & \multicolumn{2}{c|}{\textbf{Uptrain}} & \multicolumn{3}{c|}{\textbf{Perplexity\,$\downarrow$}} & \multicolumn{8}{c}{\textbf{Few-shot Accuracy\,$\uparrow$}} \\
    \cmidrule(l{2pt}r{2pt}){3-4} \cmidrule(l{2pt}r{2pt}){5-7}  \cmidrule(l{2pt}r{2pt}){8-15} 
     \textbf{Models} & N-emb & PT & $N_{tok}$ & SlimP & RedP & PG19 & LD & HS & PQ & WG & ARC-e & ARC-c & OB & Avg  \\
    \midrule
    \multirow{3}{*}{Gemma\,2B} & 1.99B & \cmark & - & 11.46 & \textbf{8.18} & 13.52 & {63.1} & \textbf{71.4} & \textbf{78.1} & \textbf{65.0} & \textbf{72.3} & \textbf{41.9} & 40.2 & \textbf{61.7}  \\
     & 1.99B &  \cmark & 15B & 10.76 & 8.47 & 13.08 & \textbf{63.5} & 68.5 & 77.0 & 63.5 & 67.6 & 38.1 & {42.6} & 60.1  \\
      & 1.99B & \cmark & 60B & \textbf{10.58} & 8.44 & \textbf{12.71} & 60.3 & 67.9 & 76.9 & 63.5 & 64.9 & 37.2 & 39.6 & 58.6  \\
     \midrule
    \multirow{3}{*}{TinyLlama\,1.1B} & 0.97B & \cmark &  - & 12.26 & 9.37 & 11.94 & 43.3 & 42.2 & 66.8 & 53.4 & 44.7 & 23.2 & 29.2 & 43.3  \\
     & 0.97B  & \cmark  & 15B & 9.87 & 8.24 & 10.73 & 49.2 & 46.3 & \textbf{68.8} & 54.0 & 48.2 & 26.0 & 32.2 & 46.4 \\
     & 0.97B  & \cmark & 60B & \textbf{9.59} & \textbf{8.12} & \textbf{10.42} & \textbf{51.6} & \textbf{48.8} & 68.6 & \textbf{54.1} & \textbf{49.9} & \textbf{26.2} & \textbf{32.8} & \textbf{47.4}  \\
     \midrule
    \multirow{3}{*}{Pythia\,1B} & 0.81B  & \cmark & - & 15.68 & 9.90 & \textbf{12.05} & \textbf{57.5} & 49.1 & 70.4 & 52.8 & \textbf{51.9} & 26.7 & \textbf{33.4} & \textbf{48.8}  \\
     & 0.81B  & \cmark & 15B & 13.46 & 9.95 & 13.38 &  55.0 & 49.0 & 71.0 & 53.6 & 51.8 & \textbf{28.2} & 32.8 & \textbf{48.8} \\
     & 0.81B  & \cmark & 60B & \textbf{12.83} & \textbf{9.76} & 13.57 & 53.0 & \textbf{50.2} & \textbf{71.1} & \textbf{54.8} & \textbf{51.9} & 27.7 & 31.6 & 48.6  \\
    \bottomrule
    \end{tabular}
    }
    \caption{
    Uptraining the pretrained models on datasets that differ significantly in quality or distribution from their pretraining datasets can lead to decreased performance. We evaluated models after uptraining on the SlimPajama dataset. We measured perplexity on test sets of the  SlimPajama, RedPajama, and PG19, and few-shot accuracy on LAMBADA, HellaSwag, PIQA, WinoGrande, ARC-easy, ARC-challenge, and OpenBookQA benchmarks.
    }
    \label{tab_rrt:target_baseline_performance}
\end{table}

\subsection{Non-Recursive Model Baselines}
\label{sec_rrt:exp_baselines}

Given that we leveraged pretrained model weights for initialization and subsequently uptrained the models, it becomes crucial to define clear performance targets for our parameter-shared models.

\paragraph{Full-size model.}

Our ultimate goal is for the Recursive Transformer to achieve performance comparable to the original, full-size pretrained model, without much uptraining. However, we observed that the distribution divergence between the pretraining and uptraining datasets can hinder achieving the desired performance. In particular, uptraining on new datasets, particularly those of comparatively lower quality, sometimes led to performance degradation on certain benchmarks. Table\,\ref{tab_rrt:target_baseline_performance} summarizes the evaluation results of full-size models based on the number of uptraining tokens. For instance, in the case of Gemma, where the pretraining dataset is unreleased but potentially well-curated\,\citep{team2024gemma}, all few-shot performance metrics gradually decreased after uptraining on the SlimPajama dataset. This suggests that the achievable upper bound performance with the SlimPajama dataset might be considerably lower than the original model performance. Therefore, we set the target performance for Gemma and Pythia models as the performance achieved by uptraining a full-size pretrained model with an equivalent number of tokens. Since TinyLlama was already pretrained on SlimPajama---which is the same dataset we use for uptraining (eliminating any distribution shift)---for slightly longer than our runs, we use the performance of the original checkpoint as reference.

\paragraph{Reduced-size model.}

To demonstrate the performance advantages of Recursive Transformers compared to models with an equivalent number of parameters, we introduce another baseline: reduced-size models. These models have either half or one-third the parameters of their full-sized counterparts, matching the parameter count of our recursive models. However, these reduced models are pretrained from scratch on the same training recipe (number of training tokens and distillation from full-size model), but without the benefits of the pretrained weights and the looping mechanism. This comparison serves to highlight the efficacy of our initialization techniques and the recursive function itself in attaining strong performance, even with a constrained model size.

\subsection{Main Results}
\label{sec_rrt:main_results}

\begin{figure}[t!]
    \centering
    \begin{subfigure}[t]{0.325\textwidth}
        \includegraphics[width=\textwidth]{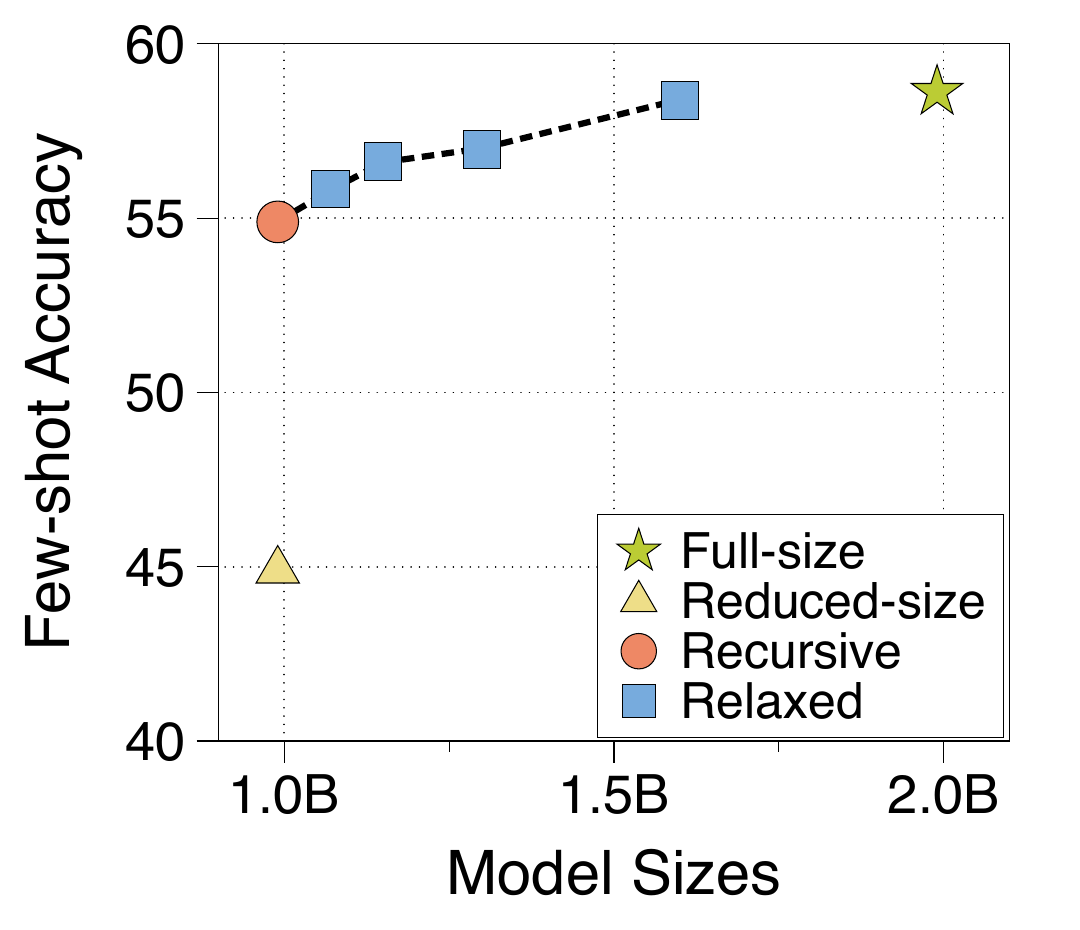}
        \subcaption{Gemma}
    \end{subfigure}
    \centering
    \begin{subfigure}[t]{0.325\textwidth}
        \includegraphics[width=\textwidth]{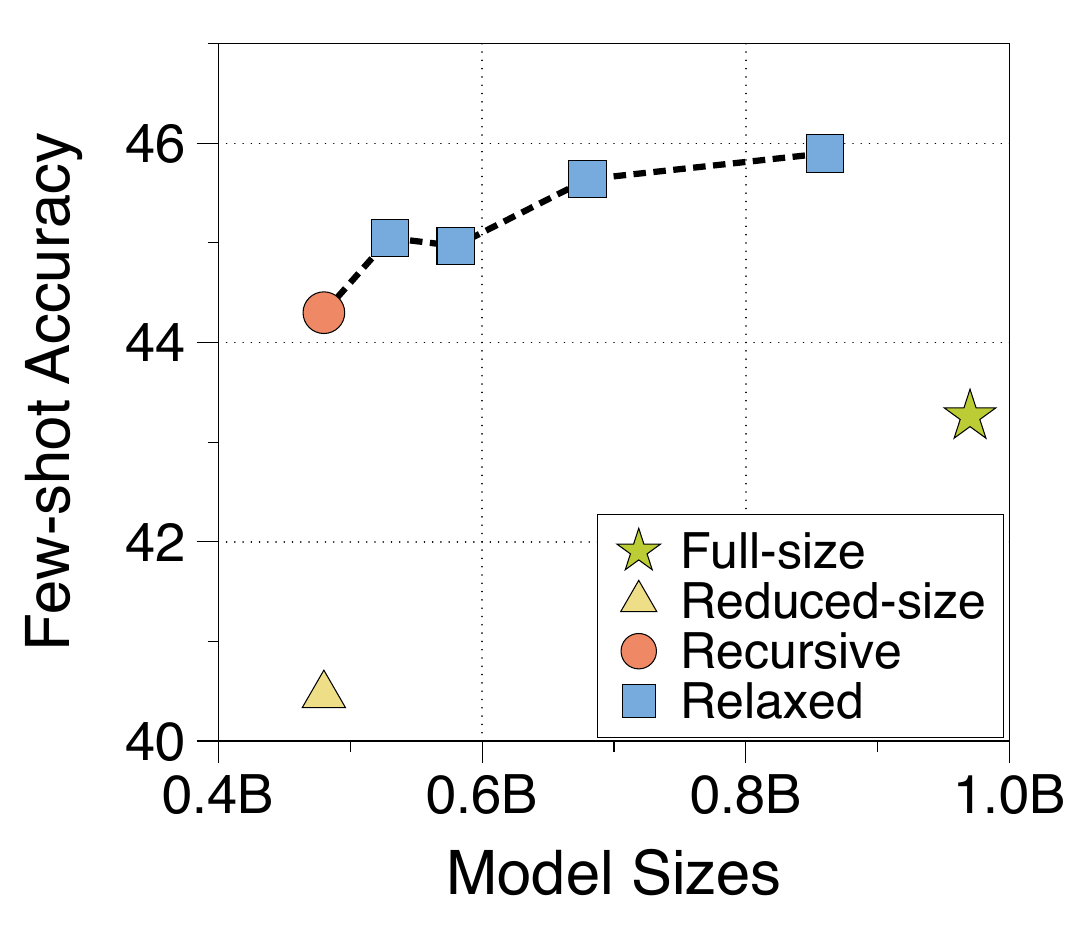}
        \subcaption{TinyLlama}
    \end{subfigure}
    \centering
    \begin{subfigure}[t]{0.325\textwidth}
        \includegraphics[width=\textwidth]{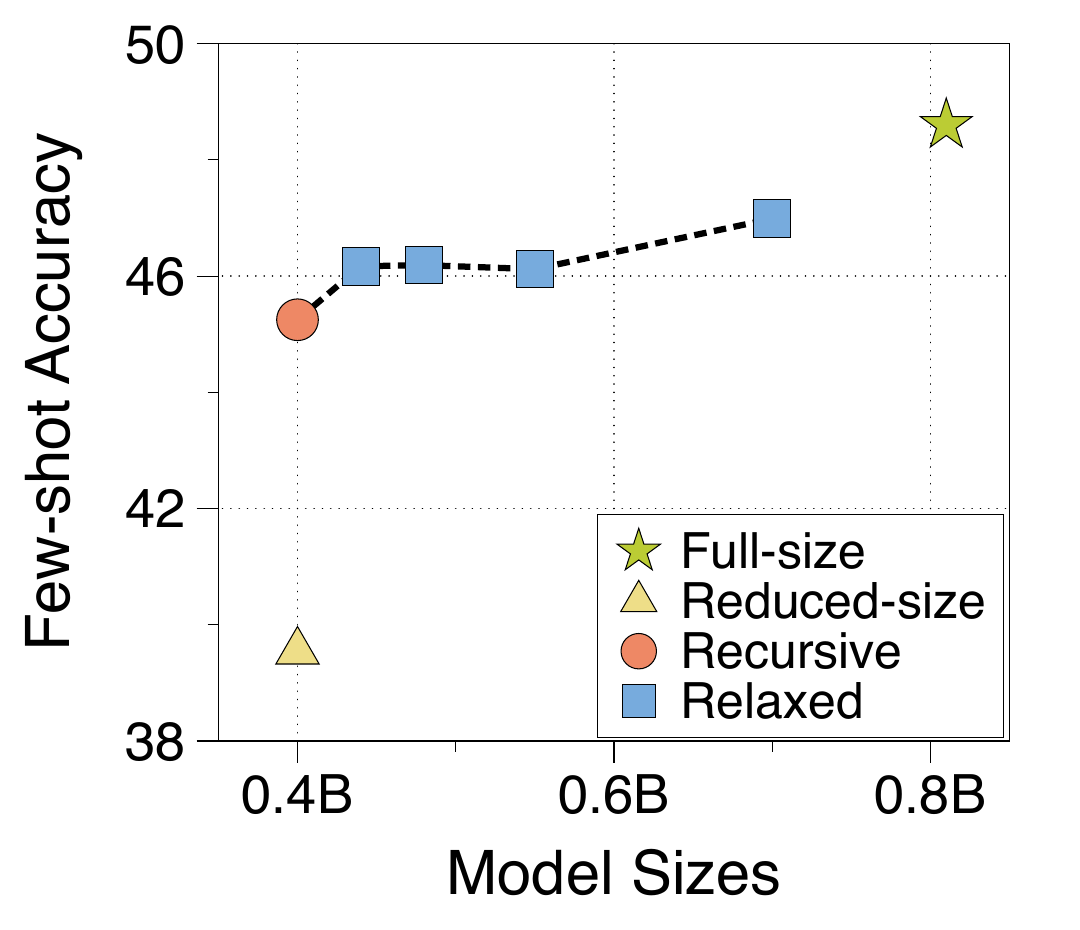}
        \subcaption{Pythia}
    \end{subfigure}
    \caption{
    Recursive and Relaxed Recursive Transformers achieve comparable performance to full-size models, and significantly outperform reduced-size models. Recursive models were initialized using the Stepwise method, while relaxed models utilized Average and SVD methods for looped layers and LoRA modules. We show the performance of four different rank values:  64, 128, 256, and 512. Recursive and reduced-size models were either uptrained (recursive model) and pretrained from scratch (reduced-size model) on 60 billion tokens using a knowledge distillation objective.
    }
    \label{fig_rrt:main_figures}
\end{figure}

Figure\,\ref{fig_rrt:main_figures} presents the few-shot performance of Recursive Transformers with two blocks and their relaxed variants. Recursive Transformers, even without relaxation, demonstrate remarkably high performance  despite having only half the parameters of the full-size model. The Gemma model achieved a 10\%p performance gain compared to the reduced-size model, which was also trained on 60 billion tokens using distillation loss.
Remarkably, the recursive TinyLlama model even surpassed the vanilla model's performance, even though the latter was pretrained on a larger corpus of 105 billion tokens. 
Our initialization techniques proved highly effective in achieving this superior result, along with the benefit of the uptraining dataset (SlimPajama) being the same as its pretraining dataset.

The relaxed models effectively interpolate between the full-size model and the Recursive Transformer, depending on the LoRA rank. As the model size increases with larger LoRA modules, SVD initialization methods allow for a more precise approximation of full-rank matrices, resulting in improved performance. Notably, the relaxed Gemma model with a rank of 512 achieves performance on par with the original model pretrained on 3 trillion tokens (58.4\% vs. 58.6\%), despite using fewer parameters and uptraining on only 60 billion tokens. This trade-off provides flexibility in selecting the best configuration for various deployment scenarios. We believe that additional uptraining and higher-quality datasets could yield better performance with even more streamlined models. 

In the subsequent sections, we provide a comprehensive overview of extensive ablation studies conducted prior to achieving this final performance. In \S\ref{sec_rrt:exp_initialization}, we delve into the analysis of various initialization methodologies for Recursive Transformers. Insights into the relaxation model are detailed in \S\ref{sec_rrt:exp_relaxed_recursive_transformer}. Finally, we explore enhanced training strategies like knowledge distillation (\S\ref{sec_rrt:kd_long_training}).

\subsection{Initialization Techniques for Looped Layers}
\label{sec_rrt:exp_initialization}

\begin{figure}[t!]
    \centering
    \begin{subfigure}[t]{0.325\textwidth}
        \includegraphics[width=\textwidth]{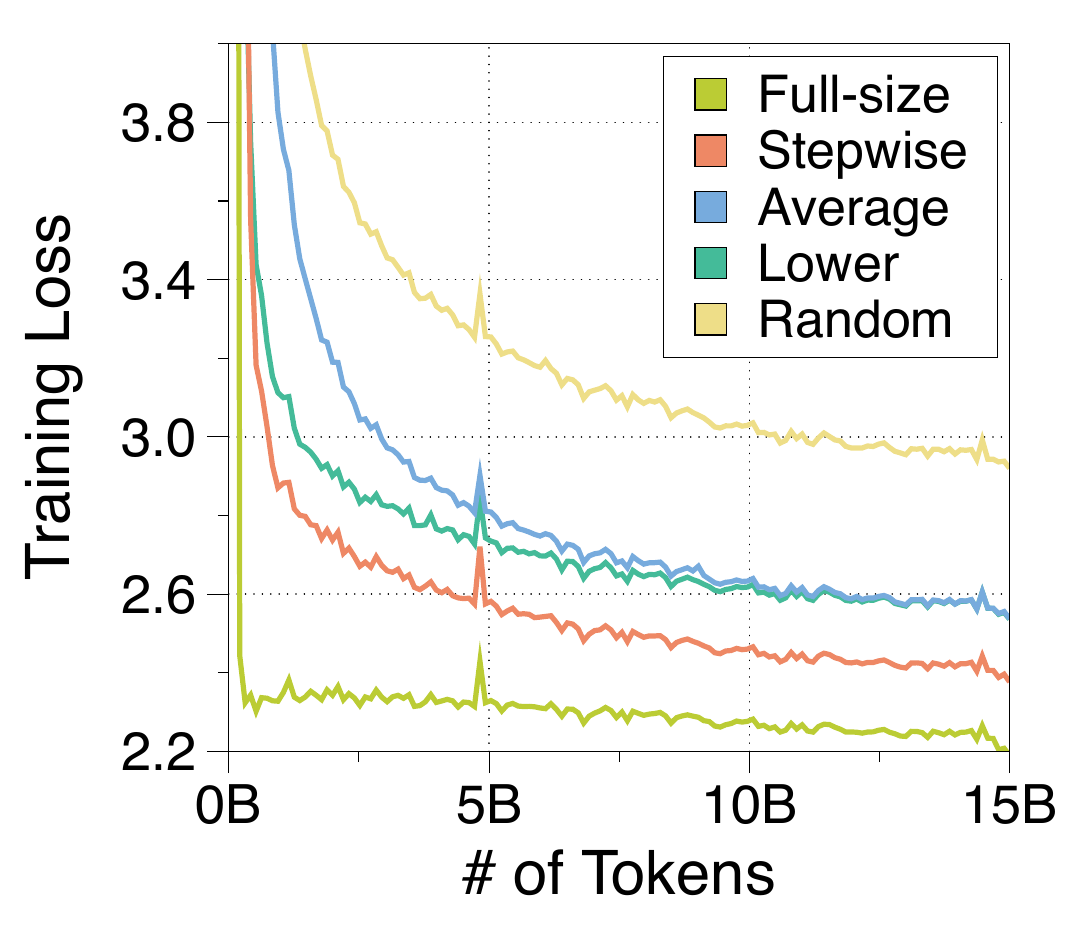}
        \subcaption{Loss curves for Gemma}
        \label{fig_rrt:initialization_gemma_training_loss}
    \end{subfigure}
    \centering
    \begin{subfigure}[t]{0.325\textwidth}
        \includegraphics[width=\textwidth]{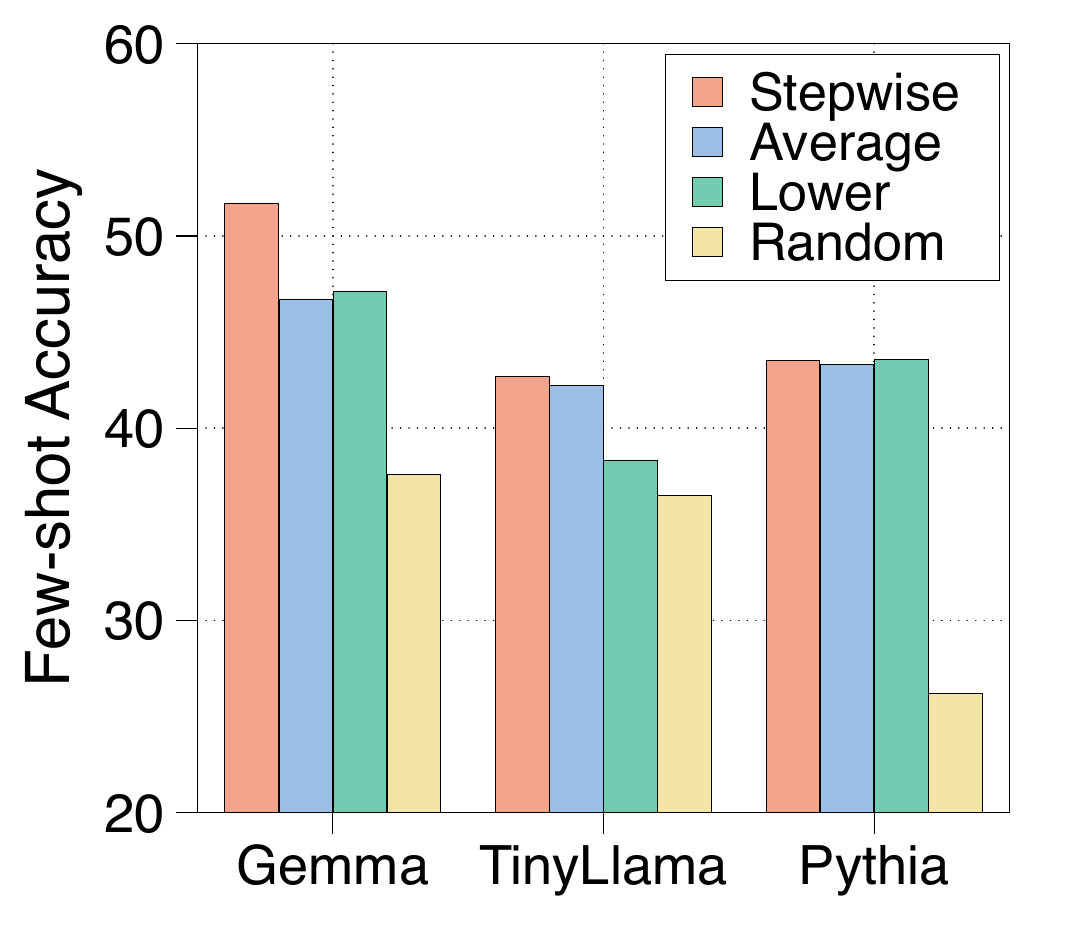}
        \subcaption{Average few-shot performance}
        \label{fig_rrt:initialization_avg_fewshot}
    \end{subfigure}
    \centering
    \begin{subfigure}[t]{0.325\textwidth}
        \includegraphics[width=\textwidth]{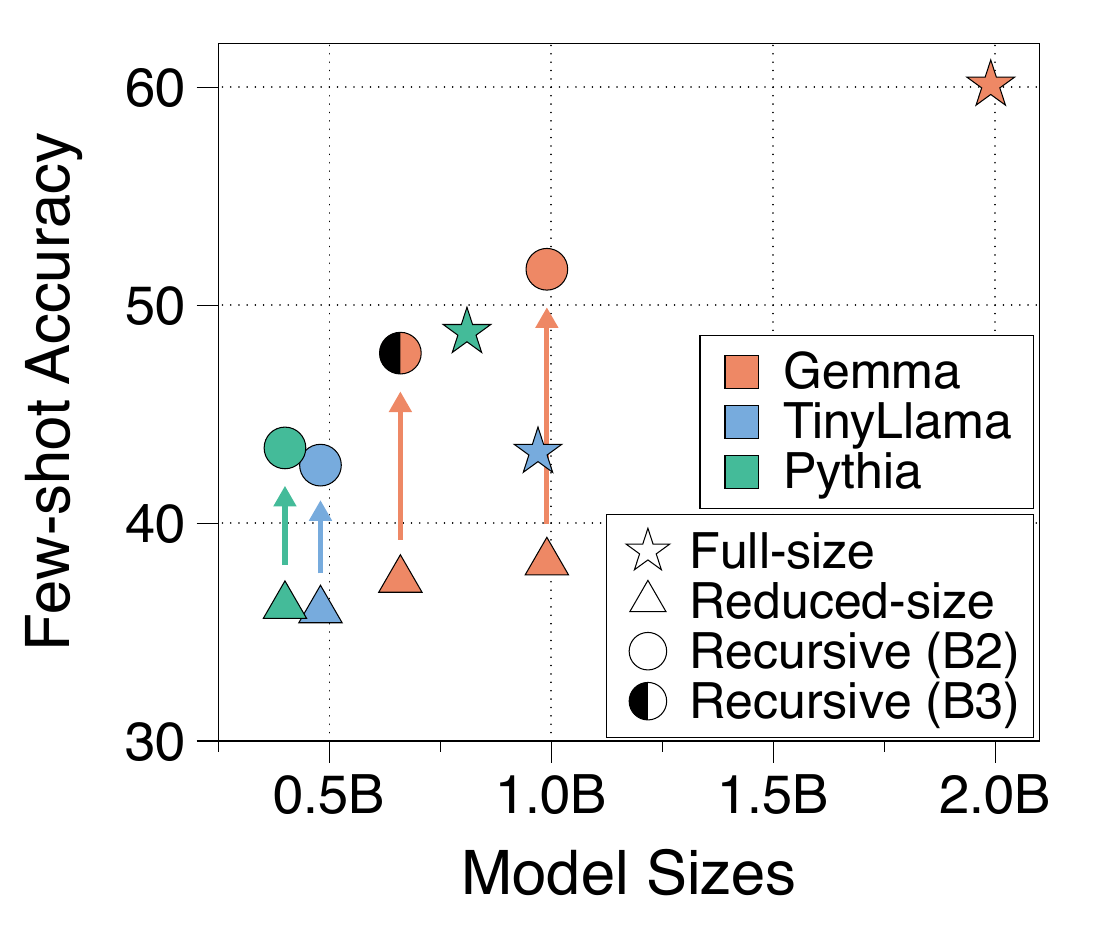} 
        \subcaption{Recursive model performance}
        \label{fig_rrt:initialization_model_size_performance}
    \end{subfigure}
    \caption{
    {(a)} Among the proposed methods, the Stepwise method obtains the lowest training loss on the SlimPajama dataset.
    {(b)} The Stepwise method consistently demonstrate the highest average few-shot accuracy across three architectures.
    {(c)} Recursive Transformers initialized with the Stepwise method demonstrated significant performance gains compared to non-recursive model baselines.
    }
    \label{fig_rrt:initialization_gemma}
\end{figure}

\paragraph{Stepwise initialization serves as the best initial point for Recursive Transformers.} We present the training loss of Gemma models initialized using three different methods in Figure\,\ref{fig_rrt:initialization_gemma_training_loss}, and their few-shot performance in Figure\,\ref{fig_rrt:initialization_avg_fewshot}. Our proposed methods significantly outperformed random initialization, which simply adds recursion to a reduced-size model, suggesting that leveraging pretrained weights in any manner is beneficial for performance boost. Moreover, the Stepwise methodology consistently demonstrated best performance, aligning with insights that LLMs can preserve performance even with a few layers skipped\,\citep{DBLP:journals/corr/abs-2404-02258, DBLP:conf/acl/Zhang00S0CM24, DBLP:conf/acl/ElhoushiSLHWL0A24}. Interestingly, as summarized in Table~\ref{tab_rrt:initialization_total_app}, the recursive TinyLlama model, uptrained on only 15 billion tokens, yields few-shot performance comparable to the original model pretrained on 105 billion tokens. This suggests that with sufficient training, even a recursive architecture can match the performance of a full-size pretrained model~\citep{DBLP:conf/iclr/DehghaniGVUK19, DBLP:conf/sustainlp/TakaseK23}.

\paragraph{Recursive Gemma 1B outperforms both pretrained TinyLlama 1.1B and Pythia 1B.}

The looped Gemma 1B model, utilizing our proposed Stepwise method, outperformed reduced-size baselines with equivalent parameter counts by up to 13.5 percentage points (51.7\% vs. 38.2\%). Furthermore, it even outperformed the full-size TinyLlama 1.1B and Pythia 1B models (see Figure\,\ref{fig_rrt:initialization_model_size_performance}). This is a noteworthy achievement given that Pythia was pretrained on 300 billion tokens, whereas the recursive Gemma was uptrained on only 15 billion tokens. 
Consequently, high-performing LLMs serve as a promising starting point, as their recursive counterparts readily outperform other ordinary vanilla models of similar size.
Further details can be found in Appendix~\ref{app_rrt:initialization}.

\begin{tcolorbox}[
    enhanced,
    colback=blue!5!white, 
    colframe=black, 
    width=\textwidth,
    coltitle=white,
    title=\textbf{Takeaways for Recursive Transformer},
    top=3mm, bottom=2mm,
    attach boxed title to top left={yshift=-2.8mm, xshift=4mm},
    boxed title style={colback=black, colframe=black, boxrule=0mm, 
    toptitle=1mm, bottomtitle=1mm}
    ]
We find that converting well-pretrained models into Recursive Transformers leads to high-performing models with minimal uptraining. Notably, initializing looped layers via the Stepwise method yields the best results. With just 15 billion tokens of uptraining, a recursive Gemma 1B model outperforms even the full-size pretrained TinyLlama and Pythia models.
\end{tcolorbox}

\subsection{Relaxation of Parameter Sharing via LoRA Modules}
\label{sec_rrt:exp_relaxed_recursive_transformer}

\begin{figure}[t!]
    \centering
    \begin{subfigure}[t]{0.325\textwidth}
        \includegraphics[width=\textwidth]{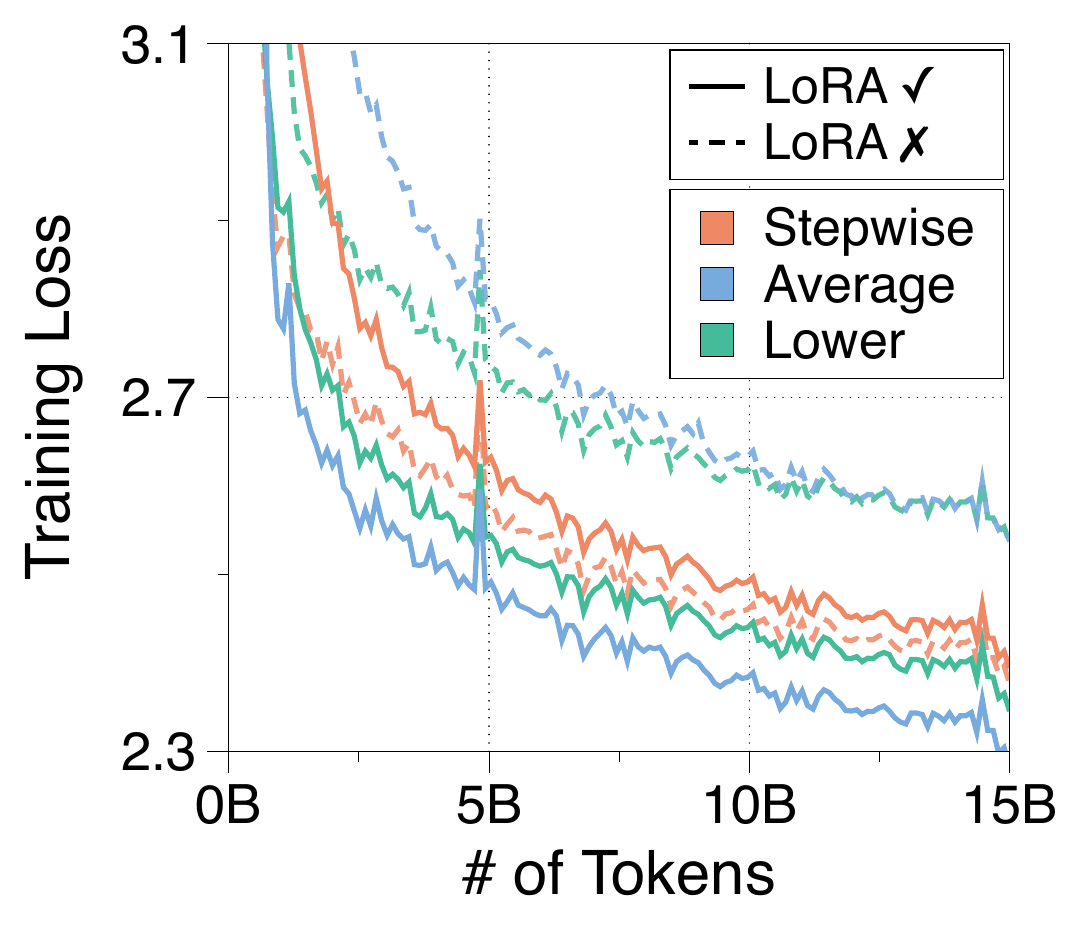}
        \subcaption{Loss changes in Gemma}
        \label{fig_rrt:lora_gemma_training_loss}
    \end{subfigure}
    \centering
    \begin{subfigure}[t]{0.325\textwidth}
        \includegraphics[width=\textwidth]{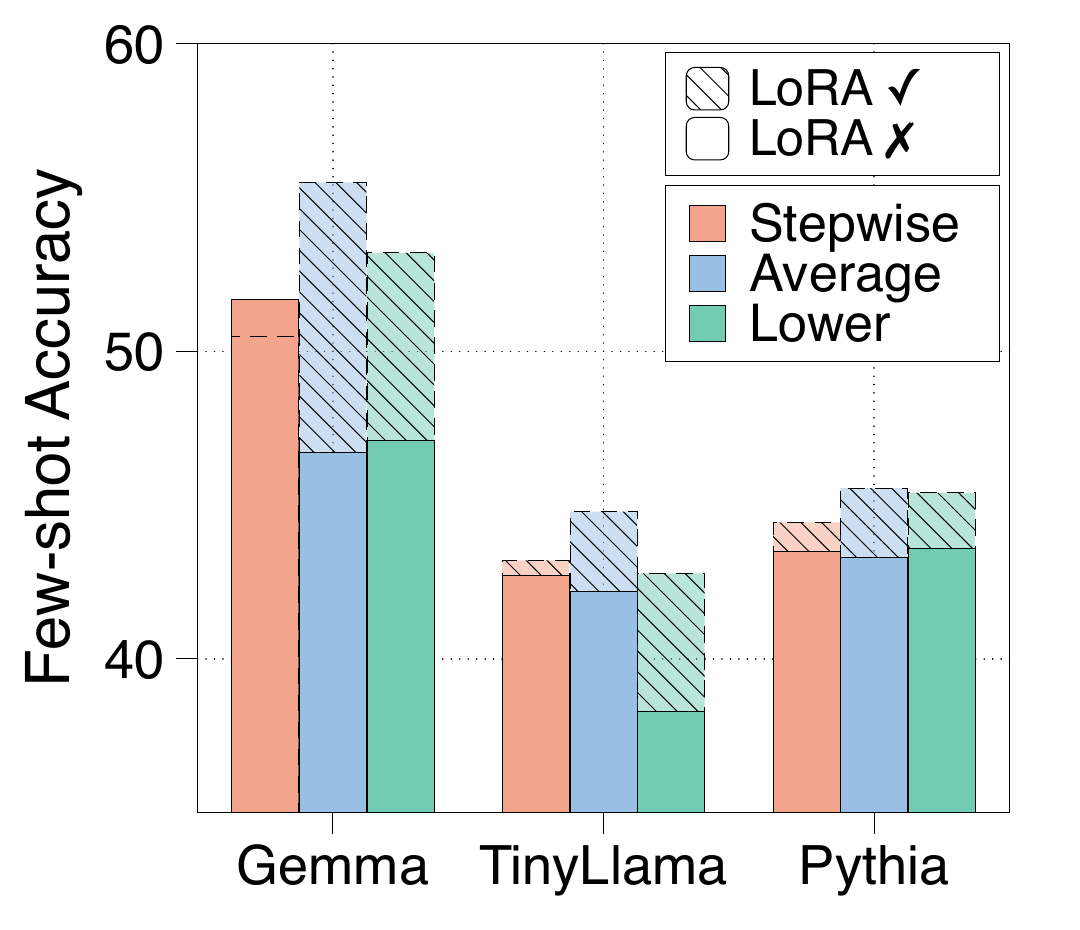}
        \subcaption{Accuracy gains from relaxation}
        \label{fig_rrt:lora_fewshot_accuracy}
    \end{subfigure}
    \centering
    \begin{subfigure}[t]{0.325\textwidth}
        \includegraphics[width=\textwidth]{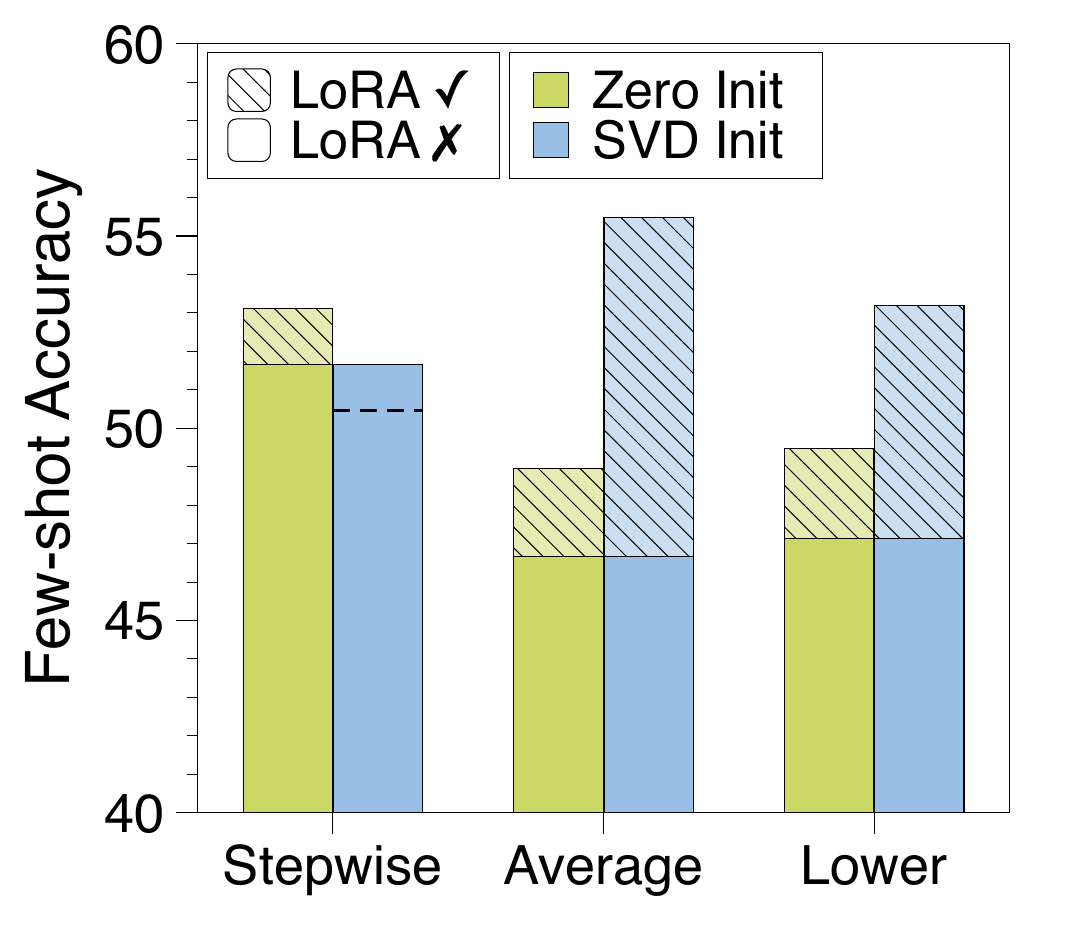} 
        \subcaption{Effects of SVD initialization}
        \label{fig_rrt:svd_zero_gemma}
    \end{subfigure}
    \caption{
    The Relaxed Recursive Transformer, with its looped layer initialized using Average method, achieved the best performance in terms of both ({a}) training loss and ({b}) few-shot accuracy. The models utilize two blocks, with the LoRA modules initialized using the SVD method at a rank of 512. ({c}) SVD initialization method significantly enhanced performance compared to zero initialization.
    }
    \label{fig_rrt:layer_wise_lora_main}
\end{figure}

\paragraph{Average initialization for looped layers is most compatible with Relaxed Recursive Models.}

Figures\,\ref{fig_rrt:lora_gemma_training_loss} and \ref{fig_rrt:lora_fewshot_accuracy} illustrate the effect of relaxing parameter sharing via layer-wise LoRA modules. Notably, initializing tied layers in relaxed models with Average method yielded substantial performance improvements, even outperforming the non-relaxed model initialized with Stepwise. Approximating residual matrices between averaged weights and their individual weights appears readily achievable using truncated SVD with low ranks. In contrast, we observed an intriguing phenomenon where our models initialized with Stepwise occasionally showed performance degradation after relaxation. This is likely because capturing the nuances between entirely distinct layer weights is challenging with an insufficient rank, leading to a suboptimal solution. Further details are provided in Appendix\,\ref{app_rrt:cross_layer_lora}.

\paragraph{SVD initialization to approximate pretrained weights outperforms zero initialization.}

LoRA modules initialized with zero values guarantee that the model begins training from the same point as the non-relaxed model. Conversely, SVD initialization positions the model closer to either the full-size model (with full-rank) or the non-relaxed model (with small rank). To emphasize the effectiveness of initializing near full-size model weights, we compared these two methods at a moderately large rank of 512, as shown in Figure\,\ref{fig_rrt:svd_zero_gemma}. Our proposed SVD strategy demonstrated an impressive performance boost of up to 6.5 points, facilitating faster convergence by updating the principal low-rank matrices (aligned with findings in Meng et al. (2024)~\cite{DBLP:journals/corr/abs-2404-02948}). For results across other architectures, refer to Figure\,\ref{fig_rrt:zero_svd_init_app}.

\paragraph{Higher rank enhances recovery of original pretrained weights.}

\begin{figure}[t!]
    \centering
    \begin{subfigure}[t]{0.325\textwidth}
        \includegraphics[width=\textwidth]{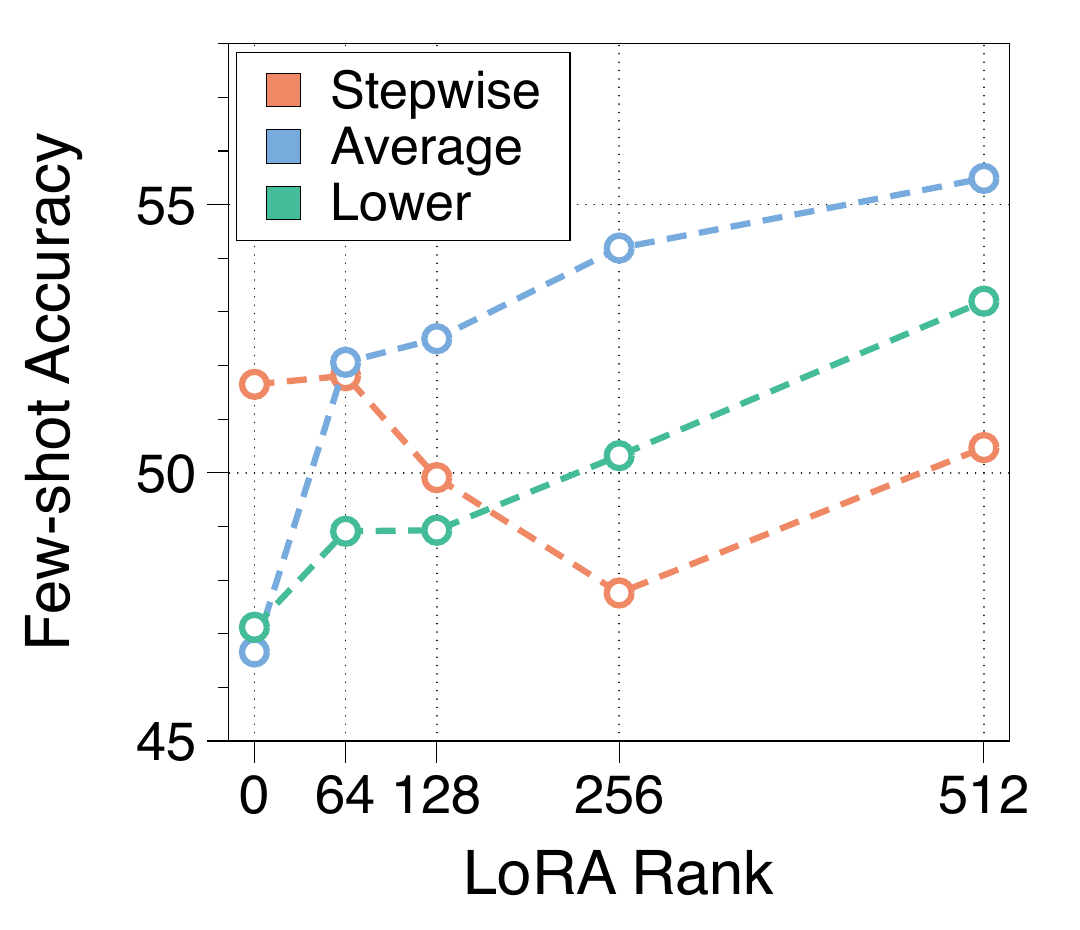}
        \subcaption{Performance by LoRA rank}
        \label{fig_rrt:lora_rank}
    \end{subfigure}
    \centering
    \begin{subfigure}[t]{0.325\textwidth}
        \includegraphics[width=\textwidth]{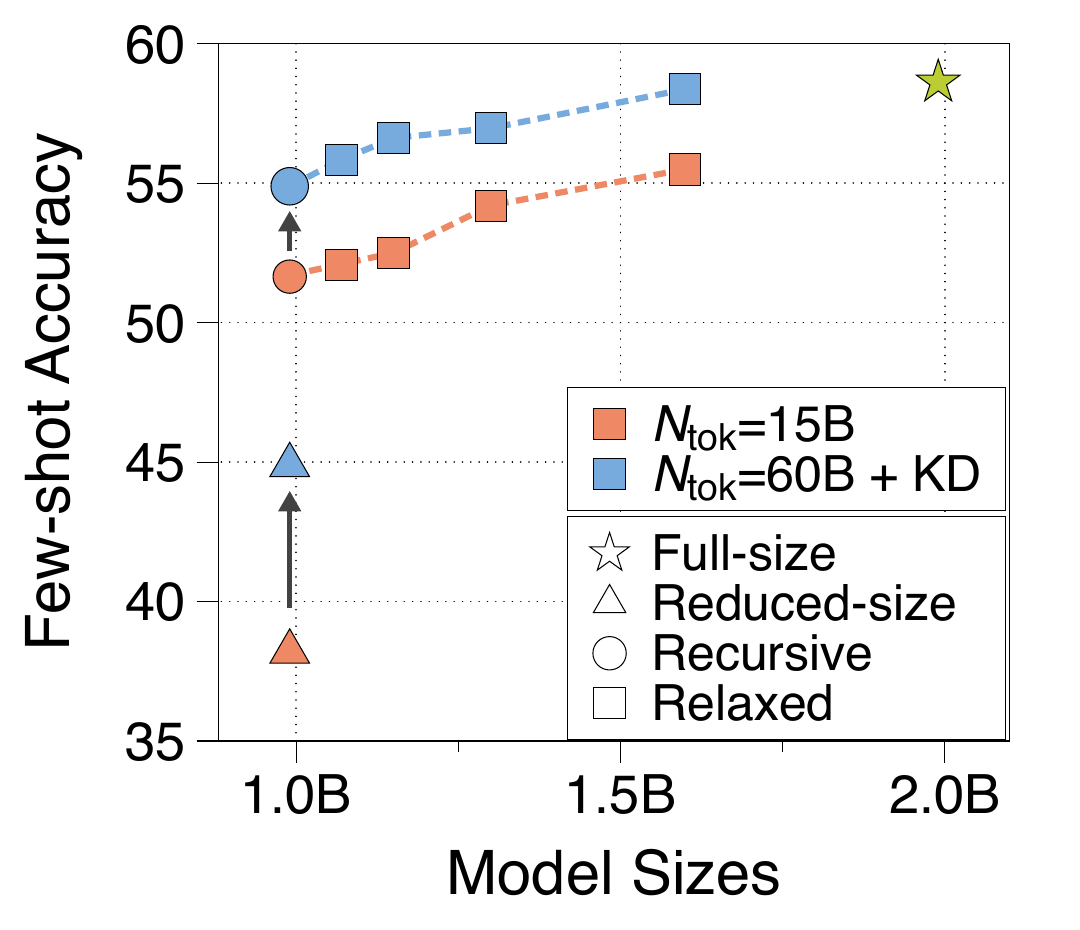} 
        \subcaption{Gains from longer KD training}
        \label{fig_rrt:kd_long_gemma}
    \end{subfigure}
    \centering
    \begin{subfigure}[t]{0.325\textwidth}
        \includegraphics[width=\textwidth]{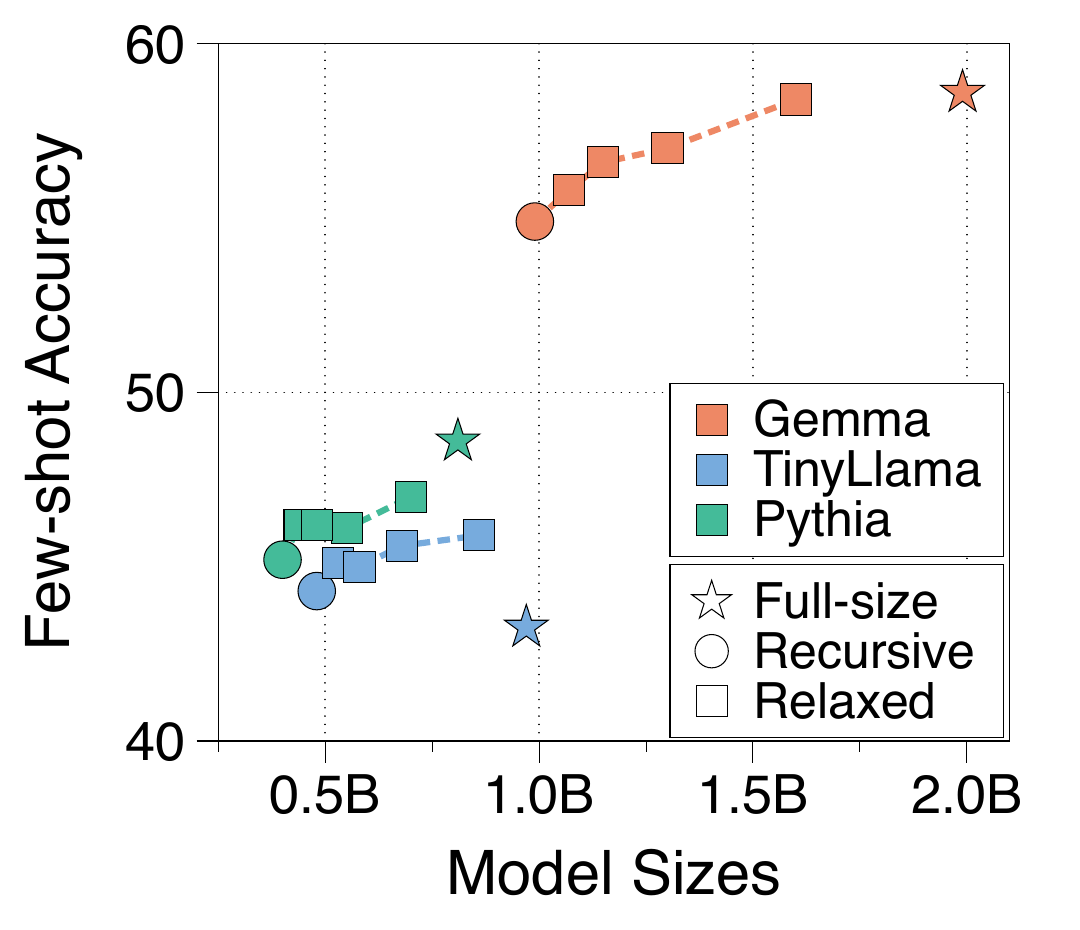}
        \subcaption{Overall performance}
        \label{fig_rrt:overall_performance}
    \end{subfigure}
    \label{fig_rrt:lora_rank_long_training}
    \caption{
    ({a}) Increasing the LoRA rank typically leads to improved performance in relaxed Gemma models, attributed to the use of SVD initialization.
    ({b}) Extended uptraining and knowledge distillation yielded substantial accuracy gains for Gemma models.
    Note that the full-size model is a pretrained model that is further uptrained on 60 billion tokens.
    ({c}) 
    Recursive and Relaxed Recursive Transformers achieve a compelling Pareto frontier with respect to model size and performance.
    Recursive and relaxed models used Stepwise and Average method to initialize looped layers, respectively.
    }
\end{figure}

At full rank, relaxed models can perfectly match full-size pretrained models. Consequently, as illustrated in Figure\,\ref{fig_rrt:lora_rank}, performance generally improves with increasing rank, resulting in a clear Pareto frontier between model size and performance. However, only Stepwise initialization showed a U-shaped performance trend: a middle-range rank resulted in poor approximation, whereas very low ranks (akin to random initialization for LoRA modules) yielded better performance. The overall results are summarized in Table\,\ref{tab_rrt:lora_total_app}.

\begin{tcolorbox}[
    enhanced,
    colback=blue!5!white, 
    colframe=black, 
    width=\textwidth,
    coltitle=white,
    title=\textbf{Takeaways for Relaxed Recursive Transformer},
    top=3mm, bottom=2mm,
    attach boxed title to top left={yshift=-2.8mm, xshift=4mm},
    boxed title style={colback=black, colframe=black, boxrule=0mm, 
    toptitle=1mm, bottomtitle=1mm}
]
Adjusting the LoRA rank in the Relaxed Recursive Transformer, together with our SVD-based initialization technique, allows for a smoother trade-off between a fully weight-tied recursive model and a vanilla model. Furthermore, we find that initializing the shared weights in the looped layers with the Average method leads to the best performance in this setting.
\end{tcolorbox}

\subsection{Extended Uptraining and Knowledge Distillation}
\label{sec_rrt:kd_long_training}

We further enhanced the performance of our low-rank models by introducing two techniques: uptraining on an extended corpus and knowledge distillation from the full-sized model. Specifically, we increased the number of uptraining tokens from 0.5\% to 2\% of the total 3 trillion tokens used for pretraining Gemma models, resulting in a total of 60 billion tokens. Additionally, we regularized the losses using a forward Kullback-Leibler divergence\,\citep{DBLP:journals/corr/HintonVD15, DBLP:conf/emnlp/KimR16}, which exhibited the best performance gains among the examined distillation losses. Table~\ref{tab_rrt:kd_ablation_study} summarizes the results of various ablation studies conducted to investigate the impact of these two techniques. 

The combined effect of these techniques is presented in Figure\,\ref{fig_rrt:kd_long_gemma}, demonstrating an improvement of up to 4.1 percentage points in few-shot accuracy compared to the previous 15 billion token uptraining results. Notably, the relaxed Gemma model with a rank of 512 nearly matched the performance of the full-size model. We also expect that further performance gains can be achieved with a much lighter recursive model by utilizing a superior teacher model or conducting more extensive training on high-quality data. Figure\,\ref{fig_rrt:overall_performance} illustrates the Pareto frontier achieved by the final models. All models exhibit competitive performance compared to the full-size model. Moreover, the superior performance of the recursive Gemma model strongly highlights the advantages of converting high-performing LLMs to a recursive architecture. Additional details can be found in Appendix\,\ref{app_rrt:further_techniques}.

\begin{table}[t!]
        \small
    \centering
    \renewcommand{\arraystretch}{0.9}
    \resizebox{\textwidth}{!}{
    \setlength{\tabcolsep}{4pt}
    \begin{tabular}{c|cc|cc|ccc|ccccccc|cc}
    \toprule
     & \multicolumn{2}{c|}{\textbf{Uptrain}} & \multicolumn{2}{c|}{\textbf{Looping}} &  \multicolumn{3}{c|}{\textbf{Early-Exit Train}} & \multicolumn{9}{c}{\textbf{Few-shot Accuracy\,$\uparrow$}} \\
    \cmidrule(l{2pt}r{2pt}){2-3} \cmidrule(l{2pt}r{2pt}){4-5}  \cmidrule(l{2pt}r{2pt}){6-8}  \cmidrule(l{2pt}r{2pt}){9-17} 
     N-emb & PT & $N_{tok}$ & Block & Init  &  $N_{tok}$ & CE & KD  & LD & HS & PQ & WG & ARC-e & ARC-c & OB & Avg & $\Delta$  \\
    \midrule
    0.99B & \cmark & 15B  & 2 & Step &  - &- & - & {53.0} & {57.3} & {73.2} & {56.2} & {56.1} & {29.2} & {36.6} & {51.7} & - \\
    \midrule
     \multirow{2}{*}{0.99B} & \multirow{2}{*}{\cmark} & \multirow{2}{*}{15B} & \multirow{2}{*}{2} & \multirow{2}{*}{Step} &  \multirow{2}{*}{15B}  &  \multirow{2}{*}{Weighted} &  \multirow{2}{*}{\xmark}  &   48.9 &  55.5 &  72.7 &  55.3 &  54.9 &  30.1 &  36.0 &  50.5 & \textcolor{custom_red}{\textbf{--\,1.2}} \\
     &  &  &  & &  &   & &   49.5 &  54.8 &  72.0 &  53.4 &  54.1 &  29.1 &  35.6 &  49.8 & - \\
     \addlinespace[-1pt]
     \cmidrule(l{2pt}r{2pt}){9-17} 
     \addlinespace[-1pt]
      &  &  &  &  &  &    &  &   53.0 &  59.1 &  73.9 &  55.4 &  57.4 &  30.6 &  37.8 &  52.5 & \textcolor{custom_green}{\textbf{+0.8}} \\
     \multirow{-2}{*}{0.99B} & \multirow{-2}{*}{\cmark} &  \multirow{-2}{*}{15B}  & \multirow{-2}{*}{2} & \multirow{-2}{*}{Step}   & \multirow{-2}{*}{15B}  & \multirow{-2}{*}{Agg\,(0.1)} & \multirow{-2}{*}{\xmark}   &  45.9 &  51.2 &  71.4 &  54.5 &  48.1 &  26.8 &  32.0 &  47.1 & - \\
     \midrule
     \multirow{2}{*}{0.99B} & \multirow{2}{*}{\cmark} & \multirow{2}{*}{15B} & \multirow{2}{*}{2} & \multirow{2}{*}{Step} &   \multirow{2}{*}{15B}   &  \multirow{2}{*}{Weighted} &  \multirow{2}{*}{\cmark}  &   47.7 &  55.1 &  73.2 &  55.6 &  54.5 &  29.1 &  37.2 &  50.4 & \textcolor{custom_red}{\textbf{--\,1.3}} \\
     &  &  &  & &  &  &  &  48.3 &  54.9 &  72.1 &  55.9 &  54.3 &  28.4 &  35.4 &  49.9 & - \\
     \addlinespace[-1pt]
     \cmidrule(l{2pt}r{2pt}){9-17} 
     \addlinespace[-1pt]
     \rowcolor[gray]{0.9}
      &  &  &  & &  &    &  &    52.9 &  58.9 &  73.7 &  55.7 &  57.5 &  31.1 &  38.2 &  52.6 & \textcolor{custom_green}{\textbf{+0.9}} \\
    \rowcolor[gray]{0.9}
     \multirow{-2}{*}{0.99B} & \multirow{-2}{*}{\cmark} &  \multirow{-2}{*}{15B}  & \multirow{-2}{*}{2} & \multirow{-2}{*}{Step}   & \multirow{-2}{*}{15B}  & \multirow{-2}{*}{Agg\,(0.1)} & \multirow{-2}{*}{\cmark} &  46.3 &  52.1 &  71.6 &  55.3 &  49.2 &  28.5 &  32.6 &  48.0 & - \\
    \bottomrule
    \end{tabular}
    }
    \caption{
    A small loss coefficient to the first loop output (intermediate output) can significantly improve intermediate performance without compromising the final performance. Performance was evaluated under a static-exiting scenario\,\citep{DBLP:conf/nips/SchusterFG0B0TM22}, where all tokens exit at either first or second loop. We further trained the previously uptrained Gemma models on 15 billion tokens (post-training). Delta\,($\Delta$) denotes the performance changes in the final outputs after early-exit training. 
    }
    \label{tab_rrt:early_exit_ablation}
\end{table}

\subsection{Early-Exit Training Strategy for Recursive Transformer}

The throughput of Recursive Transformers can be amplified by an early-exiting framework. Hence, we further train intermediate representations from fewer looping iterations to enable token prediction. We conducted an ablation study on various strategies, as summarized in Table\,\ref{tab_rrt:early_exit_ablation} (more detailed results are presented in Table\,\ref{tab_rrt:early_exit_ablation_app}). Directly applying the weighted CE loss ($\mathcal{L} = \sum_{i=1}^{B} \alpha_{i} \mathcal{L}_{i} \text{\; where \;} \alpha_{i} = i / \sum_{i} i$) commonly used in prior works~\citep{DBLP:conf/nips/SchusterFG0B0TM22, DBLP:conf/emnlp/BaeKSY23} led to an overemphasis on the training of intermediate representations. To address this, we employ an aggressive coefficient strategy that aggressively reduces the loss coefficient for intermediate outputs while maintaining a coefficient of 1 for the final output. Our experiments demonstrated that an aggressive coefficient of 0.1, utilizing knowledge distillation from the detached final outputs~\citep{DBLP:conf/emnlp/BaeKSY23}, effectively preserves final performance while enhancing intermediate performance. Notably, the first loop output yielded only a difference of 4.6 percentage points in accuracy compared to the final output. This underscores the potential to maximize the benefits of early-exiting in parameter-shared LLMs.

We applied this post-training strategy for early-exiting to our final uptrained models (shown in \S\ref{sec_rrt:main_results}), with all experimental results detailed in Appendix\,\ref{app_rrt:early_exit}. The aggressive coefficient strategy, combined with self-distillation, consistently achieved the best performance for intermediate outputs while maintaining strong performance for the final loop output across all models. However, as the optimal strategy derived from the non-relaxed models was directly applied to the relaxed models, a more tailored training approach might further enhance the performance of intermediate loop outputs in Relaxed Recursive Transformers.\looseness=-1

\subsection{Hypothetical Inference Speedup via Continuous Depth-wise Batching}
\label{exp_rrt:hypothetical_generation_speedup}

\begin{figure}[t!]
    \centering
    \begin{tabular}{cc}
        \raggedleft
        \begin{subfigure}[b]{0.49\textwidth}
            \includegraphics[width=\textwidth]{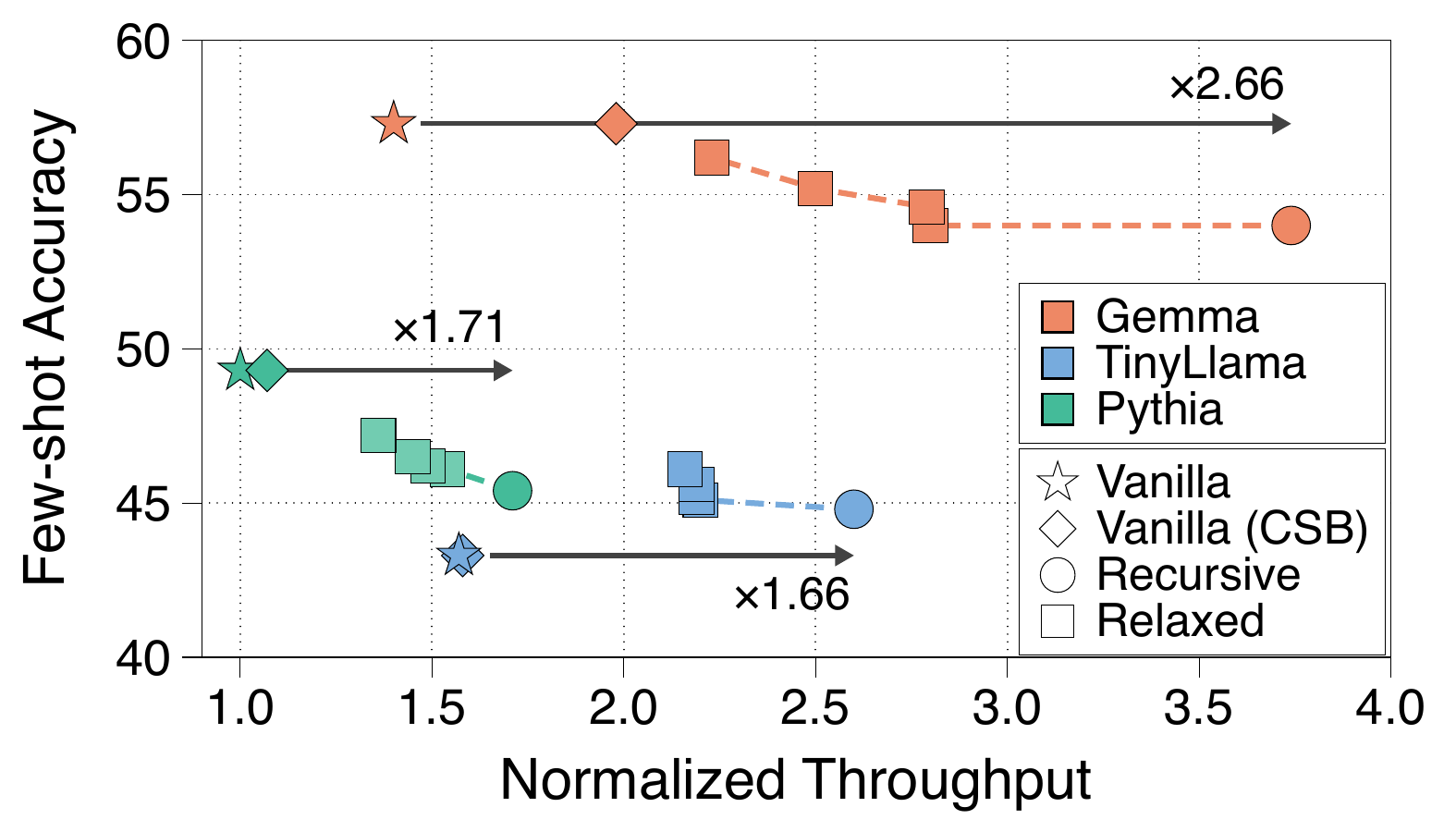}
            \vfill
        \end{subfigure} &
        \hspace{-10pt}
        \begin{minipage}[t]{0.49\textwidth}
        \vspace{-43mm}
            \raggedright
            \addtolength{\tabcolsep}{-3pt}
            \resizebox{1.0\textwidth}{!}{
            \begin{tabular}{ccc|cc|c|ccc}
            \toprule
            N-emb & Loop & LoRA & Batch & Exit & Acc. & Thr. & $\Delta_{V}$ & $\Delta_{Seq}$ \\
            \hline
            1.99B & - & - & -  & \xmark & 57.3 & 1080 & \textcolor{gray}{{$\mathbf{\times 1.00}$}} & \textcolor{custom_red}{{$\mathbf{\times 0.71}$}} \\ 
            1.99B & - & - & CSB & \xmark & 57.3 &  1528 & \textcolor{custom_green}{{$\mathbf{\times 1.41}$}} & \textcolor{gray}{{$\mathbf{\times 1.00}$}} \\ 
            \hline
            0.99B & 2 & - & CDB & \cmark  & 54.0 & 2877 & \textcolor{custom_green}{{$\mathbf{\times 2.66}$}} & \textcolor{custom_green}{{$\mathbf{\times 1.88}$}} \\ 
            1.07B & 2 & 64 & CDB & \cmark  & 54.0 & 2157 & \textcolor{custom_green}{{$\mathbf{\times 2.00}$}} & \textcolor{custom_green}{{$\mathbf{\times 1.41}$}} \\ 
            1.15B & 2 & 128 & CDB & \cmark  & 54.6 & 2149 & \textcolor{custom_green}{{$\mathbf{\times 1.99}$}} & \textcolor{custom_green}{{$\mathbf{\times 1.41}$}} \\ 
            1.30B & 2 & 256 & CDB & \cmark  & 55.2 & 1921 & \textcolor{custom_green}{{$\mathbf{\times 1.78}$}} & \textcolor{custom_green}{{$\mathbf{\times 1.26}$}} \\ 
            1.60B & 2 & 512 & CDB & \cmark  & 56.2 & 1719 & \textcolor{custom_green}{{$\mathbf{\times 1.59}$}} & \textcolor{custom_green}{{$\mathbf{\times 1.13}$}} \\ 
            \bottomrule
            \end{tabular}
            }
        \end{minipage}
    \end{tabular}
    \caption{
    Continuous depth-wise batching (CDB) with early exiting enables Recursive Transformers to theoretically achieve significant throughput improvements. Throughput\,(tokens/sec) was averaged across SlimPajama, RedPajama, and PG19, and then normalized to the throughput of the vanilla Pythia model. The accompanying table gives detailed throughout and performance measurements  for Gemma. $\Delta_V$ measures throughput relative to the vanilla Gemma model, while $\Delta_{Seq}$ measures throughput relative to the vanilla Gemma model with continuous sequence-wise batching (CSB).
    }
    \label{fig_rrt:early_exit_throughput}
\end{figure}

\paragraph{How we theoretically approximate actual throughput.}
As developing practical early-exiting algorithms is beyond the scope of this work, we present hypothetical throughput improvements based on an oracle-exiting approach\,\citep{DBLP:conf/nips/SchusterFG0B0TM22, DBLP:conf/emnlp/BaeKSY23}. This assumes that tokens exit at the earliest looping block where their prediction aligns with the final loop's prediction. We simulated the generation of language modeling datasets as if they were generated by our models, to obtain the exit trajectory for each token. Then, we measured the average per-token generation time under specific constraints, such as different memory limit or context lengths. Using these measurements and the exit trajectory data, we conducted simulations to estimate theoretical throughput. Detailed explanations and limitations are discussed in Appendix\,\ref{app_rrt:hypothetical_generation_speedup}.\looseness=-1

\paragraph{Continuous depth-wise batching paired with early-exiting can further boost throughput.}

Figure\,\ref{fig_rrt:early_exit_throughput} illustrates the throughput of our proposed models and the vanilla Transformer across three architectures. We consistently achieve higher speeds than the vanilla models by combining continuous depth-wise batching with early-exiting, even surpassing those with continuous sequence-wise batching\,\citep{DBLP:conf/osdi/YuJKKC22, DBLP:conf/sosp/KwonLZ0ZY0ZS23}. In particular, Recursive models demonstrate up to 2.66$\times$ speedup in generation compared to vanilla counterparts. Additionally, the recursive Gemma model significantly outperforms the vanilla pretrained Pythia model, with nearly 4$\times$ improvement in throughput. Relaxed recursive models show a clear trade-off between achievable few-shot performance and throughput, modulated by the degree of relaxation through the LoRA ranks. This characteristic enables flexible model selection tailored to specific deployment scenarios. Comprehensive results are presented in Tables\,\ref{tab_rrt:final_performance_throughput} and \ref{tab_rrt:final_performance_throughput_16gb}.

\begin{tcolorbox}[
    enhanced,
    colback=blue!5!white, 
    colframe=black, 
    width=\textwidth,
    coltitle=white,
    title=\textbf{Takeaways for Continuous Depth-wise Batching},
    top=3mm, bottom=2mm,
    attach boxed title to top left={yshift=-2.8mm, xshift=4mm},
    boxed title style={colback=black, colframe=black, boxrule=0mm, 
    toptitle=1mm, bottomtitle=1mm}
]
We analyze the potential for throughput improvement in the Recursive Transformer via continuous depth-wise batching, a novel inference paradigm. In theory, we find that we can achieve up to 2-3$\times$ speedup compared to a vanilla Transformer. This even outperforms the throughput gain achieved by existing continuous sequence-wise batching methods in vanilla models.
\end{tcolorbox}

\section{Conclusion and Future Work}

In this work, we introduced Recursive Transformers, in which we compress LLMs via parameter sharing across recursively looped blocks of layers. Additionally, we presented a novel relaxation strategy that allows for low-rank deltas between shared layers by integrating layer-specific LoRA modules into the fully-tied structure. Through novel initialization techniques for looped layers and LoRA modules, we achieved significant performance improvements that closely approximate the original pretrained model. Finally, by exploiting the recursive patterns and an early-exiting approach, we propose a continuous depth-wise batching paradigm tailored for efficient serving systems of Recursive Transformers. We theoretically demonstrated that an oracle-exiting strategy can yield substantial throughput gains, reaching up to 2-3$\times$ speedup. This work motivates further research on recursive patterns in modern LLMs such as:\looseness=-1

\paragraph{Compatibility with sparse designs.}

Sparsity-based approaches, such as pruning~\citep{DBLP:journals/corr/HanPTD15}, quantization~\citep{DBLP:conf/cvpr/JacobKCZTHAK18}, or layer-skipping mechanisms~\citep{DBLP:journals/corr/abs-2404-02258}, recently also give promising model compression results. In fact, many of these techniques are complementary to our approach: for example, we can seamlessly have a recursive, \textit{sparse} architecture. In this work, we rather choose to focus on recursive dense designs (a domain that remains relatively unexplored) that also have very promising, practical performance traits (i.e., allowing for continuous depth-wise batching for faster throughput). That said, while in this work we take the first step at studying Relaxed Recursive Transformer with dense Transformer layers, we do believe that incorporating Mixture-of-expert~\citep{DBLP:journals/jmlr/FedusZS22}, activation-skipping~\citep{DBLP:conf/icml/LiuWDZY0S0TRC23} and SSM components~\citep{DBLP:journals/corr/abs-2405-16712} within the looped blocks are promising directions for future research.

\paragraph{Latent reasoning via recurrent depth.}

Beyond efficiency gains through down-scaling materialized parameters with recursive patterns, an alternative research direction lies in scaling-up recurrent depth to facilitate latent reasoning. Specifically, recurrent computation can manifest thinking vertically by processing internal hidden states at each depth. One promising approach involves leveraging contemplation tokens~\citep{pfau2024let, DBLP:conf/iclr/GoyalJRMKN24} or latent (continuous) space representations~\citep{hao2024training, cheng2024compressed} to enhance reasoning in mathematical and code generation tasks. Another valuable direction focuses on enhancing the efficiency and training stability of approaches that recursively scale-up depth, building upon concepts of deep thinking~\citep{DBLP:conf/nips/SchwarzschildBG21, geiping2025scaling}.

\paragraph{Scaling up Recursive Transformers.}

Scaling our approach to larger LLMs (7B and beyond) is a promising avenue for future research. While our methodology is expected to remain effective, achieving comparable performance may require significantly higher uptraining costs. Increased model size offers the potential for a reduced memory footprint from recursive patterns; however, it is unclear whether this translates to larger batch sizes, given the corresponding increase in hidden dimensions. Nevertheless, our continuous depth-wise batching will yield considerable gains in serving efficiency.

\paragraph{Beyond hypothetical generation speedup.}

Our oracle-exiting approach assumes any intermediate prediction matching the final output can be exited. However, accurate throughput measurement requires confidence-based early-exiting algorithms~\citep{DBLP:conf/nips/SchusterFG0B0TM22, DBLP:conf/emnlp/BaeKSY23}. Moreover, practical deployment needs to address decoding bottlenecks like key-value cache computation for exited tokens in remaining loops. Nevertheless, there are potential solutions: for example, the missing KV cache computations can be addressed by leveraging continuous depth-wise batching, allowing the KV cache for exited positions in subsequent loops to be performed in parallel with the computations for the next sequence sample. Moreover, we can explore key-value cache sharing strategies~\citep{DBLP:journals/corr/abs-2405-05254, DBLP:journals/corr/abs-2405-12981} for future work.\looseness=-1

\paragraph{Efficient serving of multi-LoRA layers.}

Relaxed models require the computation of distinct LoRA modules during batched inference, akin to multi-task learning~\citep{DBLP:conf/coling/FengHZHW24, wang2023customizable}, hindering parallel computation. We concatenated LoRA weights into a single weight to improve efficiency over sequential computation, yet it introduces redundancy. To mitigate this, we can explore optimized CUDA kernels for LoRA serving~\citep{DBLP:journals/corr/abs-2311-03285, chen2024punica} and parallelization across accelerators, inspired by distributed training for Mixture of Experts~\citep{DBLP:journals/jmlr/FedusZS22, gale2023megablocks}.\looseness=-1

\clearpage

\section{Appendix}

\subsection{Expanded Results of Initialization Methods for Looped Layers}
\label{app_rrt:initialization}

\paragraph{Ablation study of Stepwise method.}

\begin{figure}[h]
    \centering
    \begin{subfigure}[t]{0.6\textwidth}
        \includegraphics[width=\textwidth]{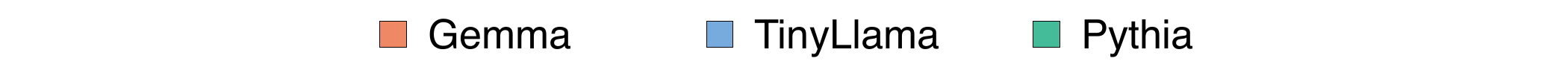}
    \end{subfigure}
    \\
    \centering
    \begin{subfigure}[t]{0.327\textwidth}
        \includegraphics[width=\textwidth]{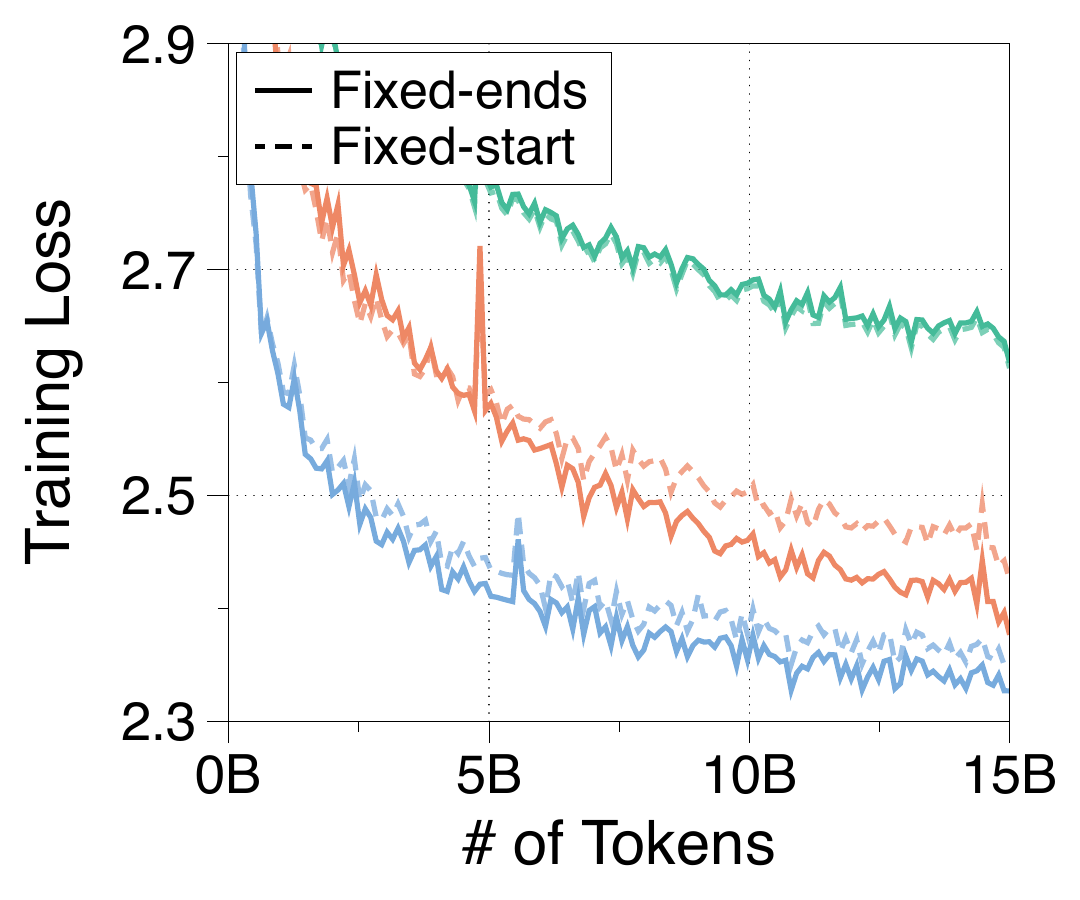}
        \subcaption{Stepwise ablations}
        \label{fig_rrt:stepwise_abl_app}
    \end{subfigure}
    \hspace{10pt}
    \centering
    \begin{subfigure}[t]{0.3\textwidth}
        \includegraphics[width=\textwidth]{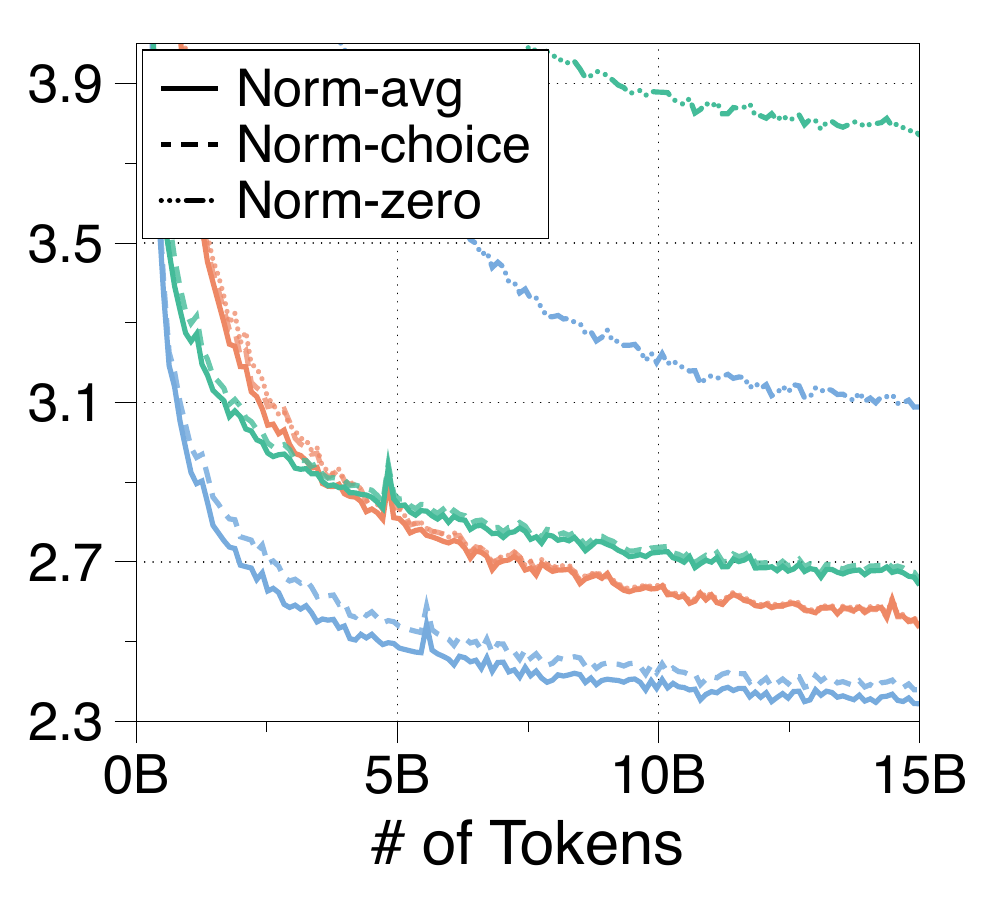}
        \subcaption{Average ablations}
        \label{fig_rrt:average_abl_app}
    \end{subfigure}
    \caption{
    Training loss curves of Stepwise and Average initialization variants across three models with two blocks. {(a)} ``Fixed-start'' indicates that the first layer of the pretrained model is selected initially, and subsequent layers are repeatedly chosen at a fixed interval. ``Fixed-ends'' means that the first and last layers are included, and intermediate layers are selected at specific step intervals. {(b)} When initializing the weights of normalization layer (RMSNorm in Gemma and TinyLlama, and LayerNorm in Pythia), we consider whether to average the weights\,(Norm-avg), select a single layer's weights\,(Norm-choice), or use zero initialization\,(Norm-zero). 
    }
    \label{fig_rrt:initialization_ablation_app}
\end{figure}

We initially hypothesized that the Stepwise method's performance could be significantly influenced by the specific rule used for layer selection from the pretrained model. To investigate this, we conducted a controlled experiment (illustrated in Figure\,\ref{fig_rrt:stepwise_abl_app}), where layers were selected at certain intervals starting from the first layer. We then varied whether the final layer of the pretrained model was included in the initialization or not. While a Pythia model showed no significant differences in training loss or few-shot performance, other models like Gemma exhibited superior results when both the first and last layers were preserved. This observation aligns well with prior work suggesting that maintaining the weights of the first and last layers during depth up-scaling for LLMs can yield performance benefits~\citep{DBLP:conf/naacl/KimKPLSKKKLKAYLPGCLK24}.

\paragraph{Ablation study of Average method.}

The Average initialization method exhibited notably poor performance, particularly when applied to the Gemma model. We hypothesized that this could be attributed to instability in the model's learned distribution, potentially arising from averaging of normalization layer weights. Relatedly, several studies~\citep{csordas2024moeut, shim2024leveraging, mohtashami2023cotformer} have explored the careful design of layer normalization in parameter-shared models. To investigate this further, we experimented with three different methods for initializing normalization weights, as outlined in Figure\,\ref{fig_rrt:average_abl_app}: averaging weights (Norm-avg), selecting weights from a single layer (Norm-choice), and zero initialization (Norm-zero). The performance trend observed among these methods varied across different model architectures. However, zero initialization of normalization layers resulted in a huge performance drop in certain architectures like TinyLlama and Pythia. Conversely, we observed no big difference between averaging and single-layer selection, suggesting that any form of distillation of the normalization weights appears to be sufficient for maintaining performance.\looseness=-1

\paragraph{Overall comparison of training perplexity.}

\begin{figure}[ht]
    \centering
    \begin{subfigure}[t]{\textwidth}
        \includegraphics[width=\textwidth]{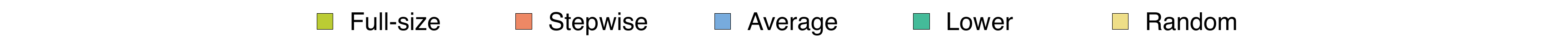}
    \end{subfigure}
    \centering
    \begin{subfigure}[t]{0.26\textwidth}
        \includegraphics[width=\textwidth]{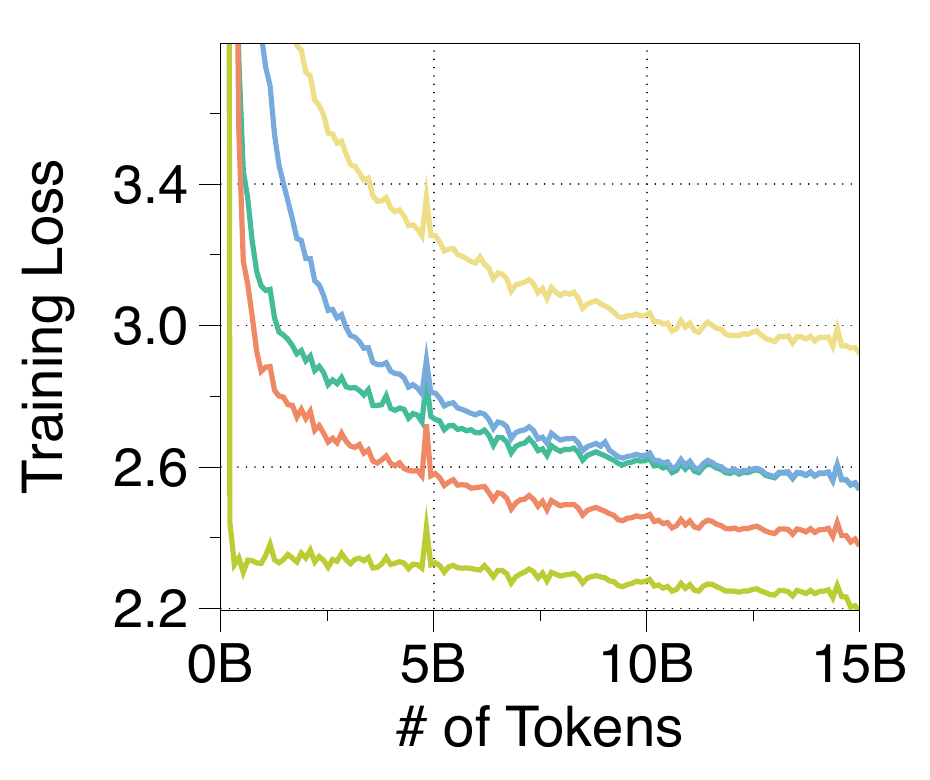}
        \subcaption{Gemma (2 blocks)}
    \end{subfigure}
    \centering
    \begin{subfigure}[t]{0.238\textwidth}
        \includegraphics[width=\textwidth]{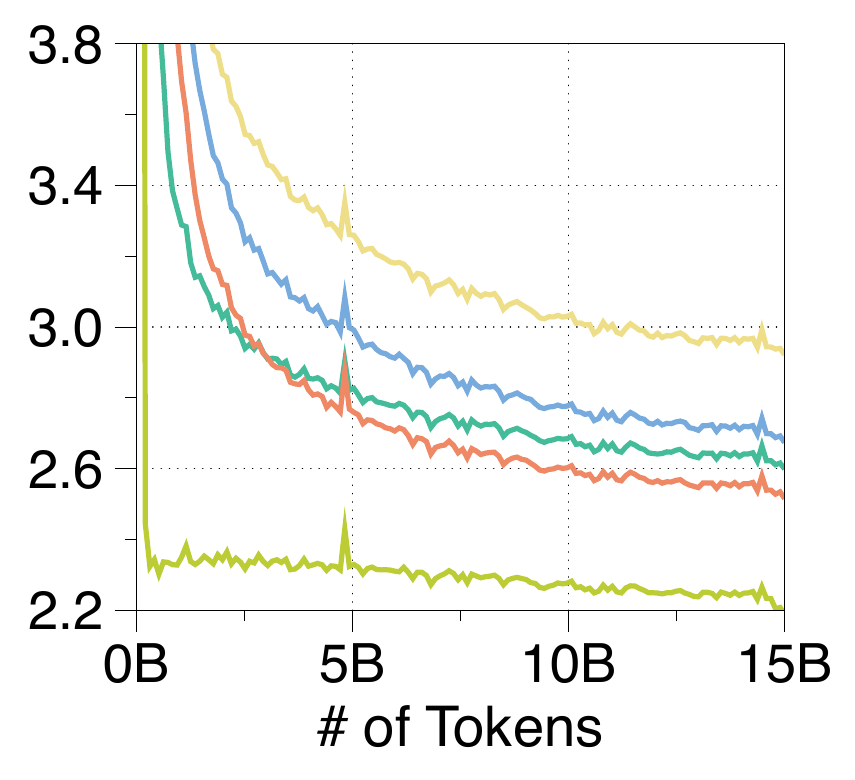}
        \subcaption{Gemma (3 blocks)}
    \end{subfigure}
    \centering
    \begin{subfigure}[t]{0.238\textwidth}
        \includegraphics[width=\textwidth]{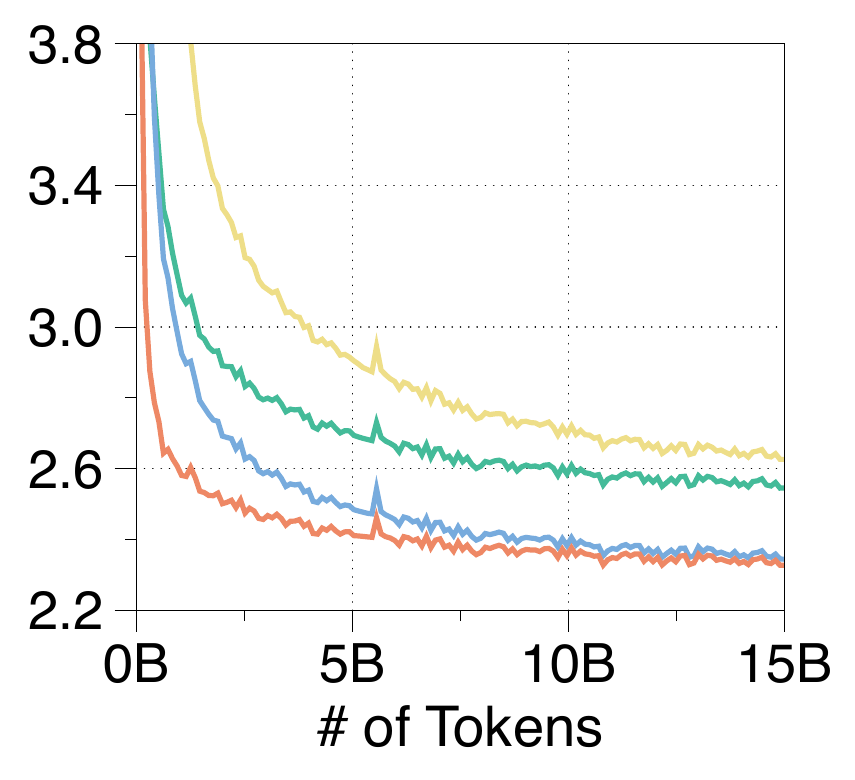}
        \subcaption{TinyLlama (2 blocks)}
    \end{subfigure}
    \centering
    \begin{subfigure}[t]{0.238\textwidth}
        \includegraphics[width=\textwidth]{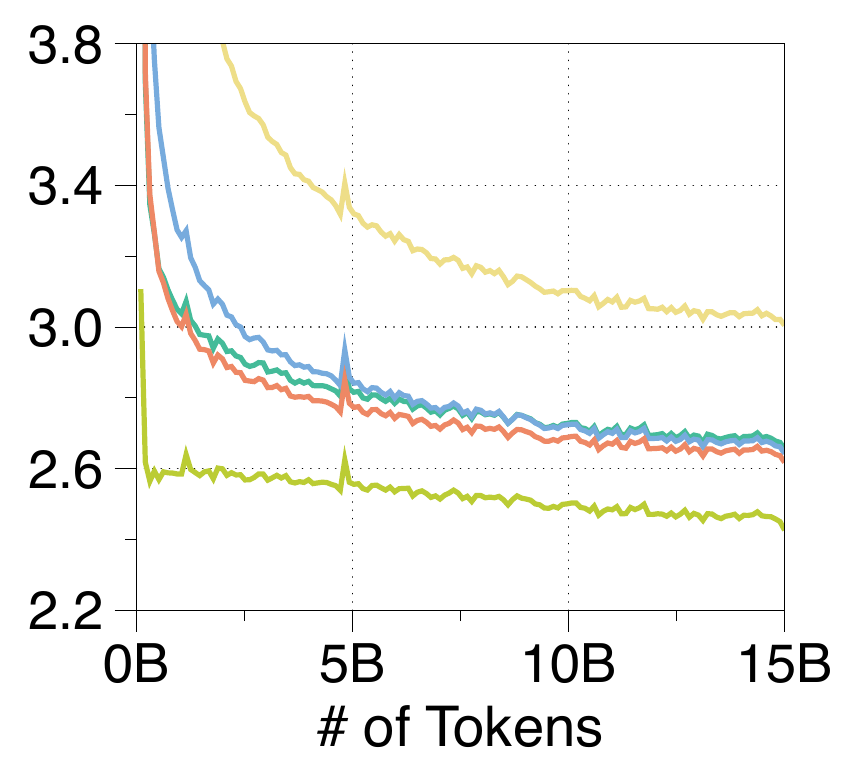}
        \subcaption{Pythia (2 blocks)}
    \end{subfigure}
    \caption{Training loss curves of Recursive Transformers using various initialization. We omitted a separate curve for the full-size TinyLlama model, as we used the original pretrained model as the full-size baseline because both pretraining and uptraining datasets are same as SlimPajama. 
    }
    \label{fig_rrt:training_loss_app}
\end{figure}

Figure\,\ref{fig_rrt:training_loss_app} presents a comparative analysis of training loss across three model architectures and varying looping blocks, incorporating our proposed initialization methodologies. To set an upper bound on performance, we utilized a full-size model further uptrained on SlimPajama, accounting for the distribution shift between uptraining and pretraining data. Additionally, we trained a Recursive Transformer with a random initialization, ensuring its exclusive reliance on the recursive architecture without leveraging any pretrained weights. While some variance was observed across architectures, all proposed methods utilizing pretrained model weights demonstrated significantly superior performance compared to random initialization. Notably, the Stepwise method consistently achieved the best performance across diverse settings. Although the full-size model's performance was considerably higher, bridging this gap with only 15 billion tokens of uptraining represents a remarkable achievement.\looseness=-1

\paragraph{Overall comparison of few-shot performance.}

\begin{figure}[ht!]
    \centering
    \begin{subfigure}[t]{\textwidth}
        \includegraphics[width=\textwidth]{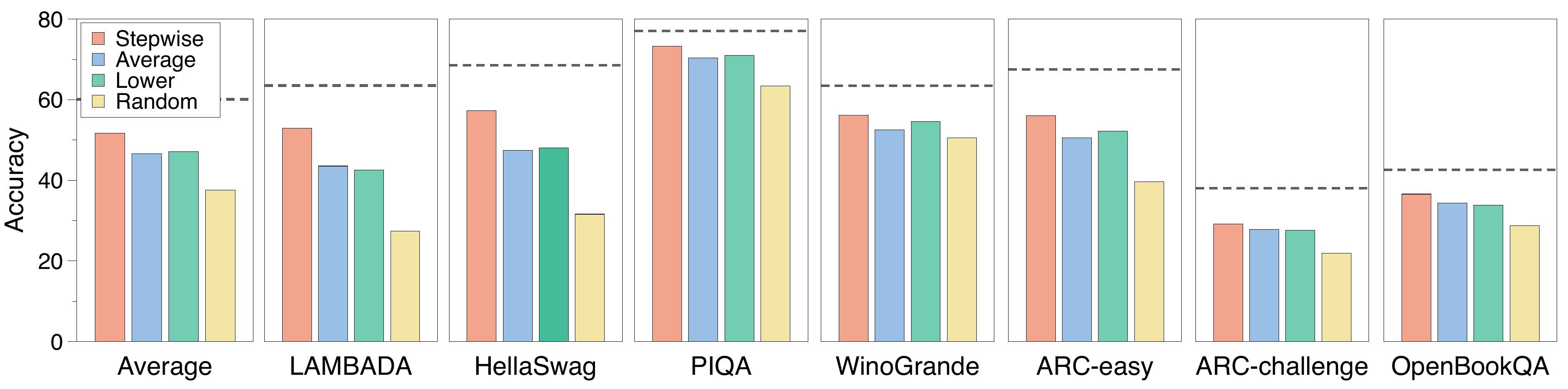}
        \caption{Recursive Gemma with 2 blocks}
        \vspace{5pt}
    \end{subfigure}
    \centering
    \begin{subfigure}[t]{\textwidth}
        \includegraphics[width=\textwidth]{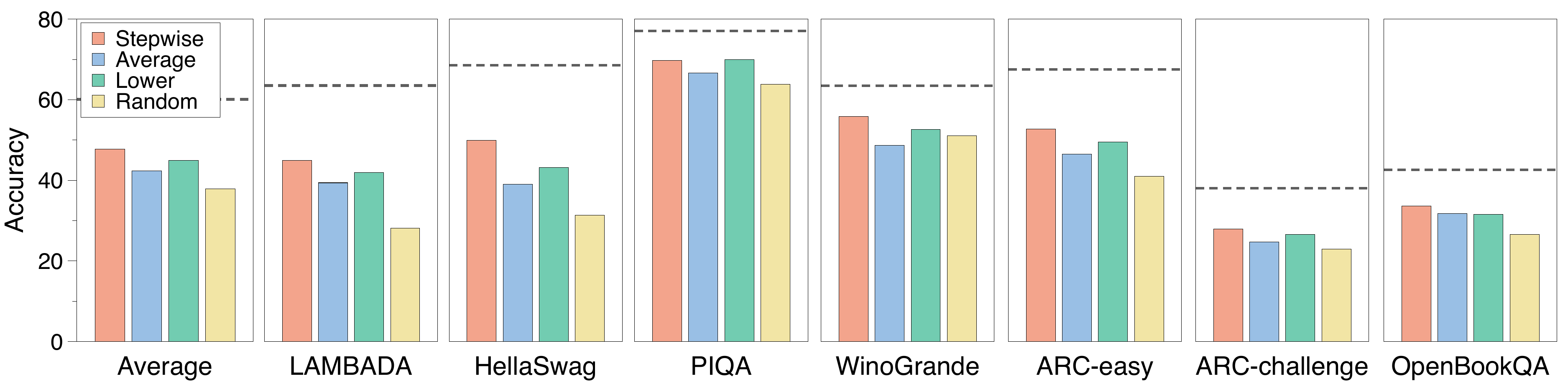}
        \caption{Recursive Gemma with 3 blocks}
        \vspace{5pt}
    \end{subfigure}
    \centering
    \begin{subfigure}[t]{\textwidth}
        \includegraphics[width=\textwidth]{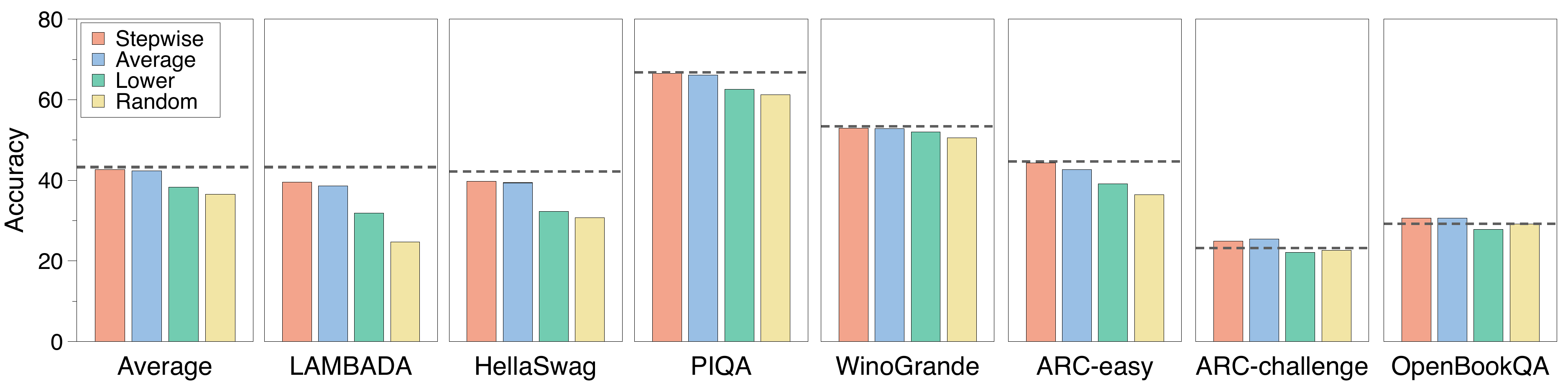}
        \caption{Recursive TinyLlama with 2 blocks}
        \vspace{5pt}
    \end{subfigure}
    \centering
    \begin{subfigure}[t]{\textwidth}
    \includegraphics[width=\textwidth]{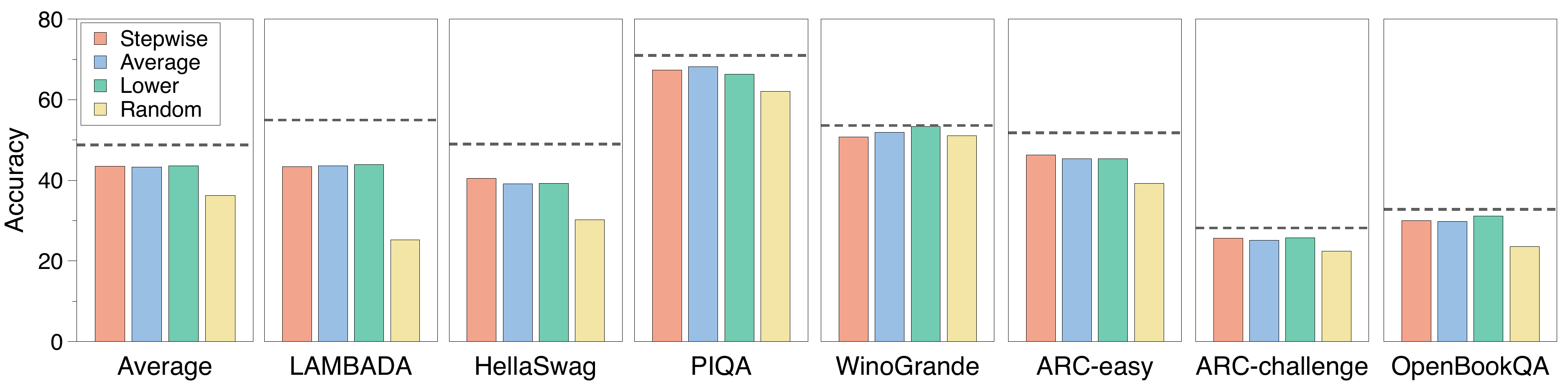}
        \caption{Recursive Pythia with 2 blocks}
        \vspace{5pt}
    \end{subfigure}
    \caption{Few-shot performance on seven benchmarks and their average accuracy based on four looping initialization methods. Full-size model performance is represented by a gray dotted line.
    }
    \label{fig_rrt:fewshot_bar_total_app}
\end{figure}

Few-shot performance exhibited a consistent trend with training perplexity. Table\,\ref{tab_rrt:initialization_total_app} provides a comparative summary of the proposed looping initialization methods against the full-size model, the reduced-size model, and Recursive Transformers utilizing random initialization. Moreover, Figure\,\ref{fig_rrt:fewshot_bar_total_app} visually illustrates the performance differences across different few-shot datasets. Notably, the Stepwise method consistently demonstrated the best performance, showing a performance improvement of up to 14.1\%p compared to random initialization.

\begin{table}[ht!]
    \small
    \centering
    \resizebox{\textwidth}{!}{
    \setlength{\tabcolsep}{4pt}
    \begin{tabular}{l|c|cc|cc|rrr|ccccccc|cc}
    \toprule
     & & \multicolumn{2}{c|}{\textbf{Uptrain}} & \multicolumn{2}{c|}{\textbf{Looping}} & \multicolumn{3}{c|}{\textbf{Perplexity\,$\downarrow$}} & \multicolumn{9}{c}{\textbf{Few-shot Accuracy\,$\uparrow$}} \\
    \cmidrule(l{2pt}r{2pt}){3-4} \cmidrule(l{2pt}r{2pt}){5-6}  \cmidrule(l{2pt}r{2pt}){7-9}  \cmidrule(l{2pt}r{2pt}){10-18} 
     \textbf{Models} & N-emb & PT & $N_{tok}$ & Block & Init & SlimP & RedP & PG19 & LD & HS & PQ & WG & ARC-e & ARC-c & OB & Avg & $\Delta$  \\
    \midrule
    \multirow{11}{*}{Gemma} & 1.99B &  \cmark & 15B & - & - & 10.76 & 8.47 & 13.08 & 63.5 & 68.5 & 77.0 & 63.5 & 67.6 & 38.1 & {42.6} & 60.1  & - \\
     & 0.99B &  \xmark & 15B & - & - & 22.63 & 20.03 & 32.60 & 28.9 & 31.6 & 63.1 & 52.3 & 41.2 & 22.5 & 27.8 &  38.2 & -  \\
     & 0.66B &  \xmark & 15B & - & - & 24.44 & 21.69 & 36.03 & 27.2 & 30.6 & 63.8 & 50.5 & 40.6 & 22.0 & 27.0  & 37.4 & -   \\
     \cmidrule(l{2pt}r{2pt}){2-18} 
     \rowcolor[gray]{0.9}
     \cellcolor{white} & 0.99B & \cmark & 15B &  2 & Step & \textbf{12.85} & \textbf{10.29} & \textbf{16.21} & \textbf{53.0} & \textbf{57.3} & \textbf{73.2} & \textbf{56.2} & \textbf{56.1} & \textbf{29.2} & \textbf{36.6} & \textbf{51.7} & \textcolor{custom_green}{\!\!\!\textbf{+14.1}} \\
     & 0.99B & \cmark & 15B &  2 & Avg & 15.15 & 12.57 & 19.86 & 43.6 & 47.4 & 70.4 & 52.6 & 50.5 & 27.8 & 34.4 & 46.7 & \textcolor{custom_green}{\textbf{+9.1}}  \\
     & 0.99B & \cmark & 15B &  2 & Lower & 15.03 & 12.46 & 19.63 & 42.5 & 48.0 & 71.0 & 54.6 & 52.2 & 27.7 & 33.8 & 47.1 & \textcolor{custom_green}{\textbf{+9.5}}  \\
     & 0.99B & \xmark & 15B &  2 & Rand & 22.66 & 20.06 & 32.86 & 27.4 & 31.6 & 63.4 & 50.5 & 39.7 & 21.9 & 28.8 & 37.6 & -  \\
     \cmidrule(l{2pt}r{2pt}){2-18} 
     \rowcolor[gray]{0.9}
     \cellcolor{white} & 0.66B & \cmark & 15B &  3 & Step & \textbf{14.75} & \textbf{12.10} & \textbf{19.32} & \textbf{45.0} & \textbf{49.9} & \textbf{69.8} & \textbf{55.8} & \textbf{52.7} & \textbf{27.9} & \textbf{33.6} & \textbf{47.8} & \textcolor{custom_green}{\textbf{+9.9}} \\
     & 0.66B & \cmark & 15B &  3 & Avg & 17.45 & 14.65 & 23.63 & 39.4 & 39.0 & 66.6 & 48.7 & 46.5 & 24.7 & 31.8 & 42.4 & \textcolor{custom_green}{\textbf{+4.5}} \\
     & 0.66B & \cmark & 15B &  3 & Lower & 15.96 & 13.24 & 20.90 & 41.9 & 43.2 & 70.0 & 52.6 & 49.5 & 26.6 & 31.6 & 45.0 & \textcolor{custom_green}{\textbf{+7.1}} \\
     & 0.66B & \xmark & 15B &  3 & Rand & 22.67 & 20.09 & 32.77 & 28.1 & 31.4 & 63.8 & 51.1 & 41.0 & 23.0 & 26.6 & 37.9 & -  \\
     \midrule
    & 0.97B & \cmark & - & - & - & 12.26 & 9.37 & 11.94 & 43.3 & 42.2 & 66.8 & 53.4 & 44.7 & 23.2 & 29.2 & 43.3 & -  \\
     & 0.48B &  \xmark & 15B & - & - & 16.61 & 15.66 & 20.27 & 22.3 & 30.0 & 60.9 & 50.6 & 37.0 & 23.0 & 28.0 & 36.0 & -  \\
     \cmidrule(l{2pt}r{2pt}){2-18} 
     \rowcolor[gray]{0.9}
     \cellcolor{white}TinyLlama &  0.48B  &   \cmark &  15B &   2 &   Step &   \textbf{11.61} &   \textbf{9.89}  &   \textbf{13.00} &  \textbf{ 39.6} &  \textbf{ 39.8} &   \textbf{66.5} &   \textbf{52.9} &   \textbf{44.3} &   \textbf{24.9} &   \textbf{30.6} &   \textbf{42.7} &   \textcolor{custom_green}{\textbf{+6.2}}  \\
     & 0.48B  & \cmark & 15B & 2 & Avg & 11.86 & 10.29  & 13.42 & 38.6 & 39.4 & 66.1 & 52.8 & 42.7 & 25.4 & \textbf{30.6} & 42.2 & \textcolor{custom_green}{\textbf{+5.7}} \\
     & 0.48B  & \cmark & 15B & 2 & Lower & 14.67 & 12.67  & 16.68 & 31.9 & 32.3 & 62.6 & 52.0 & 39.1 & 22.1 & 27.8 & 38.3 & \textcolor{custom_green}{\textbf{+1.8}} \\
     & 0.48B  & \xmark & 15B & 2 & Rand & 16.14 & 15.11 & 19.55 & 24.7 & 30.7 & 61.2 & 50.6 & 36.4 & 22.6 & 29.2 & 36.5 & - \\
     \midrule
    & 0.81B  & \cmark & 15B & - & - & 13.46 & 9.95 & 13.38 &  55.0 & 49.0 & 71.0 & 53.6 & 51.8 & {28.2} & 32.8 & {48.8} & - \\
     & 0.40B &  \xmark & 15B & - & - & 25.69 & 20.00 & 32.08 & 24.3 & 30.0 & 61.9 & 50.7 & 38.3 & 22.3 & 26.0  & 36.2 & -  \\
     \cmidrule(l{2pt}r{2pt}){2-18} 
     \rowcolor[gray]{0.9}
    \cellcolor{white}Pythia &  0.40B  &   \cmark &   15B &   2 &   Step &   \textbf{16.38} &   \textbf{12.37} &   \textbf{17.74} &   43.4 &   \textbf{40.5} &   67.4 &   50.8 &   \textbf{46.3} &   25.7 &   30.0 &   43.5 &   \textcolor{custom_green}{\textbf{+7.3}} \\
     & 0.40B  & \cmark & 15B & 2 & Avg & 16.76 & 12.76 & 18.63 & 43.6 & 39.1 & \textbf{68.2} & 51.9 & 45.4 & 25.1 & 29.8 & 43.3 & \textcolor{custom_green}{\textbf{+7.1}} \\
     & 0.40B  & \cmark & 15B & 2 & Lower & 17.04 & 12.62 & 18.44 & \textbf{43.9} & 39.2 & 66.3 & \textbf{53.4} & 45.4 & \textbf{25.8} & \textbf{31.2} & \textbf{43.6} & \textcolor{custom_green}{\textbf{+7.4}} \\
     & 0.40B  & \xmark & 15B & 2 & Rand & 24.45 & 18.93 & 29.63 & 25.2 & 30.2 & 62.1 & 51.1 & 39.2 & 22.4 & 23.6 & 36.2 & - \\
    \bottomrule
    \end{tabular}
    }
    \caption{
    Evaluation results of various initialization methods for looped layers. We indicate whether pretrained weights are used and the number of uptraining tokens. Perplexity is evaluated on test sets of three language modeling datasets, and accuracy is evaluated on seven few-shot benchmarks. Delta values\,($\Delta$) show improvements over random initialization.
    }
    \label{tab_rrt:initialization_total_app}
\end{table}

\clearpage
\paragraph{Comparison across various base model sizes.}

We observed a consistent superiority of Stepwise initialization strategy for recursive conversion across both 1B and 2B model scales. To further evaluate on a wide range of base model sizes, we additionally experimented with two smaller model sizes, Pythia 410M and 160M. Since we uptrained models on the Pile dataset~\citep{gao2020pile}, the pretraining corpus of the original Pythia model, we set the original model performance as the baseline for comparison. The results in Table~\ref{tab_rrt:pythia_160_410} further validate the superior performance of the Stepwise method for looped layer initialization. These findings reinforce the robustness of our key observations regarding initialization methods for recursive conversion, complementing our original extensive experiments.\looseness=-1

\begin{table}[ht]
    \small
    \centering
    \resizebox{\textwidth}{!}{
    \setlength{\tabcolsep}{4pt}
    \begin{tabular}{c|cc|cc|c|ccccccc|c}
    \toprule
     & \multicolumn{2}{c|}{\textbf{Uptrain}} & \multicolumn{2}{c|}{\textbf{Looping}} & \textbf{PPL}\,$\downarrow$  & \multicolumn{8}{c}{\textbf{Few-shot Accuracy\,$\uparrow$}} \\
    \cmidrule(l{2pt}r{2pt}){2-3} \cmidrule(l{2pt}r{2pt}){4-5}  \cmidrule(l{2pt}r{2pt}){6-6}  \cmidrule(l{2pt}r{2pt}){7-14} 
      N-emb & PT & $N_{tok}$ & Block & Init  & Pile & LD & HS & PQ & WG & ARC-e & ARC-c & OB & Avg  \\
    \midrule    
    300M & \cmark & - & - & - & - & 44.96 & 40.97 & 66.97 & 53.28 & 44.40 & 25.51 & 30.20 & 43.76 \\
    \midrule
    150M & \cmark & 15B & 2 & Step & \textbf{11.03} & \textbf{43.41} & \textbf{35.59} & \textbf{64.58} & \textbf{53.04} & 41.58 & 23.81 & \textbf{28.80} & \textbf{41.54} \\
    150M & \cmark & 15B & 2 & Lower & 11.47 & 42.98 & 34.32 & 63.93 & 52.41 & \textbf{42.34} & 24.15  & 25.00 & 40.73  \\
    150M & \cmark & 15B & 2 & Avg & 11.55 & 39.84  & 34.17  & 64.31  & 52.25 & 41.04  & \textbf{24.66} & 26.60 & 40.41  \\
    \midrule
    \,\,\,85M & \cmark & - & - & - & - & 13.53 & 30.67 & 58.22 & 48.62 & 36.62  & 25.00  & 28.60  & 34.47  \\
    \midrule
    \,\,\,43M & \cmark & 15B & 2 & Step & \textbf{15.93}  & 21.02  & 29.28  & 60.01  & 48.93 & 37.92 & \textbf{23.98} & \textbf{28.00} & 35.59 \\
    \,\,\,43M & \cmark & 15B & 2 & Lower & 16.19 & 21.46  &\textbf{ 29.61}  & 59.90  & \textbf{50.67} & \textbf{38.52} & 22.95  & \textbf{28.00}  & \textbf{35.87}  \\
    \,\,\,43M & \cmark & 15B & 2 & Avg & 16.12  & \textbf{22.36}  & 29.07  & \textbf{60.17}  & 49.96 & 37.24  & 23.29  & 26.60  & 35.53  \\
    \bottomrule
    \end{tabular}
    }
    \caption{
    Comparison between initialization methods for looped layers on Pythia 410M and 160M. Uptraining was performed using the Pile dataset, which was also used for pretraining the original Pythia model. In light of the inherent randomness in few-shot accuracy, a comparison based on the perplexity\,(PPL) would provide a more stable measure of performance.\looseness=-1
    }
    \label{tab_rrt:pythia_160_410}
\end{table}

\paragraph{Individual contributions of leveraging pretrained weights and recursive patterns.}

To understand the performance of our Recursive Transformer, we established two non-recursive baselines: full-size model and reduced-size model. The reduced size model performance is meant to serve as a lower bound which we can use to better judge the efficacy of (1) unique looping and parameter sharing techniques that are made possible by our approach and (2) leveraging pretrained layers. To further ablate the effect of each of two components, we conducted experiments using the Pythia 410M model presented in Table~\ref{tab_rrt:individual_effect}. Intuitively, we observed significant performance gains by leveraging pretrained layers, with further improvement achieved through recursion. We believe this additional experiment provides valuable insight into the performance contributions of the two approaches proposed for constructing Recursive Transformers.\looseness=-1

\begin{table}[h]
    \small
    \centering
    \resizebox{\textwidth}{!}{
    \setlength{\tabcolsep}{4pt}
    \begin{tabular}{c|cc|cc|c|ccccccc|c}
    \toprule
     & \multicolumn{2}{c|}{\textbf{Uptrain}} & \multicolumn{2}{c|}{\textbf{Looping}} & \textbf{PPL}\,$\downarrow$  & \multicolumn{8}{c}{\textbf{Few-shot Accuracy\,$\uparrow$}} \\
    \cmidrule(l{2pt}r{2pt}){2-3} \cmidrule(l{2pt}r{2pt}){4-5}  \cmidrule(l{2pt}r{2pt}){6-6}  \cmidrule(l{2pt}r{2pt}){7-14} 
      N-emb & PT & $N_{tok}$ & Block & Init  & Pile & LD & HS & PQ & WG & ARC-e & ARC-c & OB & Avg  \\
    \midrule    
    300M & \cmark & - & - & - & - & 44.96 & 40.97 & 66.97 & 53.28 & 44.40 & 25.51 & 30.20 & 43.76 \\
    \midrule
    150M & \xmark & 15B & - & - & 14.11 & 31.48 & 29.53 & 61.37 & \textbf{52.49} & 39.14 & 22.44  & 27.00 & 37.63 \\
    150M & \xmark & 15B & 2 & - &\textbf{ 13.81} & \textbf{31.55} & \textbf{29.94} & \textbf{62.30} & 50.88 & \textbf{40.28} & \textbf{23.98} & \textbf{28.20} & \textbf{38.02} \\
    \midrule
    150M & \cmark & 15B & - & Step & 11.48 & 40.48 & 34.19 & 63.42 & 50.99 & \textbf{41.84} & 23.12 & 28.40 & 40.35 \\
    150M & \cmark & 15B & 2 & Step & \textbf{11.03} & \textbf{43.41} & \textbf{35.59} & \textbf{64.58} & \textbf{53.04} & 41.58 & \textbf{23.81} & \textbf{28.80 }& \textbf{41.54} \\
    \bottomrule
    \end{tabular}
    }
    \caption{
    Performance of recursive and baseline models with Pythia 410M to investigate the individual contributions of pretrained weights and looping strategy. Uptraining was performed using the Pile dataset~\citep{gao2020pile}, which was also used for pretraining the original Pythia model.
    }
    \label{tab_rrt:individual_effect}
\end{table}

\subsection{Expanded Results of Relaxed Recursive Transformers}
\label{app_rrt:cross_layer_lora}

\paragraph{Training perplexity changes with LoRA modules.}

\begin{figure}[h]
    \centering
    \begin{subfigure}[t]{\textwidth}
        \includegraphics[width=\textwidth]{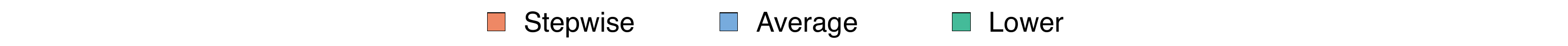}
    \end{subfigure}
    \centering
    \begin{subfigure}[t]{0.34\textwidth}
        \includegraphics[width=\textwidth]{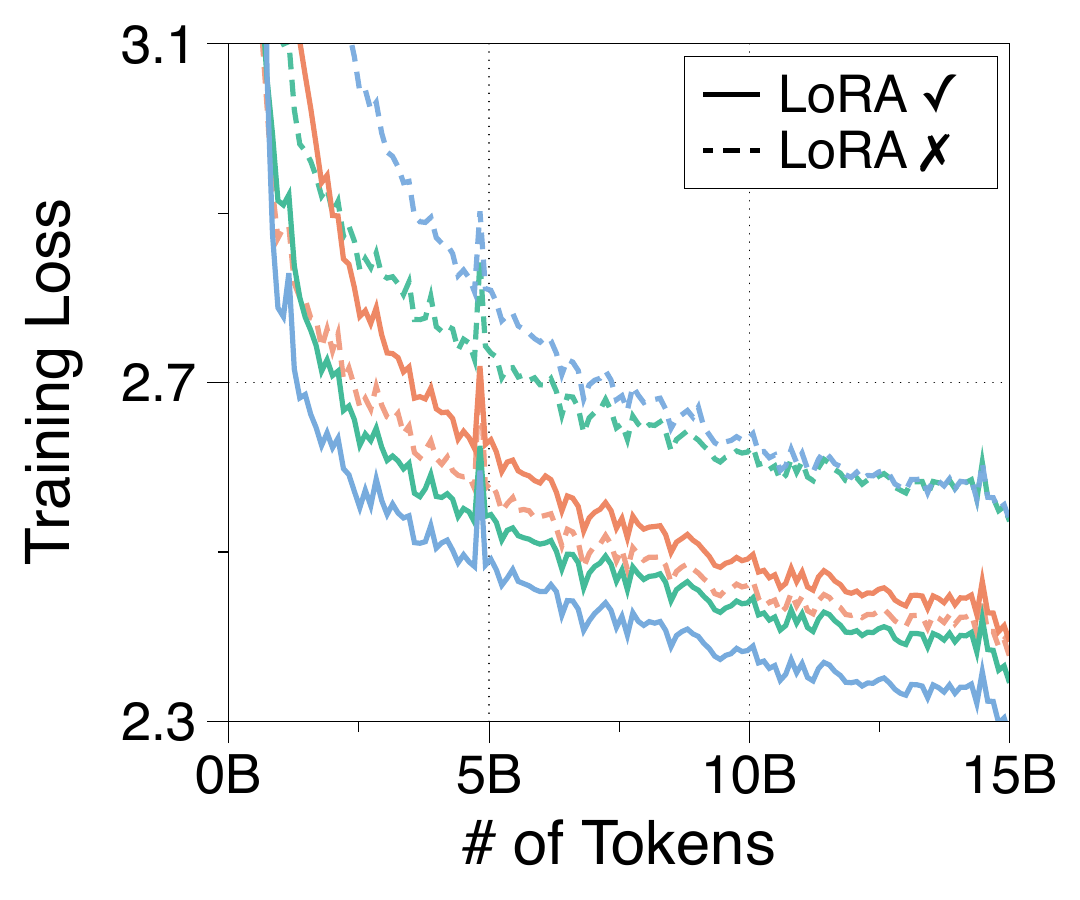}
        \subcaption{Gemma}
    \end{subfigure}
    \centering
    \begin{subfigure}[t]{0.31\textwidth}
        \includegraphics[width=\textwidth]{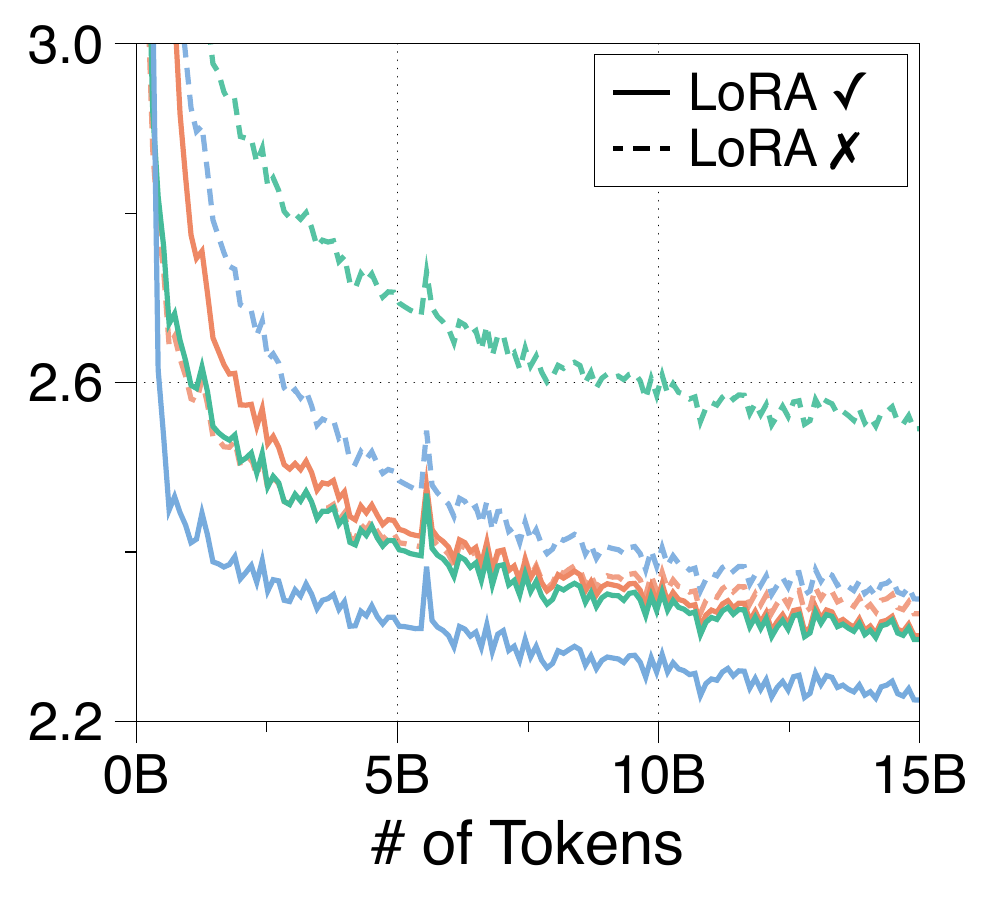}
        \subcaption{TinyLlama}
    \end{subfigure}
    \centering
    \begin{subfigure}[t]{0.31\textwidth}
        \includegraphics[width=\textwidth]{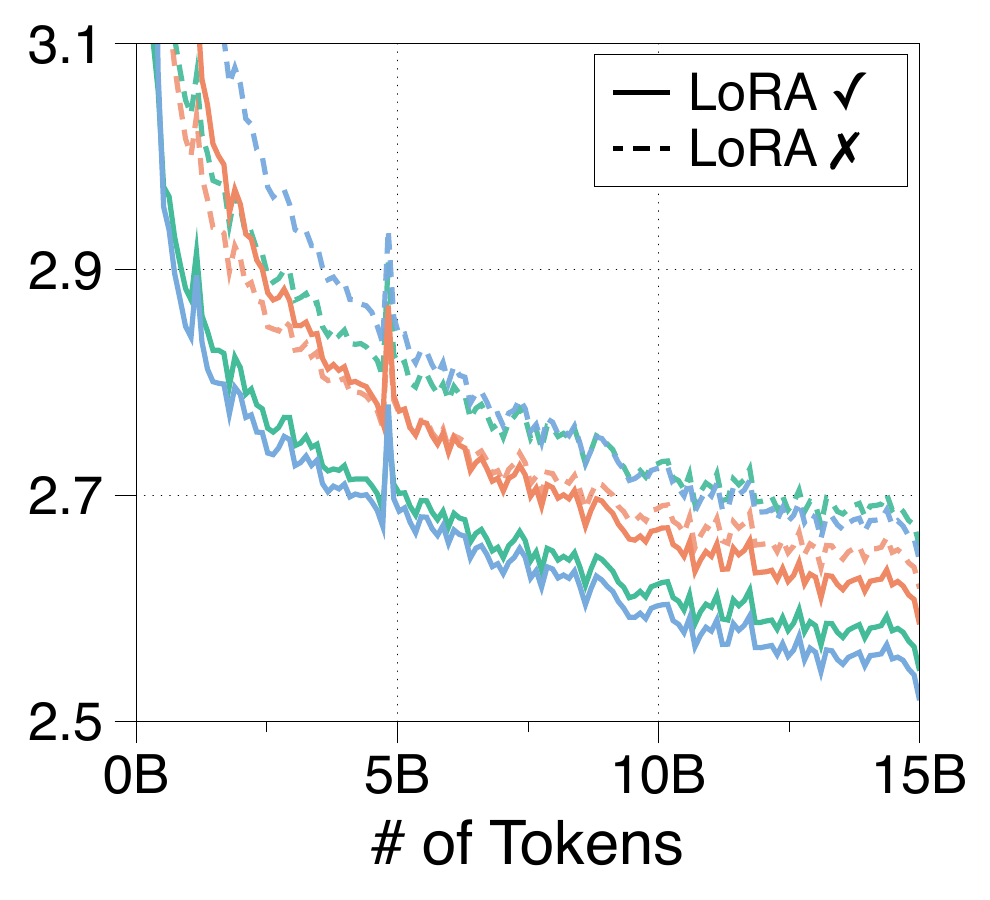}
        \subcaption{Pythia}
    \end{subfigure}
    \caption{
    Comparison of training loss for recursive and relaxed recursive models with two blocks. The LoRA rank is set to 512, and the SVD initialization method is used for LoRA modules.
    }
    \label{fig_rrt:lora_training_loss_app}
\end{figure}

Figure\,\ref{fig_rrt:lora_training_loss_app} illustrates the changes in training loss after incorporating the layer-wise LoRA modules. The Average and Lower initialization methods, when coupled with our proposed SVD-based initialization of LoRA modules, demonstrated significantly enhanced benefits. In particular, the Relaxed Recursive Transformer employing the Average method consistently outperformed the others. This suggests that it is considerably easier to learn the difference between the original pretrained weights and the averaged looped weights using low-rank matrices.

\paragraph{Comparison between SVD and zero initialization.}

\begin{figure}[h]
    \centering
    \begin{subfigure}[t]{0.34\textwidth}
        \includegraphics[width=\textwidth]{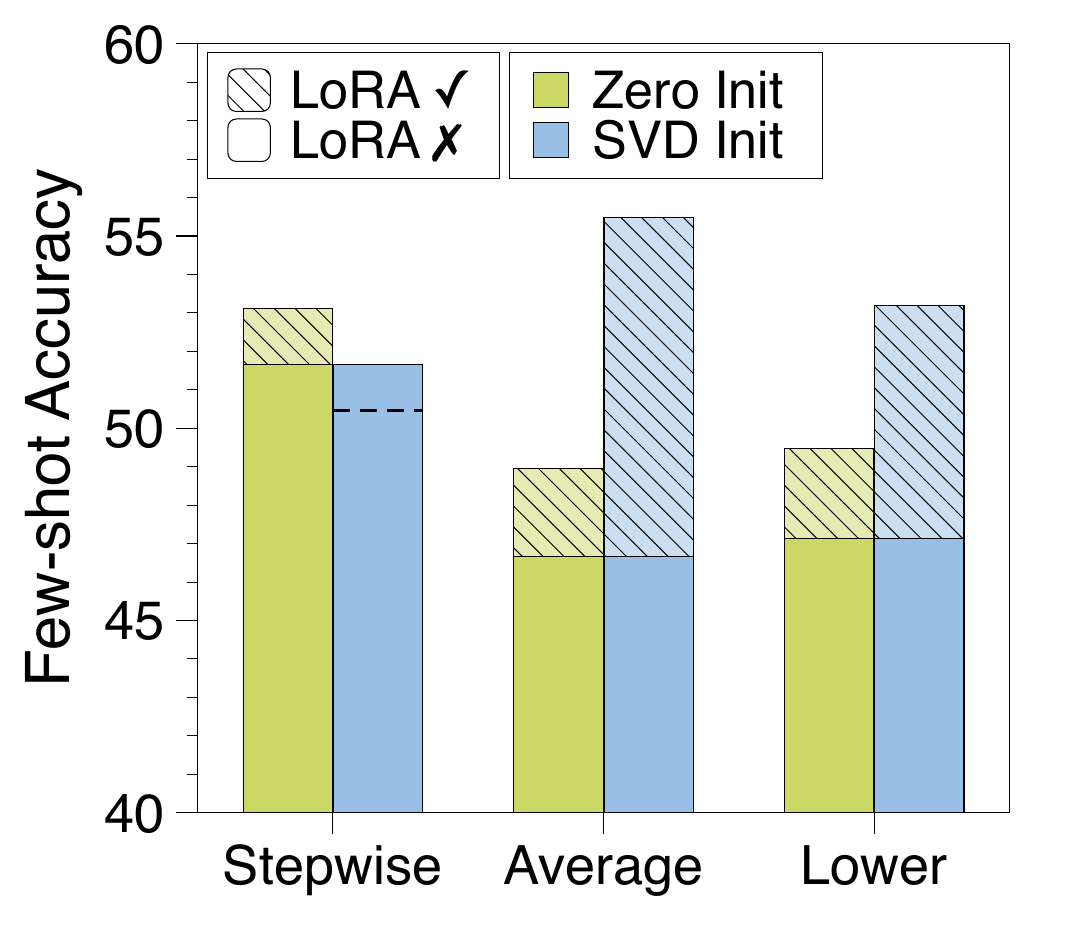}
        \subcaption{Gemma}
    \end{subfigure}
    \centering
    \begin{subfigure}[t]{0.31\textwidth}
        \includegraphics[width=\textwidth]{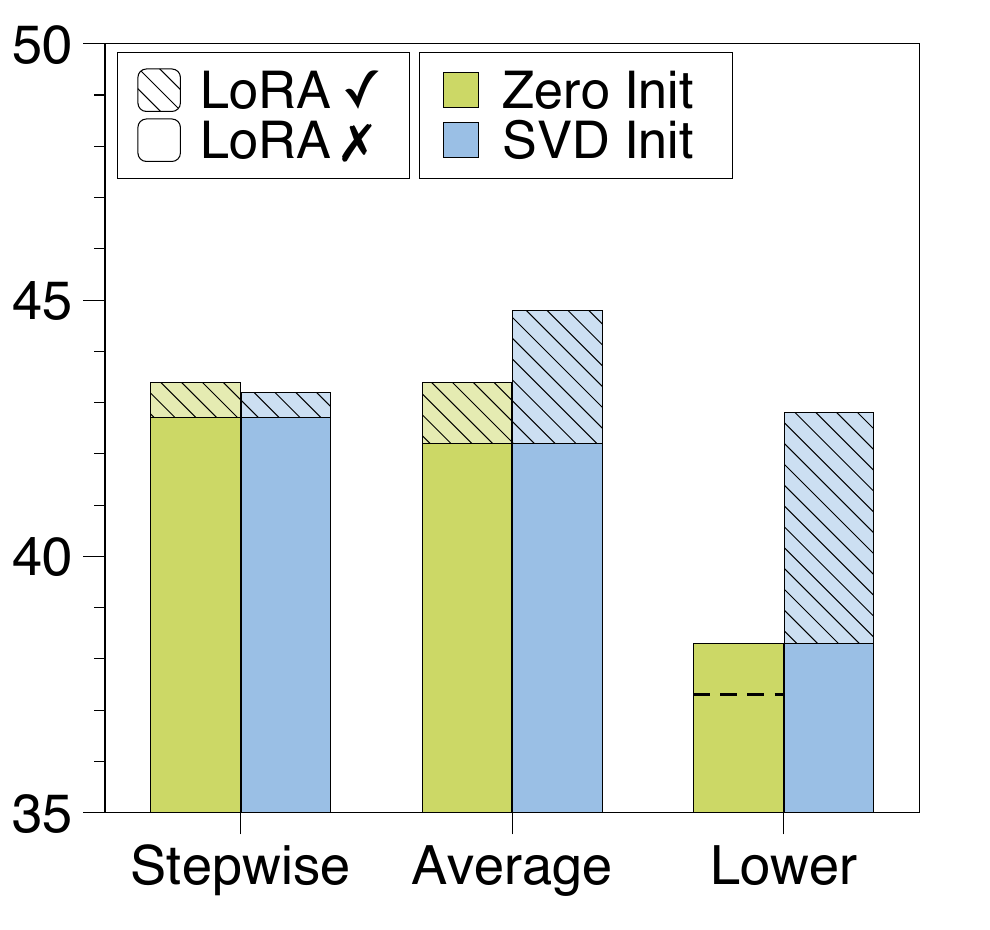}
        \subcaption{TinyLlama}
    \end{subfigure}
    \centering
    \begin{subfigure}[t]{0.31\textwidth}
        \includegraphics[width=\textwidth]{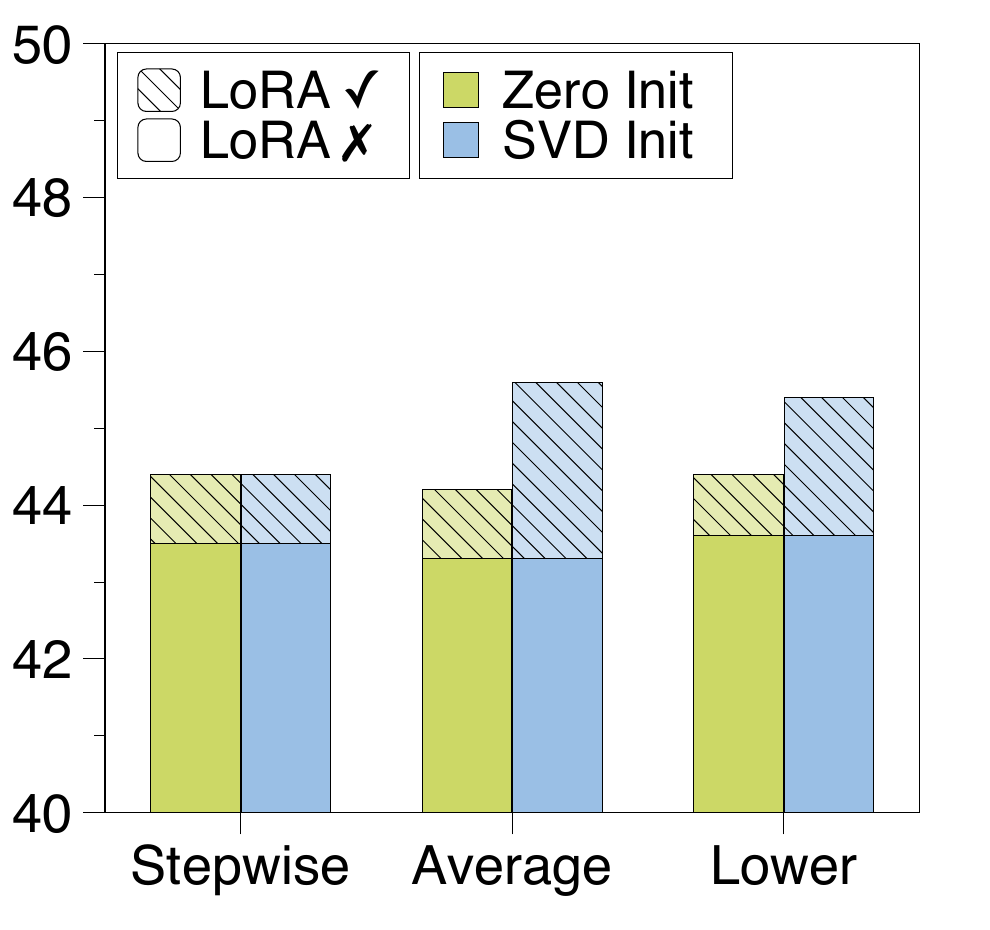}
        \subcaption{Pythia}
    \end{subfigure}
    \caption{Comparison of average few-shot accuracy between zero and SVD initialization methods across three models. Performance gains due to LoRA relaxation are indicated by hatched bars, while cases where performance is lower than the recursive counterpart (without LoRA modules) are represented by dotted lines.
    }
    \label{fig_rrt:zero_svd_init_app}
\end{figure}

The utilization of layer-wise LoRA modules enhances model capacity by introducing additional parameters and relaxation, thereby potentially improving performance. 
As depicted in Figure\,\ref{fig_rrt:zero_svd_init_app}, SVD initialization significantly amplified these performance gains compared to standard zero initialization. However, an interesting exception was observed with the Stepwise method, where the SVD initialized LoRA module surprisingly led to a performance degradation. 
This appears to be attributed to LoRA ranks being insufficient to adequately approximate the low-rank deltas across layers, resulting in initialization at a sub-optimal point.

\paragraph{Ablation study on the LoRA rank values.} 

\begin{figure}[ht!]
    \centering
    \begin{subfigure}[t]{\textwidth}
        \includegraphics[width=\textwidth]{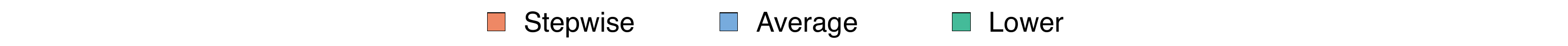}
    \end{subfigure}
    \centering
    \begin{subfigure}[t]{0.34\textwidth}
        \includegraphics[width=\textwidth]{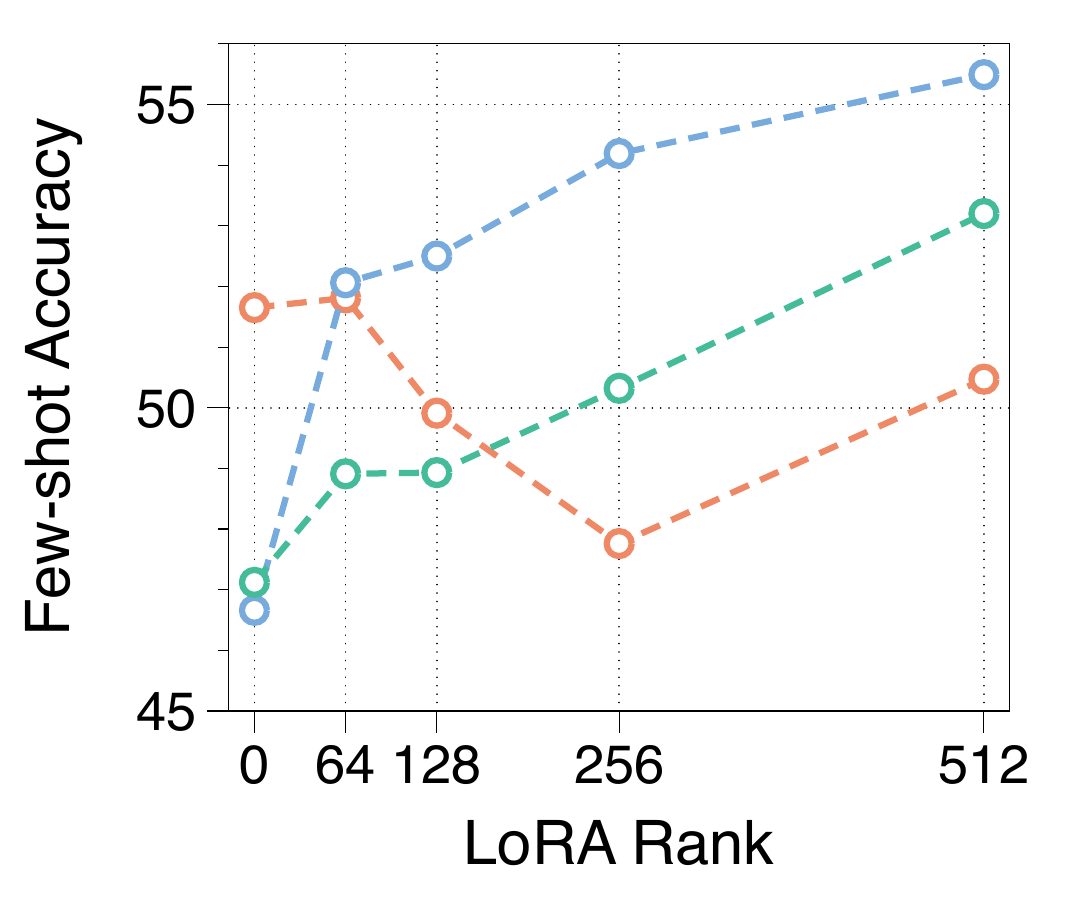}
        \subcaption{Gemma}
    \end{subfigure}
    \centering
    \begin{subfigure}[t]{0.31\textwidth}
        \includegraphics[width=\textwidth]{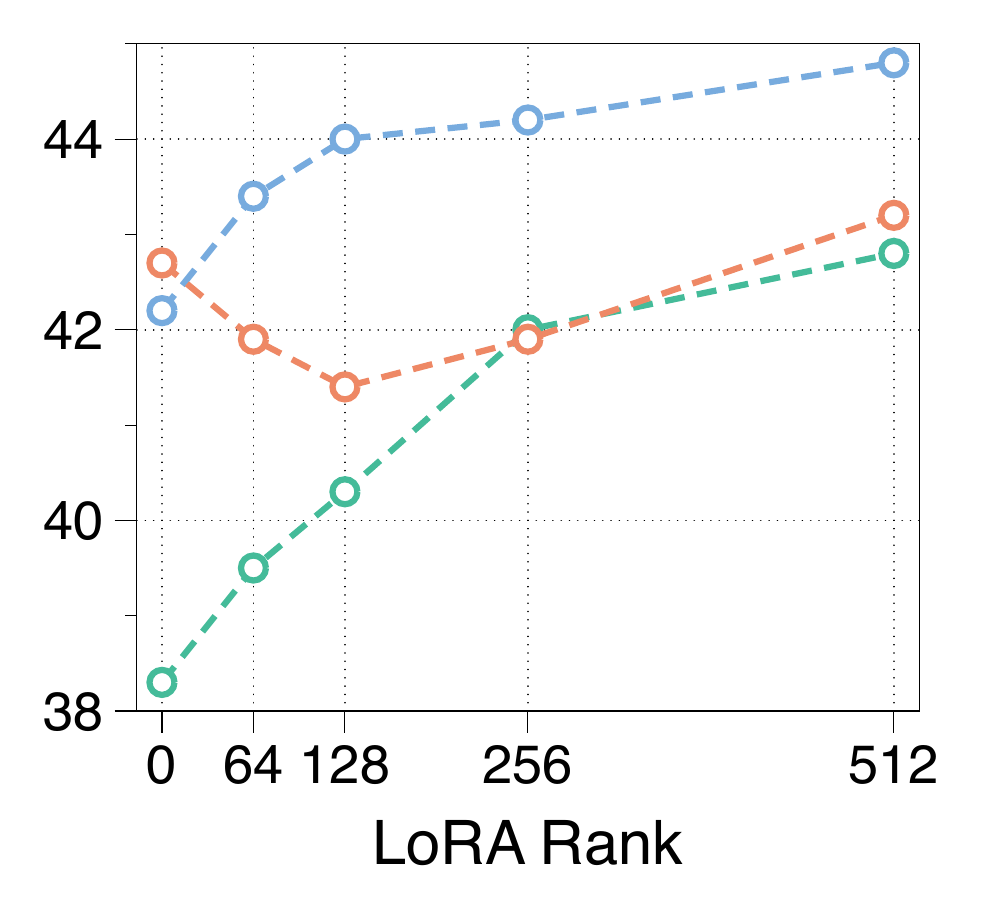}
        \subcaption{TinyLlama}
    \end{subfigure}
    \centering
    \begin{subfigure}[t]{0.31\textwidth}
        \includegraphics[width=\textwidth]{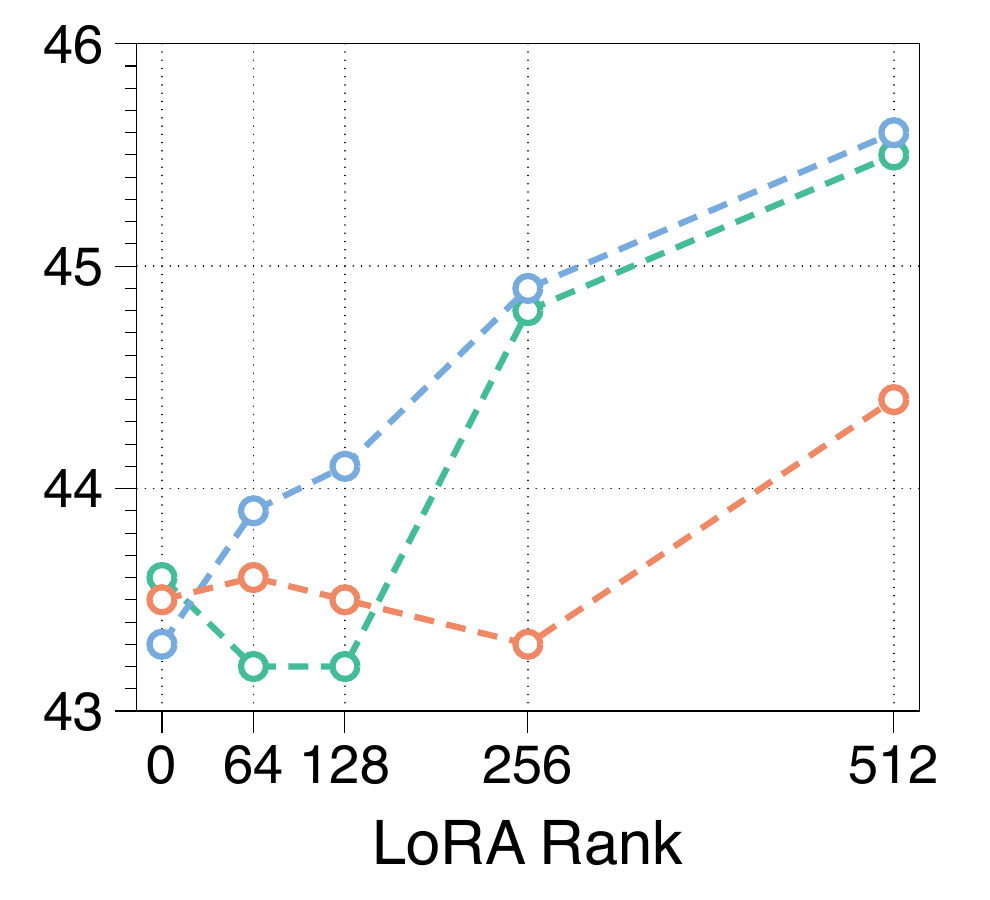}
        \subcaption{Pythia}
    \end{subfigure}
    \caption{Performance comparison with varying LoRA ranks under different initialization methods for looped layers. All LoRA weights are initialized using our proposed SVD initialization method.}
    \label{fig_rrt:rank_fewshot_app}
\end{figure}

Our proposed SVD initialization ensures that the Relaxed Recursive Transformer can function as an interpolation between vanilla and Recursive Transformers. The approximation accuracy of SVD is directly influenced by the LoRA rank value; a higher rank leads to improved restoration of the pretrained model weights. In Figure\,\ref{fig_rrt:rank_fewshot_app}, we present a summary of the performance changes observed in the relaxed models by varying the LoRA ranks. As expected, for the Average and Lower looping initialization methods, a larger rank value results in enhanced performance. The Stepwise method, consistent with previous experimental findings, exhibited a U-shaped trend: with extremely low or high ranks, a clear performance increase results. However, with mid-range values, the approximation becomes less accurate, leading to a further performance decrease.

We further experimented with assigning different ranks to LoRA modules associated with each linear weights. Given the computational overhead inherent to LoRA modules, allocating different ranks to each module can offer a better balance between performance and computational efficiency. The experimental results in Table\,\ref{tab_rrt:lora_rank_ablation_app} reveal a strong correlation between performance and overall model sizes. Due to the substantial hidden dimension of the linear weights within the FFN layer, reducing its rank led to the most significant performance drop. Conversely, the relatively smaller size of other attention weights resulted in less performance drop. It is intriguing that even minimal relaxation of key-value weights (achieved through small ranks) yielded comparable performance, despite the inherent strong sharing of key-value caches between attention heads in the Multi-Query attention structure~\citep{DBLP:conf/emnlp/AinslieLJZLS23}.

\clearpage

\begin{table}[ht]
    \small
    \centering
    \resizebox{\textwidth}{!}{
    \setlength{\tabcolsep}{3pt}
    \begin{tabular}{c|cc|cc|ccccc|rrr|ccccccc|cc}
    \toprule
     & \multicolumn{2}{c|}{\textbf{Uptrain}} & \multicolumn{2}{c|}{\textbf{Looping}} &  \multicolumn{5}{c|}{\textbf{LoRA}} & \multicolumn{3}{c|}{\textbf{Perplexity\,$\downarrow$}} & \multicolumn{8}{c}{\textbf{Few-shot Accuracy\,$\uparrow$}} \\
    \cmidrule(l{2pt}r{2pt}){2-3} \cmidrule(l{2pt}r{2pt}){4-5}  \cmidrule(l{2pt}r{2pt}){6-10} \cmidrule(l{2pt}r{2pt}){11-13}  \cmidrule(l{2pt}r{2pt}){14-21} 
    N-emb & PT & $N_{tok}$ & Block & Init  & Q & KV & Out & FFN & Init & SlimP & RedP & PG19 & LD & HS & PQ & WG & ARC-e & ARC-c & OB & Avg  \\
    \midrule
    1.99B &  \cmark & 15B & - & - & - & - & - & - & - & 10.76 & 8.47 & 13.08 & 63.5 & 68.5 & 77.0 & 63.5 & 67.6 & 38.1 & {42.6} & 60.1   \\
    0.99B &  \xmark & 15B & - & - & - & - & -  & - & - & 22.63 & 20.03 & 32.60 & 28.9 & 31.6 & 63.1 & 52.3 & 41.2 & 22.5 & 27.8 &  38.2  \\
    \midrule
    1.30B & \cmark & 15B & 2 & Avg  & 256 & 256 & 256 & 256 & SVD &  12.10 &  9.71 &  14.89 &  58.2 &  60.7 &  73.7 &  59.0 &  57.6 &  32.1 &  {38.0} &  54.2 \\
    1.28B & \cmark & 15B & 2 & Avg  & 128 & 256 & 128 & 256 & SVD &  12.27 &  9.81 &  15.10 &  57.4 &  60.2 &  72.5 &  58.9 &  58.1 &  32.6 &  37.8 &  53.9 \\
     1.29B & \cmark & 15B & 2 & Avg  & 256 & 128 & 256 & 256 & SVD &  12.33 &  9.90 &  15.25 &  58.5 &  59.7 &  73.3 &  58.3 &  56.6 &  32.0 &  40.0 &  54.1 \\
     1.18B & \cmark & 15B & 2 & Avg  & 256 & 256 & 256 & 128 & SVD &  12.56 &  10.12 &  15.59 &  57.0 &  58.7 &  73.0 &  57.4 &  57.0 &  31.6 &  38.2 &  53.3 \\
     1.27B & \cmark & 15B & 2 & Avg  & 128 & 128 & 128 & 256 & SVD &  12.36 &  9.92 &  15.31 &  57.2 &  59.2 &  73.1 &  57.3 &  58.0 &  32.2 &  38.6 &  53.7 \\
    \midrule
     1.15B & \cmark & 15B & 2 & Avg  & 128 & 128 & 128 & 128 & SVD &  12.52 &  10.07 &  15.51 &  56.1 &  58.2 &  72.3 &  55.8 &  57.1 &  30.7 &  37.2 &  52.5 \\
      1.14B & \cmark & 15B & 2 & Avg  & 64 & 128 & 64 & 128 & SVD &  12.61 &  10.14 &  15.69 &  55.0 &  57.8 &  73.0 &  57.5 &  56.7 &  30.9 &  38.8 &  52.8 \\
     1.14B & \cmark & 15B & 2 & Avg  & 128 & 64 & 128 &  128 & SVD &  12.72 &  10.18 &  15.76 &  55.5 &  57.7 &  72.7 &  57.0 &  56.9 &  30.1 &  38.2 &  52.6 \\
     1.08B& \cmark & 15B & 2 & Avg  & 128 & 128 & 128 & 64 & SVD &  12.80 &  10.33 &  15.97 &  55.3 &  56.7 &  72.9 &  57.7 &  55.0 &  29.6 &  36.0 &  51.9 \\
     1.13B & \cmark & 15B & 2 & Avg  & 64 & 64 & 64 & 128 & SVD &  12.77 &  10.29 &  15.95 &  55.2 &  57.4 &  73.0 &  56.7 &  56.5 &  30.5 &  37.2 &  52.3 \\
    \bottomrule
    \end{tabular}
    }
    \caption{
    Evaluation results of relaxed recursive Gemma models with varying LoRA ranks for different Transformer components. We adjusted the LoRA ranks attached to query, key-value, out, and FFN linear weights. Non-embedding parameter sizes include both the base layer parameters and the attached LoRA weights.
    }
    \label{tab_rrt:lora_rank_ablation_app}
\end{table}

\paragraph{Overall performance comparison of Relaxed Recursive Transformers.}

A comprehensive performance comparison is presented in Table\,\ref{tab_rrt:lora_total_app}. This encompasses an evaluation of the performance across three models and various looping initialization methods, considering the degree of relaxation induced by the layer-wise LoRA module. The configuration utilizing the Average method for looped layer initialization, paired with SVD initialization for the LoRA module, consistently outperformed all other baselines. Furthermore, a clear trend of improved performance was observed with increasing rank.\looseness=-1

\begin{table}[ht!]
    \small
    \centering
    \resizebox{\textwidth}{!}{
    \setlength{\tabcolsep}{3pt}
    \renewcommand{\arraystretch}{0.85}
    \begin{tabular}{l|c|cc|cc|cc|rrr|ccccccc|cc}
    \toprule
     & & \multicolumn{2}{c|}{\textbf{Uptrain}} & \multicolumn{2}{c|}{\textbf{Looping}} &  \multicolumn{2}{c|}{\textbf{LoRA}} & \multicolumn{3}{c|}{\textbf{Perplexity\,$\downarrow$}} & \multicolumn{9}{c}{\textbf{Few-shot Accuracy\,$\uparrow$}} \\
    \cmidrule(l{2pt}r{2pt}){3-4} \cmidrule(l{2pt}r{2pt}){5-6}  \cmidrule(l{2pt}r{2pt}){7-8} \cmidrule(l{2pt}r{2pt}){9-11}  \cmidrule(l{2pt}r{2pt}){12-20} 
     \textbf{Models} & N-emb & PT & $N_{tok}$ & Block & Init  & Rank & Init & SlimP & RedP & PG19 & LD & HS & PQ & WG & ARC-e & ARC-c & OB & Avg & $\Delta$  \\
    \midrule
    & 1.99B &  \cmark & 15B & - & - & - & - & 10.76 & 8.47 & 13.08 & 63.5 & 68.5 & 77.0 & 63.5 & 67.6 & 38.1 & {42.6} & 60.1  & - \\
     & 0.99B &  \xmark & 15B & - & - & - & - & 22.63 & 20.03 & 32.60 & 28.9 & 31.6 & 63.1 & 52.3 & 41.2 & 22.5 & 27.8 &  38.2 & -   \\  
     \cmidrule(l{2pt}r{2pt}){2-20} 
     & 0.99B & \cmark & 15B & 2 & Step & - & - &  {12.85} & {10.29} & {16.21} & {53.0} & {57.3} & {73.2} & {56.2} & {56.1} & {29.2} & {36.6} & {51.7} & - \\
    & 1.07B & \cmark & 15B & 2 & Step  & 64 & SVD &  12.76 &  10.21 &  15.99 &  52.1 &  57.2 &  73.0 &  57.8 &  56.9 &  28.9 &  36.8 &  51.8 & \textcolor{custom_green}{\textbf{+0.1}} \\
    & 1.15B & \cmark & 15B & 2 & Step  & 128 & SVD &  13.44 &  10.80 &  16.98 &  50.5 &  53.0 &  71.5 &  54.4 &  55.9 &  29.3 &  34.8 &  49.9 & \textcolor{custom_red}{\textbf{--\,1.8}} \\
    & 1.30B & \cmark & 15B & 2 & Step  & 256 & SVD &  14.02 &  11.44 &  18.09 &  46.1 &  49.1 &  71.8 &  53.2 &  52.8 &  27.8 &  33.4 &  47.8 & \textcolor{custom_red}{\textbf{--\,3.9}} \\
    & 1.60B & \cmark & 15B & 2 & Step  & 512 & SVD &  13.13 &  10.66 &  16.63 &  53.0 &  54.3 &  72.1 &  54.9 &  54.8 &  28.8 &  35.4 &  50.5 & \textcolor{custom_red}{\textbf{--\,1.2}} \\
     & 1.60B & \cmark & 15B & 2 & Step  & 512 & Zero &  12.46 &  9.97 &  15.58 &  54.9 &  58.8 &  \textbf{74.0} &  58.1 &  58.8 &  30.6 &  36.6 &  53.1 & \textcolor{custom_green}{\textbf{+1.4}} \\
     \cmidrule(l{2pt}r{2pt}){2-20} 
     Gemma & 0.99B & \cmark & 15B & 2 & Avg & - & - & 15.15 & 12.57 & 19.86 & 43.6 & 47.4 & 70.4 & 52.6 & 50.5 & 27.8 & 34.4 & 46.7 & -  \\
     \rowcolor[gray]{0.9} 
    \cellcolor{white!20} & 1.07B & \cmark & 15B & 2 & Avg  & 64 & SVD &  12.83 &  10.35 &  16.02 &  55.9 &  56.8 &  72.5 &  56.8 &  55.7 &  30.6 &  36.2 &  52.1 & \textcolor{custom_green}{\textbf{+5.4}} \\
     \rowcolor[gray]{0.9} 
    \cellcolor{white!20} & 1.15B & \cmark & 15B & 2 & Avg  & 128 & SVD &  12.52 &  10.07 &  15.51 &  56.1 &  58.2 &  72.3 &  55.8 &  57.1 &  30.7 &  37.2 &  52.5 & \textcolor{custom_green}{\textbf{+5.8}} \\
     \rowcolor[gray]{0.9} 
    \cellcolor{white!20} & 1.30B & \cmark & 15B & 2 & Avg  & 256 & SVD &  12.10 &  9.71 &  14.89 &  58.2 &  60.7 &  73.7 &  59.0 &  57.6 &  32.1 &  \textbf{38.0} &  54.2 & \textcolor{custom_green}{\textbf{+7.5}} \\
     \rowcolor[gray]{0.9} 
    \cellcolor{white!20} & 1.60B & \cmark & 15B & 2 & Avg  & 512 & SVD &  \textbf{11.83} &  \textbf{9.46} &  \textbf{14.57} &  \textbf{59.3} &  \textbf{62.8} &  \textbf{74.0 }&  \textbf{61.6} &  \textbf{60.1} &  \textbf{32.9} &  37.6 &  \textbf{55.5} & \textcolor{custom_green}{\textbf{+8.8}} \\
     & 1.60B & \cmark & 15B & 2 & Avg  & 512 & Zero &  13.78 &  11.31 &  17.71 &  49.8 &  52.4 &  71.7 &  53.3 &  51.2 &  29.4 &  35.0 &  49.0 & \textcolor{custom_green}{\textbf{+2.3}} \\
     \cmidrule(l{2pt}r{2pt}){2-20} 
     & 0.99B & \cmark & 15B & 2 & Lower & - & - & 15.03 & 12.46 & 19.63 & 42.5 & 48.0 & 71.0 & 54.6 & 52.2 & 27.7 & 33.8 & 47.1 & -  \\
    & 1.07B & \cmark & 15B & 2 & Lower  & 64 & SVD &  14.21 &  11.77 &  18.40 &  47.5 &  50.5 &  70.9 &  54.2 &  54.1 &  29.2 &  36.0 &  48.9 & \textcolor{custom_green}{\textbf{+1.8}} \\
    & 1.15B & \cmark & 15B & 2 & Lower  & 128 & SVD &  14.23 &  11.83 &  18.49 &  48.0 &  50.5 &  72.0 &  56.8 &  54.4 &  27.5 &  33.4 &  48.9 & \textcolor{custom_green}{\textbf{+1.8}} \\
    & 1.30B & \cmark & 15B & 2 & Lower  & 256 & SVD &  13.51 &  11.06 &  17.30 &  53.1 &  53.7 &  71.8 &  57.4 &  52.5 &  28.7 &  35.2 &  50.3 & \textcolor{custom_green}{\textbf{+3.2}} \\
    & 1.60B & \cmark & 15B & 2 & Lower  & 512 & SVD &  12.54 &  10.22 &  15.90 &  57.1 &  58.2 &  73.7 &  58.6 &  57.6 &  31.5 &  35.6 &  53.2 & \textcolor{custom_green}{\textbf{+6.1}} \\
     & 1.60B & \cmark & 15B & 2 & Lower  & 512 & Zero &  13.95 &  11.59 &  18.02 &  48.4 &  52.1 &  71.9 &  55.7 &  54.9 &  28.8 &  34.6 &  49.5 & \textcolor{custom_green}{\textbf{+2.4}} \\
     \midrule
    & 0.97B &  \cmark & - & - & - & - & - & 12.26 & 9.37 & 11.94 & 43.3 & 42.2 & 66.8 & 53.4 & 44.7 & 23.2 & 29.2 & 43.3  & - \\
     & 0.48B &  \xmark & 15B & - & - & - & - & 16.61 & 15.66 & 20.27 & 22.3 & 30.0 & 60.9 & 50.6 & 37.0 & 23.0 & 28.0 & 36.0 & -  \\ 
     \cmidrule(l{2pt}r{2pt}){2-20} 
     & 0.48B & \cmark & 15B & 2 & Step & - & - &  {11.61} & {9.89}  & {13.00} &{ 39.6} &{ 39.8} & {66.5} & {52.9} & {44.3} & {24.9} & {30.6} & {42.7} & -   \\
    & 0.53B & \cmark & 15B & 2 & Step  & 64 & SVD &  12.10 &  10.40 &  13.75 &  38.9 &  38.3 &  65.2 &  51.5 &  42.0 &  \textbf{26.0} &  31.0 &  41.9 & \textcolor{custom_red}{\textbf{--\,0.8}} \\
    & 0.58B & \cmark & 15B & 2 & Step  & 128 & SVD &  12.41 &  10.72 &  14.10 &  36.8 &  37.4 &  64.7 &  53.4 &  42.2 &  24.7 &  30.4 &  41.4 & \textcolor{custom_red}{\textbf{--\,1.3}} \\
    & 0.68B & \cmark & 15B & 2 & Step  & 256 & SVD &  11.96 &  10.35 &  13.48 &  38.9 &  38.3 &  65.8 &  51.9 &  43.1 &  25.4 &  29.8 &  41.9 & \textcolor{custom_red}{\textbf{--\,0.8}} \\
    & 0.86B & \cmark & 15B & 2 & Step  & 512 & SVD &  11.33 &  9.79 &  12.61 &  42.2 &  40.9 &  67.7 &  51.1 &  45.0 &  25.3 &  30.2 &  43.2 & \textcolor{custom_green}{\textbf{+0.5}} \\
     & 0.86B & \cmark & 15B & 2 & Step  & 512 & Zero &  11.24 &  9.60 &  12.56 &  42.0 &  41.0 &  67.4 &  52.2 &  44.5 &  25.9 &  \textbf{31.2} &  43.4 & \textcolor{custom_green}{\textbf{+0.7}} \\
     \cmidrule(l{2pt}r{2pt}){2-20} 
     TinyLlama & 0.48B & \cmark & 15B & 2 & Avg & - & - & 11.86 & 10.29  & 13.42 & 38.6 & 39.4 & 66.1 & 52.8 & 42.7 & 25.4 & {30.6} & 42.2 & -  \\
     \rowcolor[gray]{0.9} 
    \cellcolor{white!20} & 0.53B & \cmark & 15B & 2 & Avg  & 64 & SVD &  11.22 &  9.66 &  12.51 &  41.8 &  41.6 &  67.0 &  53.3 &  43.9 &  24.7 &  \textbf{31.2} &  43.4 & \textcolor{custom_green}{\textbf{+1.2}} \\
     \rowcolor[gray]{0.9} 
    \cellcolor{white!20} & 0.58B & \cmark & 15B & 2 & Avg  & 128 & SVD &  10.99 &  9.45 &  12.21 &  43.2 &  42.1 &  \textbf{68.3} &  53.2 &  44.8 &  25.9 &  30.4 &  44.0 & \textcolor{custom_green}{\textbf{+1.8}} \\
     \rowcolor[gray]{0.9} 
    \cellcolor{white!20} & 0.68B & \cmark & 15B & 2 & Avg  & 256 & SVD &  10.71 &  9.18 &  11.82 &  44.1 &  43.2 &  68.1 &  \textbf{53.5} &  44.4 &  25.7 &  30.4 &  44.2 & \textcolor{custom_green}{\textbf{+2.0}} \\
     \rowcolor[gray]{0.9} 
    \cellcolor{white!20} & 0.86B & \cmark & 15B & 2 & Avg  & 512 & SVD &  \textbf{10.46} &  \textbf{8.92} & \textbf{ 11.50} &  \textbf{46.0} &  \textbf{44.1} &  68.2 &  53.0 &  \textbf{45.8} &  25.1 &  \textbf{31.2} &  \textbf{44.8} & \textcolor{custom_green}{\textbf{+2.6}} \\
     & 0.86B & \cmark & 15B & 2 & Avg  & 512 & Zero &  11.28 &  9.75 &  12.69 &  41.5 &  41.0 &  66.8 &  53.2 &  44.8 &  25.5 &  \textbf{31.2} &  43.4 & \textcolor{custom_green}{\textbf{+1.2}} \\
     \cmidrule(l{2pt}r{2pt}){2-20} 
     & 0.48B & \cmark & 15B & 2 & Lower & - & - & 14.67 & 12.67  & 16.68 & 31.9 & 32.3 & 62.6 & 52.0 & 39.1 & 22.1 & 27.8 & 38.3 & -  \\
    & 0.53B & \cmark & 15B & 2 & Lower  & 64 & SVD &  13.68 &  11.77 &  15.48 &  35.5 &  34.0 &  63.8 &  51.0 &  40.0 &  24.6 &  28.0 &  39.5 & \textcolor{custom_green}{\textbf{+1.2}} \\
    & 0.58B & \cmark & 15B & 2 & Lower  & 128 & SVD &  13.00 &  11.14 &  14.61 &  37.6 &  35.4 &  65.3 &  51.5 &  40.4 &  24.5 &  27.6 &  40.3 & \textcolor{custom_green}{\textbf{+2.0}} \\
    & 0.68B & \cmark & 15B & 2 & Lower  & 256 & SVD &  12.14 &  10.39 &  13.59 &  40.0 &  37.7 &  66.1 &  52.6 &  42.5 &  24.8 &  30.0 &  42.0 & \textcolor{custom_green}{\textbf{+3.7}} \\
    & 0.86B & \cmark & 15B & 2 & Lower  & 512 & SVD &  11.31 &  9.61 &  12.49 &  43.2 &  40.5 &  66.0 &  50.8 &  43.9 &  24.8 &  30.0 &  42.8 & \textcolor{custom_green}{\textbf{+4.5}} \\
     & 0.86B & \cmark & 15B & 2 & Lower  & 512 & Zero &  14.56 &  12.69 &  16.57 &  21.2 &  32.9 &  63.9 &  52.6 &  39.5 &  22.9 &  27.8 &  37.3 & \textcolor{custom_red}{\textbf{--\,1.0}} \\
     \midrule
    & 0.81B &  \cmark & 15B & - & - & - & - & 13.46 & 9.95 & 13.38 &  55.0 & 49.0 & 71.0 & 53.6 & 51.8 & {28.2} & 32.8 & {48.8}  & - \\
     & 0.40B &  \xmark & 15B & - & - & - & - & 25.69 & 20.00 & 32.08 & 24.3 & 30.0 & 61.9 & 50.7 & 38.3 & 22.3 & 26.0  & 36.2 & -  \\ 
     \cmidrule(l{2pt}r{2pt}){2-20} 
     & 0.40B & \cmark & 15B & 2 & Step & - & - & {16.38} & {12.37} & {17.74} & 43.4 & {40.5} & 67.4 & 50.8 & {46.3} & 25.7 & 30.0 & 43.5 & -   \\
    & 0.44B & \cmark & 15B & 2 & Step  & 64 & SVD &  16.44 &  12.44 &  17.89 &  43.7 &  40.4 &  66.5 &  52.9 &  46.5 &  26.2 &  28.8 &  43.6 & \textcolor{custom_green}{\textbf{+0.1}} \\
    & 0.48B & \cmark & 15B & 2 & Step  & 128 & SVD &  16.63 &  12.61 &  18.35 &  42.4 &  39.3 &  68.0 &  51.5 &  46.3 &  26.7 &  30.6 &  43.5 & \textcolor{gray}{\textbf{+0.0}} \\
    & 0.55B & \cmark & 15B & 2 & Step  & 256 & SVD &  16.54 &  12.61 &  18.39 &  42.8 &  39.1 &  67.2 &  53.7 &  46.4 &  25.9 &  27.8 &  43.3 & \textcolor{custom_red}{\textbf{--\,0.2}} \\
    & 0.70B & \cmark & 15B & 2 & Step  & 512 & SVD &  15.68 &  11.88 &  17.25 &  45.4 &  41.3 &  \textbf{68.5} &  52.6 &  46.7 &  25.4 &  31.2 &  44.4 & \textcolor{custom_green}{\textbf{+0.9}} \\
     & 0.70B & \cmark & 15B & 2 & Step  & 512 & Zero &  15.88 &  12.01 &  17.16 &  45.5 &  41.8 &  68.0 &  52.6 &  46.6 &  \textbf{26.3} &  30.0 &  44.4 & \textcolor{custom_green}{\textbf{+0.9}} \\
     \cmidrule(l{2pt}r{2pt}){2-20} 
     Pythia & 0.40B & \cmark & 15B & 2 & Avg & - & - & 16.76 & 12.76 & 18.63 & 43.6 & 39.1 & {68.2} & 51.9 & 45.4 & 25.1 & 29.8 & 43.3 & -  \\
     \rowcolor[gray]{0.9} 
    \cellcolor{white!20} & 0.44B & \cmark & 15B & 2 & Avg  & 64 & SVD &  16.03 &  12.19 &  17.59 &  45.8 &  40.9 &  67.3 &  50.0 &  45.8 &  25.5 &  31.8 &  43.9 & \textcolor{custom_green}{\textbf{+0.6}} \\
     \rowcolor[gray]{0.9} 
    \cellcolor{white!20} & 0.48B & \cmark & 15B & 2 & Avg  & 128 & SVD &  15.67 &  11.93 &  17.10 &  46.9 &  41.9 &  67.4 &  51.2 &  45.4 &  24.8 &  31.2 &  44.1 & \textcolor{custom_green}{\textbf{+0.8}} \\
     \rowcolor[gray]{0.9} 
    \cellcolor{white!20} & 0.55B & \cmark & 15B & 2 & Avg  & 256 & SVD &  15.22 &  11.54 &  16.47 &  48.5 &  43.3 &  67.2 &  51.4 &  46.7 &  25.5 &  32.0 &  44.9 & \textcolor{custom_green}{\textbf{+1.6}} \\
     \rowcolor[gray]{0.9} 
    \cellcolor{white!20}    & 0.70B & \cmark & 15B & 2 & Avg  & 512 & SVD &  \textbf{14.70} &  \textbf{11.07} &  \textbf{15.71} &  \textbf{50.2} &  \textbf{44.7} &  68.2 &  51.6 &  \textbf{47.6} &  25.4 &  31.2 &  \textbf{45.6} & \textcolor{custom_green}{\textbf{+2.3}} \\
     & 0.70B & \cmark & 15B & 2 & Avg  & 512 & Zero &  15.97 &  12.14 &  17.65 &  45.7 &  41.5 &  68.1 &  51.7 &  46.5 &  25.7 &  30.0 &  44.2 & \textcolor{custom_green}{\textbf{+0.9}} \\
     \cmidrule(l{2pt}r{2pt}){2-20} 
     & 0.40B & \cmark & 15B & 2 & Lower & - & - & 17.04 & 12.62 & 18.44 & {43.9} & 39.2 & 66.3 & {53.4} & 45.4 & {25.8} & {31.2} & {43.6} & -  \\
    & 0.44B & \cmark & 15B & 2 & Lower  & 64 & SVD &  17.03 &  12.78 &  18.73 &  44.1 &  38.3 &  66.9 &  51.9 &  45.4 &  24.7 &  30.8 &  43.2 & \textcolor{custom_red}{\textbf{--\,0.4}} \\
    & 0.48B & \cmark & 15B & 2 & Lower  & 128 & SVD &  16.63 &  12.49 &  18.17 &  45.2 &  39.2 &  66.8 &  51.0 &  45.6 &  24.9 &  29.6 &  43.2 & \textcolor{custom_red}{\textbf{--\,0.4}} \\
    & 0.55B & \cmark & 15B & 2 & Lower  & 256 & SVD &  15.93 &  11.99 &  17.30 &  47.6 &  41.4 &  68.3 &  53.2 &  46.0 &  25.8 &  31.0 &  44.8 & \textcolor{custom_green}{\textbf{+1.2}} \\
    & 0.70B & \cmark & 15B & 2 & Lower  & 512 & SVD &  15.11 &  11.34 &  16.07 &  \textbf{50.2} &  43.5 &  67.8 &  51.8 &  47.2 &  25.2 &  32.0 &  45.4 & \textcolor{custom_green}{\textbf{+1.8}} \\
     & 0.70B & \cmark & 15B & 2 & Lower  & 512 & Zero &  16.45 &  12.25 &  17.76 &  45.2 &  40.4 &  66.4 &  \textbf{54.5} &  45.8 &  25.9 &  \textbf{32.6} &  44.4 & \textcolor{custom_green}{\textbf{+0.8}} \\
    \bottomrule
    \end{tabular}
    }
    \caption{Overall evaluation results of Relaxed Recursive Transformers. Delta\,($\Delta$) represent the accuracy differences between relaxed and non-relaxed models using the same looping initialization. 
    }
    \label{tab_rrt:lora_total_app}
\end{table}

\clearpage
\subsection{Alternative Approaches to Relaxed Parameter Sharing}
\label{app_rrt:prefix_tuning_relaxation}

To mitigate the restrictive weight-tying inherent in parameter sharing, we employed LoRA modules as discussed in \S\ref{sec_rrt:methods:relaxed}, similar to prior work~\citep{DBLP:conf/emnlp/GeCW22, shim2024leveraging}. However, efficiently computing multiple LoRA modules requires specialized kernels and sequential computations of the base layers and LoRA modules, incurring computational overhead. Consequently, we explored layer-specific prompts~\citep{DBLP:journals/corr/abs-2110-07602} as an alternative. This approach integrates prompts specific to each layer as prefix tokens, generating layer-wise key and value states for self-attention, and is significantly more amenable to parallel computation.

Table\,\ref{tab_rrt:prefix_tuning_app} summarizes performance of the prefix tuning method. While offering computational advantages, its reliance on small, learnable prompts resulted in limited performance gains. Additionally, without leveraging the original pretrained weights, performance was significantly lower (52.1\% vs. 47.6\% with the Average method in 1.07B model size). Future work will explore enhancing the effectiveness of this parallel approach, as well as other strategies such as bias term-based relaxation~\citep{DBLP:conf/emnlp/GeCW22}.

\begin{table}[h]
    \small
    \centering
    \resizebox{\textwidth}{!}{
    \setlength{\tabcolsep}{3pt}
    \begin{tabular}{l|c|cc|cc|cc|rrr|ccccccc|cc}
    \toprule
     & & \multicolumn{2}{c|}{\textbf{Uptrain}} & \multicolumn{2}{c|}{\textbf{Looping}} &  \multicolumn{2}{c|}{\textbf{Prefix}} & \multicolumn{3}{c|}{\textbf{Perplexity\,$\downarrow$}} & \multicolumn{9}{c}{\textbf{Few-shot Accuracy\,$\uparrow$}} \\
    \cmidrule(l{2pt}r{2pt}){3-4} \cmidrule(l{2pt}r{2pt}){5-6}  \cmidrule(l{2pt}r{2pt}){7-8} \cmidrule(l{2pt}r{2pt}){9-11}  \cmidrule(l{2pt}r{2pt}){12-20} 
     \textbf{Models} & N-emb & PT & $N_{tok}$ & Block & Init  & Len & Size & SlimP & RedP & PG19 & LD & HS & PQ & WG & ARC-e & ARC-c & OB & Avg & $\Delta$  \\
    \midrule
    & 1.99B &  \cmark & 15B & - & - & - & - & 10.76 & 8.47 & 13.08 & 63.5 & 68.5 & 77.0 & 63.5 & 67.6 & 38.1 & {42.6} & 60.1  & - \\
     & 0.99B &  \xmark & 15B & - & - & - & - & 22.63 & 20.03 & 32.60 & 28.9 & 31.6 & 63.1 & 52.3 & 41.2 & 22.5 & 27.8 &  38.2 & -   \\
     \cmidrule(l{2pt}r{2pt}){2-20} 
     & 0.99B & \cmark & 15B & 2 & Step & - & - &  {12.85} & {10.29} & {16.21} & {53.0} & {57.3} & {73.2} & {56.2} & {56.1} & {29.2} & {36.6} & {51.7} & - \\
     & 1.00B & \cmark & 15B & 2 & Step  & \,\,\,256 & \,\,\,9.4M &  12.62 &  10.06 &  15.80 &  53.5 &  58.3 &  73.9 &  57.6 &  57.5 &  29.3 &  35.6 &  52.2 & \textcolor{custom_green}{\textbf{+0.5}} \\
    & 1.01B & \cmark & 15B & 2 & Step  & \,\,\,512 & 18.9M &  12.67 &  10.10 &  15.85 &  54.1 &  57.8 &  73.8 &  58.4 &  57.2 &  28.7 &  35.8 &  52.3 & \textcolor{custom_green}{\textbf{+0.6}} \\
    & 1.03B & \cmark & 15B & 2 & Step  & 1024 & 37.7M &  12.89 &  10.34 &  16.22 &  53.5 &  57.1 &  72.4 &  57.2 &  56.9 &  28.6 &  36.8 &  51.8 & \textcolor{custom_green}{\textbf{+0.1}} \\
    & 1.07B & \cmark & 15B & 2 & Step  & 2048 & 75.5M &  12.75 &  10.21 &  16.09 &  55.0 &  57.3 &  73.3 &  58.2 &  56.8 &  29.2 &  37.8 &  52.5 & \textcolor{custom_green}{\textbf{+0.8}} \\
     \cmidrule(l{2pt}r{2pt}){2-20} 
     Gemma & 0.99B & \cmark & 15B & 2 & Avg & - & - & 15.15 & 12.57 & 19.86 & 43.6 & 47.4 & 70.4 & 52.6 & 50.5 & 27.8 & 34.4 & 46.7 & -  \\
     & 1.00B & \cmark & 15B & 2 & Avg  & \,\,\,256 & \,\,\,9.4M &  14.85 &  12.31 &  19.41 &  46.9 &  48.3 &  70.4 &  52.7 &  51.4 &  27.2 &  34.0 &  47.3 & \textcolor{custom_green}{\textbf{+0.6}} \\
    & 1.01B & \cmark & 15B & 2 & Avg  & \,\,\,512 & 18.9M &  15.23 &  12.64 &  19.98 &  44.5 &  47.2 &  70.7 &  54.5 &  49.5 &  28.0 &  33.2 &  46.8 & \textcolor{custom_green}{\textbf{+0.1}} \\
    & 1.03B & \cmark & 15B & 2 & Avg  & 1024 & 37.7M &  14.60 &  12.02 &  18.89 &  46.9 &  49.7 &  71.1 &  52.3 &  51.0 &  28.6 &  34.2 &  47.7 & \textcolor{custom_green}{\textbf{+1.0}} \\
    & 1.07B & \cmark & 15B & 2 & Avg  & 2048 & 75.5M &  14.63 &  12.07 &  19.03 &  47.3 &  49.5 &  70.8 &  53.1 &  50.7 &  28.2 &  33.4 &  47.6 & \textcolor{custom_green}{\textbf{+0.9}} \\
     \cmidrule(l{2pt}r{2pt}){2-20} 
     & 0.99B & \cmark & 15B & 2 & Lower & - & - & 15.03 & 12.46 & 19.63 & 42.5 & 48.0 & 71.0 & 54.6 & 52.2 & 27.7 & 33.8 & 47.1 & -  \\
     & 1.00B & \cmark & 15B & 2 & Lower  & \,\,\,256 & \,\,\,9.4M &  14.59 &  12.12 &  19.02 &  46.3 &  49.7 &  71.5 &  55.1 &  52.9 &  29.0 &  34.0 &  48.4 & \textcolor{custom_green}{\textbf{+1.3}} \\
    & 1.01B & \cmark & 15B & 2 & Lower  & \,\,\,512 & 18.9M &  14.53 &  12.03 &  18.88 &  45.7 &  49.8 &  71.9 &  56.4 &  53.6 &  29.4 &  33.2 &  48.6 & \textcolor{custom_green}{\textbf{+1.5}} \\
    & 1.03B & \cmark & 15B & 2 & Lower  & 1024 & 37.7M &  14.43 &  11.98 &  18.74 &  46.3 &  50.0 &  71.9 &  55.1 &  54.3 &  29.7 &  33.8 &  48.7 & \textcolor{custom_green}{\textbf{+1.6}} \\
    & 1.07B & \cmark & 15B & 2 & Lower  & 2048 & 75.5M &  14.79 &  12.26 &  19.23 &  46.1 &  48.7 &  71.4 &  55.4 &  51.3 &  28.2 &  34.0 &  47.9 & \textcolor{custom_green}{\textbf{+0.8}} \\
     \midrule
    & 0.97B &  \cmark & - & - & - & - & - & 12.26 & 9.37 & 11.94 & 43.3 & 42.2 & 66.8 & 53.4 & 44.7 & 23.2 & 29.2 & 43.3  & - \\
     & 0.48B &  \xmark & 15B & - & - & - & - & 16.61 & 15.66 & 20.27 & 22.3 & 30.0 & 60.9 & 50.6 & 37.0 & 23.0 & 28.0 & 36.0 & -  \\
     \cmidrule(l{2pt}r{2pt}){2-20} 
     & 0.48B & \cmark & 15B & 2 & Step & - & - &  {11.61} & {9.89}  & {13.00} &{ 39.6} &{ 39.8} & {66.5} & {52.9} & {44.3} & {24.9} & {30.6} & {42.7} & -   \\
     & 0.49B & \cmark & 15B & 2 & Step  & \,\,\,256 & 11.5M &  11.61 &  9.89 &  13.00 &  39.6 &  39.9 &  66.5 &  53.9 &  44.4 &  25.3 &  30.6 &  42.9 & \textcolor{custom_green}{\textbf{+0.2}} \\
    & 0.50B & \cmark & 15B & 2 & Step  & \,\,\,512 & 23.1M &  11.61 &  9.89 &  13.01 &  39.5 &  39.9 &  66.7 &  53.4 &  44.1 &  25.3 &  29.8 &  42.7 & \textcolor{gray}{\textbf{+0.0}} \\
    & 0.53B & \cmark & 15B & 2 & Step  & 1024 & 46.1M &  11.60 &  9.88 &  13.00 &  39.7 &  39.9 &  66.7 &  53.0 &  44.3 &  25.1 &  30.6 &  42.8 & \textcolor{custom_green}{\textbf{+0.1}} \\
    & 0.57B & \cmark & 15B & 2 & Step  & 2048 & 92.3M &  11.58 &  9.87 &  13.01 &  40.1 &  39.9 &  66.8 &  53.4 &  44.4 &  24.9 &  30.0 &  42.8 & \textcolor{custom_green}{\textbf{+0.1}} \\
     \cmidrule(l{2pt}r{2pt}){2-20} 
     TinyLlama & 0.48B & \cmark & 15B & 2 & Avg & - & - & 11.86 & 10.29  & 13.42 & 38.6 & 39.4 & 66.1 & 52.8 & 42.7 & 25.4 & {30.6} & 42.2 & -  \\
     & 0.49B & \cmark & 15B & 2 & Avg  & \,\,\,256 & 11.5M &  11.86 &  10.28 &  13.41 &  38.5 &  39.4 &  66.2 &  52.5 &  42.8 &  25.9 &  30.8 &  42.3 & \textcolor{custom_green}{\textbf{+0.1}} \\
    & 0.50B & \cmark & 15B & 2 & Avg  & \,\,\,512 & 23.1M &  11.86 &  10.28 &  13.41 &  38.1 &  39.3 &  66.3 &  52.2 &  42.6 &  25.6 &  30.8 &  42.1 & \textcolor{custom_red}{\textbf{--\,0.1}} \\
    & 0.53B & \cmark & 15B & 2 & Avg  & 1024 & 46.1M &  11.86 &  10.28 &  13.42 &  38.4 &  39.2 &  65.7 &  52.7 &  42.5 &  25.5 &  31.0 &  42.1 & \textcolor{custom_red}{\textbf{--\,0.1}} \\
    & 0.57B & \cmark & 15B & 2 & Avg  & 2048 & 92.3M &  11.86 &  10.28 &  13.42 &  38.5 &  39.5 &  65.9 &  52.7 &  42.4 &  25.7 &  31.0 &  42.2 & \textcolor{gray}{\textbf{+0.0}} \\
     \cmidrule(l{2pt}r{2pt}){2-20} 
     & 0.48B & \cmark & 15B & 2 & Lower & - & - & 14.67 & 12.67  & 16.68 & 31.9 & 32.3 & 62.6 & 52.0 & 39.1 & 22.1 & 27.8 & 38.3 & -  \\
     & 0.49B & \cmark & 15B & 2 & Lower  & \,\,\,256 & 11.5M &  14.67 &  12.67 &  16.69 &  31.9 &  32.4 &  62.7 &  51.5 &  38.9 &  22.3 &  27.8 &  38.2 & \textcolor{custom_red}{\textbf{--\,0.1}} \\
    & 0.50B & \cmark & 15B & 2 & Lower  & \,\,\,512 & 23.1M &  14.67 &  12.67 &  16.69 &  31.9 &  32.3 &  62.8 &  51.7 &  38.9 &  22.2 &  27.8 &  38.2 & \textcolor{custom_red}{\textbf{--\,0.1}} \\
    & 0.53B & \cmark & 15B & 2 & Lower  & 1024 & 46.1M &  14.67 &  12.67 &  16.68 &  31.6 &  32.3 &  63.0 &  51.9 &  38.9 &  22.1 &  28.0 &  38.3 & \textcolor{gray}{\textbf{+0.0}} \\
    & 0.57B & \cmark & 15B & 2 & Lower  & 2048 & 92.3M &  14.67 &  12.67 &  16.67 &  34.1 &  32.5 &  62.8 &  52.4 &  38.5 &  23.0 &  27.6 &  38.7 & \textcolor{custom_green}{\textbf{+0.4}} \\
    \bottomrule
    \end{tabular}
    }
    \caption{Evaluation results of relaxation through prefix tuning methods. Prefix length denotes the sequence length of trainable vectors used to generate key-value states in each self-attention layer. Non-embedding parameter sizes include the sizes of these trainable prefixes. Delta\,($\Delta$) represent the accuracy differences to non-relaxed models using the same looping initialization. 
    }
    \label{tab_rrt:prefix_tuning_app}
\end{table}

\subsection{Expanded Results of Extended Uptraining and Knowledge Distillation}
\label{app_rrt:further_techniques}

\paragraph{Ablation study on individual techniques.}

To further enhance performance through uptraining, we increased the number of uptraining tokens and employed knowledge distillation loss~\citep{DBLP:journals/corr/HintonVD15, DBLP:conf/emnlp/KimR16}. Specifically, we expanded the token number from 15 billion to 60 billion. Furthermore, we designated the teacher model as the full-size model for each architecture, uptrained on 15 billion tokens from the SlimPajama dataset. Given the huge number of uptraining tokens, we adopted an \textit{online} approach to extract logits from the teacher model. Four loss functions were utilized: forward KL (FKL; \citep{DBLP:conf/emnlp/KimR16}), reverse KL (RKL; \citep{DBLP:conf/iclr/Gu0WH24}), Jensen–Shannon divergence (JSD; \citep{DBLP:conf/iclr/AgarwalVZSGGB24}), and total variation distance (TVD; \citep{wen-etal-2023-f}). Table\,\ref{tab_rrt:kd_ablation_study} summarizes the controlled experimental results for each method. We finally selected forward KL as the distillation loss function due to its simplicity and superior performance. Especially, we observed a performance improvement of 1.7\%p attributed to the extended uptraining and up to 1.7\%p from the KD loss. This suggests that combining both techniques could yield even greater gains.

\clearpage

\begin{table}[h]
    \small
    \centering
    \resizebox{\textwidth}{!}{
    \setlength{\tabcolsep}{3pt}
    \begin{tabular}{c|cccc|cc|cc|rrr|ccccccc|cc}
    \toprule
     & \multicolumn{4}{c|}{\textbf{Uptrain}} & \multicolumn{2}{c|}{\textbf{Looping}} &  \multicolumn{2}{c|}{\textbf{LoRA}} & \multicolumn{3}{c|}{\textbf{Perplexity\,$\downarrow$}} & \multicolumn{9}{c}{\textbf{Few-shot Accuracy\,$\uparrow$}} \\
    \cmidrule(l{2pt}r{2pt}){2-5} \cmidrule(l{2pt}r{2pt}){6-7}  \cmidrule(l{2pt}r{2pt}){8-9} \cmidrule(l{2pt}r{2pt}){10-12}  \cmidrule(l{2pt}r{2pt}){13-21} 
      N-emb & PT & $N_{tok}$ & KD & Func & Block & Init  & Rank & Init & SlimP & RedP & PG19 & LD & HS & PQ & WG & ARC-e & ARC-c & OB & Avg & $\Delta$  \\
    \midrule
    0.99B &  \cmark & 15B & \xmark & - & 2 & Step & - & - &  {12.85} & {10.29} & {16.21} & {53.0} & {57.3} & {73.2} & {56.2} & {56.1} & {29.2} & {36.6} & {51.7} & - \\
    \rowcolor{gray!20}
    0.99B &  \cmark & 60B & \xmark & - & 2 & Step & - & - &  12.00 &  9.70 &  14.84 &  52.5 &  59.9 &  74.7 &  58.5 &  58.0 &  30.3 &  40.2 &  53.4 & \textcolor{custom_green}{\textbf{+1.7}} \\
    \midrule
    0.99B &  \cmark & 15B & \xmark & - & 2 & Step & - & - & {12.85} & {10.29} & {16.21} & {53.0} & {57.3} & {73.2} & {56.2} & {56.1} & {29.2} & {36.6} & {51.7} & - \\
    \rowcolor{gray!20}
    0.99B &  \cmark & 15B & \cmark & FKL & 2 & Step & - & - &  12.36 &  9.85 &  15.45 &  56.8 &  58.6 &  74.8 &  58.6 &  59.1 &  29.2 &  36.6 &  53.4 & \textcolor{custom_green}{\textbf{+1.7}} \\
    0.99B &  \cmark & 15B & \cmark & RKL & 2 & Step & - & - &  12.56 &  10.09 &  15.80 &  55.6 &  58.3 &  74.3 &  58.6 &  58.3 &  30.4 &  37.4 &  53.3 & \textcolor{custom_green}{\textbf{+1.6}} \\
    0.99B &  \cmark & 15B & \cmark & JSD & 2 & Step & - & - &  12.60 &  10.06 &  15.77 &  56.1 &  58.4 &  73.4 &  57.0 &  58.4 &  29.8 &  37.2 &  52.9 & \textcolor{custom_green}{\textbf{+1.2}} \\
    0.99B &  \cmark & 15B & \cmark & TVD & 2 & Step & - & - &  12.47 &  9.92 &  15.52 &  55.1 &  58.5 &  74.0 &  58.2 &  58.9 &  29.5 &  36.8 &  53.0 & \textcolor{custom_green}{\textbf{+1.3}} \\
    \midrule
    1.30B &  \cmark & 15B & \xmark & - & 2 & Avg & 256 & SVD & 12.10 &  9.71 &  14.89 &  58.2 &  60.7 &  73.7 &  59.0 &  57.6 &  32.1 &  {38.0} &  54.2 & - \\
    \rowcolor{gray!20}
    1.30B &  \cmark & 15B & \cmark & FKL & 2 & Avg & 256 & SVD &  11.90 &  9.52 &  14.63 &  59.9 &  62.0 &  74.1 &  60.0 &  58.6 &  33.2 &  38.0 &  55.1 & \textcolor{custom_green}{\textbf{+0.9}} \\
    1.30B &  \cmark & 15B & \cmark & RKL & 2 & Avg & 256 & SVD &  11.95 &  9.62 &  14.79 &  60.0 &  61.6 &  74.5 &  60.0 &  58.1 &  32.9 &  37.8 &  55.0 & \textcolor{custom_green}{\textbf{+0.8}} \\
    1.30B &  \cmark & 15B & \cmark & JSD & 2 & Avg & 256 & SVD &  12.09 &  9.65 &  14.81 &  58.1 &  61.1 &  73.1 &  60.8 &  59.0 &  33.2 &  38.6 &  54.8 & \textcolor{custom_green}{\textbf{+0.6}} \\
    1.30B &  \cmark & 15B & \cmark & TVD & 2 & Avg & 256 & SVD &  12.05 &  9.62 &  14.78 &  59.3 &  61.5 &  73.9 &  60.5 &  59.0 &  33.0 &  38.2 &  55.1 & \textcolor{custom_green}{\textbf{+0.9}} \\
    \bottomrule
    \end{tabular}
    }
    \caption{
    Evaluation results of ablation studies related to longer uptraining and knowledge distillation loss. Performance improvements, represented by Delta, were measured for each technique. For the knowledge distillation loss function, we experimented with four options: FKL, RKL, JSD, and TVD. 
    Forward KL was selected as the final configuration due to its simplicity and superior performance.
    }
    \label{tab_rrt:kd_ablation_study}
\end{table}

\paragraph{Overall performance after longer training with distillation loss.}

\begin{figure}[h]
    \centering
    \begin{subfigure}[t]{\textwidth}
        \includegraphics[width=\textwidth]{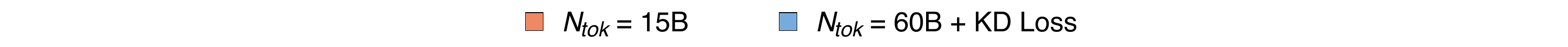}
    \end{subfigure}
    \centering
    \begin{subfigure}[t]{0.34\textwidth}
        \includegraphics[width=\textwidth]{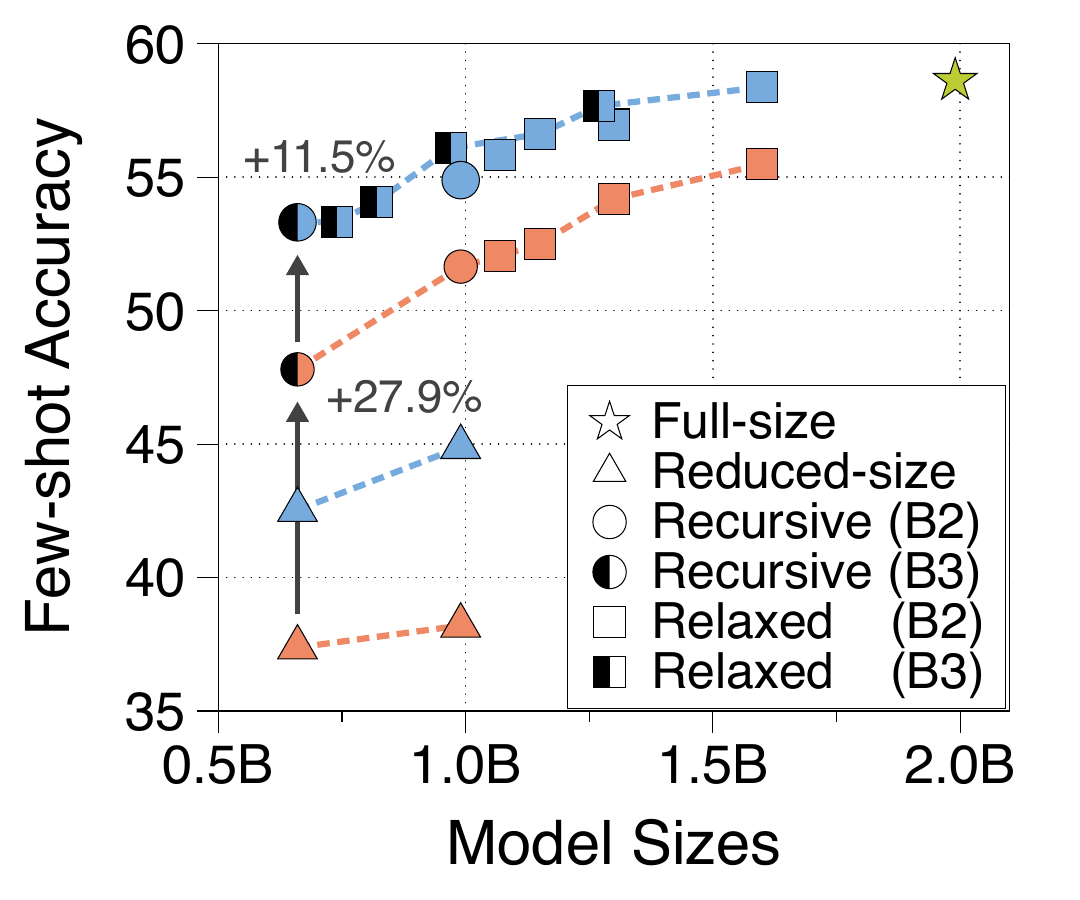}
        \subcaption{Gemma}
    \end{subfigure}
    \centering
    \begin{subfigure}[t]{0.31\textwidth}
        \includegraphics[width=\textwidth]{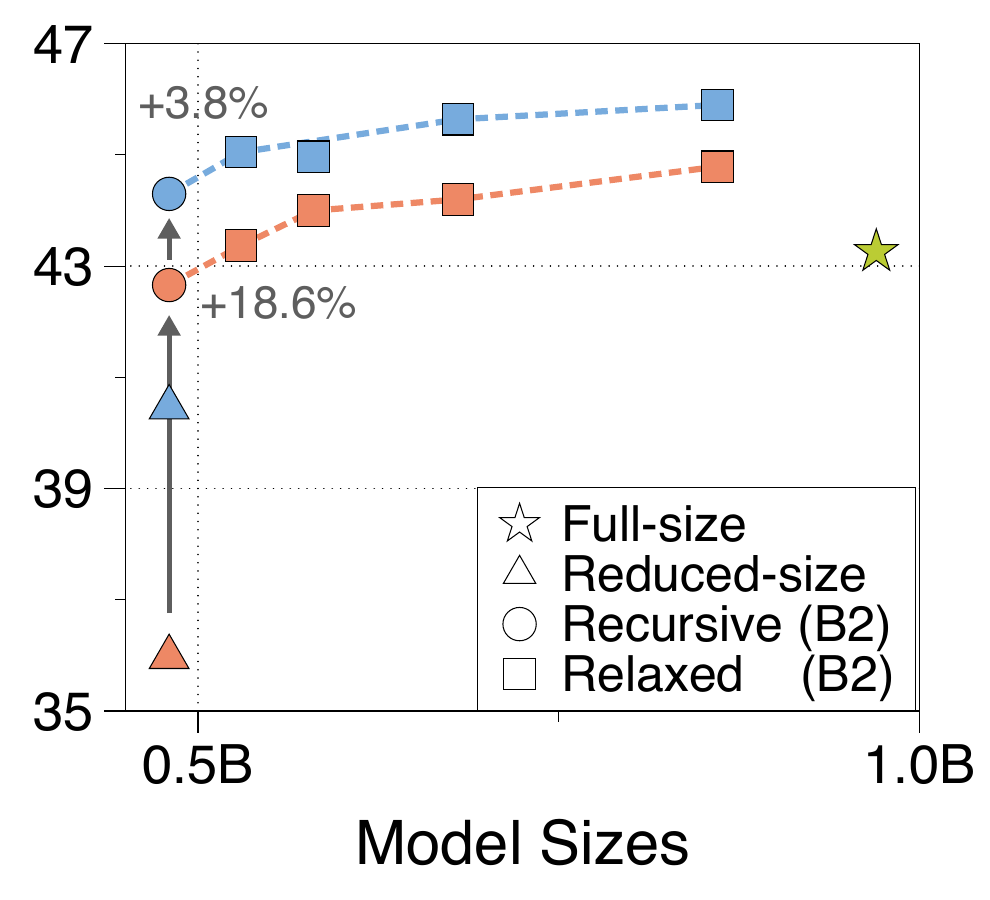}
        \subcaption{TinyLlama}
    \end{subfigure}
    \centering
    \begin{subfigure}[t]{0.31\textwidth}
        \includegraphics[width=\textwidth]{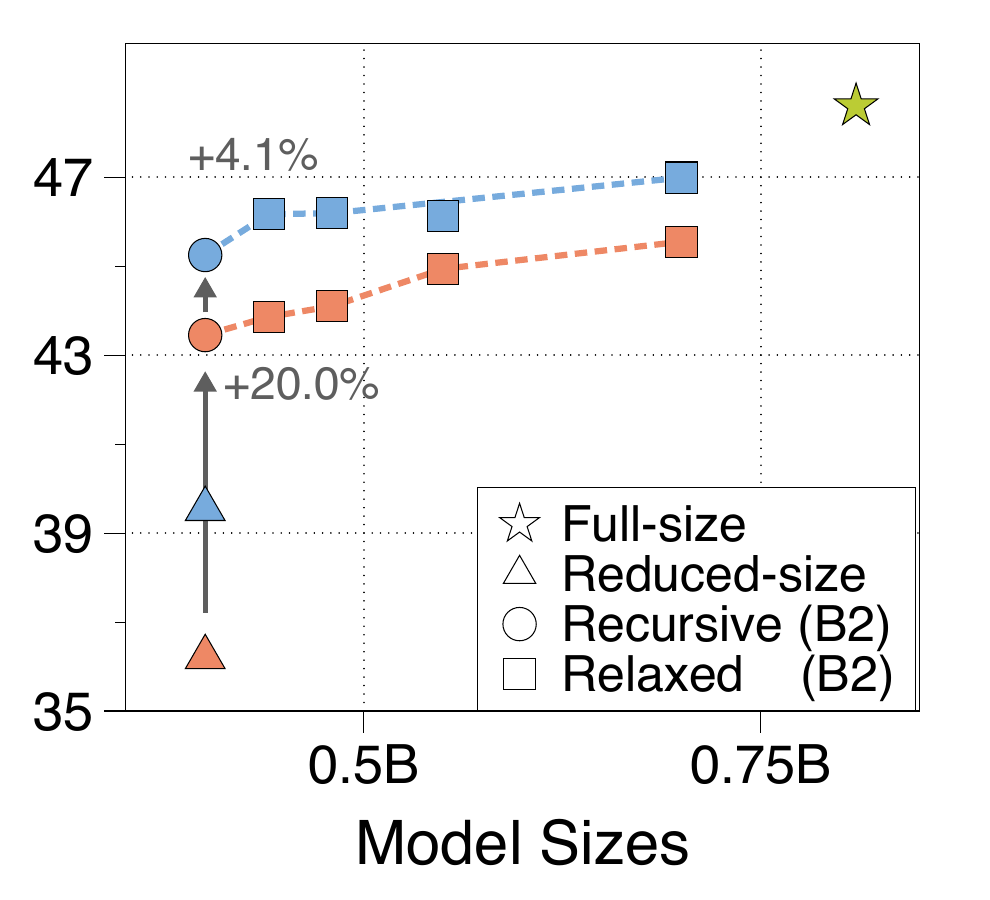}
        \subcaption{Pythia}
    \end{subfigure}
    \caption{
    Few-shot performance of three models with extended uptraining and knowledge distillation. Optimal configurations are used for each model size (Stepwise for recursive models and Average for relaxed models). 
    Dotted lines represent the Pareto frontier, showing the optimal trade-offs between model size and performance for each setting.
    }
    \label{fig_rrt:60b_kd_model_size_performance_app}
\end{figure}

Figure\,\ref{fig_rrt:60b_kd_model_size_performance_app} and Table\,\ref{tab_rrt:long_training_kd_app} summarize the performance gains achieved by incorporating these advanced training techniques.
We consistently observed substantial improvements in few-shot performance across all architectures and with varying numbers of looping blocks. We anticipate that further performance enhancements can be achieved by utilizing a superior teacher model and increasing the uptraining cost.

\begin{table}[ht!]
    \small
    \centering
    \resizebox{\textwidth}{!}{
    \setlength{\tabcolsep}{3pt}
    \renewcommand{\arraystretch}{0.95}
    \begin{tabular}{l|c|ccc|cc|cc|rrr|ccccccc|cc}
    \toprule
     & & \multicolumn{3}{c|}{\textbf{Uptrain}} & \multicolumn{2}{c|}{\textbf{Looping}} &  \multicolumn{2}{c|}{\textbf{LoRA}} & \multicolumn{3}{c|}{\textbf{Perplexity\,$\downarrow$}} & \multicolumn{9}{c}{\textbf{Few-shot Accuracy\,$\uparrow$}} \\
    \cmidrule(l{2pt}r{2pt}){3-5} \cmidrule(l{2pt}r{2pt}){6-7}  \cmidrule(l{2pt}r{2pt}){8-9} \cmidrule(l{2pt}r{2pt}){10-12}  \cmidrule(l{2pt}r{2pt}){13-21} 
     \textbf{Models} & N-emb & PT & $N_{tok}$ & KD & Block & Init  & Rank & Init & SlimP & RedP & PG19 & LD & HS & PQ & WG & ARC-e & ARC-c & OB & Avg & $\Delta$  \\
    \midrule
    & 1.99B &  \cmark & 60B & \xmark & - & - & - & - & 10.58 & 8.44 & 12.71 & 60.3 & 67.9 & 76.9 & 63.5 & 64.9 & 37.2 & {39.6} & 58.6  & - \\
     & 0.99B & \xmark & 60B & \cmark & - & - & - & - & 15.33 & 13.04 & 20.37 & 42.3 & 43.0 & 68.8 & 53.4 & 49.4 & 26.3 & 31.0 & 44.9 & - \\ 
     & 0.99B & \xmark & 15B & \xmark & - & - & - & - & 22.63 & 20.03 & 32.60 & 28.9 & 31.6 & 63.1 & 52.3 & 41.2 & 22.5 & 27.8 &  38.2 & - \\  
     & 0.66B & \xmark & 60B & \cmark & - & - & - & - & 16.79 & 14.39 & 22.85 & 37.5 & 38.4 & 68.7 & 50.4 & 46.5 & 24.6 & 31.6 & 42.5 & - \\ 
     & 0.66B & \xmark & 15B & \xmark & - & - & - & - & 24.44 & 21.69 & 36.03 & 27.2 & 30.6 & 63.8 & 50.5 & 40.6 & 22.0 & 27.0  & 37.4 & - \\ 
     \cmidrule(l{2pt}r{2pt}){2-21} 
     & 0.99B & \cmark & 15B & \xmark & 2 & Step & - & - &  {12.85} & {10.29} & {16.21} & {53.0} & {57.3} & {73.2} & {56.2} & {56.1} & {29.2} & {36.6} & {51.7} & - \\
     & 0.66B & \cmark & 15B & \xmark & 3 & Step & - & - &  {14.75} & {12.10} & {19.32} & {45.0} & {49.9} & {69.8} & {55.8} & {52.7} & {27.9} & {33.6} & {47.8} & - \\
     & 1.07B & \cmark & 15B & \xmark & 2 & Avg  & 64 & SVD &  12.83 &  10.35 &  16.02 &  55.9 &  56.8 &  72.5 &  56.8 &  55.7 &  30.6 &  36.2 &  52.1 & - \\
     & 1.15B & \cmark & 15B & \xmark & 2 & Avg  & 128 & SVD &  12.52 &  10.07 &  15.51 &  56.1 &  58.2 &  72.3 &  55.8 &  57.1 &  30.7 &  37.2 &  52.5 & - \\
     & 1.30B & \cmark & 15B & \xmark & 2 & Avg  & 256 & SVD &  12.10 &  9.71 &  14.89 &  58.2 &  60.7 &  73.7 &  59.0 &  57.6 &  32.1 &  {38.0} &  54.2 & - \\
    Gemma & 1.60B & \cmark & 15B & \xmark & 2 & Avg  & 512 & SVD &  {11.83} &  {9.46} &  {14.57} &  {59.3} &  {62.8} &  {74.0 }&  {61.6} &  {60.1} &  {32.9} &  37.6 &  {55.5} & - \\
     \cmidrule(l{2pt}r{2pt}){2-21} 
     \rowcolor[gray]{0.9} \cellcolor{white!20} 
     & 0.99B & \cmark & 60B & \cmark & 2 & Step  & - & - &  11.44 &  9.14 &  13.98 &  56.5 &  62.1 &  75.2 &  59.4 &  59.8 &  32.5 &  38.6 &  54.9 & \textcolor{custom_green}{\textbf{+3.2}} \\
     \rowcolor[gray]{0.9} \cellcolor{white!20} 
    & 1.07B & \cmark & 60B & \cmark & 2 & Avg  & 64 & SVD &  11.36 &  9.14 &  13.82 &  58.9 &  62.8 &  75.1 &  61.5 &  61.2 &  33.7 &  37.6 &  55.8 & \textcolor{custom_green}{\textbf{+3.7}} \\
    \rowcolor[gray]{0.9} \cellcolor{white!20} 
    & 1.15B & \cmark & 60B & \cmark & 2 & Avg  & 128 & SVD &  11.25 &  9.04 &  13.64 &  58.7 &  63.6 &  76.5 &  61.2 &  62.6 &  34.6 &  39.0 &  56.6 & \textcolor{custom_green}{\textbf{+4.1}} \\
    \rowcolor[gray]{0.9} \cellcolor{white!20} 
    & 1.30B & \cmark & 60B & \cmark & 2 & Avg  & 256 & SVD &  11.05 &  8.88 &  13.35 &  60.6 &  64.7 &  75.3 &  62.5 &  61.6 &  35.3 &  38.8 &  57.0 & \textcolor{custom_green}{\textbf{+2.8}} \\
    \rowcolor[gray]{0.9} \cellcolor{white!20} 
    & 1.60B & \cmark & 60B & \cmark & 2 & Avg  & 512 & SVD &  \textbf{10.81} &  \textbf{8.63} &  \textbf{12.94} &  61.4 &  \textbf{65.8} &  \textbf{76.3} & \textbf{63.5} &  \textbf{65.1} &  \textbf{37.1} &  39.4 & \textbf{58.4} & \textcolor{custom_green}{\textbf{+2.9}} \\
     \cmidrule(l{2pt}r{2pt}){2-21} 
     \rowcolor[gray]{0.9} \cellcolor{white!20} 
     & 0.66B & \cmark & 60B & \cmark & 3 & Step  & - & - &  12.27 &  9.90 &  15.24 &  55.6 &  58.1 &  73.1 &  60.2 &  58.8 &  30.2 &  37.2 &  53.3 & \textcolor{custom_green}{\textbf{+5.5}} \\
     \rowcolor[gray]{0.9} \cellcolor{white!20} 
    & 0.74B & \cmark & 60B & \cmark & 3 & Avg  & 64 & SVD &  12.13 &  9.80 &  14.95 &  55.5 &  58.3 &  73.5 &  60.1 &  58.0 &  31.1 &  36.8 &  53.3 & - \\
    \rowcolor[gray]{0.9} \cellcolor{white!20} 
    & 0.82B & \cmark & 60B & \cmark & 3 & Avg  & 128 & SVD &  11.83 &  9.53 &  14.51 &  56.7 &  60.2 &  74.2 &  59.8 &  59.1 &  33.0 &  35.4 &  54.1 & - \\
    \rowcolor[gray]{0.9} \cellcolor{white!20} 
    & 0.97B & \cmark & 60B & \cmark & 3 & Avg  & 256 & SVD &  11.43 &  9.17 &  13.87 &  59.3 &  62.6 &  74.7 &  61.2 &  61.6 &  32.9 &  \textbf{40.2} &  56.1 & - \\
    \rowcolor[gray]{0.9} \cellcolor{white!20} 
    & 1.27B & \cmark & 60B & \cmark & 3 & Avg  & 512 & SVD &  11.01 &  8.80 &  13.25 &  \textbf{61.5} &  64.9 &  76.2 &  62.0 &  64.3 &  35.6 &  39.2 &  57.7 & - \\
     \midrule
    & 0.97B &  \cmark & - & - & - & - & - & - & 12.26 & 9.37 & 11.94 & 43.3 & 42.2 & 66.8 & 53.4 & 44.7 & 23.2 & 29.2 & 43.3  & - \\
     & 0.48B &  \xmark & 60B & \cmark & - & - & - & - & 11.93 & 10.86 & 13.93 & 33.3 & 37.3 & 66.8 & 50.1 & 41.7 & 23.9 & 30.2 & 40.5 & - \\  
     & 0.48B &  \xmark & 15B & \xmark & - & - & - & - & 16.61 & 15.66 & 20.27 & 22.3 & 30.0 & 60.9 & 50.6 & 37.0 & 23.0 & 28.0 & 36.0 & - \\  
     \cmidrule(l{2pt}r{2pt}){2-21} 
     & 0.48B & \cmark & 15B & \xmark & 2 & Step & - & - &  {11.61} & {9.89}  & {13.00} &{ 39.6} &{ 39.8} & {66.5} & {52.9} & {44.3} & {24.9} & {30.6} & {42.7} & -   \\
      & 0.53B & \cmark & 15B & \xmark &  2 & Avg  & 64 & SVD &  11.22 &  9.66 &  12.51 &  41.8 &  41.6 &  67.0 &  53.3 &  43.9 &  24.7 &  {31.2} &  43.4 & - \\
    & 0.58B & \cmark & 15B &  \xmark & 2 & Avg  & 128 & SVD &  10.99 &  9.45 &  12.21 &  43.2 &  42.1 &  {68.3} &  53.2 &  44.8 &  25.9 &  30.4 &  44.0 & - \\
    TinyLlama & 0.68B & \cmark & 15B & \xmark &  2 & Avg  & 256 & SVD &  10.71 &  9.18 &  11.82 &  44.1 &  43.2 &  68.1 &  \textbf{53.5} &  44.4 &  25.7 &  30.4 &  44.2 & - \\
    & 0.86B & \cmark & 15B & \xmark &  2 & Avg  & 512 & SVD &  {10.46} &  {8.92} & { 11.50} &  {46.0} &  {44.1} &  68.2 &  53.0 &  {45.8} &  25.1 &  {31.2} &  {44.8} & - \\
     \cmidrule(l{2pt}r{2pt}){2-21} 
     \rowcolor[gray]{0.9} \cellcolor{white!20} 
    & 0.48B & \cmark & 60B & \cmark & 2 & Step  & - & - &  10.51 &  9.01 &  11.60 &  44.2 &  43.1 &  68.2 &  52.4 &  44.7 &  25.3 &  32.2 &  44.3 & \textcolor{custom_green}{\textbf{+1.6}} \\
    \rowcolor[gray]{0.9} \cellcolor{white!20} 
    & 0.53B & \cmark & 60B & \cmark & 2 & Avg  & 64 & SVD &  10.14 &  8.77 &  11.19 &  44.3 &  44.9 &  69.5 &  52.5 &  46.5 &  26.1 &  \textbf{31.6} &  45.1 & \textcolor{custom_green}{\textbf{+1.6}} \\
    \rowcolor[gray]{0.9} \cellcolor{white!20} 
    & 0.58B & \cmark & 60B & \cmark & 2 & Avg  & 128 & SVD &  10.07 &  8.68 &  11.07 &  45.9 &  45.1 &  69.4 &  50.5 &  46.8 &  25.4 &  \textbf{31.6} &  45.0 & \textcolor{custom_green}{\textbf{+1.0}} \\
    \rowcolor[gray]{0.9} \cellcolor{white!20} 
    & 0.68B & \cmark & 60B & \cmark & 2 & Avg  & 256 & SVD &  9.96 &  8.56 &  10.93 &  46.2 &  45.7 &  69.0 &  53.2 &  \textbf{47.9} &  25.9 &  \textbf{31.6} &  45.6 & \textcolor{custom_green}{\textbf{+1.4}} \\
    \rowcolor[gray]{0.9} \cellcolor{white!20} 
    & 0.86B & \cmark & 60B & \cmark & 2 & Avg  & 512 & SVD &  \textbf{9.85} &  \textbf{8.44} &  \textbf{10.76} &  \textbf{47.4} &  \textbf{46.3} &  \textbf{69.7} &  52.8 &  47.5 &  \textbf{26.3} &  31.4 &  \textbf{45.9} & \textcolor{custom_green}{\textbf{+1.1}} \\
     \midrule
    & 0.81B &  \cmark & 60B & \xmark & - & - & - & - & 12.83 & 9.76 & 13.57 &  53.0 & 50.2 & 71.1 & 54.8 & 51.9 & {27.7} & 31.6 & {48.6}  & - \\
     & 0.40B &  \xmark & 60B & \cmark & - & - & - & - & 18.27 & 14.39 & 21.93 & 32.1 & 35.0 & 66.1 & 49.6 & 42.9 & 24.2 & 27.0  & 39.5 & -  \\
     & 0.40B &  \xmark & 15B & \xmark & - & - & - & - & 25.69 & 20.00 & 32.08 & 24.3 & 30.0 & 61.9 & 50.7 & 38.3 & 22.3 & 26.0  & 36.2 & -  \\ 
     \cmidrule(l{2pt}r{2pt}){2-21} 
     & 0.40B & \cmark & 15B & \xmark & 2 & Step & - & - & {16.38} & {12.37} & {17.74} & 43.4 & {40.5} & 67.4 & 50.8 & {46.3} & 25.7 & 30.0 & 43.5 & -   \\
     & 0.44B & \cmark & 15B & \xmark  & 2 & Avg  & 64 & SVD &  16.03 &  12.19 &  17.59 &  45.8 &  40.9 &  67.3 &  50.0 &  45.8 &  25.5 &  31.8 &  43.9 & - \\
     & 0.48B & \cmark & 15B & \xmark  & 2 & Avg  & 128 & SVD &  15.67 &  11.93 &  17.10 &  46.9 &  41.9 &  67.4 &  51.2 &  45.4 &  24.8 &  31.2 &  44.1 & - \\
    Pythia & 0.55B & \cmark & 15B & \xmark & 2 & Avg  & 256 & SVD &  15.22 &  11.54 &  16.47 &  48.5 &  43.3 &  67.2 &  51.4 &  46.7 &  25.5 &  32.0 &  44.9 & - \\
    & 0.70B & \cmark & 15B & \xmark & 2 & Avg  & 512 & SVD &  {14.70} &  {11.07} &  {15.71} &  {50.2} &  {44.7} &  68.2 &  51.6 &  {47.6} &  25.4 &  31.2 &  {45.6} & - \\
     \cmidrule(l{2pt}r{2pt}){2-21} 
    \rowcolor[gray]{0.9} \cellcolor{white!20} 
     & 0.40B & \cmark & 60B & \cmark & 2 & Step  & - & - &  14.59 &  11.13 &  15.79 &  47.8 &  43.8 &  69.3 &  52.0 &  48.1 &  25.4 &  30.4 &  45.2 & \textcolor{custom_green}{\textbf{+1.7}} \\
     \rowcolor[gray]{0.9} \cellcolor{white!20} 
    & 0.44B & \cmark & 60B & \cmark & 2 & Avg  & 64 & SVD &  14.24 &  10.89 &  15.52 &  50.0 &  44.5 &  68.9 &  \textbf{54.1} &  48.0 &  26.5 &  31.2 &  46.2 & \textcolor{custom_green}{\textbf{+2.3}} \\
    \rowcolor[gray]{0.9} \cellcolor{white!20} 
    & 0.48B & \cmark & 60B & \cmark & 2 & Avg  & 128 & SVD &  14.10 &  10.79 &  15.27 &  50.1 &  45.5 &  69.0 &  52.6 &  48.3 &  25.8 &  32.0 &  46.2 & \textcolor{custom_green}{\textbf{+2.1}} \\
    \rowcolor[gray]{0.9} \cellcolor{white!20} 
    & 0.55B & \cmark & 60B & \cmark & 2 & Avg  & 256 & SVD &  13.91 &  10.61 &  14.91 &  50.5 &  45.6 &  68.7 &  51.2 &  48.4 &  25.7 &  \textbf{32.8} &  46.1 & \textcolor{custom_green}{\textbf{+1.2}} \\
    \rowcolor[gray]{0.9} \cellcolor{white!20} 
    & 0.70B & \cmark & 60B & \cmark & 2 & Avg  & 512 & SVD &  \textbf{13.59} &  \textbf{10.38} &  \textbf{14.43} &  \textbf{52.0} &  \textbf{47.0} &  \textbf{69.6} &  53.4 &  \textbf{48.9} &  \textbf{26.9} &  31.2 &  \textbf{47.0} & \textcolor{custom_green}{\textbf{+1.4}} \\
    \bottomrule
    \end{tabular}
    }
    \caption{
    Evaluation results of our Recursive Transformers with 60 billion token uptraining and knowledge distillation loss. We utilized the forward KL loss as the knowledge distillation loss function. Full-size model baselines for Gemma and Pythia are the pretrained models further uptrained on 60 billion tokens, accounting for distribution shifts between Slimapajama and their pretraining datasets. Delta\,($\Delta$) represents the accuracy differences between the longer uptrained models with KD and their 15 billion uptrained counterparts. 
    }
    \label{tab_rrt:long_training_kd_app}
\end{table}

\clearpage
\subsection{Expanded Results of Early-Exit Training}
\label{app_rrt:early_exit}

\paragraph{Ablation study on early-exit training strategy.}

To enable early-exiting capabilities, all models require additional training to align intermediate representations with classifier heads. In this study, we conduct ablation studies on various strategies, demonstrating Recursive Transformers can be transformed into early-exiting models without compromising final loop output's performance. Table\,\ref{tab_rrt:early_exit_ablation_app} presents a comprehensive summary of experimental results across various categories, including training procedures, loss functions, and early-exit training data. Our key findings are as follows:
\begin{itemize}[leftmargin=*]
    \item Post-training after uptraining is essential for preserving final loop performance. Jointly training intermediate loop output during the uptraining phase (co-training) significantly degraded the final output performance, even with an aggressive loss coefficient strategy.
    \item While freezing all parameters, we attempted to train intermediate loop outputs by attaching trainable LoRA modules to frozen classifier head. However, we found that this proved ineffective.
    \item The aggressive coefficient strategy successfully maintained final loop output performance while enhancing intermediate output performance. Moreover, incorporating knowledge distillation from detached final outputs further enhanced intermediate performance.
    \item No significant performance differences were observed when using the overlapped uptraining dataset versus new SlimPajama tokens for post-training.
\end{itemize}
Finally, we opted to perform post-training with new tokens sourced from the same SlimPajama dataset. Moreover, we incorporated a distillation loss from the final loop output, while aggressively reducing the loss coefficients of intermediate outputs.

\paragraph{Early-exit training results on final models.}

We applied the aggressive coefficient strategy with distillation loss to our final models (uptrained on 60 billion tokens with knowledge distillation). Tables\,\ref{tab_rrt:final_early_exit_gemma_app} and \ref{tab_rrt:final_early_exit_tinyllama_pythia_app} summarize the performance of intermediate loops and the final loop across three models. For fair comparison, the full-size models (Gemma and Pythia) were also uptrained with 60 billion tokens and then post-trained with 15 billion tokens. Consistent with previous findings, the aggressive coefficient strategy yielded the best performance across both intermediate and final outputs.

However, we find that intermediate loop outputs in LoRA-relaxed models underperformed their non-relaxed counterparts (recursive models). This could potentially reduce throughput gain, as early loop performance directly influences the number of tokens eligible for early-exit. In perfectly tied looping blocks, intermediate outputs seem to be distilled from the last, as all gradients are backpropagated to the same parameters. Conversely, since LoRA modules allow each layer to specialize based on its depth, intermediate representations appear to be optimized to facilitate performance of the final output, not for their own sake. Hence, relaxation introduces a trade-off between final performance and early-exiting benefits. As the optimal strategy derived from the non-relaxed models was directly applied to the relaxed models, further exploration of optimal strategies specifically for relaxed models is left for future work.

\begin{table}[ht]
    \small
    \centering
    \resizebox{\textwidth}{!}{
    \setlength{\tabcolsep}{3pt}
    \begin{tabular}{c|cc|cc|cccccc|rrr|ccccccc|cc}
    \toprule
     & \multicolumn{2}{c|}{\textbf{Uptrain}} & \multicolumn{2}{c|}{\textbf{Looping}} &  \multicolumn{6}{c|}{\textbf{Early-Exit Train}} & \multicolumn{3}{c|}{\textbf{Perplexity\,$\downarrow$}} & \multicolumn{9}{c}{\textbf{Few-shot Accuracy\,$\uparrow$}} \\
    \cmidrule(l{2pt}r{2pt}){2-3} \cmidrule(l{2pt}r{2pt}){4-5}  \cmidrule(l{2pt}r{2pt}){6-11} \cmidrule(l{2pt}r{2pt}){12-14}  \cmidrule(l{2pt}r{2pt}){15-23} 
     N-emb & PT & $N_{tok}$ & Block & Init  & Train & Freeze& $N_{tok}$ & CE & KD & Data & SlimP & RedP & PG19 & LD & HS & PQ & WG & ARC-e & ARC-c & OB & Avg & $\Delta$  \\
    \midrule
    1.99B &  \cmark & 15B  & - & - & - & - & - & - & - & - & 10.76 & 8.47 & 13.08 & 63.5 & 68.5 & 77.0 & 63.5 & 67.6 & 38.1 & {42.6} & 60.1  & - \\
     0.99B & \xmark & 15B  & - & - & - & - & - & - & - & - & 22.63 & 20.03 & 32.60 & 28.9 & 31.6 & 63.1 & 52.3 & 41.2 & 22.5 & 27.8 &  38.2 & - \\ 
     0.99B & \cmark & 15B  & 2 & Step &  - &- & - & - & - & - & {12.85} & {10.29} & {16.21} & {53.0} & {57.3} & {73.2} & {56.2} & {56.1} & {29.2} & {36.6} & {51.7} & - \\
    \midrule
     \multirow{2}{*}{0.99B} & \multirow{2}{*}{\cmark} & \multirow{2}{*}{15B}  & \multirow{2}{*}{2} & \multirow{2}{*}{Step}  &  \multirow{2}{*}{Post-} &  \multirow{2}{*}{\xmark} & \multirow{2}{*}{15B}&  \multirow{2}{*}{Weighted} &  \multirow{2}{*}{\xmark} &  \multirow{2}{*}{Ovlp}  &  12.97 &  10.51 &  16.55 &  48.9 &  55.5 &  72.7 &  55.3 &  54.9 &  30.1 &  36.0 &  50.5 & \textcolor{custom_red}{\textbf{--\,1.2}} \\
     &  &  &  & &  &  &  & &    &  &  13.11 &  10.59 &  16.71 &  49.5 &  54.8 &  72.0 &  53.4 &  54.1 &  29.1 &  35.6 &  49.8 & - \\
     \addlinespace[-1pt]
     \cmidrule(l{2pt}r{2pt}){12-23} 
     \addlinespace[-1pt]
     \multirow{2}{*}{0.99B} & \multirow{2}{*}{\cmark} & \multirow{2}{*}{15B} & \multirow{2}{*}{2} & \multirow{2}{*}{Step} &  \multirow{2}{*}{Post-} &  \multirow{2}{*}{\xmark}& \multirow{2}{*}{15B} &  \multirow{2}{*}{Agg\,(0.3)} &  \multirow{2}{*}{\xmark} &  \multirow{2}{*}{Ovlp} &  12.60 &  10.21 &  15.75 &  51.8 &  58.2 &  73.7 &  56.8 &  57.0 &  29.9 &  37.8 &  52.2 & \textcolor{custom_green}{\textbf{+0.5}} \\
     &  &  &  & &  &  &   &  & &  &  13.63 &  11.02 &  17.55 &  47.5 &  53.0 &  71.2 &  54.9 &  50.2 &  28.2 &  34.8 &  48.5 & - \\
     \addlinespace[-1pt]
     \cmidrule(l{2pt}r{2pt}){12-23} 
     \addlinespace[-1pt]
     \rowcolor[gray]{0.9}
      &  &  &  &  &  &   &   &  &   &  &  12.37 &  9.94 &  15.37 &  53.0 &  59.1 &  73.9 &  55.4 &  57.4 &  30.6 &  37.8 &  52.5 & \textcolor{custom_green}{\textbf{+0.8}} \\
      \rowcolor[gray]{0.9}
     \multirow{-2}{*}{0.99B} & \multirow{-2}{*}{\cmark} &  \multirow{-2}{*}{15B} & \multirow{-2}{*}{2} & \multirow{-2}{*}{Step} & \multirow{-2}{*}{Post-} & \multirow{-2}{*}{\xmark} &  \multirow{-2}{*}{15B}  & \multirow{-2}{*}{Agg\,(0.1)} & \multirow{-2}{*}{\xmark} & \multirow{-2}{*}{Ovlp} &  14.55 &  11.87 &  19.00 &  45.9 &  51.2 &  71.4 &  54.5 &  48.1 &  26.8 &  32.0 &  47.1 & - \\
     \addlinespace[-1pt]
     \cmidrule(l{2pt}r{2pt}){12-23} 
     \addlinespace[-1pt]
     \multirow{2}{*}{0.99B} & \multirow{2}{*}{\cmark} & \multirow{2}{*}{15B} & \multirow{2}{*}{2} & \multirow{2}{*}{Step} &  \multirow{2}{*}{Post-} &  \multirow{2}{*}{\xmark} &  \multirow{2}{*}{15B}  &  \multirow{2}{*}{Agg\,(0.05)} &  \multirow{2}{*}{\xmark} &  \multirow{2}{*}{Ovlp} &  12.33 &  9.90 &  15.31 &  52.8 &  59.2 &  73.6 &  57.5 &  57.7 &  30.5 &  37.2 &  52.6 & \textcolor{custom_green}{\textbf{+0.9}} \\
     &  &  &  & &  &  &   &  & & &  15.70 &  12.93 &  20.69 &  43.1 &  49.8 &  69.8 &  55.2 &  46.0 &  26.9 &  31.2 &  46.0 & - \\
     \addlinespace[-1pt]
     \cmidrule(l{2pt}r{2pt}){12-23} 
     \addlinespace[-1pt]
      &  & &  &  &  &   &   &  &  &   &  12.28 &  9.80 &  15.23 &  52.9 &  59.5 &  73.3 &  56.5 &  57.2 &  30.1 &  37.2 &  52.4 & \textcolor{custom_green}{\textbf{+0.7}} \\
     \multirow{-2}{*}{0.99B} & \multirow{-2}{*}{\cmark} &  \multirow{-2}{*}{15B} & \multirow{-2}{*}{2} & \multirow{-2}{*}{Step} & \multirow{-2}{*}{Post-} & \multirow{-2}{*}{\xmark} &  \multirow{-2}{*}{15B}  & \multirow{-2}{*}{Agg\,(0.01)} & \multirow{-2}{*}{\xmark} & \multirow{-2}{*}{Ovlp} &  22.76 &  20.37 &  30.39 &  32.2 &  45.2 &  67.5 &  53.9 &  40.3 &  26.3 &  29.2 &  42.1 & - \\
     \midrule
     \multirow{2}{*}{0.99B} & \multirow{2}{*}{\cmark} & \multirow{2}{*}{15B} & \multirow{2}{*}{2} & \multirow{2}{*}{Step} &  \multirow{2}{*}{Post-} &  \multirow{2}{*}{\xmark} &  \multirow{2}{*}{15B}  &  \multirow{2}{*}{Weighted} &  \multirow{2}{*}{\cmark} &  \multirow{2}{*}{Ovlp} &  13.04 &  10.57 &  16.66 &  47.7 &  55.1 &  73.2 &  55.6 &  54.5 &  29.1 &  37.2 &  50.4 & \textcolor{custom_red}{\textbf{--\,1.3}} \\
     &  &  &  & &  &  &   &  & & &  13.04 &  10.54 &  16.66 &  48.3 &  54.9 &  72.1 &  55.9 &  54.3 &  28.4 &  35.4 &  49.9 & - \\
     \addlinespace[-1pt]
     \cmidrule(l{2pt}r{2pt}){12-23} 
     \addlinespace[-1pt]
     \rowcolor[gray]{0.9}
      &  &  &  & &  &   &   &  & &    &  12.40 &  9.97 &  15.42 &  52.9 &  58.9 &  73.7 &  55.7 &  57.5 &  31.1 &  38.2 &  52.6 & \textcolor{custom_green}{\textbf{+0.9}} \\
    \rowcolor[gray]{0.9}
     \multirow{-2}{*}{0.99B} & \multirow{-2}{*}{\cmark} &  \multirow{-2}{*}{15B} & \multirow{-2}{*}{2} & \multirow{-2}{*}{Step} & \multirow{-2}{*}{Post-} & \multirow{-2}{*}{\xmark} &  \multirow{-2}{*}{15B}  & \multirow{-2}{*}{Agg\,(0.1)} & \multirow{-2}{*}{\cmark} & \multirow{-2}{*}{Ovlp} &  14.11 &  11.47 &  18.32 &  46.3 &  52.1 &  71.6 &  55.3 &  49.2 &  28.5 &  32.6 &  48.0 & - \\
     \midrule
     \multirow{2}{*}{0.99B} & \multirow{2}{*}{\cmark} & \multirow{2}{*}{15B} & \multirow{2}{*}{2} & \multirow{2}{*}{Step} &  \multirow{2}{*}{Post-} &  \multirow{2}{*}{\cmark} &  \multirow{2}{*}{15B}  &  \multirow{2}{*}{Standard} &  \multirow{2}{*}{\xmark} &  \multirow{2}{*}{Ovlp} & {12.85} & {10.29} & {16.21} & {53.0} & {57.3} & {73.2} & {56.2} & {56.1} & {29.2} & {36.6} & {51.7} & \textcolor{gray}{\textbf{+0.0}} \\
     &  &  &  & &  &  &   &  & & & 43.74 &  41.63 &  56.78 &  5.3 &  37.9 &  61.4 &  52.6 &  35.3 &  24.0 &  29.0 &  35.0 & - \\
     \addlinespace[-1pt]
     \cmidrule(l{2pt}r{2pt}){12-23} 
     \addlinespace[-1pt]
     \multirow{2}{*}{0.99B} & \multirow{2}{*}{\cmark} & \multirow{2}{*}{15B}  & \multirow{2}{*}{2} & \multirow{2}{*}{Step} &  \multirow{2}{*}{Post-} &  \multirow{2}{*}{\cmark} &  \multirow{2}{*}{15B}  &  \multirow{2}{*}{Standard} &  \multirow{2}{*}{\cmark} &  \multirow{2}{*}{Ovlp} & {12.85} & {10.29} & {16.21} & {53.0} & {57.3} & {73.2} & {56.2} & {56.1} & {29.2} & {36.6} & {51.7} & \textcolor{gray}{\textbf{+0.0}} \\
     &  &  &  & &  &  &   &  & &  &  43.09 &  39.97 &  55.37 &  5.6 &  37.7 &  62.5 &  52.7 &  34.5 &  24.7 &  29.2 &  35.3 & - \\
     \midrule
     \multirow{2}{*}{0.99B} & \multirow{2}{*}{\cmark} & \multirow{2}{*}{15B}  & \multirow{2}{*}{2} & \multirow{2}{*}{Step} &  \multirow{2}{*}{Co-} &  \multirow{2}{*}{\xmark} &  \multirow{2}{*}{15B}  &  \multirow{2}{*}{Agg\,(0.1)} &  \multirow{2}{*}{\xmark} &  \multirow{2}{*}{Ovlp} &  13.24 &  10.67 &  16.98 &  50.1 &  54.2 &  72.2 &  53.7 &  54.7 &  28.9 &  37.4 &  50.2 & \textcolor{custom_red}{\textbf{--\,1.5}} \\
     &  &  &  & &  &  &   &  & &  &  13.59 &  10.89 &  17.42 &  50.6 &  52.7 &  71.2 &  54.4 &  53.0 &  27.5 &  35.0 &  49.2 & - \\
     \midrule
     \rowcolor[gray]{0.9}
      &  &  &  &  &  &    &  &   &  & &  12.34 &  9.92 &  15.31 &  52.3 &  59.0 &  73.8 &  57.6 &  55.5 &  30.4 &  37.2 &  52.3 & \textcolor{custom_green}{\textbf{+0.6}} \\
     \rowcolor[gray]{0.9}
    \multirow{-2}{*}{0.99B} & \multirow{-2}{*}{\cmark} &  \multirow{-2}{*}{15B} & \multirow{-2}{*}{2} & \multirow{-2}{*}{Step} & \multirow{-2}{*}{Post-} & \multirow{-2}{*}{\xmark} &  \multirow{-2}{*}{15B}  & \multirow{-2}{*}{Agg\,(0.1)} & \multirow{-2}{*}{\xmark} & \multirow{-2}{*}{New} &  14.49 &  11.86 &  18.89 &  43.9 &  51.3 &  71.0 &  54.9 &  48.1 &  27.5 &  31.4 &  46.9 & - \\
    \bottomrule
    \end{tabular}
    }
    \caption{
    Ablation studies on early-exit training for recursive Gemma models. We evaluated performance in a static-exiting scenario\,\citep{DBLP:conf/nips/SchusterFG0B0TM22, DBLP:conf/emnlp/BaeKSY23}, where all tokens exit at either first or second iteration loops (9th or 18th depths). We explored post-training (after uptraining) and co-training (during uptraining) approaches. Moreover, we explored freezing uptrained weights and adding LoRA with the rank of 128 to the classifier head. Different coefficient values were tested for the aggressive CE loss function. Early-exit training utilized 15 billion tokens, either overlapping with uptraining data or entirely new. Delta\,($\Delta$) indicates the performance changes of the final loop outputs. We highlight the final configuration: post-training with aggressive CE and KD loss on 15 billion new tokens.
    }
    \label{tab_rrt:early_exit_ablation_app}
\end{table}

\begin{table}[ht!]
    \small
    \centering
    \resizebox{\textwidth}{!}{
    \setlength{\tabcolsep}{3pt}
    \renewcommand{\arraystretch}{0.85}
    \begin{tabular}{c|ccc|cc|cc|ccc|rrr|ccccccc|cc}
    \toprule
     & \multicolumn{3}{c|}{\textbf{Uptrain}} & \multicolumn{2}{c|}{\textbf{Looping}} &  \multicolumn{2}{c|}{\textbf{LoRA}} &  \multicolumn{3}{c|}{\textbf{Early-exit\,Train}} & \multicolumn{3}{c|}{\textbf{Perplexity\,$\downarrow$}} & \multicolumn{9}{c}{\textbf{Few-shot Accuracy\,$\uparrow$}} \\
    \cmidrule(l{2pt}r{2pt}){2-4} \cmidrule(l{2pt}r{2pt}){5-6}  \cmidrule(l{2pt}r{2pt}){7-8} \cmidrule(l{2pt}r{2pt}){9-11} \cmidrule(l{2pt}r{2pt}){12-14}  \cmidrule(l{2pt}r{2pt}){15-23} 
     N-emb & PT & $N_{tok}$ & KD & Block & Init  & Rank & Init & $N_{tok}$ & CE & KD & SlimP & RedP & PG19 & LD & HS & PQ & WG & ARC-e & ARC-c & OB & Avg & $\Delta$  \\
    \midrule
    1.99B & \cmark & 60B & \xmark & - & -  & - & - &  - & -  & - &  10.58 & 8.44 & 12.71 & 60.3 & 67.9 & 76.9 & 63.5 & 64.9 & 37.2 & {39.6} & 58.6  & - \\
    1.99B & \cmark & 75B & \xmark & - & -  & - & - &  - & -  & - &  11.03 &  8.88 &  13.33 &  57.0 &  65.9 &  76.2 &  63.9 &  63.0 &  35.9 &  38.8 &  57.3 & \textcolor{custom_red}{\textbf{--\,1.3}} \\
    \midrule
    0.99B & \cmark & 60B & \cmark & 2 & Step  & - & - &  - & -  & -&  11.44 &  9.14 &  13.98 &  56.5 &  62.1 &  75.2 &  59.4 &  59.8 &  32.5 &  38.6 &  54.9 & - \\
    1.07B & \cmark & 60B & \cmark & 2 & Avg  & 64 & SVD &  - & - & - &  11.36 &  9.14 &  13.82 &  58.9 &  62.8 &  75.1 &  61.5 &  61.2 &  33.7 &  37.6 &  55.8 & - \\
    1.15B & \cmark & 60B & \cmark & 2 & Avg  & 128 & SVD &  - & - & - &  11.25 &  9.04 &  13.64 &  58.7 &  63.6 &  76.5 &  61.2 &  62.6 &  34.6 &  39.0 &  56.6 & - \\
    1.30B & \cmark & 60B & \cmark & 2 & Avg  & 256 & SVD &  - & - & - &  11.05 &  8.88 &  13.35 &  60.6 &  64.7 &  75.3 &  62.5 &  61.6 &  35.3 &  38.8 &  57.0 & - \\
    1.60B & \cmark & 60B & \cmark & 2 & Avg  & 512 & SVD &  - & - & - &  {10.81} &  {8.63} &  {12.94} &  61.4 &  {65.8} &  {76.3} & { 63.5} &  {65.1} &  {37.1} &  39.4 & {58.4} & - \\
    \midrule
       & & & & & & & & &  & &  11.71 &  9.56 &  14.46 &  54.0 &  61.7 &  75.1 &  58.9 &  58.6 &  31.9 &  37.6 &  54.0 & \textcolor{custom_red}{\textbf{--\,0.9}} \\
    \multirow{-2}{*}{0.99B} & \multirow{-2}{*}{\cmark} & \multirow{-2}{*}{60B} & \multirow{-2}{*}{\cmark} & \multirow{-2}{*}{2} & \multirow{-2}{*}{Step} & \multirow{-2}{*}{-} & \multirow{-2}{*}{-} & \multirow{-2}{*}{15B}  & \multirow{-2}{*}{Agg\,(0.1)} & \multirow{-2}{*}{\cmark}  &  13.68 &  11.39 &  17.60 &  45.0 &  54.1 &  71.9 &  58.5 &  49.8 &  28.8 &  33.4 &  48.8 & - \\
    \addlinespace[-1pt] \cmidrule(l{2pt}r{2pt}){12-23} \addlinespace[-1pt]
      & & & & & & & & & & & 11.79 &  9.70 &  14.52 &  53.7 &  60.8 &  73.6 &  61.1 &  58.7 &  32.9 &  37.2 &  54.0 &\textcolor{custom_red}{\textbf{--\,1.8}} \\
    \multirow{-2}{*}{1.07B} & \multirow{-2}{*}{\cmark} & \multirow{-2}{*}{60B} & \multirow{-2}{*}{\cmark} & \multirow{-2}{*}{2} & \multirow{-2}{*}{Avg} & \multirow{-2}{*}{64} & \multirow{-2}{*}{SVD}  & \multirow{-2}{*}{15B}  & \multirow{-2}{*}{Agg\,(0.1)} & \multirow{-2}{*}{\cmark}  &  19.45 &  16.46 &  26.10 &  30.7 &  37.9 &  66.5 &  55.3 &  42.2 &  25.3 &  27.6 &  40.8 & - \\
    \addlinespace[-1pt] \cmidrule(l{2pt}r{2pt}){12-23} \addlinespace[-1pt]
      & & & & & & & & & & &  11.66 &  9.59 &  14.32 &  53.3 &  62.1 &  74.9 &  60.0 &  59.9 &  33.4 &  38.8 &  54.6 &\textcolor{custom_red}{\textbf{--\,2.0}} \\
    \multirow{-2}{*}{1.15B} & \multirow{-2}{*}{\cmark} & \multirow{-2}{*}{60B} & \multirow{-2}{*}{\cmark} & \multirow{-2}{*}{2} & \multirow{-2}{*}{Avg} & \multirow{-2}{*}{128} & \multirow{-2}{*}{SVD} & \multirow{-2}{*}{15B}   & \multirow{-2}{*}{Agg\,(0.1)} & \multirow{-2}{*}{\cmark}  &  19.65 &  16.77 &  26.44 &  29.7 &  37.7 &  66.8 &  52.6 &  41.4 &  25.3 &  28.0 &  40.2 & - \\
    \addlinespace[-1pt] \cmidrule(l{2pt}r{2pt}){12-23} \addlinespace[-1pt]
      & & & & & & & & & & &  11.47 &  9.39 &  14.03 &  54.9 &  63.0 &  74.5 &  61.7 &  60.5 &  33.1 &  38.4 &  55.2 & \textcolor{custom_red}{\textbf{--\,1.8}} \\
    \multirow{-2}{*}{1.30B} & \multirow{-2}{*}{\cmark} & \multirow{-2}{*}{60B} & \multirow{-2}{*}{\cmark} & \multirow{-2}{*}{2} & \multirow{-2}{*}{Avg} & \multirow{-2}{*}{256} & \multirow{-2}{*}{SVD} & \multirow{-2}{*}{15B}   & \multirow{-2}{*}{Agg\,(0.1)} & \multirow{-2}{*}{\cmark}  &  19.67 &  16.82 &  26.40 &  29.7 &  38.3 &  66.4 &  53.1 &  43.8 &  24.7 &  27.6 &  40.5 & - \\
    \addlinespace[-1pt] \cmidrule(l{2pt}r{2pt}){12-23} \addlinespace[-1pt]
      & & & & & & & & & & &  11.20 &  9.14 &  13.58 &  57.2 &  64.1 &  75.2 &  61.7 &  62.1 &  34.6 &  38.2 &  56.2 & \textcolor{custom_red}{\textbf{--\,2.2}} \\
    \multirow{-2}{*}{1.60B} & \multirow{-2}{*}{\cmark} & \multirow{-2}{*}{60B} & \multirow{-2}{*}{\cmark} & \multirow{-2}{*}{2} & \multirow{-2}{*}{Avg} & \multirow{-2}{*}{512} & \multirow{-2}{*}{SVD}  & \multirow{-2}{*}{15B}  & \multirow{-2}{*}{Agg\,(0.1)} & \multirow{-2}{*}{\cmark}  &  19.29 &  16.47 &  25.73 &  32.0 &  39.6 &  67.6 &  53.3 &  43.2 &  25.8 &  30.2 &  41.7 & -\\
    \midrule
       & & & & & & & & & & & 12.11 &  9.98 &  14.97 &  52.6 &  59.8 &  74.4 &  59.4 &  57.6 &  31.1 &  37.0 &  53.1 & \textcolor{custom_red}{\textbf{--\,2.7}}\\
    \multirow{-2}{*}{1.07B} & \multirow{-2}{*}{\cmark} & \multirow{-2}{*}{60B} & \multirow{-2}{*}{\cmark} & \multirow{-2}{*}{2} & \multirow{-2}{*}{Avg} & \multirow{-2}{*}{64} & \multirow{-2}{*}{SVD} & \multirow{-2}{*}{15B}   & \multirow{-2}{*}{Agg\,(0.3)} & \multirow{-2}{*}{\cmark}  &  16.09 &  13.54 &  21.19 &  35.4 &  42.8 &  69.8 &  52.8 &  45.8 &  25.8 &  31.0 &  43.3 & -\\
    \addlinespace[-1pt] \cmidrule(l{2pt}r{2pt}){12-23} \addlinespace[-1pt]
      & & & & & & & & & & & 11.96 &  9.87 &  14.76 &  52.3 &  60.5 &  74.2 &  59.1 &  58.9 &  33.0 &  37.2 &  53.6 & \textcolor{custom_red}{\textbf{--\,3.0}} \\
    \multirow{-2}{*}{1.15B} & \multirow{-2}{*}{\cmark} & \multirow{-2}{*}{60B} & \multirow{-2}{*}{\cmark} & \multirow{-2}{*}{2} & \multirow{-2}{*}{Avg} & \multirow{-2}{*}{128} & \multirow{-2}{*}{SVD}  & \multirow{-2}{*}{15B}  & \multirow{-2}{*}{Agg\,(0.3)} & \multirow{-2}{*}{\cmark}  &  16.28 &  13.77 &  21.45 &  35.2 &  42.1 &  69.8 &  53.5 &  46.5 &  25.8 &  31.2 &  43.4 & -\\
    \addlinespace[-1pt] \cmidrule(l{2pt}r{2pt}){12-23} \addlinespace[-1pt]
      & & & & & & & & & & & 11.73 &  9.63 &  14.43 &  54.3 &  61.4 &  75.0 &  60.7 &  58.8 &  33.1 &  38.6 &  54.6 & \textcolor{custom_red}{\textbf{--\,2.4}} \\
    \multirow{-2}{*}{1.30B} & \multirow{-2}{*}{\cmark} & \multirow{-2}{*}{60B} & \multirow{-2}{*}{\cmark} & \multirow{-2}{*}{2} & \multirow{-2}{*}{Avg} & \multirow{-2}{*}{256} & \multirow{-2}{*}{SVD}  & \multirow{-2}{*}{15B}  & \multirow{-2}{*}{Agg\,(0.3)} & \multirow{-2}{*}{\cmark}  &  16.41 &  13.89 &  21.68 &  35.6 &  42.3 &  69.0 &  52.7 &  46.8 &  26.4 &  29.8 &  43.2 & - \\
    \addlinespace[-1pt] \cmidrule(l{2pt}r{2pt}){12-23} \addlinespace[-1pt]
      & & & & & & & & & & & 11.47 &  9.36 &  13.93 &  56.2 &  62.7 &  75.4 &  60.9 &  60.4 &  34.0 &  37.0 &  55.2  & \textcolor{custom_red}{\textbf{--\,3.2}}\\
    \multirow{-2}{*}{1.60B} & \multirow{-2}{*}{\cmark} & \multirow{-2}{*}{60B} & \multirow{-2}{*}{\cmark} & \multirow{-2}{*}{2} & \multirow{-2}{*}{Avg} & \multirow{-2}{*}{512} & \multirow{-2}{*}{SVD}  & \multirow{-2}{*}{15B}  & \multirow{-2}{*}{Agg\,(0.3)} & \multirow{-2}{*}{\cmark}  &  16.24 &  13.72 &  21.42 &  37.8 &  43.6 &  69.0 &  54.4 &  45.5 &  26.4 &  31.2 &  44.0 & - \\
    \midrule
    0.66B & \cmark & 60B & \cmark & 3 & Step  & - & - &  - & - & - & 12.27 &  9.90 &  15.24 &  55.6 &  58.1 &  73.1 &  60.2 &  58.8 &  30.2 &  37.2 &  53.3 & -\\
    0.74B & \cmark & 60B & \cmark & 3 & Avg  & 64 & SVD &  - & - & - & 12.13 &  9.80 &  14.95 &  55.5 &  58.3 &  73.5 &  60.1 &  58.0 &  31.1 &  36.8 &  53.3 & - \\
    0.82B & \cmark & 60B & \cmark & 3 & Avg  & 128 & SVD &  - & - & - & 11.83 &  9.53 &  14.51 &  56.7 &  60.2 &  74.2 &  59.8 &  59.1 &  33.0 &  35.4 &  54.1 & -\\
    0.97B & \cmark & 60B & \cmark & 3 & Avg  & 256 & SVD &  - & - & - & 11.43 &  9.17 &  13.87 &  59.3 &  62.6 &  74.7 &  61.2 &  61.6 &  32.9 &  {40.2} &  56.1 & - \\
    1.27B & \cmark & 60B & \cmark & 3 & Avg  & 512 & SVD &  - & - & - & 11.01 &  8.80 &  13.25 &  {61.5} &  64.9 &  76.2 &  62.0 &  64.3 &  35.6 &  39.2 &  57.7  & - \\
    \midrule
      & & & & & & & & & & & 12.75 &  10.48 &  16.01 &  50.2 &  57.0 &  72.7 &  58.6 &  56.7 &  30.0 &  38.2 &  51.9 & \textcolor{custom_red}{\textbf{--\,1.4}} \\
      & & & & & & & & & & & 13.81 &  11.47 &  17.80 &  48.4 &  53.0 &  72.4 &  55.6 &  51.6 &  27.2 &  35.2 &  49.0 & - \\
    \multirow{-3}{*}{0.66B} & \multirow{-3}{*}{\cmark} & \multirow{-3}{*}{60B} & \multirow{-3}{*}{\cmark} & \multirow{-3}{*}{3} & \multirow{-3}{*}{Step} & \multirow{-3}{*}{-} & \multirow{-3}{*}{-}  & \multirow{-3}{*}{15B}  & \multirow{-3}{*}{Agg\,(0.1)} & \multirow{-3}{*}{\cmark}  &  16.72 &  14.23 &  22.97 &  37.7 &  44.2 &  69.8 &  53.6 &  44.2 &  24.6 &  30.2 &  43.5 & -\\
    \addlinespace[-1pt] \cmidrule(l{2pt}r{2pt}){12-23} \addlinespace[-1pt]
      & & & & & & & & & & & 12.64 &  10.43 &  15.81 &  51.4 &  56.3 &  72.2 &  57.9 &  56.7 &  30.4 &  35.0 &  51.4 & \textcolor{custom_red}{\textbf{--\,1.9}} \\
      & & & & & & & & & & & 19.90 &  16.88 &  26.26 &  30.4 &  39.3 &  66.3 &  54.1 &  41.2 &  24.8 &  29.2 &  40.8 & - \\
    \multirow{-3}{*}{0.74B} & \multirow{-3}{*}{\cmark} & \multirow{-3}{*}{60B} & \multirow{-3}{*}{\cmark} & \multirow{-3}{*}{3} & \multirow{-3}{*}{Avg} & \multirow{-3}{*}{64} & \multirow{-3}{*}{SVD}  & \multirow{-3}{*}{15B} & \multirow{-3}{*}{Agg\,(0.1)} & \multirow{-3}{*}{\xmark}  &  26.31 &  22.49 &  36.10 &  20.9 &  31.2 &  62.6 &  50.8 &  37.2 &  22.0 &  28.0 &  36.1 & - \\
    \addlinespace[-1pt] \cmidrule(l{2pt}r{2pt}){12-23} \addlinespace[-1pt]
      & & & & & & & & & & &  12.37 &  10.21 &  15.38 &  52.0 &  58.0 &  72.0 &  56.5 &  58.4 &  30.0 &  35.2 &  51.7 & \textcolor{custom_red}{\textbf{--\,2.4}}\\
      & & & & & & & & & & & 20.07 &  17.09 &  26.47 &  30.9 &  40.5 &  66.3 &  55.4 &  40.8 &  24.4 &  29.6 &  41.1 & - \\
    \multirow{-3}{*}{0.82B} & \multirow{-3}{*}{\cmark} & \multirow{-3}{*}{60B} & \multirow{-3}{*}{\cmark} & \multirow{-3}{*}{3} & \multirow{-3}{*}{Avg} & \multirow{-3}{*}{128} & \multirow{-3}{*}{SVD}  & \multirow{-3}{*}{15B} & \multirow{-3}{*}{Agg\,(0.1)} & \multirow{-3}{*}{\xmark}  &  26.15 &  22.46 &  35.98 &  21.3 &  31.2 &  62.7 &  51.8 &  36.4 &  22.9 &  26.2 &  36.1 & - \\
    \addlinespace[-1pt] \cmidrule(l{2pt}r{2pt}){12-23} \addlinespace[-1pt]
      & & & & & & & & & & & 11.92 &  9.78 &  14.71 &  54.8 &  60.6 &  74.6 &  60.1 &  60.1 &  31.8 &  36.6 &  54.1 & \textcolor{custom_red}{\textbf{--\,2.0}}\\
      & & & & & & & & & & & 19.29 &  16.49 &  25.51 &  35.2 &  42.5 &  65.8 &  55.6 &  41.5 &  25.6 &  29.4 &  42.2 & - \\
    \multirow{-3}{*}{0.97B} & \multirow{-3}{*}{\cmark} & \multirow{-3}{*}{60B} & \multirow{-3}{*}{\cmark} & \multirow{-3}{*}{3} & \multirow{-3}{*}{Avg} & \multirow{-3}{*}{256} & \multirow{-3}{*}{SVD}  & \multirow{-3}{*}{15B} & \multirow{-3}{*}{Agg\,(0.1)} & \multirow{-3}{*}{\xmark}  &  25.12 &  21.53 &  34.53 &  23.1 &  32.0 &  63.2 &  49.7 &  36.1 &  23.0 &  25.2 &  36.1 & - \\
    \addlinespace[-1pt] \cmidrule(l{2pt}r{2pt}){12-23} \addlinespace[-1pt]
      & & & & & & & & & & & 11.49 &  9.38 &  14.00 &  56.1 &  62.7 &  74.4 &  60.5 &  62.1 &  34.9 &  38.8 &  55.7 & \textcolor{custom_red}{\textbf{--\,3.0}}\\
      & & & & & & & & & & & 18.52 &  15.79 &  24.34 &  36.7 &  44.9 &  67.2 &  55.3 &  43.8 &  26.0 &  30.4 &  43.5 & - \\
    \multirow{-3}{*}{1.27B} & \multirow{-3}{*}{\cmark} & \multirow{-3}{*}{60B} & \multirow{-3}{*}{\cmark} & \multirow{-3}{*}{3} & \multirow{-3}{*}{Avg} & \multirow{-3}{*}{512} & \multirow{-3}{*}{SVD}  & \multirow{-3}{*}{15B} & \multirow{-3}{*}{Agg\,(0.1)} & \multirow{-3}{*}{\xmark}  &  24.19 &  20.70 &  33.20 &  24.4 &  32.4 &  63.9 &  50.8 &  37.9 &  21.9 &  27.4 &  37.0 & - \\
    \midrule
      & & & & & & & & & & & 13.07 &  10.84 &  16.49 &  47.7 &  54.4 &  71.7 &  56.1 &  55.9 &  29.4 &  35.2 &  50.1 & \textcolor{custom_red}{\textbf{--\,3.2}}\\
      & & & & & & & & & & & 16.68 &  14.08 &  21.86 &  35.4 &  42.4 &  68.2 &  53.8 &  44.6 &  26.3 &  29.4 &  42.9 & - \\
    \multirow{-3}{*}{0.74B} & \multirow{-3}{*}{\cmark} & \multirow{-3}{*}{60B} & \multirow{-3}{*}{\cmark} & \multirow{-3}{*}{3} & \multirow{-3}{*}{Avg} & \multirow{-3}{*}{64} & \multirow{-3}{*}{SVD} & \multirow{-3}{*}{15B}  & \multirow{-3}{*}{Agg\,(0.3)} & \multirow{-3}{*}{\xmark}  &  21.43 &  18.26 &  29.12 &  24.4 &  34.1 &  64.3 &  50.5 &  40.7 &  22.3 &  27.8 &  37.7 & - \\
    \addlinespace[-1pt] \cmidrule(l{2pt}r{2pt}){12-23} \addlinespace[-1pt]
      & & & & & & & & & & & 12.71 &  10.54 &  15.92 &  50.4 &  55.9 &  73.1 &  57.5 &  56.8 &  30.1 &  34.8 &  51.2 & \textcolor{custom_red}{\textbf{--\,2.9}}\\
      & & & & & & & & & & & 16.90 &  14.32 &  22.18 &  37.6 &  43.5 &  67.6 &  54.5 &  45.0 &  25.3 &  29.0 &  43.2 & - \\
    \multirow{-3}{*}{0.82B} & \multirow{-3}{*}{\cmark} & \multirow{-3}{*}{60B} & \multirow{-3}{*}{\cmark} & \multirow{-3}{*}{3} & \multirow{-3}{*}{Avg} & \multirow{-3}{*}{128} & \multirow{-3}{*}{SVD}  & \multirow{-3}{*}{15B} & \multirow{-3}{*}{Agg\,(0.3)} & \multirow{-3}{*}{\xmark}  &  21.23 &  18.13 &  28.88 &  25.3 &  34.0 &  64.6 &  51.7 &  40.7 &  23.0 &  26.4 &  38.0 & - \\
    \addlinespace[-1pt] \cmidrule(l{2pt}r{2pt}){12-23} \addlinespace[-1pt]
      & & & & & & & & & & & 12.26 &  10.15 &  15.23 &  53.5 &  58.5 &  73.5 &  58.8 &  58.3 &  30.6 &  37.6 &  53.0 & \textcolor{custom_red}{\textbf{--\,3.1}}\\
      & & & & & & & & & & & 16.56 &  14.09 &  21.68 &  42.6 &  45.1 &  68.2 &  57.7 &  45.7 &  25.9 &  28.8 &  44.8 & - \\
    \multirow{-3}{*}{0.97B} & \multirow{-3}{*}{\cmark} & \multirow{-3}{*}{60B} & \multirow{-3}{*}{\cmark} & \multirow{-3}{*}{3} & \multirow{-3}{*}{Avg} & \multirow{-3}{*}{256} & \multirow{-3}{*}{SVD}  & \multirow{-3}{*}{15B} & \multirow{-3}{*}{Agg\,(0.3)} & \multirow{-3}{*}{\xmark}  &  20.78 &  17.72 &  28.29 &  27.9 &  34.3 &  66.3 &  52.2 &  39.7 &  23.6 &  26.8 &  38.7 & - \\
    \addlinespace[-1pt] \cmidrule(l{2pt}r{2pt}){12-23} \addlinespace[-1pt]
      & & & & & & & & & & & 11.80 &  9.68 &  14.45 &  54.1 &  61.2 &  74.0 &  59.0 &  59.9 &  32.9 &  38.0 &  54.1 & \textcolor{custom_red}{\textbf{--\,3.6}}\\
      & & & & & & & & & & & 16.02 &  13.53 &  20.86 &  43.5 &  47.5 &  68.3 &  56.2 &  47.1 &  27.0 &  30.4 &  45.7 & - \\
    \multirow{-3}{*}{1.27B} & \multirow{-3}{*}{\cmark} & \multirow{-3}{*}{60B} & \multirow{-3}{*}{\cmark} & \multirow{-3}{*}{3} & \multirow{-3}{*}{Avg} & \multirow{-3}{*}{512} & \multirow{-3}{*}{SVD}  & \multirow{-3}{*}{15B} & \multirow{-3}{*}{Agg\,(0.3)} & \multirow{-3}{*}{\xmark}  &  20.20 &  17.21 &  27.50 &  28.9 &  35.2 &  65.6 &  52.9 &  41.9 &  23.2 &  26.6 &  39.2 & - \\
    \bottomrule
    \end{tabular}
    }
    \caption{
    Evaluation results of Gemma models after early-exit training. Delta\,($\Delta$) represent the accuracy changes in original last loop outputs after early-exit post-training. These changes should be compared in reference to the performance drops observed in 75B and 60B uptraining for the full-size model. The relaxed model with three blocks shows a more significant performance drop because KD loss (from final loop output) could not be utilized due to out-of-memory issues.
    }
    \label{tab_rrt:final_early_exit_gemma_app}
\end{table}

\begin{table}[ht!]
    \small
    \centering
    \resizebox{\textwidth}{!}{
    \setlength{\tabcolsep}{3pt}
    \begin{tabular}{l|c|ccc|cc|cc|ccc|rrr|ccccccc|cc}
    \toprule
     & & \multicolumn{3}{c|}{\textbf{Uptrain}} & \multicolumn{2}{c|}{\textbf{Looping}} &  \multicolumn{2}{c|}{\textbf{LoRA}} &  \multicolumn{3}{c|}{\textbf{Early-exit\,Train}} & \multicolumn{3}{c|}{\textbf{Perplexity\,$\downarrow$}} & \multicolumn{9}{c}{\textbf{Few-shot Accuracy\,$\uparrow$}} \\
    \cmidrule(l{2pt}r{2pt}){3-5} \cmidrule(l{2pt}r{2pt}){6-7}  \cmidrule(l{2pt}r{2pt}){8-9} \cmidrule(l{2pt}r{2pt}){10-12} \cmidrule(l{2pt}r{2pt}){13-15}  \cmidrule(l{2pt}r{2pt}){16-24} 
    \textbf{Models} & N-emb & PT & $N_{tok}$ & KD & Block & Init  & Rank & Init & $N_{tok}$ & CE & KD & SlimP & RedP & PG19 & LD & HS & PQ & WG & ARC-e & ARC-c & OB & Avg & $\Delta$  \\
    \midrule
    & 0.97B & \cmark & - & \xmark & - & -  & - & - &  - & - & - &  12.26 & 9.37 & 11.94 & 43.3 & 42.2 & 66.8 & 53.4 & 44.7 & 23.2 & 29.2 & 43.3  & - \\
    \cmidrule(l{2pt}r{2pt}){2-24}
      & 0.48B & \cmark & 60B & \cmark & 2 & Step  & - & - &  - & - & - &  10.51 &  9.01 &  11.60 &  44.2 &  43.1 &  68.2 &  52.4 &  44.7 &  25.3 &  32.2 &  44.3 & - \\
    & 0.53B & \cmark & 60B & \cmark & 2 & Avg  & 64 & SVD &  - & - & - & 10.14 &  8.77 &  11.19 &  44.3 &  44.9 &  69.5 &  52.5 &  46.5 &  26.1 &  {31.6} &  45.0 & - \\
    & 0.58B & \cmark & 60B & \cmark & 2 & Avg  & 128 & SVD &  - & - & - & 10.07 &  8.68 &  11.07 &  45.9 &  45.1 &  69.4 &  50.5 &  46.8 &  25.4 &  {31.6} &  45.0 & - \\
    & 0.68B & \cmark & 60B & \cmark & 2 & Avg  & 256 & SVD &  - & - & - & 9.96 &  8.56 &  10.93 &  46.2 &  45.7 &  69.0 &  53.2 &  {47.9} &  25.9 &  {31.6} &  45.6 & - \\
    & 0.86B & \cmark & 60B & \cmark & 2 & Avg  & 512 & SVD &  - & - & - & {9.85} &  {8.44} &  {10.76} &  {47.4} &  {46.3} &  {69.7} &  52.8 &  47.5 &  {26.3} &  31.4 &  {45.9} & - \\
    \cmidrule(l{2pt}r{2pt}){2-24}
      & & & & & & & & & & & & 10.55 &  9.16 &  11.68 &  45.0 &  43.7 &  68.9 &  53.4 &  44.8 &  25.3 &  32.2 &  44.8 & \textcolor{custom_green}{\textbf{+\,0.5}} \\
     & \multirow{-2}{*}{0.48B} & \multirow{-2}{*}{\cmark} & \multirow{-2}{*}{60B} & \multirow{-2}{*}{\cmark} & \multirow{-2}{*}{2} & \multirow{-2}{*}{Step} & \multirow{-2}{*}{-} & \multirow{-2}{*}{-}  & \multirow{-2}{*}{15B} & \multirow{-2}{*}{Agg\,(0.1)} & \multirow{-2}{*}{\cmark}  &  12.28 &  10.62 &  13.83 &  38.2 &  39.4 &  65.8 &  52.3 &  41.5 &  24.7 &  30.6 &  41.8 & -\\
    \addlinespace[-1pt] \cmidrule(l{2pt}r{2pt}){13-24} \addlinespace[-1pt]
     & & & & & & & & & & & & 10.34 &  9.08 &  11.50 &  43.4 &  44.8 &  69.5 &  53.4 &  46.9 &  25.6 &  32.0 &  45.1 & \textcolor{custom_green}{\textbf{+\,0.1}}\\
     & \multirow{-2}{*}{0.53B} & \multirow{-2}{*}{\cmark} & \multirow{-2}{*}{60B} & \multirow{-2}{*}{\cmark} & \multirow{-2}{*}{2} & \multirow{-2}{*}{Avg} & \multirow{-2}{*}{64} & \multirow{-2}{*}{SVD}  & \multirow{-2}{*}{15B} & \multirow{-2}{*}{Agg\,(0.1)} & \multirow{-2}{*}{\cmark}  &  21.23 &  18.63 &  24.85 &  16.8 &  29.0 &  57.6 &  48.9 &  33.2 &  23.1 &  27.0 &  33.7 & - \\
    \addlinespace[-1pt] \cmidrule(l{2pt}r{2pt}){13-24} \addlinespace[-1pt]
     & & & & & & & & & & & & 10.25 &  8.97 &  11.36 &  45.2 &  45.5 &  68.8 &  54.0 &  46.5 &  25.0 &  31.6 &  45.2 & \textcolor{custom_green}{\textbf{+\,0.2}}\\
     & \multirow{-2}{*}{0.58B} & \multirow{-2}{*}{\cmark} & \multirow{-2}{*}{60B} & \multirow{-2}{*}{\cmark} & \multirow{-2}{*}{2} & \multirow{-2}{*}{Avg} & \multirow{-2}{*}{128} & \multirow{-2}{*}{SVD}  & \multirow{-2}{*}{15B} & \multirow{-2}{*}{Agg\,(0.1)} & \multirow{-2}{*}{\cmark}  &  21.30 &  18.56 &  24.75 &  18.5 &  28.9 &  58.4 &  48.0 &  34.1 &  21.8 &  27.4 &  33.9 & - \\
    \addlinespace[-1pt] \cmidrule(l{2pt}r{2pt}){13-24} \addlinespace[-1pt]
    TinyLlama & & & & & & & & & & & &  10.13 &  8.84 &  11.23 &  45.2 &  45.9 &  69.6 &  53.6 &  46.9 &  25.9 &  32.0 &  45.6 & \textcolor{gray}{\textbf{+\,0.0}}\\
     & \multirow{-2}{*}{0.68B} & \multirow{-2}{*}{\cmark} & \multirow{-2}{*}{60B} & \multirow{-2}{*}{\cmark} & \multirow{-2}{*}{2} & \multirow{-2}{*}{Avg} & \multirow{-2}{*}{256} & \multirow{-2}{*}{SVD}  & \multirow{-2}{*}{15B} & \multirow{-2}{*}{Agg\,(0.1)} & \multirow{-2}{*}{\cmark}  &  20.95 &  18.16 &  24.22 &  20.1 &  28.8 &  57.8 &  48.9 &  33.8 &  22.5 &  25.0 &  33.9 & - \\
    \addlinespace[-1pt] \cmidrule(l{2pt}r{2pt}){13-24} \addlinespace[-1pt]
     & & & & & & & & & & & & 10.02 &  8.74 &  11.04 &  46.6 &  46.5 &  68.6 &  54.5 &  47.9 &  26.3 &  32.2 &  46.1 & \textcolor{custom_green}{\textbf{+\,0.2}}\\
     & \multirow{-2}{*}{0.86B} & \multirow{-2}{*}{\cmark} & \multirow{-2}{*}{60B} & \multirow{-2}{*}{\cmark} & \multirow{-2}{*}{2} & \multirow{-2}{*}{Avg} & \multirow{-2}{*}{512} & \multirow{-2}{*}{SVD}  & \multirow{-2}{*}{15B} & \multirow{-2}{*}{Agg\,(0.1)} & \multirow{-2}{*}{\cmark}  &  20.38 &  17.70 &  23.57 &  19.9 &  28.8 &  58.2 &  49.0 &  34.7 &  22.8 &  25.8 &  34.2 & - \\
    \cmidrule(l{2pt}r{2pt}){2-24}
     & & & & & & & & & & & & 10.61 &  9.36 &  11.87 &  42.1 &  43.7 &  68.6 &  54.1 &  46.1 &  26.0 &  31.2 &  44.6 & \textcolor{custom_red}{\textbf{--\,0.4}}\\
     & \multirow{-2}{*}{0.53B} & \multirow{-2}{*}{\cmark} & \multirow{-2}{*}{60B} & \multirow{-2}{*}{\cmark} & \multirow{-2}{*}{2} & \multirow{-2}{*}{Avg} & \multirow{-2}{*}{64} & \multirow{-2}{*}{SVD}  & \multirow{-2}{*}{15B} & \multirow{-2}{*}{Agg\,(0.3)} & \multirow{-2}{*}{\cmark}  &  16.83 &  14.88 &  19.77 &  22.0 &  30.3 &  60.7 &  50.7 &  36.9 &  24.1 &  27.8 &  36.1 & - \\
    \addlinespace[-1pt] \cmidrule(l{2pt}r{2pt}){13-24} \addlinespace[-1pt]
     & & & & & & & & & & & & 10.50 &  9.22 &  11.71 &  44.2 &  44.2 &  69.2 &  53.0 &  46.0 &  25.5 &  31.2 &  44.8 & \textcolor{custom_red}{\textbf{--\,0.2}}\\
     & \multirow{-2}{*}{0.58B} & \multirow{-2}{*}{\cmark} & \multirow{-2}{*}{60B} & \multirow{-2}{*}{\cmark} & \multirow{-2}{*}{2} & \multirow{-2}{*}{Avg} & \multirow{-2}{*}{128} & \multirow{-2}{*}{SVD}  & \multirow{-2}{*}{15B} & \multirow{-2}{*}{Agg\,(0.3)} & \multirow{-2}{*}{\cmark}  &  17.10 &  15.03 &  19.99 &  23.5 &  30.1 &  60.8 &  51.3 &  36.5 &  23.8 &  26.4 &  36.0 & - \\
    \addlinespace[-1pt] \cmidrule(l{2pt}r{2pt}){13-24} \addlinespace[-1pt]
     & & & & & & & & & & & & 10.34 &  9.07 &  11.51 &  44.0 &  45.0 &  68.4 &  53.0 &  45.8 &  26.0 &  31.2 &  44.8 & \textcolor{custom_red}{\textbf{--\,0.8}}\\
     & \multirow{-2}{*}{0.68B} & \multirow{-2}{*}{\cmark} & \multirow{-2}{*}{60B} & \multirow{-2}{*}{\cmark} & \multirow{-2}{*}{2} & \multirow{-2}{*}{Avg} & \multirow{-2}{*}{256} & \multirow{-2}{*}{SVD}  & \multirow{-2}{*}{15B} & \multirow{-2}{*}{Agg\,(0.3)} & \multirow{-2}{*}{\cmark}  &  17.06 &  14.92 &  19.82 &  24.2 &  30.4 &  59.9 &  51.7 &  36.2 &  23.9 &  27.2 &  36.2 & -\\
    \addlinespace[-1pt] \cmidrule(l{2pt}r{2pt}){13-24} \addlinespace[-1pt]
     & & & & & & & & & & & & 10.21 &  8.94 &  11.28 &  45.1 &  45.8 &  69.3 &  54.5 &  46.7 &  25.9 &  33.4 &  45.8 & \textcolor{custom_green}{\textbf{--\,0.1}}\\
     & \multirow{-2}{*}{0.86B} & \multirow{-2}{*}{\cmark} & \multirow{-2}{*}{60B} & \multirow{-2}{*}{\cmark} & \multirow{-2}{*}{2} & \multirow{-2}{*}{Avg} & \multirow{-2}{*}{512} & \multirow{-2}{*}{SVD}  & \multirow{-2}{*}{15B} & \multirow{-2}{*}{Agg\,(0.3)} & \multirow{-2}{*}{\cmark}  &  16.76 &  14.68 &  19.43 &  24.4 &  30.0 &  61.1 &  51.9 &  37.1 &  22.9 &  28.2 &  36.5 & -\\
    \midrule
    & 0.81B & \cmark & 60B & \xmark & - & -  & - & - &  - & - & - & 12.83 & 9.76 & 13.57 &  53.0 & 50.2 & 71.1 & 54.8 & 51.9 & {27.7} & 31.6 & {48.6}  & - \\
    & 0.81B & \cmark & 75B & \xmark & - & -  & - & - &  - & - & - & 12.86 & 9.86 & 13.74 &  54.8 & 50.3 & 70.5 & 55.3 & 52.2 & {28.8} & 33.0 & {49.3}  & \textcolor{custom_green}{\textbf{+\,0.7}} \\
    \cmidrule(l{2pt}r{2pt}){2-24}
    & 0.40B & \cmark & 60B & \cmark & 2 & Step  & - & - &  - & - & - & 14.59 &  11.13 &  15.79 &  47.8 &  43.8 &  69.3 &  52.0 &  48.1 &  25.4 &  30.4 &  45.2 & -\\
    & 0.44B & \cmark & 60B & \cmark & 2 & Avg  & 64 & SVD &  - & - & - & 14.24 &  10.89 &  15.52 &  50.0 &  44.5 &  68.9 &  {54.1} &  48.0 &  26.5 &  31.2 &  46.2 & - \\
    & 0.48B & \cmark & 60B & \cmark & 2 & Avg  & 128 & SVD &  - & - & - & 14.10 &  10.79 &  15.27 &  50.1 &  45.5 &  69.0 &  52.6 &  48.3 &  25.8 &  32.0 &  46.2 & -\\
    & 0.55B & \cmark & 60B & \cmark & 2 & Avg  & 256 & SVD &  - & - & - & 13.91 &  10.61 &  14.91 &  50.5 &  45.6 &  68.7 &  51.2 &  48.4 &  25.7 &  {32.8} &  46.1 & - \\
    & 0.70B & \cmark & 60B & \cmark & 2 & Avg  & 512 & SVD &  - & - & - & {13.59} &  {10.38} &  {14.43} &  {52.0} &  {47.0} &  {69.6} &  53.4 &  {48.9} &  {26.9} &  31.2 &  {47.0}  & - \\
    \cmidrule(l{2pt}r{2pt}){2-24}
      & & & & & & & & & & & & 14.72 &  11.38 &  16.31 &  47.0 &  44.2 &  69.2 &  53.4 &  48.6 &  24.7 &  30.4 &  45.4 & \textcolor{custom_green}{\textbf{+\,0.2}} \\
     & \multirow{-2}{*}{0.40B} & \multirow{-2}{*}{\cmark} & \multirow{-2}{*}{60B} & \multirow{-2}{*}{\cmark} & \multirow{-2}{*}{2} & \multirow{-2}{*}{Step} & \multirow{-2}{*}{-} & \multirow{-2}{*}{-}  & \multirow{-2}{*}{15B} & \multirow{-2}{*}{Agg\,(0.1)} & \multirow{-2}{*}{\cmark}  &  18.61 &  14.11 &  20.96 &  38.4 &  38.1 &  67.0 &  53.7 &  43.3 &  24.4 &  29.0 &  42.0 & -\\
    \addlinespace[-1pt] \cmidrule(l{2pt}r{2pt}){13-24} \addlinespace[-1pt]
     & & & & & & & & & & & & 14.49 &  11.22 &  16.12 &  49.1 &  43.9 &  69.8 &  53.8 &  48.6 &  26.1 &  31.2 &  46.1 & \textcolor{custom_red}{\textbf{--\,0.1}}\\
     & \multirow{-2}{*}{0.44B} & \multirow{-2}{*}{\cmark} & \multirow{-2}{*}{60B} & \multirow{-2}{*}{\cmark} & \multirow{-2}{*}{2} & \multirow{-2}{*}{Avg} & \multirow{-2}{*}{64} & \multirow{-2}{*}{SVD}  & \multirow{-2}{*}{15B} & \multirow{-2}{*}{Agg\,(0.1)} & \multirow{-2}{*}{\cmark}  &  24.43 &  18.19 &  27.89 &  26.7 &  31.6 &  61.6 &  50.8 &  38.2 &  22.9 &  27.6 &  37.1 & - \\
    \addlinespace[-1pt] \cmidrule(l{2pt}r{2pt}){13-24} \addlinespace[-1pt]
     & & & & & & & & & & & & 14.35 &  11.17 &  15.93 &  50.1 &  44.7 &  69.0 &  52.1 &  49.9 &  25.3 &  32.6 &  46.2 & \textcolor{gray}{\textbf{+\,0.0}}\\
     & \multirow{-2}{*}{0.48B} & \multirow{-2}{*}{\cmark} & \multirow{-2}{*}{60B} & \multirow{-2}{*}{\cmark} & \multirow{-2}{*}{2} & \multirow{-2}{*}{Avg} & \multirow{-2}{*}{128} & \multirow{-2}{*}{SVD}  & \multirow{-2}{*}{15B} & \multirow{-2}{*}{Agg\,(0.1)} & \multirow{-2}{*}{\cmark}  &  24.33 &  18.09 &  27.96 &  28.2 &  32.3 &  61.1 &  53.0 &  38.8 &  23.7 &  27.4 &  37.8 & -\\
    \addlinespace[-1pt] \cmidrule(l{2pt}r{2pt}){13-24} \addlinespace[-1pt]
     Pythia & & & & & & & & & & & & 14.14 &  10.96 &  15.54 &  50.8 &  45.5 &  68.2 &  53.9 &  48.8 &  25.3 &  32.8 &  46.5 & \textcolor{custom_green}{\textbf{+\,0.4}}\\
     & \multirow{-2}{*}{0.55B} & \multirow{-2}{*}{\cmark} & \multirow{-2}{*}{60B} & \multirow{-2}{*}{\cmark} & \multirow{-2}{*}{2} & \multirow{-2}{*}{Avg} & \multirow{-2}{*}{256} & \multirow{-2}{*}{SVD}  & \multirow{-2}{*}{15B} & \multirow{-2}{*}{Agg\,(0.1)} & \multirow{-2}{*}{\cmark}  &  24.18 &  17.87 &  27.48 &  28.1 &  32.3 &  61.9 &  54.1 &  38.1 &  22.9 &  28.6 &  38.0 & -\\
    \addlinespace[-1pt] \cmidrule(l{2pt}r{2pt}){13-24} \addlinespace[-1pt]
     & & & & & & & & & & & & 13.81 &  10.72 &  15.11 &  52.4 &  47.0 &  69.3 &  52.7 &  50.1 &  26.9 &  32.0 &  47.2 & \textcolor{custom_green}{\textbf{+\,0.2}}\\
     & \multirow{-2}{*}{0.70B} & \multirow{-2}{*}{\cmark} & \multirow{-2}{*}{60B} & \multirow{-2}{*}{\cmark} & \multirow{-2}{*}{2} & \multirow{-2}{*}{Avg} & \multirow{-2}{*}{512} & \multirow{-2}{*}{SVD}  & \multirow{-2}{*}{15B} & \multirow{-2}{*}{Agg\,(0.1)} & \multirow{-2}{*}{\cmark}  &  23.50 &  17.49 &  26.72 &  29.5 &  32.8 &  63.2 &  52.3 &  38.8 &  22.8 &  27.8 &  38.2 & -\\
    \cmidrule(l{2pt}r{2pt}){2-24}
     & & & & & & & & & & & & 14.87 &  11.53 &  16.61 &  47.0 &  43.1 &  68.7 &  53.0 &  47.4 &  25.7 &  31.0 &  45.1 & \textcolor{custom_red}{\textbf{--\,0.9}}\\
     & \multirow{-2}{*}{0.44B} & \multirow{-2}{*}{\cmark} & \multirow{-2}{*}{60B} & \multirow{-2}{*}{\cmark} & \multirow{-2}{*}{2} & \multirow{-2}{*}{Avg} & \multirow{-2}{*}{64} & \multirow{-2}{*}{SVD}  & \multirow{-2}{*}{15B} & \multirow{-2}{*}{Agg\,(0.3)} & \multirow{-2}{*}{\cmark}  &  20.62 &  15.60 &  23.57 &  32.6 &  33.6 &  63.4 &  51.2 &  40.7 &  23.3 &  28.0 &  39.0 & - \\
    \addlinespace[-1pt] \cmidrule(l{2pt}r{2pt}){13-24} \addlinespace[-1pt]
     & & & & & & & & & & & & 14.69 &  11.46 &  16.36 &  48.9 &  43.8 &  68.4 &  53.0 &  49.1 &  26.2 &  31.6 &  45.9 & \textcolor{custom_red}{\textbf{--\,0.3}}\\
     & \multirow{-2}{*}{0.48B} & \multirow{-2}{*}{\cmark} & \multirow{-2}{*}{60B} & \multirow{-2}{*}{\cmark} & \multirow{-2}{*}{2} & \multirow{-2}{*}{Avg} & \multirow{-2}{*}{128} & \multirow{-2}{*}{SVD}  & \multirow{-2}{*}{15B} & \multirow{-2}{*}{Agg\,(0.3)} & \multirow{-2}{*}{\cmark}  &  20.60 &  15.56 &  23.63 &  33.2 &  33.6 &  62.7 &  51.1 &  41.3 &  23.6 &  27.8 &  39.0 & - \\
    \addlinespace[-1pt] \cmidrule(l{2pt}r{2pt}){13-24} \addlinespace[-1pt]
     & & & & & & & & & & & & 14.44 &  11.20 &  15.94 &  50.0 &  44.7 &  69.2 &  52.3 &  48.1 &  25.4 &  32.2 &  46.0 & \textcolor{custom_red}{\textbf{--\,0.1}}\\
     & \multirow{-2}{*}{0.55B} & \multirow{-2}{*}{\cmark} & \multirow{-2}{*}{60B} & \multirow{-2}{*}{\cmark} & \multirow{-2}{*}{2} & \multirow{-2}{*}{Avg} & \multirow{-2}{*}{256} & \multirow{-2}{*}{SVD}  & \multirow{-2}{*}{15B} & \multirow{-2}{*}{Agg\,(0.3)} & \multirow{-2}{*}{\cmark}  &  20.61 &  15.48 &  23.45 &  33.3 &  34.2 &  63.4 &  52.2 &  40.8 &  23.0 &  28.8 &  39.4 & - \\
    \addlinespace[-1pt] \cmidrule(l{2pt}r{2pt}){13-24} \addlinespace[-1pt]
     & & & & & & & & & & & & 14.08 &  10.94 &  15.44 &  51.1 &  46.4 &  68.7 &  52.2 &  50.0 &  26.9 &  31.6 &  46.7 & \textcolor{custom_red}{\textbf{--\,0.3}} \\
     & \multirow{-2}{*}{0.70B} & \multirow{-2}{*}{\cmark} & \multirow{-2}{*}{60B} & \multirow{-2}{*}{\cmark} & \multirow{-2}{*}{2} & \multirow{-2}{*}{Avg} & \multirow{-2}{*}{512} & \multirow{-2}{*}{SVD}  & \multirow{-2}{*}{15B} & \multirow{-2}{*}{Agg\,(0.3)} & \multirow{-2}{*}{\cmark}  &  20.20 &  15.25 &  22.98 &  34.6 &  34.1 &  63.5 &  53.0 &  41.5 &  23.6 &  27.6 &  39.7 & - \\
    \bottomrule
    \end{tabular}
    }
    \caption{
    Evaluation results of TinyLlama and Pythia models after early-exit training. Delta\,($\Delta$) represents the accuracy change in the original last loop outputs after early-exit post-training. 
    }
    \label{tab_rrt:final_early_exit_tinyllama_pythia_app}
\end{table}

\clearpage
\subsection{Expanded Results of Hypothetical Generation Speedup}
\label{app_rrt:hypothetical_generation_speedup}

\paragraph{Measuring the average per-token generation time.}

First, we measured the generation time with various model configurations using dummy weights and inputs. We measured the elapsed time for each components, such as embedding matrices, Transformer blocks, and the classifier head. Especially, we calculated the time per token by dividing the total time by the decoding length. Default prefix and decoding lengths are set to 512 and 2048, but we also used shorter context lengths, like 64 and 256 to simulate scenarios where parameter memory sizes become limiting. We measured decoding speed using FlashDecoding~\citep{DBLP:conf/nips/DaoFERR22}, a technique that has recently become standard in serving LLMs. Using a single A100 40GB GPU, we measured generation times by increasing batch sizes until an out-of-memory error occurred or memory usuage reached the predefined limit.

In Table\,\ref{tab_rrt:measured_generation_time_a100}, generation time was measured up to the maximum batch size that a single A100 GPU could accommodate before encountering out-of-memory errors, with prefix and decoding lengths set to 512 and 2048, respectively. Meanwhile, Table\,\ref{tab_rrt:measured_generation_time_a100_16gb} presents generation times measured in a more memory-constrained deployment scenario, where the prefix and decoding lengths were reduced to 64 and 256, and the memory limit was set to 16GB. As anticipated, under severe memory constraints, the reduced parameter memory footprint of Recursive Transformers enabled substantially larger batch sizes. This observation indicates that Recursive Transformers, even without continuous batching techniques, can achieve higher throughput than vanilla Transformers due to their memory efficiency.

When comparing the speed of the three models, Gemma 2B was the fastest, followed by TinyLlama 1.1B and then Pythia 1B. This order is the exact inverse of their non-embedding parameter sizes. This speed difference is attributed to the Grouped-Query~\citep{DBLP:journals/corr/abs-1911-02150} and Multi-Query attention~\citep{DBLP:conf/emnlp/AinslieLJZLS23}. The main decoding bottleneck in Transformers is memory access of heavy key-value caches. Hence, Gemma that effectively reduces the key-value cache size through the MQA mechanism, achieves fastest speeds. Despite using GQA, TinyLlama 1.1B has a similar speed to Gemma 2B due to its shallow and deep architecture (22 layers compared to Gemma's 18 layers). This deeper architecture likely offsets the speed gains from the attention mechanism.

\paragraph{Comparison of hypothetical generation throughput.}

We conducted early-exiting simulations using language modeling datasets (test sets of SlimPajama, RedPajama, and PG19), assuming our models generated those tokens. We used 20K samples to obtain their exit trajectories, and we employed an oracle-exiting algorithm to determine the earliest possible exit point for each token. 
Combining this trajectory data with previously measured per-token generation times across various batch sizes (considering only Transformer block computations), we estimated the hypothetical throughput for various settings and datasets.
The results are detailed in Tables\,\ref{tab_rrt:final_performance_throughput} and \ref{tab_rrt:final_performance_throughput_16gb}.

Our analysis reveals that Recursive Transformers achieve a 2-3$\times$ throughput gain over vanilla model counterparts. Relaxed models also demonstrated significant speedup despite their unoptimized LoRA computations. Currently, we merge multiple LoRAs into a single, large LoRA module to enable parallel computation of samples at different looping iterations. 
However, this introduces extra overhead due to redundant computations, resulting in reduced throughput gains in memory-constrained scenarios (shorter context lengths and lower memory limits). This degradation stems from the increased proportion of LoRA computation time relative to overall processing time. Since attention computation has quadratic complexity with respect to sequence length, it becomes less expensive at shorter lengths, while the complexity of LoRA computation remains constant. This highlights the need for highly optimized LoRA computations to achieve substantial throughput gains in all scenarios.  Nevertheless, these findings also suggest that relaxed models will yield even greater performance and throughput improvements with longer contexts where attention computation dominates.

\paragraph{Approximation errors in our hypothetical throughput}

Since our throughput estimations are based on theoretical estimation, they may introduce certain approximation errors as follows:
\begin{itemize}[leftmargin=*]
    \item 
    As our models are not fine-tuned for any downstream task, we simulated the exit trajectories of language modeling datasets by assuming they were generated by our models. While this approach is expected to closely approximate actual generation, empirical validation is necessary to confirm its accuracy.
    \item Throughput gains should be measured using realistic (confidence-based) early-exiting algorithms, rather than relying on the oracle-exiting algorithm. While early-exiting algorithms can introduce performance degradation due to inherent errors in confidence estimation, they also incur additional computational costs for estimating prediction confidence, necessitating further efficiency improvements.
    \item Our analysis solely focused on speed improvements within Transformer blocks. However, upon early exiting, the exited tokens require separate processing through the embedding layer or the classifier head for subsequent sequence generation. This necessitates non-exited tokens to wait for others, potentially reducing efficiency as the embedding layer computation may not fully utilize the maximum batch size.
    \item 
    Early-exiting architectures require computing key-value caches in remaining layers for already exited tokens to prevent performance degradation~\citep{DBLP:conf/emnlp/BaeKSY23}. While this adds negligible overhead in memory-bound scenarios, it inevitably increases overhead in compute-bound scenarios where the maximum batch size is fully utilized.  Our throughput estimation, however, excludes the computation time for these key-value caches in later loops (though we did account for their memory size).  Incorporating these computations into a more realistic analysis of early-exiting generation is a direction for future work.
\end{itemize}

\definecolor{Gray}{gray}{0.9}
\newcolumntype{a}{>{\columncolor{Gray}}c}

\begin{table}[ht!]
    \small
    \centering
    \resizebox{\textwidth}{!}{
    \setlength{\tabcolsep}{6pt}
     \renewcommand{\arraystretch}{0.85}
    \begin{tabular}{l|ccccc|c|cc|c|ccac}
    \toprule
      &  \multicolumn{5}{c|}{\textbf{Model Architecture}} &  & \multicolumn{2}{c|}{\textbf{Recursive}} & & \multicolumn{4}{c}{\textbf{Time\,(ms) per token}} \\
    \cmidrule(l{2pt}r{2pt}){2-6} \cmidrule(l{2pt}r{2pt}){8-9}
     \cmidrule(l{2pt}r{2pt}){11-14}
     \textbf{Models} & $N_L$ & $d_{model}$ & $N_{head}$ & $N_{KV}$ & Vocab & N-emb & Block & Rank & Batch & Total & Emb & \cellcolor{white} Transformer & Head  \\
    \midrule
     & & & & & & & & & 1 & 22.994 & 0.087 & 21.344 & 0.803 \\
     & \multirow{-2}{*}{18} & \multirow{-2}{*}{2048} & \multirow{-2}{*}{8} & \multirow{-2}{*}{1} & \multirow{-2}{*}{256K} & \multirow{-2}{*}{1.98B} &  \multirow{-2}{*}{-}  & \multirow{-2}{*}{-} & \textbf{43} & \,\,\,0.657 & 0.002 & \,\,\,0.616 & 0.023  \\
    \addlinespace[-1pt] \cmidrule(l{2pt}r{2pt}){2-14} \addlinespace[-1pt]
     & & & & & & & & & 1 & 13.918 & 0.088 & 11.059 & 0.827 \\
     & \multirow{-2}{*}{18} & \multirow{-2}{*}{2048} & \multirow{-2}{*}{8} & \multirow{-2}{*}{1} & \multirow{-2}{*}{256K} & \multirow{-2}{*}{0.99B}  & \multirow{-2}{*}{2}  & \multirow{-2}{*}{-} & \textbf{43} & \,\,\,0.336 & 0.002 & \,\,\,0.265 & 0.023  \\
    \addlinespace[-1pt] \cmidrule(l{2pt}r{2pt}){2-14} \addlinespace[-1pt]
     & & & & & & & & & 1 & 15.858 & 0.080 & 13.096 & 0.825 \\
     & \multirow{-2}{*}{18} & \multirow{-2}{*}{2048} & \multirow{-2}{*}{8} & \multirow{-2}{*}{1} & \multirow{-2}{*}{256K} & \multirow{-2}{*}{1.07B}  & \multirow{-2}{*}{2}  & \multirow{-2}{*}{64} & \textbf{41} & \,\,\,0.398 & 0.002 & \,\,\,0.323 & 0.024  \\
    \addlinespace[-1pt] \cmidrule(l{2pt}r{2pt}){2-14} \addlinespace[-1pt]
     & & & & & & & & & 1 & 15.708 & 0.080 & 12.969 & 0.822 \\
     & \multirow{-2}{*}{18} & \multirow{-2}{*}{2048} & \multirow{-2}{*}{8} & \multirow{-2}{*}{1} & \multirow{-2}{*}{256K} & \multirow{-2}{*}{1.15B}  & \multirow{-2}{*}{2} & \multirow{-2}{*}{128} & \textbf{41} & \,\,\,0.398 & 0.002 & \,\,\,0.324 & 0.024  \\
    \addlinespace[-1pt] \cmidrule(l{2pt}r{2pt}){2-14} \addlinespace[-1pt]
     & & & & & & & & & 1 & 15.456 & 0.083 & 12.721 & 0.818 \\
     & \multirow{-2}{*}{18} & \multirow{-2}{*}{2048} & \multirow{-2}{*}{8} & \multirow{-2}{*}{1} & \multirow{-2}{*}{256K} & \multirow{-2}{*}{1.30B}  & \multirow{-2}{*}{2}  & \multirow{-2}{*}{256} & \textbf{39} & \,\,\,0.450 & 0.002 & \,\,\,0.372 & 0.025  \\
    \addlinespace[-1pt] \cmidrule(l{2pt}r{2pt}){2-14} \addlinespace[-1pt]
    Gemma & & & & & & & & & 1 & 15.489 & 0.078 & 12.775 & 0.817 \\
     & \multirow{-2}{*}{18} & \multirow{-2}{*}{2048} & \multirow{-2}{*}{8} & \multirow{-2}{*}{1} & \multirow{-2}{*}{256K} & \multirow{-2}{*}{1.60B} & \multirow{-2}{*}{2} & \multirow{-2}{*}{512} & \textbf{39} & \,\,\,0.499 & 0.002 & \,\,\,0.422 & 0.025  \\
    \addlinespace[-1pt] \cmidrule(l{2pt}r{2pt}){2-14} \addlinespace[-1pt]
     & & & & & & & & & 1 & 10.546 & 0.081 & \,\,\,7.394 & 0.827 \\
     & \multirow{-2}{*}{18} & \multirow{-2}{*}{2048} & \multirow{-2}{*}{8} & \multirow{-2}{*}{1} & \multirow{-2}{*}{256K} & \multirow{-2}{*}{0.66B} &  \multirow{-2}{*}{3}  & \multirow{-2}{*}{-} & \textbf{43} & \,\,\,0.263 & 0.002 & \,\,\,0.182 & 0.023  \\
    \addlinespace[-1pt] \cmidrule(l{2pt}r{2pt}){2-14} \addlinespace[-1pt]
     & & & & & & & & & 1 & 11.871 & 0.080 & \,\,\,8.724 & 0.827 \\
     & \multirow{-2}{*}{18} & \multirow{-2}{*}{2048} & \multirow{-2}{*}{8} & \multirow{-2}{*}{1} & \multirow{-2}{*}{256K} & \multirow{-2}{*}{0.74B} &  \multirow{-2}{*}{3}  & \multirow{-2}{*}{64} & \textbf{43} & \,\,\,0.306 & 0.002 & \,\,\,0.182 & 0.023  \\
    \addlinespace[-1pt] \cmidrule(l{2pt}r{2pt}){2-14} \addlinespace[-1pt]
     & & & & & & & & & 1 & 11.768 & 0.080 & \,\,\,8.649 & 0.825 \\
     & \multirow{-2}{*}{18} & \multirow{-2}{*}{2048} & \multirow{-2}{*}{8} & \multirow{-2}{*}{1} & \multirow{-2}{*}{256K} & \multirow{-2}{*}{0.82B} &  \multirow{-2}{*}{3}  & \multirow{-2}{*}{128} & \textbf{43} & \,\,\,0.294 & 0.002 & \,\,\,0.221 & 0.023  \\
    \addlinespace[-1pt] \cmidrule(l{2pt}r{2pt}){2-14} \addlinespace[-1pt]
     & & & & & & & & & 1 & 12.018 & 0.081 & \,\,\,8.848 & 0.823 \\
     & \multirow{-2}{*}{18} & \multirow{-2}{*}{2048} & \multirow{-2}{*}{8} & \multirow{-2}{*}{1} & \multirow{-2}{*}{256K} & \multirow{-2}{*}{0.97B} &  \multirow{-2}{*}{3}  & \multirow{-2}{*}{256} & \textbf{41} & \,\,\,0.311 & 0.002 & \,\,\,0.226 & 0.024  \\
    \addlinespace[-1pt] \cmidrule(l{2pt}r{2pt}){2-14} \addlinespace[-1pt]
     & & & & & & & & & 1 & 12.087 & 0.082 & \,\,\,8.932 & 0.822 \\
     & \multirow{-2}{*}{18} & \multirow{-2}{*}{2048} & \multirow{-2}{*}{8} & \multirow{-2}{*}{1} & \multirow{-2}{*}{256K} & \multirow{-2}{*}{1.27B} &  \multirow{-2}{*}{3}  & \multirow{-2}{*}{512} & \textbf{39} & \,\,\,0.325 & 0.002 & \,\,\,0.237 & 0.025  \\
    \midrule
     & & & & & & & & & 1 & 22.016 & 0.082 & 21.010 & 0.188 \\
     & \multirow{-2}{*}{22} & \multirow{-2}{*}{2048} & \multirow{-2}{*}{32} & \multirow{-2}{*}{4} & \multirow{-2}{*}{32K} & \multirow{-2}{*}{0.97B} &  \multirow{-2}{*}{-}  & \multirow{-2}{*}{-} & \textbf{329} & \,\,\,0.819 & 0.000 & \,\,\,0.815 & 0.001  \\
    \addlinespace[-1pt] \cmidrule(l{2pt}r{2pt}){2-14} \addlinespace[-1pt]
     & & & & & & & & & 1 & 12.657 & 0.077 & 10.370 & 0.209 \\
     & \multirow{-2}{*}{22} & \multirow{-2}{*}{2048} & \multirow{-2}{*}{32} & \multirow{-2}{*}{4} & \multirow{-2}{*}{32K} & \multirow{-2}{*}{0.48B} &  \multirow{-2}{*}{2}  & \multirow{-2}{*}{-} & \textbf{233} & \,\,\,0.446 & 0.000 & \,\,\,0.413 & 0.001  \\
    \addlinespace[-1pt] \cmidrule(l{2pt}r{2pt}){2-14} \addlinespace[-1pt]
     & & & & & & & & & 1 & 15.243 & 0.079 & 12.908 & 0.211 \\
     TinyLlama & \multirow{-2}{*}{22} & \multirow{-2}{*}{2048} & \multirow{-2}{*}{32} & \multirow{-2}{*}{4} & \multirow{-2}{*}{32K} & \multirow{-2}{*}{0.53B} &  \multirow{-2}{*}{2}  & \multirow{-2}{*}{64} & \textbf{211} & \,\,\,0.454 & 0.000 & \,\,\,0.421 & 0.002  \\
    \addlinespace[-1pt] \cmidrule(l{2pt}r{2pt}){2-14} \addlinespace[-1pt]
     & & & & & & & & & 1 & 15.456 & 0.082 & 13.118 & 0.213 \\
     & \multirow{-2}{*}{22} & \multirow{-2}{*}{2048} & \multirow{-2}{*}{32} & \multirow{-2}{*}{4} & \multirow{-2}{*}{32K} & \multirow{-2}{*}{0.58B} &  \multirow{-2}{*}{2}  & \multirow{-2}{*}{128} & \textbf{209} & \,\,\,0.454 & 0.000 & \,\,\,0.421 & 0.002  \\
    \addlinespace[-1pt] \cmidrule(l{2pt}r{2pt}){2-14} \addlinespace[-1pt]
     & & & & & & & & & 1 & 15.223 & 0.081 & 12.908 & 0.208 \\
     & \multirow{-2}{*}{22} & \multirow{-2}{*}{2048} & \multirow{-2}{*}{32} & \multirow{-2}{*}{4} & \multirow{-2}{*}{32K} & \multirow{-2}{*}{0.68B} &  \multirow{-2}{*}{2}  & \multirow{-2}{*}{256} & \textbf{209} & \,\,\,0.457 & 0.000 & \,\,\,0.423 & 0.002  \\
    \addlinespace[-1pt] \cmidrule(l{2pt}r{2pt}){2-14} \addlinespace[-1pt]
     & & & & & & & & & 1 & 15.383 & 0.080 & 13.062 & 0.211 \\
     & \multirow{-2}{*}{22} & \multirow{-2}{*}{2048} & \multirow{-2}{*}{32} & \multirow{-2}{*}{4} & \multirow{-2}{*}{32K} & \multirow{-2}{*}{0.86B} &  \multirow{-2}{*}{2}  & \multirow{-2}{*}{512} & \textbf{209} & \,\,\,0.461 & 0.000 & \,\,\,0.428 & 0.002  \\
    \midrule
     & & & & & & & & & 1 & 13.280 & 0.080 & 12.286 & 0.235 \\
     & \multirow{-2}{*}{16} & \multirow{-2}{*}{2048} & \multirow{-2}{*}{8} & \multirow{-2}{*}{8} & \multirow{-2}{*}{50K} & \multirow{-2}{*}{0.81B} &  \multirow{-2}{*}{-}  & \multirow{-2}{*}{-} & \textbf{53} & \,\,\,1.227 & 0.002 & \,\,\,1.206 & 0.005  \\
    \addlinespace[-1pt] \cmidrule(l{2pt}r{2pt}){2-14} \addlinespace[-1pt]
     & & & & & & & & & 1 & \,\,\,8.423 & 0.081 & \,\,\,6.378 & 0.262 \\
     & \multirow{-2}{*}{16} & \multirow{-2}{*}{2048} & \multirow{-2}{*}{8} & \multirow{-2}{*}{8} & \multirow{-2}{*}{50K} & \multirow{-2}{*}{0.40B} &  \multirow{-2}{*}{2}  & \multirow{-2}{*}{-} & \textbf{61} & \,\,\,0.856 & 0.001 & \,\,\,0.606 & 0.005  \\
    \addlinespace[-1pt] \cmidrule(l{2pt}r{2pt}){2-14} \addlinespace[-1pt]
     & & & & & & & & & 1 & 10.554 & 0.082 & \,\,\,8.519 & 0.260 \\
    Pythia & \multirow{-2}{*}{16} & \multirow{-2}{*}{2048} & \multirow{-2}{*}{8} & \multirow{-2}{*}{8} & \multirow{-2}{*}{50K} & \multirow{-2}{*}{0.44B} &  \multirow{-2}{*}{2}  & \multirow{-2}{*}{64} & \textbf{63} & \,\,\,0.875 & 0.001 & \,\,\,0.626 & 0.005  \\
    \addlinespace[-1pt] \cmidrule(l{2pt}r{2pt}){2-14} \addlinespace[-1pt]
     & & & & & & & & & 1 & 10.167 & 0.076 & \,\,\,8.196 & 0.256 \\
     & \multirow{-2}{*}{16} & \multirow{-2}{*}{2048} & \multirow{-2}{*}{8} & \multirow{-2}{*}{8} & \multirow{-2}{*}{50K} & \multirow{-2}{*}{0.48B} &  \multirow{-2}{*}{2}  & \multirow{-2}{*}{128} & \textbf{59} & \,\,\,0.892 & 0.001 & \,\,\,0.642 & 0.005  \\
    \addlinespace[-1pt] \cmidrule(l{2pt}r{2pt}){2-14} \addlinespace[-1pt]
     & & & & & & & & & 1 & 10.410 & 0.079 & \,\,\,8.402 & 0.258 \\
     & \multirow{-2}{*}{16} & \multirow{-2}{*}{2048} & \multirow{-2}{*}{8} & \multirow{-2}{*}{8} & \multirow{-2}{*}{50K} & \multirow{-2}{*}{0.55B} &  \multirow{-2}{*}{2}  & \multirow{-2}{*}{256} & \textbf{59} & \,\,\,0.913 & 0.001 & \,\,\,0.662 & 0.005  \\
    \addlinespace[-1pt] \cmidrule(l{2pt}r{2pt}){2-14} \addlinespace[-1pt]
     & & & & & & & & & 1 & 12.609 & 0.091 & 10.311 & 0.267 \\
     & \multirow{-2}{*}{16} & \multirow{-2}{*}{2048} & \multirow{-2}{*}{8} & \multirow{-2}{*}{8} & \multirow{-2}{*}{50K} & \multirow{-2}{*}{0.70B} &  \multirow{-2}{*}{2}  & \multirow{-2}{*}{512} & \textbf{53} & \,\,\,0.956 & 0.002 & \,\,\,0.702 & 0.006  \\
    \bottomrule
    \end{tabular}
    }
    \caption{
    Measurements of generation time across three models using a single A100 40GB GPU. We measured time per token for both a batch size of 1 and the maximum batch size achievable by each model. The prefix length was set to 512 tokens, and the decoded output length to 2048 tokens. We then averaged the total elapsed time by the output length of 2048. Dummy input and dummy tensors were used for measurement. Both Gemma, employing multi-query attention, and TinyLlama, utilizing grouped-query attention, demonstrated fast generation speeds and large maximum batch sizes relative to their model sizes. TinyLlama's deep and narrow architecture allowed for a significantly large maximum batch size, although its generation speed was slower due to the increased number of layers.
    }
    \label{tab_rrt:measured_generation_time_a100}
\end{table}

\begin{table}[ht!]
    \small
    \centering
    \resizebox{\textwidth}{!}{
    \setlength{\tabcolsep}{3pt}
    \begin{tabular}{l|c|ccc|cc|cc|ccc|cc|ccc|c|rrrcc}
    \toprule
     & &  \multicolumn{3}{c|}{\textbf{Uptrain}} & \multicolumn{2}{c|}{\textbf{Looping}} &  \multicolumn{2}{c|}{\textbf{LoRA}} &  \multicolumn{3}{c|}{\textbf{Early-Exit\,Train}}  &  \multicolumn{2}{c|}{\textbf{Batching}} & \multicolumn{3}{c|}{\textbf{Few-shot Accuracy}} &  & \multicolumn{5}{c}{\textbf{Throughput\,$\uparrow$}} \\
     \cmidrule(l{2pt}r{2pt}){3-5} \cmidrule(l{2pt}r{2pt}){6-7}  \cmidrule(l{2pt}r{2pt}){8-9} \cmidrule(l{2pt}r{2pt}){10-12} \cmidrule(l{2pt}r{2pt}){13-14} \cmidrule(l{2pt}r{2pt}){15-17} \cmidrule(l{2pt}r{2pt}){19-23}
    \textbf{Models} & N-emb & PT & $N_{tok}$ & KD & Block & Init  & Rank & Init & $N_{tok}$ & CE & KD & Type & Exit & Last & Mid\,1 & Mid\,2 & Batch & SlimP & RedP & PG19 & $\Delta_{V}$ & $\Delta_{Seq}$ \\
    \midrule
      & 1.99B & \cmark & 75B & \xmark & - & -  & - & - & - & - & - & - & \xmark & 57.3 & - & - & 43 & 655 &  1228 &  1357 & \textcolor{gray}{{$\mathbf{\times 1.00}$}} & \textcolor{custom_red}{{$\mathbf{\times 0.71}$}} \\
      & 1.99B & \cmark & 75B & \xmark & - & -  & - & - &  - & - & - & CSB & \xmark & 57.3 & - & - & 43  &  1622 &  1604 &  1357 & \textcolor{custom_green}{{$\mathbf{\times 1.41}$}} & \textcolor{gray}{{$\mathbf{\times 1.00}$}} \\
      \cmidrule(l{2pt}r{2pt}){2-23}
      & 0.99B & \cmark & 60B & \cmark & 2 & Step  & - & - &  15B & Agg\,(0.1) & \cmark & CDB & \cmark & 54.0 & 48.8 & - & 43  &  3159 &  3050 &  2421 & \textcolor{custom_green}{{$\mathbf{\times {2.66}}$}} & \textcolor{custom_green}{{$\mathbf{\times {1.88}}$}} \\
      & 1.07B & \cmark & 60B & \cmark & 2 & Avg  & 64 & SVD &  15B & Agg\,(0.1) & \cmark & CDB & \cmark & 54.0 & 40.8 & - &  41  & 2357 &  2255 &  1858 & \textcolor{custom_green}{{$\mathbf{\times {2.00}}$}} & \textcolor{custom_green}{{$\mathbf{\times {1.41}}$}} \\
      & 1.15B & \cmark & 60B & \cmark & 2 & Avg  & 128 & SVD &  15B & Agg\,(0.1) & \cmark & CDB & \cmark & 54.6 & 40.2 & - & 41  &  2355 &  2250 &  1844 
 & \textcolor{custom_green}{{$\mathbf{\times {1.99}}$}} & \textcolor{custom_green}{{$\mathbf{\times {1.41}}$}} \\
      & 1.30B & \cmark & 60B & \cmark & 2 & Avg  & 256 & SVD &  15B & Agg\,(0.1) & \cmark & CDB & \cmark & 55.2 & 40.5 & - & 39  &  2047 &  1976 &  1740 
 & \textcolor{custom_green}{{$\mathbf{\times {1.78}}$}} & \textcolor{custom_green}{{$\mathbf{\times {1.26}}$}} \\
      & 1.60B & \cmark & 60B & \cmark & 2 & Avg  & 512 & SVD &  15B & Agg\,(0.1) & \cmark & CDB & \cmark & 56.2 & 41.7 & - & 39  &  1806 &  1754 &  1598 
 & \textcolor{custom_green}{{$\mathbf{\times {1.59}}$}} & \textcolor{custom_green}{{$\mathbf{\times {1.13}}$}} \\
      \cmidrule(l{2pt}r{2pt}){2-23}
      & 1.07B & \cmark & 60B & \cmark & 2 & Avg  & 64 & SVD &  15B & Agg\,(0.3) & \cmark & CDB & \cmark & 53.1 & 43.3 & - & 41  &  2454 &  2357 &  1929 
 & \textcolor{custom_green}{{$\mathbf{\times {2.08}}$}} & \textcolor{custom_green}{{$\mathbf{\times {1.47}}$}} \\
      & 1.15B & \cmark & 60B & \cmark & 2 & Avg  & 128 & SVD &  15B & Agg\,(0.3) & \cmark & CDB & \cmark & 53.6 & 43.4 & - & 41  &  2445 &  2346 &  1926 
 & \textcolor{custom_green}{{$\mathbf{\times {2.07}}$}} & \textcolor{custom_green}{{$\mathbf{\times {1.47}}$}} \\
      & 1.30B & \cmark & 60B & \cmark & 2 & Avg  & 256 & SVD &  15B & Agg\,(0.3) & \cmark & CDB & \cmark & 54.6 & 43.2 & - & 39  &  2123 &  2056 &  1804 
 & \textcolor{custom_green}{{$\mathbf{\times {1.85}}$}} & \textcolor{custom_green}{{$\mathbf{\times {1.31}}$}} \\
      Gemma & 1.60B & \cmark & 60B & \cmark & 2 & Avg  & 512 & SVD &  15B & Agg\,(0.3) & \cmark & CDB & \cmark & 55.2 & 44.0 & - & 39  &  1870 &  1819 &  1655 
 & \textcolor{custom_green}{{$\mathbf{\times {1.65}}$}} & \textcolor{custom_green}{{$\mathbf{\times {1.17}}$}} \\
      \cmidrule(l{2pt}r{2pt}){2-23}
      & 0.66B & \cmark & 60B & \cmark & 3 & Step  & - & - &  15B & Agg\,(0.1) & \cmark & CDB & \cmark & 51.9 & 49.0 & 43.5 & 43  &  3120 &  3041 &  2729 
 & \textcolor{custom_green}{{$\mathbf{\times {2.74}}$}} & \textcolor{custom_green}{{$\mathbf{\times {1.94}}$}} \\
      & 0.74B & \cmark & 60B & \cmark & 3 & Avg  & 64 & SVD &  15B & Agg\,(0.1) & \xmark & CDB & \cmark & 51.4 & 40.8 & 36.1 & 43  &  2334 &  2274 &  2059 
 & \textcolor{custom_green}{{$\mathbf{\times {2.06}}$}} & \textcolor{custom_green}{{$\mathbf{\times {1.45}}$}} \\
      & 0.82B & \cmark & 60B & \cmark & 3 & Avg  & 128 & SVD &  15B & Agg\,(0.1) & \xmark & CDB & \cmark & 51.7 & 41.1 & 36.1 & 43  &  2290 &  2230 &  2007 
 & \textcolor{custom_green}{{$\mathbf{\times {2.02}}$}} & \textcolor{custom_green}{{$\mathbf{\times {1.42}}$}} \\
      & 0.97B & \cmark & 60B & \cmark & 3 & Avg  & 256 & SVD &  15B & Agg\,(0.1) & \xmark & CDB & \cmark & 54.1 & 42.2 & 36.1 & 41  &  2281 &  2219 &  1984 
 & \textcolor{custom_green}{{$\mathbf{\times {2.00}}$}} & \textcolor{custom_green}{{$\mathbf{\times {1.41}}$}} \\
      & 1.27B & \cmark & 60B & \cmark & 3 & Avg  & 512 & SVD &  15B & Agg\,(0.1) & \xmark & CDB & \cmark & 55.7 & 43.5 & 37.0 &  39  & 2181 &  2122 &  1900 
 & \textcolor{custom_green}{{$\mathbf{\times {1.91}}$}} & \textcolor{custom_green}{{$\mathbf{\times {1.35}}$}} \\
      \cmidrule(l{2pt}r{2pt}){2-23}
      & 0.74B & \cmark & 60B & \cmark & 3 & Avg  & 64 & SVD &  15B & Agg\,(0.3) & \xmark & CDB & \cmark & 50.1 & 42.9 & 37.7 & 43  &  2427 &  2372 &  2143 
 & \textcolor{custom_green}{{$\mathbf{\times {2.14}}$}} & \textcolor{custom_green}{{$\mathbf{\times {1.51}}$}} \\
      & 0.82B & \cmark & 60B & \cmark & 3 & Avg  & 128 & SVD &  15B & Agg\,(0.3) & \xmark & CDB & \cmark & 51.2 & 43.2 & 38.0 & 43  &  2376 &  2321 &  2084 
 & \textcolor{custom_green}{{$\mathbf{\times {2.09}}$}} & \textcolor{custom_green}{{$\mathbf{\times {1.48}}$}} \\
      & 0.97B & \cmark & 60B & \cmark & 3 & Avg  & 256 & SVD &  15B & Agg\,(0.3) & \xmark & CDB & \cmark & 53.0 & 44.8 & 38.7 &  41  & 2359 &  2300 &  2039 
 & \textcolor{custom_green}{{$\mathbf{\times {2.07}}$}} & \textcolor{custom_green}{{$\mathbf{\times {1.46}}$}} \\
      & 1.27B & \cmark & 60B & \cmark & 3 & Avg  & 512 & SVD &  15B & Agg\,(0.3) & \xmark & CDB & \cmark & 54.1 & 45.7 & 39.2 & 39  &  2251 &  2191 &  1975 
 & \textcolor{custom_green}{{$\mathbf{\times {1.98}}$}} & \textcolor{custom_green}{{$\mathbf{\times {1.40}}$}} \\
      \midrule
      & 0.97B & \cmark & - & - & - & -  & - & - & - & - & - & - & \xmark & 43.3 & - & - & 329  &  1205 &  1220 &  1194 & \textcolor{gray}{{$\mathbf{\times 1.00}$}} & \textcolor{custom_red}{{$\mathbf{\times 0.99}$}}\\
      & 0.97B & \cmark & - & - & - & -  & - & - &  - & - & - & CSB & \xmark & 43.3 & - & - & 329  &  1227 &  1225 &  1194 & \textcolor{custom_green}{{$\mathbf{\times 1.01}$}} & \textcolor{gray}{{$\mathbf{\times 1.00}$}}\\
      \cmidrule(l{2pt}r{2pt}){2-23}
      & 0.48B & \cmark & 60B & \cmark & 2 & Step  & - & - &  15B & Agg\,(0.1) & \cmark & CDB & \cmark & 44.8 & 41.8 & - & 233  &  2038 &  2023 &  1933 
 & \textcolor{custom_green}{{$\mathbf{\times {1.66}}$}} & \textcolor{custom_green}{{$\mathbf{\times {1.64}}$}} \\
      & 0.53B & \cmark & 60B & \cmark & 2 & Avg  & 64 & SVD &  15B & Agg\,(0.1) & \cmark & CDB & \cmark & 45.1 & 33.7 & - & 211  &  1733 &  1719 &  1617  
 & \textcolor{custom_green}{{$\mathbf{\times {1.40}}$}} & \textcolor{custom_green}{{$\mathbf{\times {1.39}}$}} \\
      & 0.58B & \cmark & 60B & \cmark & 2 & Avg  & 128 & SVD &  15B & Agg\,(0.1) & \cmark & CDB & \cmark & 45.2 & 33.9 & - & 209  &  1733 &  1717 &  1609 
 & \textcolor{custom_green}{{$\mathbf{\times {1.40}}$}} & \textcolor{custom_green}{{$\mathbf{\times {1.39}}$}} \\
      TinyLlama & 0.68B & \cmark & 60B & \cmark & 2 & Avg  & 256 & SVD &  15B & Agg\,(0.1) & \cmark & CDB & \cmark & 45.6 & 33.9 & - & 209  &  1728 &  1714 &  1606  & \textcolor{custom_green}{{$\mathbf{\times {1.39}}$}} & \textcolor{custom_green}{{$\mathbf{\times {1.38}}$}} \\
      & 0.86B & \cmark & 60B & \cmark & 2 & Avg  & 512 & SVD &  15B & Agg\,(0.1) & \cmark & CDB & \cmark & 46.1 & 34.2 & - & 209  &  1716 &  1702 &  1581 
    & \textcolor{custom_green}{{$\mathbf{\times {1.38}}$}} & \textcolor{custom_green}{{$\mathbf{\times {1.37}}$}} \\    
      \cmidrule(l{2pt}r{2pt}){2-23}
      & 0.53B & \cmark & 60B & \cmark & 2 & Avg  & 64 & SVD &  15B & Agg\,(0.3) & \cmark & CDB & \cmark & 44.6 & 36.1 & - &  211  & 1810 &  1796 &  1688 
 & \textcolor{custom_green}{{$\mathbf{\times {1.46}}$}} & \textcolor{custom_green}{{$\mathbf{\times {1.45}}$}} \\
      & 0.58B & \cmark & 60B & \cmark & 2 & Avg  & 128 & SVD &  15B & Agg\,(0.3) & \cmark & CDB & \cmark & 44.8 & 36.0 & - & 209  &  1802 &  1787 &  1668  
 & \textcolor{custom_green}{{$\mathbf{\times {1.45}}$}} & \textcolor{custom_green}{{$\mathbf{\times {1.44}}$}} \\
      & 0.68B & \cmark & 60B & \cmark & 2 & Avg  & 256 & SVD &  15B & Agg\,(0.3) & \cmark & CDB & \cmark & 44.8 & 36.2 & - & 209  &  1793 &  1779 &  1668 
 & \textcolor{custom_green}{{$\mathbf{\times {1.45}}$}} & \textcolor{custom_green}{{$\mathbf{\times {1.44}}$}} \\
      & 0.86B & \cmark & 60B & \cmark & 2 & Avg  & 512 & SVD &  15B & Agg\,(0.3) & \cmark & CDB & \cmark & 45.8 & 36.5 & - & 209  &  1778 &  1763 &  1637  
 & \textcolor{custom_green}{{$\mathbf{\times {1.43}}$}} & \textcolor{custom_green}{{$\mathbf{\times {1.42}}$}} \\
      \midrule
      & 0.81B & \cmark & 75B & \xmark & - & -  & - & - & - & - & - & - & \xmark & 49.3 & - & - & 53  &  702 &  785 &  822 &\textcolor{gray}{{$\mathbf{\times 1.00}$}} & \textcolor{custom_red}{{$\mathbf{\times 0.93}$}} \\
      & 0.81B & \cmark & 75B & \xmark & - & -  & - & - &  - & - & - & CSB & \xmark & 49.3 & - & - & 53  &  829 &  827 &  822 & \textcolor{custom_green}{{$\mathbf{\times 1.07}$}} & \textcolor{gray}{{$\mathbf{\times 1.00}$}} \\
      \cmidrule(l{2pt}r{2pt}){2-23}
      & 0.40B & \cmark & 60B & \cmark & 2 & Step  & - & - &  15B & Agg\,(0.1) & \cmark & CDB & \cmark & 45.4 & 42.0 & - & 61  &  1339 &  1333 &  1281 
 & \textcolor{custom_green}{{$\mathbf{\times {1.71}}$}} & \textcolor{custom_green}{{$\mathbf{\times {1.60}}$}} \\
      & 0.44B & \cmark & 60B & \cmark & 2 & Avg  & 64 & SVD &  15B & Agg\,(0.1) & \cmark & CDB & \cmark & 46.1 & 37.1 & - & 63  &  1205 &  1203 &  1140 
 & \textcolor{custom_green}{{$\mathbf{\times {1.54}}$}} & \textcolor{custom_green}{{$\mathbf{\times {1.43}}$}} \\
      & 0.48B & \cmark & 60B & \cmark & 2 & Avg  & 128 & SVD &  15B & Agg\,(0.1) & \cmark & CDB & \cmark & 46.2 & 37.8 & - & 59  &  1156 &  1180 &  1108 
 & \textcolor{custom_green}{{$\mathbf{\times {1.49}}$}} & \textcolor{custom_green}{{$\mathbf{\times {1.39}}$}} \\
      Pythia & 0.55B & \cmark & 60B & \cmark & 2 & Avg  & 256 & SVD &  15B & Agg\,(0.1) & \cmark & CDB & \cmark & 46.5 & 38.0 & - & 59  &  1138 &  1139 &  1071 
 & \textcolor{custom_green}{{$\mathbf{\times {1.45}}$}} & \textcolor{custom_green}{{$\mathbf{\times {1.35}}$}} \\
      & 0.70B & \cmark & 60B & \cmark & 2 & Avg  & 512 & SVD &  15B & Agg\,(0.1) & \cmark & CDB & \cmark & 47.2 & 38.2 & - & 53  &  1051 &  1077 &  1021  
 & \textcolor{custom_green}{{$\mathbf{\times {1.36}}$}} & \textcolor{custom_green}{{$\mathbf{\times {1.27}}$}} \\
      \cmidrule(l{2pt}r{2pt}){2-23}
      & 0.44B & \cmark & 60B & \cmark & 2 & Avg  & 64 & SVD &  15B & Agg\,(0.3) & \cmark & CDB & \cmark & 45.1 & 39.0 & - & 63  &  1254 &  1252 &  1190  
 & \textcolor{custom_green}{{$\mathbf{\times {1.60}}$}} & \textcolor{custom_green}{{$\mathbf{\times {1.49}}$}} \\
      & 0.48B & \cmark & 60B & \cmark & 2 & Avg  & 128 & SVD &  15B & Agg\,(0.3) & \cmark & CDB & \cmark & 45.9 & 39.0 & - & 59  &  1200 &  1226 &  1153 
 & \textcolor{custom_green}{{$\mathbf{\times {1.55}}$}} & \textcolor{custom_green}{{$\mathbf{\times {1.45}}$}} \\
      & 0.55B & \cmark & 60B & \cmark & 2 & Avg  & 256 & SVD &  15B & Agg\,(0.3) & \cmark & CDB & \cmark & 46.0 & 39.4 & - & 59  &  1180 &  1180 &  1112 
 & \textcolor{custom_green}{{$\mathbf{\times {1.50}}$}} & \textcolor{custom_green}{{$\mathbf{\times {1.40}}$}} \\
      & 0.70B & \cmark & 60B & \cmark & 2 & Avg  & 512 & SVD &  15B & Agg\,(0.3) & \cmark & CDB & \cmark & 46.7 & 39.7 & - & 53  &  1088 &  1114 &  1058 
 & \textcolor{custom_green}{{$\mathbf{\times {1.41}}$}} & \textcolor{custom_green}{{$\mathbf{\times {1.32}}$}} \\
    \bottomrule
    \end{tabular}
    }
    \caption{
    Hypothetical generation speedup of Recursive Transformers across three models. We utilized the measurements of per-token generation time calculated in Table\,\ref{tab_rrt:measured_generation_time_a100}, which were calculated using an A100 40GB GPU, a prefix length of 512, and a decoding length of 2048. We only considered the time spent within Transformer blocks, simulating generation on the SlimPajama, RedPajama, and PG19 test sets. We used vanilla Transformer models, both with and without continuous sequence-wise batching (CSB), as our baselines. Our recursive models further enhance throughput by applying continuous depth-wise batching (CDB), leveraging looping and early-exiting techniques. The throughput improvements over the vanilla Transformer without and with sequence-wise batching are denoted as $\Delta_V$ and $\Delta_{Seq}$, respectively. To aid in understanding the speedup, we also provide the performance of intermediate layers and the maximum batch size.
    }
    \label{tab_rrt:final_performance_throughput}
\end{table}

\begin{table}[ht!]
    \small
    \centering
    \resizebox{\textwidth}{!}{
    \setlength{\tabcolsep}{6pt}
    \renewcommand{\arraystretch}{0.85}
    \begin{tabular}{l|ccccc|c|cc|c|ccac}
    \toprule
    &  \multicolumn{5}{c|}{\textbf{Model Architecture}} &  & \multicolumn{2}{c|}{\textbf{Recursive}} & & \multicolumn{4}{c}{\textbf{Time\,(ms) per token}} \\
    \cmidrule(l{2pt}r{2pt}){2-6} \cmidrule(l{2pt}r{2pt}){8-9} \cmidrule(l{2pt}r{2pt}){11-14}
    \textbf{Models} & $N_L$ & $d_{model}$ & $N_{head}$ & $N_{KV}$ & Vocab & N-emb & Block & Rank & Batch & Total & Emb & \cellcolor{white} Transformer & Head  \\
    \midrule
    & & & & & & & & & 1 & 22.577 & 0.084 & 20.937 & 0.801\\
    &  \multirow{-2}{*}{18} & \multirow{-2}{*}{2048} & \multirow{-2}{*}{8} & \multirow{-2}{*}{1} & \multirow{-2}{*}{256K} & \multirow{-2}{*}{1.98B} &  \multirow{-2}{*}{-}  & \multirow{-2}{*}{-} & \textbf{111} & 
    \,\,\,0.207 & 0.001 & \,\,\,0.188 & 0.010\\
    \addlinespace[-1pt] \cmidrule(l{2pt}r{2pt}){2-14} \addlinespace[-1pt]
    &  & & & & & & & & 1 & 13.576 & 0.079 & 10.819 & 0.815\\
    &  \multirow{-2}{*}{18} & \multirow{-2}{*}{2048} & \multirow{-2}{*}{8} & \multirow{-2}{*}{1} & \multirow{-2}{*}{256K} & \multirow{-2}{*}{0.99B} &  \multirow{-2}{*}{2}  & \multirow{-2}{*}{-} & \textbf{123} & 
    \,\,\,0.118 & 0.001 & \,\,\,0.091 & 0.009\\
    \addlinespace[-1pt] \cmidrule(l{2pt}r{2pt}){2-14} \addlinespace[-1pt]
    & & & & & & & & & 1 & 15.372 & 0.080 & 12.675 & 0.813\\
    & \multirow{-2}{*}{18} & \multirow{-2}{*}{2048} & \multirow{-2}{*}{8} & \multirow{-2}{*}{1} & \multirow{-2}{*}{256K} & \multirow{-2}{*}{1.07B} &  \multirow{-2}{*}{2}  & \multirow{-2}{*}{64} & \textbf{117} & 
    \,\,\,0.140 & 0.001 & \,\,\,0.112 & 0.009\\
    \addlinespace[-1pt] \cmidrule(l{2pt}r{2pt}){2-14} \addlinespace[-1pt]
    & & & & & & & & & 1 & 15.631 & 0.082 & 12.899 & 0.816\\
    & \multirow{-2}{*}{18} & \multirow{-2}{*}{2048} & \multirow{-2}{*}{8} & \multirow{-2}{*}{1} & \multirow{-2}{*}{256K} & \multirow{-2}{*}{1.15B} &  \multirow{-2}{*}{2}  & \multirow{-2}{*}{128} & \textbf{115} & 
    \,\,\,0.141 & 0.001 & \,\,\,0.113 & 0.010\\
    \addlinespace[-1pt] \cmidrule(l{2pt}r{2pt}){2-14} \addlinespace[-1pt]
    &  & & & & & & & & 1 & 15.317 & 0.079 & 12.639 & 0.811\\
    & \multirow{-2}{*}{18} & \multirow{-2}{*}{2048} & \multirow{-2}{*}{8} & \multirow{-2}{*}{1} & \multirow{-2}{*}{256K} & \multirow{-2}{*}{1.30B} &  \multirow{-2}{*}{2}  & \multirow{-2}{*}{256} & \textbf{111} & 
    \,\,\,0.143 & 0.001 & \,\,\,0.115 & 0.010\\
    \addlinespace[-1pt] \cmidrule(l{2pt}r{2pt}){2-14} \addlinespace[-1pt]
    Gemma &  & & & & & & & & 1 & 15.379 & 0.080 & 12.692 & 0.807\\
    & \multirow{-2}{*}{18} & \multirow{-2}{*}{2048} & \multirow{-2}{*}{8} & \multirow{-2}{*}{1} & \multirow{-2}{*}{256K} & \multirow{-2}{*}{1.60B} &  \multirow{-2}{*}{2}  & \multirow{-2}{*}{512} & \textbf{103} & 
    \,\,\,0.158 & 0.001 & \,\,\,0.127 & 0.011\\
    \addlinespace[-1pt] \cmidrule(l{2pt}r{2pt}){2-14} \addlinespace[-1pt]
    &  & & & & & & & & 1 & 10.528 & 0.080 & \,\,\,7.411 & 0.817\\
    &  \multirow{-2}{*}{18} & \multirow{-2}{*}{2048} & \multirow{-2}{*}{8} & \multirow{-2}{*}{1} & \multirow{-2}{*}{256K} & \multirow{-2}{*}{0.66B} &  \multirow{-2}{*}{3}  & \multirow{-2}{*}{-} & \textbf{131} & 
    \,\,\,0.087 & 0.001 & \,\,\,0.058 & 0.010\\
    \addlinespace[-1pt] \cmidrule(l{2pt}r{2pt}){2-14} \addlinespace[-1pt]
    &  & & & & & & & & 1 & 11.957 & 0.081 & \,\,\,8.855 & 0.815\\
    & \multirow{-2}{*}{18} & \multirow{-2}{*}{2048} & \multirow{-2}{*}{8} & \multirow{-2}{*}{1} & \multirow{-2}{*}{256K} & \multirow{-2}{*}{0.74B} &  \multirow{-2}{*}{3}  & \multirow{-2}{*}{64} & \textbf{123} & 
    \,\,\,0.105 & 0.001 & \,\,\,0.075 & 0.009\\
    \addlinespace[-1pt] \cmidrule(l{2pt}r{2pt}){2-14} \addlinespace[-1pt]
    & & & & & & & & & 1 & 11.898 & 0.080 & \,\,\,8.787 & 0.816\\
    & \multirow{-2}{*}{18} & \multirow{-2}{*}{2048} & \multirow{-2}{*}{8} & \multirow{-2}{*}{1} & \multirow{-2}{*}{256K} & \multirow{-2}{*}{0.82B} &  \multirow{-2}{*}{3}  & \multirow{-2}{*}{128} & \textbf{121} & 
    \,\,\,0.103 & 0.001 & \,\,\,0.074 & 0.009\\
    \addlinespace[-1pt] \cmidrule(l{2pt}r{2pt}){2-14} \addlinespace[-1pt]
    & & & & & & & & & 1 & 11.734 & 0.079 & \,\,\,8.654 & 0.813\\
    & \multirow{-2}{*}{18} & \multirow{-2}{*}{2048} & \multirow{-2}{*}{8} & \multirow{-2}{*}{1} & \multirow{-2}{*}{256K} & \multirow{-2}{*}{0.97B} &  \multirow{-2}{*}{3}  & \multirow{-2}{*}{256} & \textbf{117} & 
    \,\,\,0.106 & 0.001 & \,\,\,0.076 & 0.009\\
    \addlinespace[-1pt] \cmidrule(l{2pt}r{2pt}){2-14} \addlinespace[-1pt]
    & & & & & & & & & 1 & 11.986 & 0.080 & \,\,\,8.856 & 0.809\\
    & \multirow{-2}{*}{18} & \multirow{-2}{*}{2048} & \multirow{-2}{*}{8} & \multirow{-2}{*}{1} & \multirow{-2}{*}{256K} & \multirow{-2}{*}{1.27B} &  \multirow{-2}{*}{3}  & \multirow{-2}{*}{512} & \textbf{107} & 
    \,\,\,0.125 & 0.001 & \,\,\,0.090 & 0.010\\
    \midrule
    & & & & & & & & & 1 & 23.898 & 0.080 & 22.909 & 0.189\\
    & \multirow{-2}{*}{22} & \multirow{-2}{*}{2048} & \multirow{-2}{*}{32} & \multirow{-2}{*}{4} & \multirow{-2}{*}{32K} & \multirow{-2}{*}{0.97B} &  \multirow{-2}{*}{-}  & \multirow{-2}{*}{-} & \textbf{1049} & 
    \,\,\,0.131 & 0.000 & \,\,\,0.129 & 0.001\\
    \addlinespace[-1pt] \cmidrule(l{2pt}r{2pt}){2-14} \addlinespace[-1pt]
    & & & & & & & & & 1 & 14.129 & 0.080 & 11.846 & 0.202\\
    & \multirow{-2}{*}{22} & \multirow{-2}{*}{2048} & \multirow{-2}{*}{32} & \multirow{-2}{*}{4} & \multirow{-2}{*}{32K} & \multirow{-2}{*}{0.48B} &  \multirow{-2}{*}{2}  & \multirow{-2}{*}{-} & \textbf{1121} & 
    \,\,\,0.070 & 0.000 & \,\,\,0.064 & 0.001\\
    \addlinespace[-1pt] \cmidrule(l{2pt}r{2pt}){2-14} \addlinespace[-1pt]
    & & & & & & & & & 1 & 14.897 & 0.080 & 12.627 & 0.202\\
    TinyLlama & \multirow{-2}{*}{22} & \multirow{-2}{*}{2048} & \multirow{-2}{*}{32} & \multirow{-2}{*}{4} & \multirow{-2}{*}{32K} & \multirow{-2}{*}{0.53B} &  \multirow{-2}{*}{2}  & \multirow{-2}{*}{64} & \textbf{1105} & 
    \,\,\,0.073 & 0.000 & \,\,\,0.068 & 0.001\\
    \addlinespace[-1pt] \cmidrule(l{2pt}r{2pt}){2-14} \addlinespace[-1pt]
    & & & & & & & & & 1 & 15.090 & 0.081 & 12.778 & 0.205\\
    & \multirow{-2}{*}{22} & \multirow{-2}{*}{2048} & \multirow{-2}{*}{32} & \multirow{-2}{*}{4} & \multirow{-2}{*}{32K} & \multirow{-2}{*}{0.58B} &  \multirow{-2}{*}{2}  & \multirow{-2}{*}{128} & \textbf{1089} & 
    \,\,\,0.074 & 0.000 & \,\,\,0.069 & 0.001\\
    \addlinespace[-1pt] \cmidrule(l{2pt}r{2pt}){2-14} \addlinespace[-1pt]
    & & & & & & & & & 1 & 14.962 & 0.081 & 12.659 & 0.201\\
    & \multirow{-2}{*}{22} & \multirow{-2}{*}{2048} & \multirow{-2}{*}{32} & \multirow{-2}{*}{4} & \multirow{-2}{*}{32K} & \multirow{-2}{*}{0.68B} &  \multirow{-2}{*}{2}  & \multirow{-2}{*}{256} & \textbf{1065} & 
    \,\,\,0.076 & 0.000 & \,\,\,0.071 & 0.001\\
    \addlinespace[-1pt] \cmidrule(l{2pt}r{2pt}){2-14} \addlinespace[-1pt]
    & & & & & & & & & 1 & 15.284 & 0.083 & 12.950 & 0.206\\
    & \multirow{-2}{*}{22} & \multirow{-2}{*}{2048} & \multirow{-2}{*}{32} & \multirow{-2}{*}{4} & \multirow{-2}{*}{32K} & \multirow{-2}{*}{0.86B} &  \multirow{-2}{*}{2}  & \multirow{-2}{*}{512} & \textbf{1017} & 
    \,\,\,0.080 & 0.000 & \,\,\,0.075 & 0.001\\
    \midrule
    & & & & & & & & & 1 & 13.341 & 0.081 & 12.326 & 0.239\\
    & \multirow{-2}{*}{16} & \multirow{-2}{*}{2048} & \multirow{-2}{*}{8} & \multirow{-2}{*}{8} & \multirow{-2}{*}{50K} & \multirow{-2}{*}{0.81B} &  \multirow{-2}{*}{-}  & \multirow{-2}{*}{-} & \textbf{229} & 
    \,\,\,0.176 & 0.000 & \,\,\,0.171 & 0.002\\
    \addlinespace[-1pt] \cmidrule(l{2pt}r{2pt}){2-14} \addlinespace[-1pt]
    & & & & & & & & & 1 & \,\,\,8.336 & 0.079 & \,\,\,6.303 & 0.261\\
    & \multirow{-2}{*}{16} & \multirow{-2}{*}{2048} & \multirow{-2}{*}{8} & \multirow{-2}{*}{8} & \multirow{-2}{*}{50K} & \multirow{-2}{*}{0.40B} &  \multirow{-2}{*}{2}  & \multirow{-2}{*}{-} & \textbf{241} & 
    \,\,\,0.121 & 0.000 & \,\,\,0.086 & 0.002\\
    \addlinespace[-1pt] \cmidrule(l{2pt}r{2pt}){2-14} \addlinespace[-1pt]
    & & & & & & & & & 1 & 10.408 & 0.081 & \,\,\,8.353 & 0.262\\
    Pythia & \multirow{-2}{*}{16} & \multirow{-2}{*}{2048} & \multirow{-2}{*}{8} & \multirow{-2}{*}{8} & \multirow{-2}{*}{50K} & \multirow{-2}{*}{0.44B} &  \multirow{-2}{*}{2}  & \multirow{-2}{*}{64} & \textbf{233} & 
    \,\,\,0.133 & 0.000 & \,\,\,0.097 & 0.002\\
    \addlinespace[-1pt] \cmidrule(l{2pt}r{2pt}){2-14} \addlinespace[-1pt]
    & & & & & & & & & 1 & 10.426 & 0.082 & \,\,\,8.378 & 0.259\\
    & \multirow{-2}{*}{16} & \multirow{-2}{*}{2048} & \multirow{-2}{*}{8} & \multirow{-2}{*}{8} & \multirow{-2}{*}{50K} & \multirow{-2}{*}{0.48B} &  \multirow{-2}{*}{2}  & \multirow{-2}{*}{128} & \textbf{221} & 
    \,\,\,0.137 & 0.000 & \,\,\,0.101 & 0.002\\
    \addlinespace[-1pt] \cmidrule(l{2pt}r{2pt}){2-14} \addlinespace[-1pt]
    & & & & & & & & & 1 & 10.509 & 0.080 & \,\,\,8.471 & 0.256\\
    & \multirow{-2}{*}{16} & \multirow{-2}{*}{2048} & \multirow{-2}{*}{8} & \multirow{-2}{*}{8} & \multirow{-2}{*}{50K} & \multirow{-2}{*}{0.55B} &  \multirow{-2}{*}{2}  & \multirow{-2}{*}{256} & \textbf{205} & 
    \,\,\,0.151 & 0.000 & \,\,\,0.115 & 0.002\\
    \addlinespace[-1pt] \cmidrule(l{2pt}r{2pt}){2-14} \addlinespace[-1pt]
    & & & & & & & & & 1 & 11.254 & 0.080 & \,\,\,9.241 & 0.257\\
    & \multirow{-2}{*}{16} & \multirow{-2}{*}{2048} & \multirow{-2}{*}{8} & \multirow{-2}{*}{8} & \multirow{-2}{*}{50K} & \multirow{-2}{*}{0.70B} &  \multirow{-2}{*}{2}  & \multirow{-2}{*}{512} & \textbf{165} & 
    \,\,\,0.177 & 0.001 & \,\,\,0.139 & 0.002\\
    \bottomrule
    \end{tabular}
    }
    \caption{
    Generation time measurements of Gemma models on a single A100 40GB GPU with 16GB memory constraint. We measured generation time per token for both a batch size of 1 and the maximum batch size achievable by each model. The prefix length was set to 64 tokens, and the decoded output length to 256 tokens. We then averaged the total elapsed time by the output length of 256. Dummy input and dummy tensors were used for measurement. 
    }
    \label{tab_rrt:measured_generation_time_a100_16gb}
\end{table}

\begin{table}[ht!]
    \small
    \centering
    \resizebox{\textwidth}{!}{
    \setlength{\tabcolsep}{3pt}
    \begin{tabular}{l|c|ccc|cc|cc|ccc|cc|ccc|c|rrrcc}
    \toprule
     & &  \multicolumn{3}{c|}{\textbf{Uptrain}} & \multicolumn{2}{c|}{\textbf{Looping}} &  \multicolumn{2}{c|}{\textbf{LoRA}} &  \multicolumn{3}{c|}{\textbf{Early-Exit\,Train}}  &  \multicolumn{2}{c|}{\textbf{Batching}} & \multicolumn{3}{c|}{\textbf{Few-shot Accuracy}} &  & \multicolumn{5}{c}{\textbf{Throughput\,$\uparrow$}} \\
     \cmidrule(l{2pt}r{2pt}){3-5} \cmidrule(l{2pt}r{2pt}){6-7}  \cmidrule(l{2pt}r{2pt}){8-9} \cmidrule(l{2pt}r{2pt}){10-12} \cmidrule(l{2pt}r{2pt}){13-14} \cmidrule(l{2pt}r{2pt}){15-17} \cmidrule(l{2pt}r{2pt}){19-23}
    \textbf{Models} & N-emb & PT & $N_{tok}$ & KD & Block & Init  & Rank & Init & $N_{tok}$ & CE & KD & Type & Exit & Last & Mid\,1 & Mid\,2 & Batch & SlimP & RedP & PG19 & $\Delta_{V}$ & $\Delta_{Seq}$ \\
    \midrule
      & 1.99B & \cmark & 75B & \xmark & - & -  & - & - & - & - & - & - & \xmark & 57.3 & - & - & 111 & 1740 &  3059 &  4796 & \textcolor{gray}{{$\mathbf{\times 1.00}$}} & \textcolor{custom_red}{{$\mathbf{\times 0.63}$}} \\
      & 1.99B & \cmark & 75B & \xmark & - & -  & - & - &  - & - & - & CSB & \xmark & 57.3 & - & - & 111  &  5287 &  5060 &  4796 & \textcolor{custom_green}{{$\mathbf{\times 1.58}$}} & \textcolor{gray}{{$\mathbf{\times 1.00}$}} \\
      \cmidrule(l{2pt}r{2pt}){2-23}
      & 0.99B & \cmark & 60B & \cmark & 2 & Step  & - & - &  15B & Agg\,(0.1) & \cmark & CDB & \cmark & 54.0 & 48.8 & - & 43  &  3159 &  3050 &  2421 & \textcolor{custom_green}{{$\mathbf{\times {2.50}}$}} & \textcolor{custom_green}{{$\mathbf{\times {1.59}}$}} \\
      & 1.07B & \cmark & 60B & \cmark & 2 & Avg  & 64 & SVD &  15B & Agg\,(0.1) & \cmark & CDB & \cmark & 54.0 & 40.8 & - &  41  & 2357 &  2255 &  1858 
 & \textcolor{custom_green}{{$\mathbf{\times {1.87}}$}} & \textcolor{custom_green}{{$\mathbf{\times {1.19}}$}} \\
      & 1.15B & \cmark & 60B & \cmark & 2 & Avg  & 128 & SVD &  15B & Agg\,(0.1) & \cmark & CDB & \cmark & 54.6 & 40.2 & - & 41  &  2355 &  2250 &  1844 
 & \textcolor{custom_green}{{$\mathbf{\times {1.87}}$}} & \textcolor{custom_green}{{$\mathbf{\times {1.19}}$}} \\
      & 1.30B & \cmark & 60B & \cmark & 2 & Avg  & 256 & SVD &  15B & Agg\,(0.1) & \cmark & CDB & \cmark & 55.2 & 40.5 & - & 39  &  2047 &  1976 &  1740 
 & \textcolor{custom_green}{{$\mathbf{\times {1.86}}$}} & \textcolor{custom_green}{{$\mathbf{\times {1.18}}$}} \\
      & 1.60B & \cmark & 60B & \cmark & 2 & Avg  & 512 & SVD &  15B & Agg\,(0.1) & \cmark & CDB & \cmark & 56.2 & 41.7 & - & 39  &  1806 &  1754 &  1598 
 & \textcolor{custom_green}{{$\mathbf{\times {1.73}}$}} & \textcolor{custom_green}{{$\mathbf{\times {1.10}}$}} \\
      \cmidrule(l{2pt}r{2pt}){2-23}
      & 1.07B & \cmark & 60B & \cmark & 2 & Avg  & 64 & SVD &  15B & Agg\,(0.3) & \cmark & CDB & \cmark & 53.1 & 43.3 & - & 41  &  2454 &  2357 &  1929 
 & \textcolor{custom_green}{{$\mathbf{\times {1.95}}$}} & \textcolor{custom_green}{{$\mathbf{\times {1.24}}$}} \\
      & 1.15B & \cmark & 60B & \cmark & 2 & Avg  & 128 & SVD &  15B & Agg\,(0.3) & \cmark & CDB & \cmark & 53.6 & 43.4 & - & 41  &  2445 &  2346 &  1926 
 & \textcolor{custom_green}{{$\mathbf{\times {1.95}}$}} & \textcolor{custom_green}{{$\mathbf{\times {1.24}}$}} \\
      & 1.30B & \cmark & 60B & \cmark & 2 & Avg  & 256 & SVD &  15B & Agg\,(0.3) & \cmark & CDB & \cmark & 54.6 & 43.2 & - & 39  &  2123 &  2056 &  1804 
 & \textcolor{custom_green}{{$\mathbf{\times {1.93}}$}} & \textcolor{custom_green}{{$\mathbf{\times {1.22}}$}} \\
      Gemma & 1.60B & \cmark & 60B & \cmark & 2 & Avg  & 512 & SVD &  15B & Agg\,(0.3) & \cmark & CDB & \cmark & 55.2 & 44.0 & - & 39  &  1870 &  1819 &  1655 
 & \textcolor{custom_green}{{$\mathbf{\times {1.79}}$}} & \textcolor{custom_green}{{$\mathbf{\times {1.14}}$}} \\
      \cmidrule(l{2pt}r{2pt}){2-23}
      & 0.66B & \cmark & 60B & \cmark & 3 & Step  & - & - &  15B & Agg\,(0.1) & \cmark & CDB & \cmark & 51.9 & 49.0 & 43.5 & 43  &  3120 &  3041 &  2729 
 & \textcolor{custom_green}{{$\mathbf{\times {2.62}}$}} & \textcolor{custom_green}{{$\mathbf{\times {1.66}}$}} \\
      & 0.74B & \cmark & 60B & \cmark & 3 & Avg  & 64 & SVD &  15B & Agg\,(0.1) & \xmark & CDB & \cmark & 51.4 & 40.8 & 36.1 & 43  &  2334 &  2274 &  2059 
 & \textcolor{custom_green}{{$\mathbf{\times {1.87}}$}} & \textcolor{custom_green}{{$\mathbf{\times {1.19}}$}} \\
      & 0.82B & \cmark & 60B & \cmark & 3 & Avg  & 128 & SVD &  15B & Agg\,(0.1) & \xmark & CDB & \cmark & 51.7 & 41.1 & 36.1 & 43  &  2290 &  2230 &  2007 
 & \textcolor{custom_green}{{$\mathbf{\times {1.90}}$}} & \textcolor{custom_green}{{$\mathbf{\times {1.20}}$}} \\
      & 0.97B & \cmark & 60B & \cmark & 3 & Avg  & 256 & SVD &  15B & Agg\,(0.1) & \xmark & CDB & \cmark & 54.1 & 42.2 & 36.1 & 41  &  2281 &  2219 &  1984 
 & \textcolor{custom_green}{{$\mathbf{\times {1.86}}$}} & \textcolor{custom_green}{{$\mathbf{\times {1.18}}$}} \\
      & 1.27B & \cmark & 60B & \cmark & 3 & Avg  & 512 & SVD &  15B & Agg\,(0.1) & \xmark & CDB & \cmark & 55.7 & 43.5 & 37.0 &  39  & 2181 &  2122 &  1900 
 & \textcolor{custom_green}{{$\mathbf{\times {1.62}}$}} & \textcolor{custom_green}{{$\mathbf{\times {1.03}}$}} \\
      \cmidrule(l{2pt}r{2pt}){2-23}
      & 0.74B & \cmark & 60B & \cmark & 3 & Avg  & 64 & SVD &  15B & Agg\,(0.3) & \xmark & CDB & \cmark & 50.1 & 42.9 & 37.7 & 43  &  2427 &  2372 &  2143 
 & \textcolor{custom_green}{{$\mathbf{\times {1.94}}$}} & \textcolor{custom_green}{{$\mathbf{\times {1.23}}$}} \\
      & 0.82B & \cmark & 60B & \cmark & 3 & Avg  & 128 & SVD &  15B & Agg\,(0.3) & \xmark & CDB & \cmark & 51.2 & 43.2 & 38.0 & 43  &  2376 &  2321 &  2084
 & \textcolor{custom_green}{{$\mathbf{\times {1.97}}$}} & \textcolor{custom_green}{{$\mathbf{\times {1.25}}$}} \\
      & 0.97B & \cmark & 60B & \cmark & 3 & Avg  & 256 & SVD &  15B & Agg\,(0.3) & \xmark & CDB & \cmark & 53.0 & 44.8 & 38.7 &  41  & 2359 &  2300 &  2039 
 & \textcolor{custom_green}{{$\mathbf{\times {1.92}}$}} & \textcolor{custom_green}{{$\mathbf{\times {1.22}}$}} \\
      & 1.27B & \cmark & 60B & \cmark & 3 & Avg  & 512 & SVD &  15B & Agg\,(0.3) & \xmark & CDB & \cmark & 54.1 & 45.7 & 39.2 & 39  &  2251 &  2191 &  1975
 & \textcolor{custom_green}{{$\mathbf{\times {1.67}}$}} & \textcolor{custom_green}{{$\mathbf{\times {1.06}}$}} \\
      \midrule
      & 0.97B & \cmark & - & - & - & -  & - & - & - & - & - & - & \xmark & 43.3 & - & - & 1049  &  6856 &  7481 &  4090 & \textcolor{gray}{{$\mathbf{\times 1.00}$}} & \textcolor{custom_red}{{$\mathbf{\times 0.96}$}}\\
      & 0.97B & \cmark & - & - & - & -  & - & - &  - & - & - & CSB & \xmark & 43.3 & - & - & 1049  &  7709 &  7481 &  4090 & \textcolor{custom_green}{{$\mathbf{\times 1.05}$}} & \textcolor{gray}{{$\mathbf{\times 1.00}$}}\\
      \cmidrule(l{2pt}r{2pt}){2-23}
      & 0.48B & \cmark & 60B & \cmark & 2 & Step  & - & - &  15B & Agg\,(0.1) & \cmark & CDB & \cmark & 44.8 & 41.8 & - & 233  &  2038 &  2023 &  1933 
 & \textcolor{custom_green}{{$\mathbf{\times {1.70}}$}} & \textcolor{custom_green}{{$\mathbf{\times {1.62}}$}} \\
      & 0.53B & \cmark & 60B & \cmark & 2 & Avg  & 64 & SVD &  15B & Agg\,(0.1) & \cmark & CDB & \cmark & 45.1 & 33.7 & - & 211  &  1733 &  1719 &  1617  
 & \textcolor{custom_green}{{$\mathbf{\times {1.38}}$}} & \textcolor{custom_green}{{$\mathbf{\times {1.32}}$}} \\
      & 0.58B & \cmark & 60B & \cmark & 2 & Avg  & 128 & SVD &  15B & Agg\,(0.1) & \cmark & CDB & \cmark & 45.2 & 33.9 & - & 209  &  1733 &  1717 &  1609 
 & \textcolor{custom_green}{{$\mathbf{\times {1.36}}$}} & \textcolor{custom_green}{{$\mathbf{\times {1.30}}$}} \\
      TinyLlama & 0.68B & \cmark & 60B & \cmark & 2 & Avg  & 256 & SVD &  15B & Agg\,(0.1) & \cmark & CDB & \cmark & 45.6 & 33.9 & - & 209  &  1728 &  1714 &  1606 
 & \textcolor{custom_green}{{$\mathbf{\times {1.34}}$}} & \textcolor{custom_green}{{$\mathbf{\times {1.28}}$}} \\
      & 0.86B & \cmark & 60B & \cmark & 2 & Avg  & 512 & SVD &  15B & Agg\,(0.1) & \cmark & CDB & \cmark & 46.1 & 34.2 & - & 209  &  1716 &  1702 &  1581 
 & \textcolor{custom_green}{{$\mathbf{\times {1.28}}$}} & \textcolor{custom_green}{{$\mathbf{\times {1.23}}$}} \\
      \cmidrule(l{2pt}r{2pt}){2-23}
      & 0.53B & \cmark & 60B & \cmark & 2 & Avg  & 64 & SVD &  15B & Agg\,(0.3) & \cmark & CDB & \cmark & 44.6 & 36.1 & - &  211  & 1810 &  1796 &  1688 
 & \textcolor{custom_green}{{$\mathbf{\times {1.45}}$}} & \textcolor{custom_green}{{$\mathbf{\times {1.38}}$}} \\
      & 0.58B & \cmark & 60B & \cmark & 2 & Avg  & 128 & SVD &  15B & Agg\,(0.3) & \cmark & CDB & \cmark & 44.8 & 36.0 & - & 209  &  1802 &  1787 &  1668  
 & \textcolor{custom_green}{{$\mathbf{\times {1.41}}$}} & \textcolor{custom_green}{{$\mathbf{\times {1.35}}$}} \\
      & 0.68B & \cmark & 60B & \cmark & 2 & Avg  & 256 & SVD &  15B & Agg\,(0.3) & \cmark & CDB & \cmark & 44.8 & 36.2 & - & 209  &  1793 &  1779 &  1668 
 & \textcolor{custom_green}{{$\mathbf{\times {1.39}}$}} & \textcolor{custom_green}{{$\mathbf{\times {1.33}}$}} \\
      & 0.86B & \cmark & 60B & \cmark & 2 & Avg  & 512 & SVD &  15B & Agg\,(0.3) & \cmark & CDB & \cmark & 45.8 & 36.5 & - & 209  &  1778 &  1763 &  1637  
 & \textcolor{custom_green}{{$\mathbf{\times {1.33}}$}} & \textcolor{custom_green}{{$\mathbf{\times {1.27}}$}} \\
      \midrule
      & 0.81B & \cmark & 75B & \xmark & - & -  & - & - & - & - & - & - & \xmark & 49.3 & - & - & 229  &  4273 &  5346 &  5149 &\textcolor{gray}{{$\mathbf{\times 1.00}$}} & \textcolor{custom_red}{{$\mathbf{\times 0.89}$}} \\
      & 0.81B & \cmark & 75B & \xmark & - & -  & - & - &  - & - & - & CSB & \xmark & 49.3 & - & - & 229  &  5813 &  5724 &  5149 & \textcolor{custom_green}{{$\mathbf{\times 1.13}$}} & \textcolor{gray}{{$\mathbf{\times 1.00}$}} \\
      \cmidrule(l{2pt}r{2pt}){2-23}
      & 0.40B & \cmark & 60B & \cmark & 2 & Step  & - & - &  15B & Agg\,(0.1) & \cmark & CDB & \cmark & 45.4 & 42.0 & - & 61  &  1339 &  1333 &  1281 
 & \textcolor{custom_green}{{$\mathbf{\times {1.77}}$}} & \textcolor{custom_green}{{$\mathbf{\times {1.57}}$}} \\
      & 0.44B & \cmark & 60B & \cmark & 2 & Avg  & 64 & SVD &  15B & Agg\,(0.1) & \cmark & CDB & \cmark & 46.1 & 37.1 & - & 63  &  1205 &  1203 &  1140 
 & \textcolor{custom_green}{{$\mathbf{\times {1.44}}$}} & \textcolor{custom_green}{{$\mathbf{\times {1.28}}$}} \\
      & 0.48B & \cmark & 60B & \cmark & 2 & Avg  & 128 & SVD &  15B & Agg\,(0.1) & \cmark & CDB & \cmark & 46.2 & 37.8 & - & 59  &  1156 &  1180 &  1108 
 & \textcolor{custom_green}{{$\mathbf{\times {1.32}}$}} & \textcolor{custom_green}{{$\mathbf{\times {1.17}}$}} \\
      Pythia & 0.55B & \cmark & 60B & \cmark & 2 & Avg  & 256 & SVD &  15B & Agg\,(0.1) & \cmark & CDB & \cmark & 46.5 & 38.0 & - & 59  &  1138 &  1139 &  1071
 & \textcolor{custom_green}{{$\mathbf{\times {1.22}}$}} & \textcolor{custom_green}{{$\mathbf{\times {1.08}}$}} \\
      & 0.70B & \cmark & 60B & \cmark & 2 & Avg  & 512 & SVD &  15B & Agg\,(0.1) & \cmark & CDB & \cmark & 47.2 & 38.2 & - & 53  &  1051 &  1077 &  1021  
 & \textcolor{custom_red}{{$\mathbf{\times {0.98}}$}} & \textcolor{custom_red}{{$\mathbf{\times {0.87}}$}} \\
      \cmidrule(l{2pt}r{2pt}){2-23}
      & 0.44B & \cmark & 60B & \cmark & 2 & Avg  & 64 & SVD &  15B & Agg\,(0.3) & \cmark & CDB & \cmark & 45.1 & 39.0 & - & 63  &  1254 &  1252 &  1190  
 & \textcolor{custom_green}{{$\mathbf{\times {1.50}}$}} & \textcolor{custom_green}{{$\mathbf{\times {1.33}}$}} \\
      & 0.48B & \cmark & 60B & \cmark & 2 & Avg  & 128 & SVD &  15B & Agg\,(0.3) & \cmark & CDB & \cmark & 45.9 & 39.0 & - & 59  &  1200 &  1226 &  1153 
 & \textcolor{custom_green}{{$\mathbf{\times {1.37}}$}} & \textcolor{custom_green}{{$\mathbf{\times {1.22}}$}} \\
      & 0.55B & \cmark & 60B & \cmark & 2 & Avg  & 256 & SVD &  15B & Agg\,(0.3) & \cmark & CDB & \cmark & 46.0 & 39.4 & - & 59  &  1180 &  1180 &  1112 
 & \textcolor{custom_green}{{$\mathbf{\times {1.27}}$}} & \textcolor{custom_green}{{$\mathbf{\times {1.12}}$}} \\
      & 0.70B & \cmark & 60B & \cmark & 2 & Avg  & 512 & SVD &  15B & Agg\,(0.3) & \cmark & CDB & \cmark & 46.7 & 39.7 & - & 53  &  1088 &  1114 &  1058 
 & \textcolor{custom_green}{{$\mathbf{\times {1.02}}$}} & \textcolor{custom_red}{{$\mathbf{\times {0.90}}$}} \\
    \bottomrule
    \end{tabular}
    }
    \caption{
    Hypothetical generation speedup of Recursive Transformers across three models. We utilized the measurements of per-token generation time calculated in Table\,\ref{tab_rrt:measured_generation_time_a100_16gb}, which were calculated using an A100 GPU with 16GB memory constraint, a prefix length of 64, and a decoding length of 256. We only considered the time spent within Transformer blocks, simulating generation on the SlimPajama, RedPajama, and PG19 test sets. We used vanilla transformer models, both with and without continuous sequence-wise batching (CSB), as our baselines. Our recursive models further enhance throughput by applying continuous depth-wise batching (CDB), leveraging looping and early-exiting techniques. The throughput improvements over the vanilla Transformer without and with sequence-wise batching are denoted as $\Delta_V$ and $\Delta_{Seq}$, respectively. To aid in understanding the speedup, we also provide the performance of intermediate layers and the maximum batch size.
    }
    \label{tab_rrt:final_performance_throughput_16gb}
\end{table}

\chapter{Mixture-of-Recursions: Learning Dynamic Recursive
Depths for Adaptive Token-Level Computation}
\label{chapter:mor}

\begin{tcolorbox}[colback=gray!10, colframe=black, arc=3mm, boxrule=1pt]
    \textbf{Publication Note:} This chapter is based on the paper accepted to the Conference on Neural Information Processing Systems (NeurIPS 2025)~\citep{bae2025mixture}.
    \\
    
    \textbf{Abstract:} Scaling language models unlocks impressive capabilities, but the accompanying computational and memory demands make both training and deployment expensive. Existing efficiency efforts typically target either parameter sharing or adaptive computation, leaving open the question of how to attain both simultaneously. We introduce \textbf{Mixture-of-Recursions (MoR)}, a unified framework that combines the two axes of efficiency inside a single Recursive Transformer. MoR reuses a shared stack of layers across recursion steps to achieve parameter efficiency, while lightweight routers enable adaptive token-level thinking by dynamically assign recursion depth to tokens, thereby focusing quadratic attention computation only where it is most useful. Further enhancing its efficiency, MoR incorporates a recursion-wise key-value caching mechanism that eliminates redundant memory access across recursion steps by selectively storing only the key-value caches for designated tokens. Across pretraining runs at model scales ranging from 135M to 1.7B parameters, MoR forms a new Pareto frontier: at equal training FLOPs and smaller model sizes, it significantly lowers validation perplexity and improves few-shot accuracy, while delivering higher throughput compared with vanilla and existing recursive baselines. These gains demonstrate that MoR is an effective path towards large-model quality without incurring large-model cost.
    \looseness=-1
\end{tcolorbox}

\vspace{-4pt}
\section{Introduction}
\label{sec:intro}

\begin{figure}[h]
    \centering
    \includegraphics[width=0.23\columnwidth]{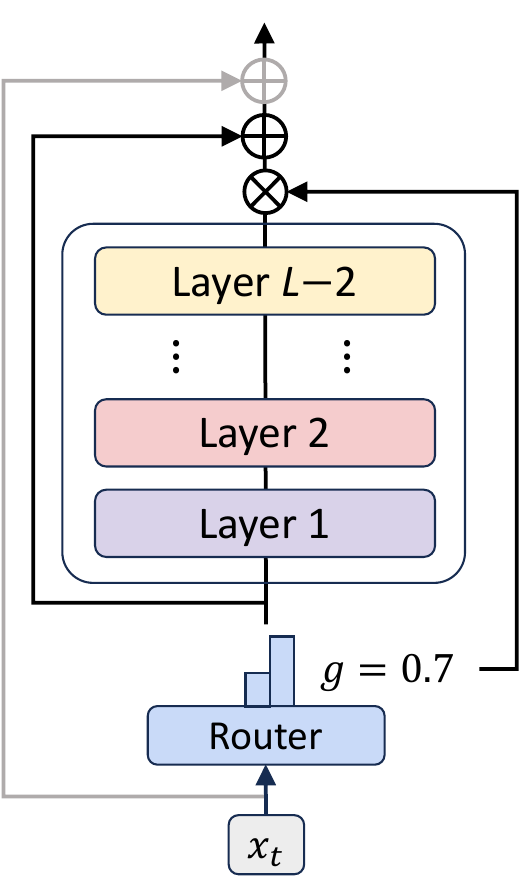}
    \hspace{23pt}
    \includegraphics[width=0.238\columnwidth]{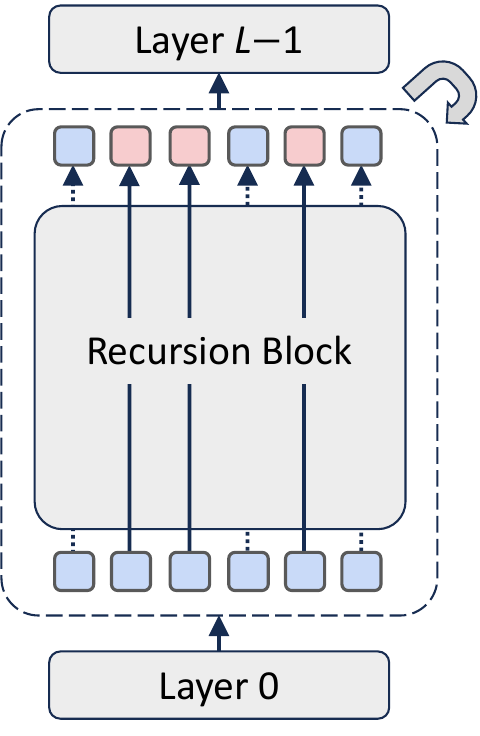}
    \hspace{7pt}
    \includegraphics[width=0.43\columnwidth]{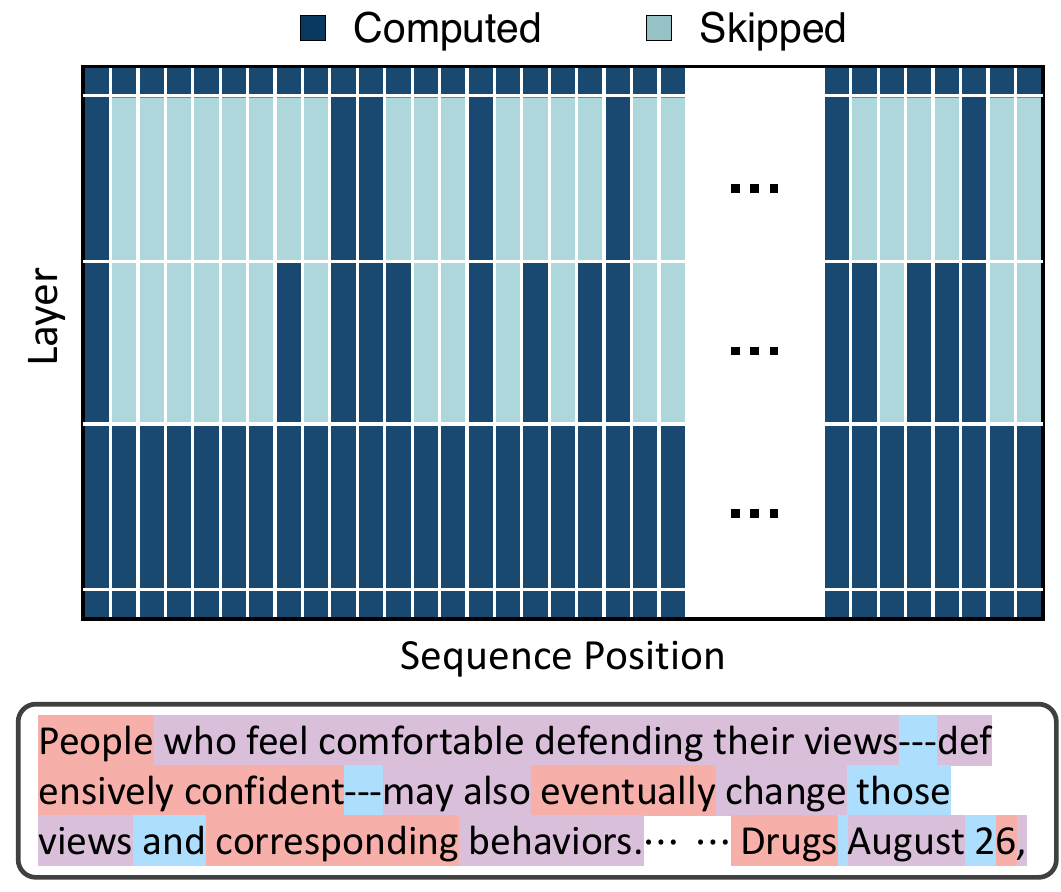}
    \caption{
    Overview of Mixture-of-Recursions (MoR). (\textit{Left}) Each recursion step consists of a fixed stack of layers and a router that determines whether each token should pass through or exit. This recursion block corresponds to the gray box in the middle.
    (\textit{Middle}) The full model structure, where the shared recursion step is applied up to $N_r$ times for each token depending on the router decision.
    (\textit{Right}) An example routing pattern showing token-wise recursion depth, where darker cells indicate active computation through the recursion block. Below shows the number of recursion steps that each subword token undergoes to predict the \textit{next} token, shown in colors: \raisebox{0pt}[0.5em][0.1em]{\colorbox{DarkOrchid!25}{1}}, \raisebox{0pt}[0.5em][0.1em]{\colorbox{SkyBlue!55}{2}}, and \raisebox{0pt}[0.5em][0.1em]{\colorbox{Salmon!60}{3}}. 
    }
    \label{fig_mor:mor_overview}
\end{figure}

Scaling Transformer networks to hundreds of billions of parameters has unlocked impressive few-shot generalization and reasoning abilities~\citep{brown2020language, chowdhery2023palm, grattafiori2024llama, openai2023gpt4, reid2024gemini, deepseek-ai2024deepseek0v3, comanici2025gemini25pushingfrontier}.
However, the accompanying memory footprint and computational requirements make both training and deployment outside hyperscale data centers challenging~\citep{patterson2021carbon, momeni2024training}.
This has motivated researchers to seek alternative ``efficient'' designs~\citep{tay2022efficient, wan2023efficient}. 
Among the different axes of efficiency, \textit{parameter efficiency}~\citep{dehghani2018universal, bae2024relaxed, shazeer2017outrageously, fedus2022switch, lecun1989optimal}—reducing or sharing model weights—and \textit{adaptive computation}~\citep{raposo2024mixture, schuster2022confident, fedus2022switch, leviathan2023fast, kim2021length}—spending more compute only when it is needed—are promising, actively studied research directions.

One proven route to parameter efficiency is \textit{layer tying}, in which a shared set of weights is reused across multiple layers~\citep{dehghani2018universal, lan2019albert0, takase2021lessons, gholami2023generative, bae2024relaxed}.
For adaptive computation, a common approach is \textit{early-exiting}, which dynamically allocates compute by exiting  earlier in the network when predicting simpler tokens~\citep{elhoushi2024layerskip, schuster2022confident, Elbayad2020Depth-Adaptive, bae-etal-2023-fast}.
Despite the progress achieved along each of these individual efficiency axes, an architecture that effectively unifies both parameter efficiency and adaptive computation is still missing.
Recursive Transformers~\citep{bae2024relaxed, fan2024looped, giannou2023looped, yang2023looped, saunshi2025reasoning, geiping2025scaling, aleksandrov2025abbie}, models that repeatedly apply the same set of shared layers multiple times, offer a strong foundation due to their built-in weight sharing.
However, prior attempts at dynamic recursion have often been constrained by practical hurdles, such as requiring additional specialized training procedures or facing challenges in efficient deployment. This has led most approaches to still default to a simpler fixed-depth recursion, which applies the same amount of computation to every token and is thus incapable of delivering truly adaptive token-level compute allocation.

In this work, we introduce \textit{Mixture-of-Recursions} (MoR), a unified framework that fully leverages the potential of Recursive Transformers (see Figure\,\ref{fig_mor:mor_overview}).
MoR trains lightweight routers end-to-end to assign token-specific recursion depths: it decides how many times a shared parameter block is applied to each token according to its required depth of ``thinking'', thereby directing computation to where it is most needed. 
This dynamic, token-level recursion inherently facilitates recursion-wise key–value (KV) caching, selectively storing and retrieving key–value pairs corresponding to each token’s assigned recursion depth. This targeted caching strategy reduces memory traffic, thereby improving throughput without relying on post-hoc modifications.
Therefore, MoR simultaneously (i) ties weights to cut parameters, (ii) routes tokens to cut redundant FLOPs, and (iii) caches key-values recursion-wise to cut memory traffic—all inside a single architecture.

Conceptually, MoR provides a \textit{pre-training} framework for latent space reasoning—performing non-verbal thinking by iteratively applying a single parameter block~\citep{hao2024training, geiping2025scaling, goyal2023think}. However, unlike approaches that deliberate on augmented continuous prompts before generation~\citep{liu2024deliberation, goyal2023think, hao2024training, shen2025codi}, MoR enables this latent thinking directly during the decoding of each token~\citep{zelikman2024quiet}. Furthermore, routing mechanism facilitates adaptive reasoning along the model's vertical axis\footnote{Thinking along the depth axis is like generating thoughts along the horizontal axis.}, moving beyond the uniform, fixed thinking depth common in prior work~\citep{geiping2025scaling, tack2025llm}. In essence, MoR enables models to efficiently adjust their thinking depth on a per-token basis, unifying parameter efficiency with adaptive computation.\looseness=-1

\vspace{-6pt}
\paragraph{Contributions.}
In summary, our key contributions in this paper are as follows.
\vspace{-3pt}
\begin{itemize}[leftmargin=*, itemsep=2pt]
    \item \textbf{Unified framework for efficient language modeling:} We present \emph{Mixture-of-Recursions} (MoR), the first architecture to unify efficiency paradigms—parameter sharing\,(\S\ref{subsec:preliminary}), token-level adaptive thinking depth\,(\S\ref{subsec:routing}), and memory-efficient KV caching\,(\S\ref{subsec:kv_cache})—within a single framework.

     \item \textbf{Dynamic recursion routing:} We introduce a router trained from scratch to assign dynamic per-token recursion depths. This aligns training with inference-time behavior and eliminates the need for costly, performance-degrading post-hoc routing stages used in conventional early-exit methods.

    \item \textbf{Extensive empirical validation:} Across models from 135M to 1.7B parameters\footnote{These are base model sizes, while MoR models have fewer unique parameters due to parameter sharing.} under equal compute budgets, MoR establishes a new Pareto frontier by improving validation loss and few-shot accuracy relative to vanilla and recursive baselines (\S\ref{subsec:main_results}, \S\ref{subsec:scaling_laws}).
    
    \item \textbf{Efficient architecture:} MoR dramatically reduces training FLOPs by selectively engaging only essential sequences in attention operations. Simultaneously, reduction in KV cache sizes leads to enhanced inference throughput itself, further boosted by continuous depth-wise batching (\S\ref{subsec:efficiency}).
    
\end{itemize}
\section{Related Work}
\label{sec:related_work}

\paragraph{Recursive Transformers.}
Parameter sharing provides an orthogonal path to efficiency~\citep{dehghani2018universal,lan2019albert0,xia2019tied,jaegle2021perceiver,takase2021lessons,tan2023sparse,ng2024loop,bae2024relaxed}.
The \textit{Universal Transformer} first showed that repeatedly applying a single block can match the representational power of deep, non-shared stacks~\citep{dehghani2018universal}. 
\textit{Looped Transformers} shown to be effective, can act as programmable computers~\citep{giannou2023looped}, learn iterative data-fitting algorithms~\citep{yang2023looped}, generalize to much longer inputs on algorithmic tasks~\citep{fan2024looped}, and illuminate few-shot learning by mimicking multi-step optimizers~\citep{gatmiry2024can}.
Furthermore, Bae et al. (2024)~\citep{bae2024relaxed} mitigate the accuracy loss often associated with weight tying by adding low-rank adaptation (LoRA) adapters~\citep{hu2022lora} in each loop, yielding \emph{Relaxed Recursive Transformers}. 
Recent work further demonstrates that Recursive Transformers excel at latent reasoning via recurrent depth~\citep{geiping2025scaling}.  
While most prior studies focus on the efficiency gains from weight tying, the \emph{recursive} architecture itself offers a second level: inspired by early-exiting~\citep{schuster2022confident} and compute routing~\citep{raposo2024mixture}, one can vary the number of recursions per input (e.g., per token), allocating compute only where it is most beneficial during both training and inference.\looseness=-1

\vspace{-3pt}
\paragraph{Adaptive computation.}
Many works have shown that \emph{dynamic} compute allocation can markedly reduce the cost of training and inference, from traditional neural networks~\citep{bengio2015conditional, DBLP:conf/eccv/HuangSLSW16, teerapittayanon2016branchynet, panda2016conditional, graves2016adaptive} to large language models~\citep{hou2020dynabert,Elbayad2020Depth-Adaptive,fedus2022switch,bae-etal-2023-fast, elhoushi2024layerskip, raposo2024mixture}.
Early exiting methods learn to halt processing for ``easy'' samples (e.g., tokens or sequences in language modeling) by skipping the remaining layers~\citep{Elbayad2020Depth-Adaptive, schuster2022confident, dehghani2018universal,mofakhami2024performance}. Alternatively, early exits can be combined with speculative decoding techniques~\citep{chen2023accelerating,leviathan2023fast} during inference by leveraging lower layers for fast drafting~\citep{bae-etal-2023-fast, elhoushi2024layerskip}.
Recently, \textit{Mixture-of-Depths} (MoD)~\citep{raposo2024mixture} reframed adaptive depth as a routing problem: a router at each layer selects a \emph{subset} of tokens to receive the full computation, while the rest bypass the layer, yielding finer-grained conditional compute.  
This new form of adaptive allocation is well suited to Transformer architectures and has already been extended to other modalities~\citep{zhang2024p,luo2024gamma}, highlighting a promising paradigm of dynamic compute at token-level granularity.
MoR applies this routing idea to \emph{recursive} Transformers: tokens are dynamically sent through repeated calls of a single, weight‑tied block instead of through distinct layers. This shift keeps parameter count constant, allows arbitrarily deep (adaptive) compute beyond the model’s physical depth.\looseness=-1

\vspace{-3pt}
\paragraph{Routing mechanism.} 
LLMs have increasingly employed routers to enable adaptive computation, primarily in sparse Mixture-of-Experts (MoE) frameworks~\citep{shazeer2017outrageously, lepikhin2020gshard, dai2022stablemoe, zoph2022st}, i.e., each token is processed by a subset of expert networks chosen by a learned router, dramatically increasing model capacity without a computational overhead. Early MoE architectures~\citep{lepikhin2020gshard, fedus2022switch, jiang2024mixtral} adopted a \textit{token-choice} routing strategy, wherein the router selects the top-k experts for each token based on its hidden state. While effective, this approach often leads to load imbalance across experts, necessitating auxiliary balancing losses. 
To address this, \textit{expert-choice} routing~\citep{zhou2022mixture, guo2025deepseek} has been proposed, wherein each expert selects the tokens to serve, ensuring perfect load balancing and improved efficiency. 
Building on this, a few works employed trainable routers to determine which layers to skip~\citep{zeng2023learning, raposo2024mixture, gadhikar2024attention}. Unlike traditional early-exit methods, these expert-choice routing mechanisms enforce a static compute budget by capping the number of tokens processed per layer (or depth).\looseness=-1

\vspace{-3pt}
\paragraph{Key-value caching.}
Key–value (KV) caching stores the per-token key and value tensors produced at each layer during autoregressive decoding; reusing them eliminates quadratic-time recomputation and boosts throughput~\citep{shazeer2019fast, ge2023model, liu2024minicache, xiao2023efficient, pope2022efficiently, kang2024gear0, brandon2024reducing}.
Unfortunately, retaining these tensors quickly saturates GPU memory, especially for long contexts and large batches~\citep{chowdhery2023palm, brandon2024reducing}.
Prior work tackles this issue by quantizing KV activations to lower precision~\citep{hooper2024kvquant0, zhang2024kv}, discarding entries that contribute little to the final output~\citep{zhang2023h2o, liu2023scissorhands}, and sharing keys and values across attention heads~\citep{shazeer2019fast, ainslie2023gqa}.
Brandon et al. (2024)~\citep{brandon2024reducing} push this idea further, allowing adjacent layers to share the same key and value tensors and achieving additional memory savings with negligible quality loss.
Our Mixture-of-Recursions offer a complementary avenue: KV caches generated in early recursions can be reused in later ones, potentially reducing memory consumption even further.
This provides the advantage of only needing to run the first recursion during prefill phase~\citep{sun2024you}, promising significant speedups for prompt settings over 1 million tokens. Two caching strategies in MoR can be optimized based on their distinct benefits to suit various deployment settings.\looseness=-1

\paragraph{Latent reasoning.}
An emerging line of work enables LLMs to perform reasoning internally within hidden states rather than through explicit verbalization~\citep{goyal2023think, pfau2024let, cheng2024compressed, tack2025llm, konglatent}. Many approaches adopt a \textit{fixed} latent reasoning depth: they insert special tokens or structured prompts (e.g., a learnable ``pause” token~\citep{goyal2023think} or filler punctuation~\citep{pfau2024let}) that allow the model to execute a predetermined number of hidden reasoning passes before producing an answer. Others reuse the model’s hidden states in a closed loop for a fixed number of iterations by feeding final hidden states back as input to simulate chain-of-thought~\citep{hao2024training, shen2025codi, saunshi2025reasoning}. Another line of research enhances latent reasoning by augmenting hidden states with intermediate semantic signals~\citep{zelikman2024quiet, tack2025llm}. However, these methods lack the flexibility to allocate computation where it is most needed, leading to unnecessary overhead on easy inputs and insufficient reasoning on complex ones. 
This motivates leveraging looping mechanisms for more adaptive latent reasoning~\citep{chen2025inner, geiping2025scaling, saunshi2025reasoning, zeng2025pretraining}.

\section{Method}

We propose \textit{Mixture-of-Recursions} (MoR)—a framework that dynamically adjusts recursion step for each token during \textit{pretraining} and \textit{inference}. The core of MoR lies in two components: a routing mechanism that assigns token-specific recursion steps to adaptively concentrate computation on more challenging tokens, and a KV caching strategy that defines how KV pairs are stored and selectively utilized for attention at each recursive step.\looseness=-1

\subsection{Preliminary}\label{subsec:preliminary}

\paragraph{Recursive Transformers.}
The standard Transformer \citep{vaswani2017attention} constructs token representations through a stack of $L$ unique layers, each with a self-attention and a feed-forward network. At time step $t$, the hidden state $h$ evolves as: $\mathbf{h}_t^{\ell+1} = f\bigl(\mathbf{h}_t^{\ell}; \Phi_\ell\bigr),$ where $\ell = 0,\dots,L{-}1$ and $\Phi_\ell$ represents the parameters of the $\ell$-th layer.
Recursive Transformers~\citep{bae2024relaxed, fan2024looped, giannou2023looped, yang2023looped, saunshi2025reasoning} aim to reduce parameter count by reusing layers across depth. Instead of having $L$ distinct sets of weights, they partition the model into $N_r$ recursion \emph{blocks}, where each block uses a shared pool of parameters $\Phi'$. This design allows for more computation (by increasing the effective network depth) without increasing parameter size.

\newcolumntype{M}[1]{>{\centering\arraybackslash}m{#1}}

\begin{table*}[t]
\centering
\caption{
Parameter-sharing strategies in Recursive Transformers. This table shows \textit{Cycle} and \textit{Middle-Cycle} schemes with cyclic layer reuse, where Middle-Cycle retains unique first and last layers. 
}

\label{tab_mor:sharing_strategy}
\renewcommand{\arraystretch}{1.1}
\resizebox{\textwidth}{!}{%
\setlength{\tabcolsep}{7pt}
\begin{tabular}{
    l
  | M{0.32\textwidth} 
  | M{0.16\textwidth} 
  | M{0.32\textwidth} 
  | >{\raggedright\arraybackslash}m{0.16\textwidth} 
}
\toprule
& \multicolumn{2}{c|}{\textbf{Cycle Strategy}} & \multicolumn{2}{c}{\textbf{Middle-Cycle Strategy}}\\
\cmidrule(l{2pt}r{2pt}){2-3}\cmidrule(l{2pt}r{2pt}){4-5}
\textbf{Layers} & Equation & Figure & Equation & \multicolumn{1}{M{0.15\textwidth}}{Figure} \\
\midrule
Last & -- & & $f\bigl(\mathbf{h}_t^{L-1}; \Phi_{L-1}\bigr)$ &\\[14pt]
Recursion & $f\!\Bigl(\mathbf{h}_t^{\ell}; \Phi'_{\ell \bmod (L/N_r)}\Bigr)$ & \multirow{-2}{*}{\includegraphics[width=0.91\linewidth]{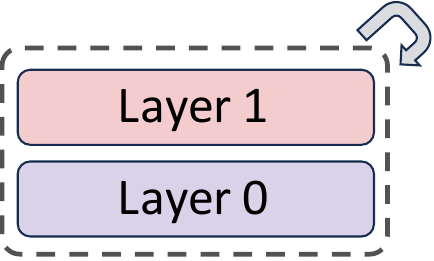}}
  & $f\!\Bigl(\mathbf{h}_t^{\ell}; \Phi'_{(\ell - 1 \bmod ((L-2)/N_r)) + 1}\Bigr)$ & \,\,\,\multirow{-2.3}{*}{\includegraphics[width=0.89\linewidth]{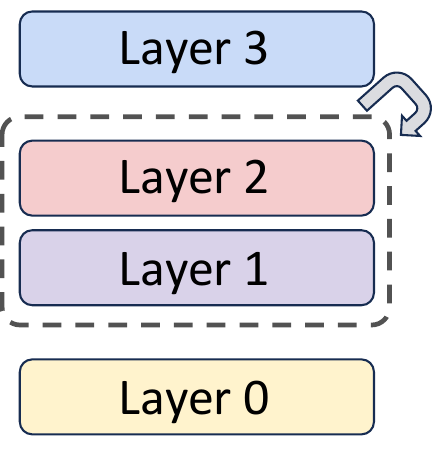}}  \\[14pt]
First   & --  &  & $f\bigl(\mathbf{h}_t^{0}; \Phi_0\bigr)$ \\
[4pt]
\bottomrule
\end{tabular}
}
\end{table*}

\vspace{-5pt}
\paragraph{Parameter-sharing strategies.}
We examine four parameter-sharing strategies: \textit{Cycle}, \textit{Sequence}, and their variants \textit{Middle-Cycle} and \textit{Middle-Sequence}. Table\,\ref{tab_mor:sharing_strategy} summarizes two main designs, and the full list is provided in Table\,\ref{tab_mor:sharing_strategy_middle} in the Appendix. In Cycle sharing, recursion blocks are reused cyclically. For example, consider an original non-recursive model with $L{=}9$ layers and its recursive counterpart using $N_r{=}3$ recursions. Under the ``Cycle'' strategy, the layers are shared and unrolled as [$(0,1,2),(0,1,2),(0,1,2)$]. In ``Sequence'' sharing, each recursion block reuses the same layer consecutively before moving to the next, resulting in [$(0,0,0),(1,1,1),(2,2,2)$] for the same configuration. Both have the same effective number of layers when unrolled ($L{=}9$), but with a different order. Furthermore, the ``Middle'' variants preserve full-capacity parameters at the first and last layers ($\Phi_0$ and $\Phi_{L-1}$), while sharing weights among the intermediate layers.

\vspace{-5pt}
\paragraph{Enhanced training and inference efficiency in recursive models.}

Parameter sharing strategies can reduce the number of unique trainable parameters by a factor of the recursion number, effectively amortizing the memory footprint of the model. From a distributed training perspective, this becomes highly efficient when using Fully Sharded Data Parallel (FSDP)~\citep{zhao2023pytorch}. While a single \texttt{all-gather} operation would only support one iteration previously (i.e., 1 iter/gather), a recursive model reuses the same gathered parameters across all recursive steps (i.e., $N_r$ iter/gather). Furthermore, recursive architectures enable a novel inference paradigm, continuous depth-wise batching~\citep{bae2024relaxed, hooper2023speed}. This technique allows tokens at different stages to be grouped into a single batch, as they all utilize the same block of parameters. This can eliminate the bubbles—idle periods spent waiting for other samples to complete—thereby leading to significant throughput gains.

\vspace{-5pt}
\paragraph{Limitations in prior works.}

Although model parameters are tied, the distinct KV caches are typically used for each depth. This design fails to reduce the cache sizes, meaning the high retrieval latency still remains a severe inference bottleneck. Moreover, most existing recursive models simply apply a fixed recursion depth to all tokens, ignoring the varying complexity. While post-hoc methods like early-exiting methods can introduce some adaptivity, they often require separate training phases that can degrade performance~\citep{schuster2022confident, elhoushi2024layerskip, bae2024relaxed}. Ideally, the recursion depth should be learned dynamically during pretraining, allowing the model to adapt its computational path to each token's difficulty in a data-driven manner. However, such dynamic paths introduce a new challenge: exited tokens will have missing KV pairs at subsequent recursion depths. Addressing this would require a parallel decoding mechanism~\citep{bae-etal-2023-fast, elhoushi2024layerskip, kim2023speculative} to efficiently compute actual KV pairs, but this requires separate, complex engineering.\looseness=-1

\begin{figure}[t!]
    \vspace{-13pt}
    \centering
    \begin{subfigure}[t]{0.31\textwidth}
    \centering
    \includegraphics[width=0.7\textwidth]{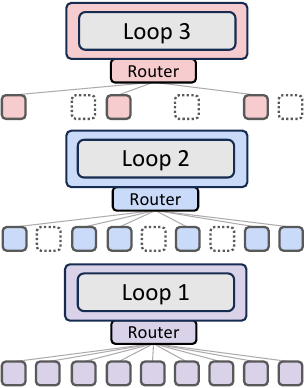}
        \centering
        \captionsetup{justification=centering}
        \subcaption{Expert-choice routing}
        \label{fig_mor:method_overview-a}
    \end{subfigure}
    \hfill
    \begin{subfigure}[t]{0.41\textwidth}
    \centering
    \includegraphics[width=\textwidth]{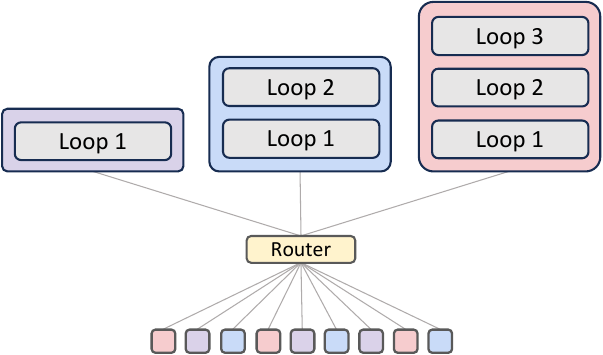}
        \centering
        \captionsetup{justification=centering}
        \subcaption{Token-choice routing}
        \label{fig_mor:method_overview-b}
    \end{subfigure}
    \centering
    \hfill
    \begin{subfigure}[t]{0.245\textwidth}
        \centering\includegraphics[width=0.65\textwidth]{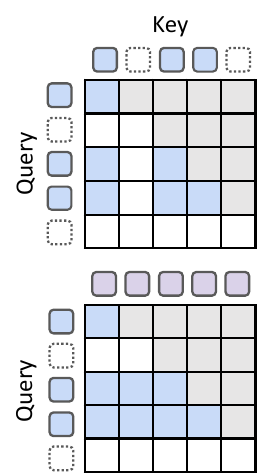}
        \captionsetup{justification=centering}
        \subcaption{Caching mechanism}
        \label{fig_mor:method_overview-c}
    \end{subfigure}
    \caption{
        Architectural components of Mixture-of-Recursions (MoR).  
        (\textit{a}) {Expert-choice routing:} At each recursion step, a router selects top-$k$ tokens to continue, progressively narrowing the set of active tokens with depth.  
        (\textit{b}) {Token-choice routing:} Each token is assigned a fixed recursion step at the outset via a single routing decision, defining its complete compute path through the model.  
        (\textit{c}) {KV caching strategies:} Each square in the matrix represents whether a token (row) attends to another token’s cached key (column).  
        In ``recursion-wise KV caching'' (\textit{Top}), only the keys of currently selected (non-dropped) tokens at each recursion step are cached (\raisebox{0pt}[0.1em][0.1em]{\colorbox{custom_blue}{blue}}), and attention is restricted only to these entries.  
        In ``recursive KV sharing'' (\textit{Bottom}), all keys of previous tokens are cached at the first recursion step (\raisebox{0pt}[0.1em][0.1em]{\colorbox{custom_purple}{purple}}) and shared across subsequent recursion steps for attention operations.\looseness=-1
        }
    \label{fig_mor:method_overview}
\end{figure}

\subsection{Routing Strategies: Expert-choice vs. Token-choice}
\label{subsec:routing}

\paragraph{Expert-choice routing. 
(Figure\,\ref{fig_mor:method_overview-a})}
Inspired by top-$k$ gating in MoD models~\citep{raposo2024mixture}, 
in expert-choice routing, each recursion depth becomes an expert and selects their preferred top-$k$ tokens (e.g., for $N_r = 3$, we have three experts: Expert 1 applies the first recursion step, Expert 2 applies the second recursion step, and so on).
At each recursion step $r$, the corresponding router uses the hidden state $\mathcal{H}_t^r$ (input to the $r$-th recursion block) and its routing parameters $\theta_r$ to compute a scalar score $g_t^r = \mathcal{G}(\theta_r^\top \mathcal{H}_t^r$) for token $t$. Here, $\mathcal{G}$ represents an activation function like \texttt{sigmoid} or \texttt{tanh}.
Then, the top-$k$ tokens are selected to pass through the recursion block:\looseness=-1
\begin{align}
\mathcal{H}_t^{r+1} =
\begin{cases}
g_t^r f(\mathcal{H}_t^r,\, \Phi') + \mathcal{H}_t^r, & \text{if } g_t^r > P_\beta({G}^r) \\
\mathcal{H}_t^r, & \text{otherwise}
\end{cases}
\end{align}
where $P_\beta({G}^r)$ is the $\beta$-percentile threshold over all scores at recursion step $r$.

To ensure coherent progression through steps, we adopt \textit{hierarchical filtering}: only tokens selected at recursion step $r$ can be re-evaluated at $r{+}1$. This simulates early-exit behavior while learning from scratch. As deeper layers tend to encode increasingly abstract and sparse information~\citep{li2022lazy, yang2024pyramidinfer, nawrot2024dynamic}, this mechanism prioritizes computation for only the most demanding tokens.

\vspace{-5pt}
\paragraph{Token-choice routing. 
(Figure\,\ref{fig_mor:method_overview-b})}  Unlike expert-choice, where token selection is made at each recursion step, token-choice commits each token to a full sequence of recursion blocks from the start.
Formally, given the hidden state $\mathcal{H}_t^1$ (in Middle-Cycle strategy, $\mathcal{H}_t^1 = h_t^1$), the router computes a non-linear function (\texttt{softmax} or \texttt{sigmoid}) over experts: ${g}_t = \mathcal{G}(\theta_r^\top \mathcal{H}_t^1)$, where $g_t^j$ denotes the routing score for expert $j \in \{1,\dots,N_r\}$. The token is assigned to expert $i = \arg\max_j g_t^j$ (top-1 gating), which corresponds to sequentially applying the recursion $i$ times. The hidden state is then updated recursively as:
\begin{align}
\mathcal{H}_t^{r+1} =
\begin{cases}
g_t^r f(\mathcal{H}_t^r,\, \Phi') + \mathcal{H}_t^1, & \text{if } r = i  \\
g_t^r f(\mathcal{H}_t^r,\, \Phi'), & \text{otherwise} \\
\end{cases}
\end{align}

To compare routing strategies under equal compute, we align the token allocation budgets of expert-choice with that of token-choice. Specifically, we calibrate token capacity (i.e., top-\textit{k}) of expert-choice to match the expected token distribution of token-choice routing with perfect load balancing. 
In perfectly balanced token-choice, each token is assigned to recursion depth $i \in \{1, \dots, N_r\}$ with equal probability $1/N_r$. Thus, recursion step $j$ processes a fraction $(N_r - j + 1)/N_r$ of the tokens. For example, when $N_r = 3$, recursion steps 1, 2, and 3 handle $\{3/3, 2/3, 1/3\}$ of tokens, respectively. Therefore, we apply this same fractional allocation in the top-$k$ selection of the expert-choice routing (i.e., $k$ is sequenced like $N_r / N_r$, $\cdots$, $1 / N_r$ over $N_r$ recursion steps).\looseness=-1

\vspace{-5pt}
\paragraph{Strengths and limitations. (Table\,\ref{tab_mor:method_comparison}--\textit{Left})} 
Although expert-choice routing guarantees perfect load balancing with static top-$k$ selection, it suffers from information leakage~\citep{zhou2022mixture, wang2024auxiliary, raposo2024mixture}. This violation of causality during training forces to exploit an auxiliary router or a regularization loss~\citep{zhou2022mixture, raposo2024mixture}, aiming to precisely detect top-$k$ tokens at inference without access to future token information. Meanwhile, token-choice is free from such leakage, but typically requires a balancing loss or loss-free algorithms~\citep{wang2024auxiliary, fedus2022switch, zoph2022st} due to its inherent load balancing challenges. We explore each of these components for MoR in further detail (\S\ref{subsec:ablation_routing}).\looseness=-1

\begin{table*}[t!]
    \caption{
    Comparison of routing strategies and key-value caching strategies.
    (\textit{Left}) Summary of two routing strategies: expert-choice and token-choice, highlighting their pros, cons, and mitigating solutions from previous works~\citep{raposo2024mixture, wang2024auxiliary, zoph2022st}. 
    (\textit{Right}) Relative cost efficiency of caching strategies against a vanilla Transformer (normalized to 1). Here, $N_r$ denotes the number of recursions, and $k$ ($k < N_{\mathrm{ctx}}$) denotes the number of selected tokens per layer. KV cache memory and IO are measured across the entire model, whereas attention FLOPs are reported per layer.\looseness=-1
    }
    \label{tab_mor:method_comparison}
    \small
    \begin{subtable}[t]{0.478\textwidth}
    \renewcommand{\arraystretch}{1.1}
    \centering
    \resizebox{\linewidth}{!}{
    \setlength{\tabcolsep}{7pt}
    \begin{tabular}{l|c|c}
    \toprule
      &  \textbf{Expert-choice} & \textbf{Token-choice} \\
    \midrule
    Pros & Static\,compute\,budget & No\,leakage \\[3pt]
    Cons & Causality\,violation  & Load\,imbalance\\[1pt]
    $\llcorner$\,Sol & Aux\,Rout, Aux\,Loss & Bal\,Loss, Loss-free \\
    \bottomrule
    \end{tabular}
    }
    \end{subtable}
    \hspace{-4pt}
    \begin{subtable}[t]{0.522\textwidth}
    \renewcommand{\arraystretch}{1.1}
    \centering
    \resizebox{\linewidth}{!}{
    \setlength{\tabcolsep}{3.5pt}
    \begin{tabular}{l|c|c}
    \toprule
     & \textbf{\,Recursion-wise\,Caching\,} & \textbf{\,Recursive\,Sharing\,}\\
    \midrule
    KV\,Memory  & $(N_r + 1)/2N_r$ & $1/N_r$ \\[3pt]
     KV\,Cache\,IO  & $(N_r + 1) / 2N_r$ & 1 \\[3pt]
    Attn\,FLOPs  & $k^2/N_{\mathrm{ctx}}^2$  & $k/ N_{\mathrm{ctx}}$ \\
    \bottomrule
    \end{tabular}
    }
    \end{subtable}
\end{table*}

\subsection{KV Caching Strategies: Recursion-wise Caching vs. Recursive Sharing}
\label{subsec:kv_cache}

Dynamic-depth models often struggle with KV cache consistency during autoregressive decoding. When a token exits early, its corresponding keys and values in deeper layers will be missing, which can be crucial information for subsequent tokens.
Some methods attempt to reuse stale entries~\citep{schuster2022confident} or run parallel decoding~\citep{bae-etal-2023-fast}, but these solutions still introduce overhead and complexity. To this end, we design and explore two KV cache strategies tailored to MoR models: \emph{recursion-wise caching} and \emph{recursive sharing}.

\paragraph{Recursion-wise KV caching. (Figure\,\ref{fig_mor:method_overview-c}--\textit{Top})}

Inspired by Raposo et al. (2024)\,\citep{raposo2024mixture}, we cache KV pairs selectively: only tokens routed to a given recursion step store their key–value entries at that level. Thereby, the KV cache size at each recursion depth is determined exactly by the capacity factor in expert-choice,  or according to actual balancing ratios in token-choice. Attention is then restricted to those locally cached tokens. This design promotes block-local computation, which improves memory efficiency and reduces IO demands.

\paragraph{Recursive KV sharing. (Figure\,\ref{fig_mor:method_overview-c}--\textit{Bottom})}

A key design choice for our MoR model is that all tokens traverse at least the first recursion block\footnote{Though this is not a strict requirement of the MoR framework itself.}. We leverage this by caching KV pairs exclusively at this initial step and reusing them across all subsequent recursions. Therefore, the query length might get shorter at each recursion depth based on the selection capacity, but the key and value lengths will consistently maintain the full sequence. This ensures that all tokens can access to past context without recomputation, despite any distribution mismatch.\looseness=-1

\paragraph{Strengths and limitations. (Table\,\ref{tab_mor:method_comparison}--\textit{Right})}

Recursion-wise caching cuts KV memory and IO to approximately $(N_r + 1)/2N_r$ times across the entire model (when assuming capacity factors follow a sequence like $N_r / N_r$, $\cdots$, $1 / N_r$ over $N_r$ recursion steps). It also reduces per-layer attention FLOPs to a factor of $(k / N_{\text{ctx}})^2$ of those in vanilla models, resulting in substantially improved efficiency for both training and inference phases.
Meanwhile, recursive sharing can yield maximal memory savings by globally reusing context. Speedups can be further achieved by skipping KV projection and prefill at shared depths (compatible only with Cycle strategy~\citep{sun2024you}). However, attention FLOPs only decrease by $k / N_{\text{ctx}}$, and high volume of KV IO still leads to a decoding bottleneck.\looseness=-1

\section{Experiments}

We pretrain our models from scratch using a Llama-based Transformer architecture\footnote{Experiments on Llama are conducted without direction or involvement from Google advisors.}~\citep{grattafiori2024llama}, referring to the configurations of SmolLM open-source models~\citep{allal2024SmolLM}, on a deduplicated subset of the FineWeb-Edu dataset~\citep{penedo2024the} in SmolLM-Corpus~\citep{benallal2024smollmcorpus}.
We evaluate the models on validation set of FineWeb-edu and six few-shot benchmarks~\citep{eval-harness}. Detailed training and evaluation procedures, as well as throughput measurement protocols, are described in Appendix\,\ref{app:experimental_setup}.

\begin{table*}[!t]
    \caption{
    Comparison of MoR, Recursive, and Vanilla Transformers under both fixed FLOPs (16.5e18) and token (20B) settings. All models are trained on FineWeb-Edu and evaluated by validation negative log-likelihood\,(NLL) and few-shot accuracy. For the isoFLOP rows, the number of training tokens\,($N_{tok}$) varies by model efficiency. For the fixed-token rows, we report the effective FLOPs consumed.
    For the model sizes, we report non-embedding parameter counts. For the KV mechanisms, we distinguish between Cache (recursion-wise caching) and Share (recursive sharing). $\text{}^\dagger$In recursive models, all tokens go through fixed recursion depths ($N_r$), instead of adaptive depths.\looseness=-1
    }
    \label{tab_mor:main_results}
    \small
    \centering
    \addtolength{\tabcolsep}{-1pt}
    \resizebox{\textwidth}{!}{
    \begin{tabular}{l|cc|cc|ccc|c|cccccc|c}
    \toprule
      &  \multicolumn{2}{c|}{\textbf{MoR}} &  \multicolumn{2}{c|}{\textbf{Recursion}} & \multicolumn{3}{c|}{\textbf{Pretrain}} & \textbf{NLL\,$\downarrow$} & \multicolumn{7}{c}{\textbf{Few-shot Accuracy\,$\uparrow$}} \\
    \cmidrule(l{2pt}r{2pt}){2-3} \cmidrule(l{2pt}r{2pt}){4-5} \cmidrule(l{2pt}r{2pt}){6-8} \cmidrule(l{2pt}r{2pt}){9-9} \cmidrule(l{2pt}r{2pt}){10-16} 
     \textbf{Models} & Type & KV & Share & $N_R$ & Param & FLOPs & $N_{tok}$ & FineWeb & LD & HS & PQ & WG & ARC & \!MMLU\! & Avg  \\
    \midrule
    Vanilla & - & - & - & -  & 315M &  16.5 & 20B & 2.7824 & 32.0 & 37.8 & 65.6 & 50.5 & 39.6 & 28.0 & 42.3 \\
    \midrule
    \multirow{3}{*}{$\text{Recursive}^{\dagger}$} & - & -  & M-Cyc  & 2 & 167M & 16.5& 20B  & 2.8079 & 31.0 & 37.1 & 66.7 & 52.3 & \textbf{40.8} & 27.5 & 42.6 \\
     & - & - & M-Cyc & 3 & 118M & 16.5 & 20B & 2.8466 & 29.8 & 35.9 & 65.0 & 52.3 & 39.0 & 27.2 & 41.5 \\
     & - & - & M-Cyc & 4 & \,\,\,98M & 16.5& 19B  & 2.8781  & 28.2 & 35.4 & 65.5 & {52.5} & 38.0 & 26.8 & 41.0 \\
     \midrule
     \rowcolor[gray]{0.9}
    \cellcolor{white}& Expert & Cache  & M-Cyc  & 2 & 167M & 16.5 & 27B & \textbf{2.7511} & \textbf{34.4} & \textbf{39.3} & 65.7 & 51.2 & 39.6 & \textbf{28.1} & \textbf{43.1} \\
    \rowcolor[gray]{0.9}
     \cellcolor{white}& Expert & Cache & M-Cyc & 3& 118M & 16.5 & 30B  & 2.7925 & 33.1 & 37.9 & \textbf{66.9} & 52.1 & 38.3 & 27.4 & 42.6 \\
     \rowcolor[gray]{0.9}
     \cellcolor{white}& Expert & Cache & M-Cyc & 4 & \,\,\,98M & 16.5 & 30B & 2.8204 & 30.1 & 37.3 & 65.0 & 51.1 & 38.9 & 27.4 & 41.6 \\
     \cmidrule(l{2pt}r{2pt}){2-16} 
     \rowcolor[gray]{0.9}
     \cellcolor{white}& Expert & Cache  & M-Cyc  & 2 & 167M & 12.3 & 20B & {2.7749} & {33.2} & {38.3} & 65.2 & \textbf{52.6} & 40.1 & \textbf{28.1} & {42.9} \\
      \cellcolor{white}& Expert & Cache & M-Cyc & 3& 118M & 11.0 & 20B  & 2.8246 & 31.9 & 37.0 & 65.7 & 50.5 & 38.3 & 27.4 & 41.8 \\
      \cellcolor{white}& Expert & Cache & M-Cyc & 4 & \,\,\,98M & 11.0 & 20B & 2.8519 & 30.2 & 36.5 & 64.3 & 52.3 & 38.6 & 27.2 & 41.5 \\
     \cmidrule(l{2pt}r{2pt}){2-16} 
     \cellcolor{white}& Token & Cache & M-Cyc & 3& 118M & 16.5  & 30B & 2.9163 & 27.6 & 34.1 & 63.8 & 50.6 & 37.4 & 26.8 & 40.0 \\
     \cmidrule(l{2pt}r{2pt}){2-16} 
     \cellcolor{white}\multirow{-9}{*}{MoR\,(ours)}
     & Expert & Share & M-Cyc & 3 & 118M & 16.5 & 31B & 2.7983 & 31.7 & 37.2 & 65.1 & 51.0 & 39.0 & 27.1  & 41.9 \\
    \bottomrule
    \end{tabular}
    }
\end{table*}

\subsection{Main Results}
\label{subsec:main_results}

\paragraph{MoR outperforms baselines with fewer parameters under equal train compute.}
Under an equal training budget of 16.5e18 FLOPs, we compared our Mixture-of-Recursions (MoR) model against both Vanilla and Recursive Transformers. As shown in Table\,\ref{tab_mor:main_results}, the MoR model, using an expert-choice router and two recursions, achieves a lower validation loss and surpasses the vanilla baseline in average few-shot accuracy (43.1\% vs. 42.3\%). Remarkably, this superior performance is achieved despite using nearly 50\% fewer parameters. This is attributed to MoR's higher computational efficiency, which allows it to process more training tokens within the same FLOPs budget. Furthermore, as $N_r$ increases to 3 or 4, MoR maintains its competitive accuracy, consistently outperforming the recursive baselines while remaining within a tight margin of the full-capacity vanilla model.\looseness=-1

\vspace{-5pt}
\paragraph{MoR outperforms baselines with less compute at equal data.}

To isolate architectural differences, we analyze performance under a fixed number of training tokens (20B). Specifically, our MoR model with $N_r=2$ outperforms both vanilla and recursive baselines—achieving lower validation loss and higher accuracy—despite using 25\% fewer training FLOPs. This theoretical efficiency translates into significant practical gains: compared to the vanilla baseline, our model reduces training time by 19\% and cuts peak memory usage by 25\%. These improvements stem from our hierarchical filtering and recursion-wise attention mechanism, which shortens sequence lengths to achieve a superior compute-accuracy trade-off, even during pretraining.

\vspace{-5pt}
\paragraph{MoR performance varies with routing and caching strategies.}

We also evaluate a few design variants within MoR framework, specifically with $N_r=3$ that is lightweight and still comparable with Vanilla. In this case, using token-choice routing yields lower performance (40.0\%) compared to expert-choice routing (42.6\%), indicating that routing granularity plays a pivotal role in model performance. Additionally, applying KV cache sharing slightly reduces performance compared to independent caching, while providing improved memory efficiency. This trade-off remains favorable for practical deployment when memory usage is a key concern.

\begin{figure}[t]
    \centering
    \begin{subfigure}[t]{\textwidth}
        \centering
        \captionsetup{justification=centering}
        \includegraphics[width=0.57\textwidth]{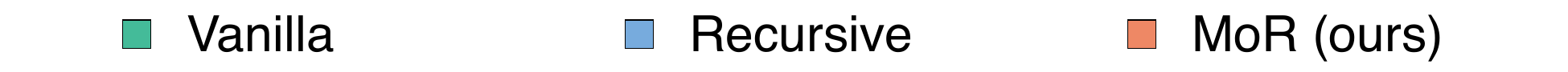}
    \end{subfigure}
    \centering
    \begin{subfigure}[t]{0.265\textwidth}
    \captionsetup{justification=centering}
        \includegraphics[width=\textwidth]{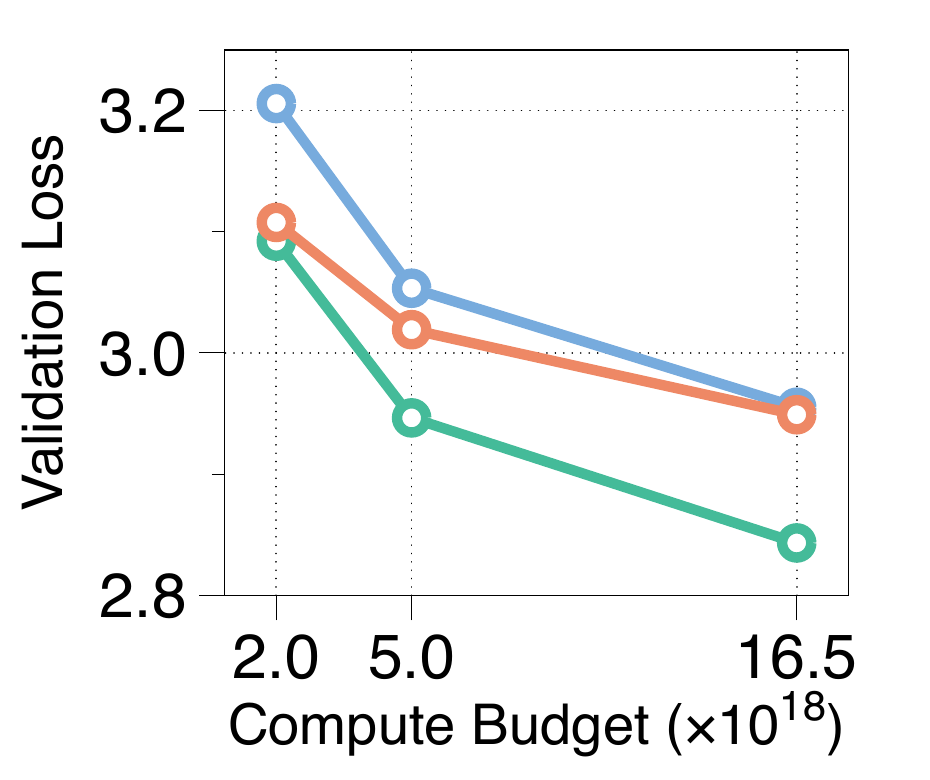}
        \subcaption{135M-based model}
    \end{subfigure}
    \hspace{-8pt}
    \centering
    \begin{subfigure}[t]{0.24\textwidth}
    \captionsetup{justification=centering}
        \includegraphics[width=\textwidth]{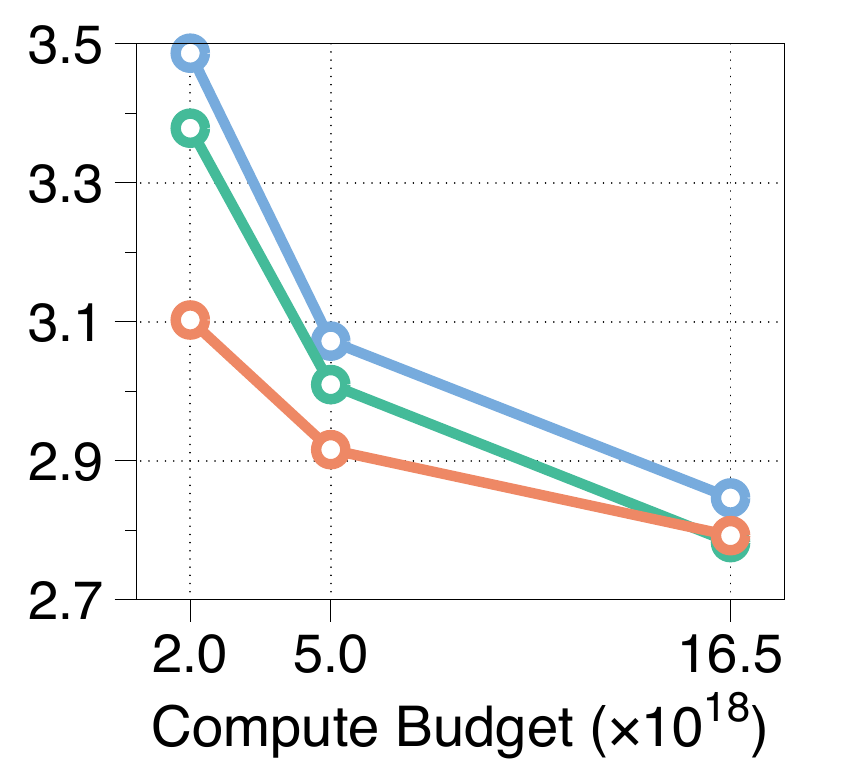}
        \subcaption{360M-based model}
    \end{subfigure}
    \hspace{-8pt}
    \centering
    \begin{subfigure}[t]{0.24\textwidth}
    \captionsetup{justification=centering}
        \includegraphics[width=\textwidth]{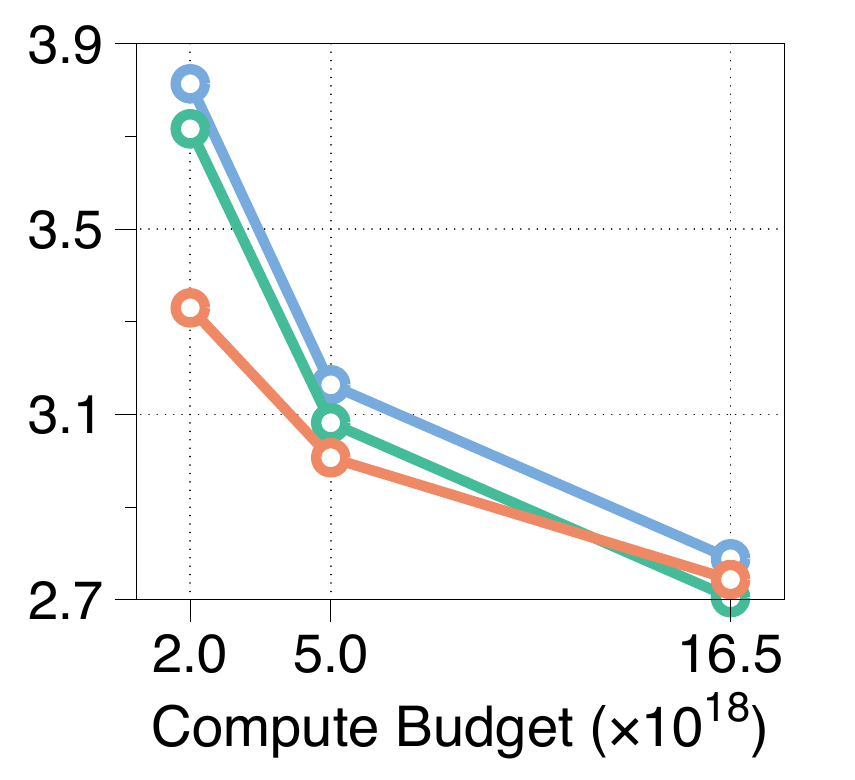}
        \subcaption{730M-based model}
    \end{subfigure}
    \hspace{-8pt}
    \centering
    \begin{subfigure}[t]{0.24\textwidth}
    \captionsetup{justification=centering}
        \includegraphics[width=\textwidth]{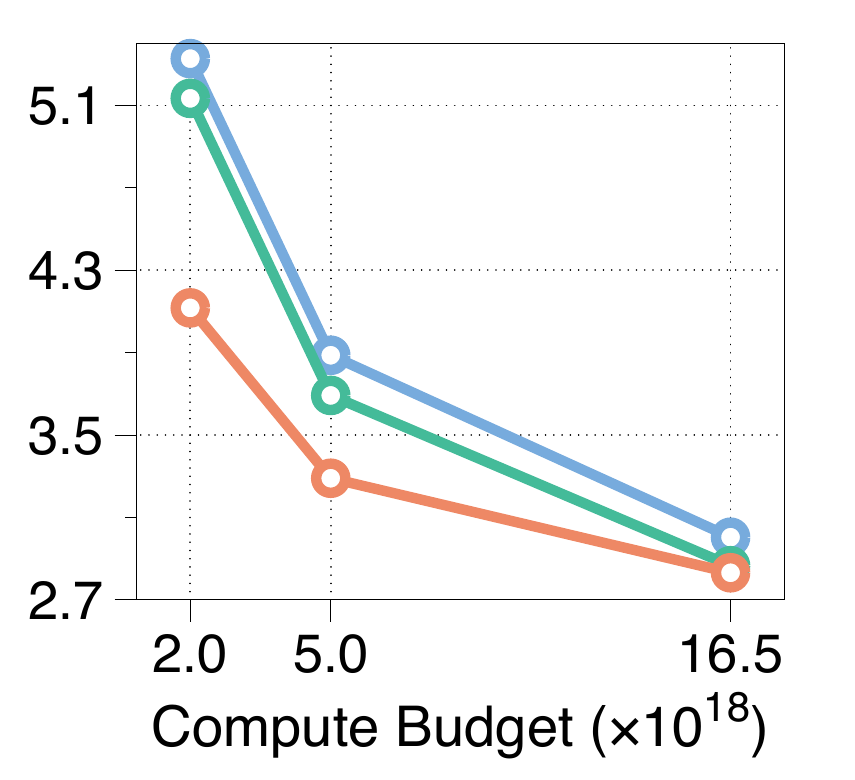}
        \subcaption{1.7B-based model}
    \end{subfigure}
    \caption{
    Validation loss across different compute budgets across four model sizes: 135M, 360M, 730M, and 1.7B parameters. For MoR models, we use expert-choice routing and recursion-wise caching. MoR consistently outperforms recursive baselines and matches or exceeds the standard Transformers at larger scales, despite using significantly fewer parameters (approximately one-third due to layer tying with $N_R=3$).\looseness=-1
    }
    \label{fig_mor:scaling_laws}
\end{figure}

\subsection{IsoFLOP Analysis}
\label{subsec:scaling_laws}

A core criterion for evaluating a new model architectural design is whether performance continues to improve as model and compute scales grow~\citep{kaplan2020scaling}. Therefore, we evaluate MoR against both Vanilla and Recursive Transformers across a wide range of model sizes and computational budgets to show that it maintains competitive or superior predictive performance as the scale increases.

\vspace{-5pt}
\paragraph{Experimental Setup.}

We experiment with four scales (135M, 360M, 730M, and 1.7B parameters) fixing the number of recursions to three for both Recursive and MoR configurations, resulting in roughly one-third the number of unique parameters. Each model is pretrained under three FLOPs budgets: 2e18, 5e18, and 16.5e18.\looseness=-1

\vspace{-5pt}
\paragraph{MoR is a scalable and parameter-efficient architecture.}

As shown in Figure\,\ref{fig_mor:scaling_laws}, MoR consistently outperforms recursive baselines across all model sizes and compute budgets. While it underperforms the vanilla model at the smallest model size (135M)—likely due to a recursive capacity bottleneck—this gap closes rapidly at scale. For >360M parameters, MoR not only matches but often exceeds the Vanilla Transformer, particularly under low and mid-range budgets. Overall, these results highlight that MoR is a scalable and efficient alternative to standard Transformers. It achieves strong validation performance with significantly lower parameter counts, making it a strong candidate for both pretraining and large‑scale deployment. Further details are presented in Appendix\,\ref{app:scaling_laws}.\looseness=-1

\subsection{Inference Throughput Evaluation}
\label{subsec:efficiency}

As a parameter-shared architecture, MoR can leverage continuous depth-wise batching~\citep{bae2024relaxed} to dramatically boost inference throughput compared to Vanilla Transformers. This maintains high and consistent GPU utilization by immediately replacing completed sequences with incoming tokens during decoding. The early-exiting mechanism in MoR further eliminates bubbles in the computational batch.\looseness=-1

\vspace{-5pt}
\paragraph{Experimental Setup.}

We measure throughput for 360M scale-based MoR models with recursion depths of 2, 3, and 4, trained under a 16.5e18 FLOPs budget. Throughput (tokens/second) is measured based on the generation time for tokens per sample, where the number of tokens is sampled from a normal distribution with a mean of 256, starting without any input prefix. We examine two batching configurations: a fixed batch size of 32 and a (relative) maximum batch size derived by multiplying 32 by the ratio of the maximum batch sizes of vanilla and MoR models. Further details on the experimental setup are provided in Appendix\,\ref{app:throughput}.\looseness=-1

\begin{figure}[t!]
    \centering
    \begin{subfigure}[t]{0.325\textwidth}
        \includegraphics[width=\textwidth]{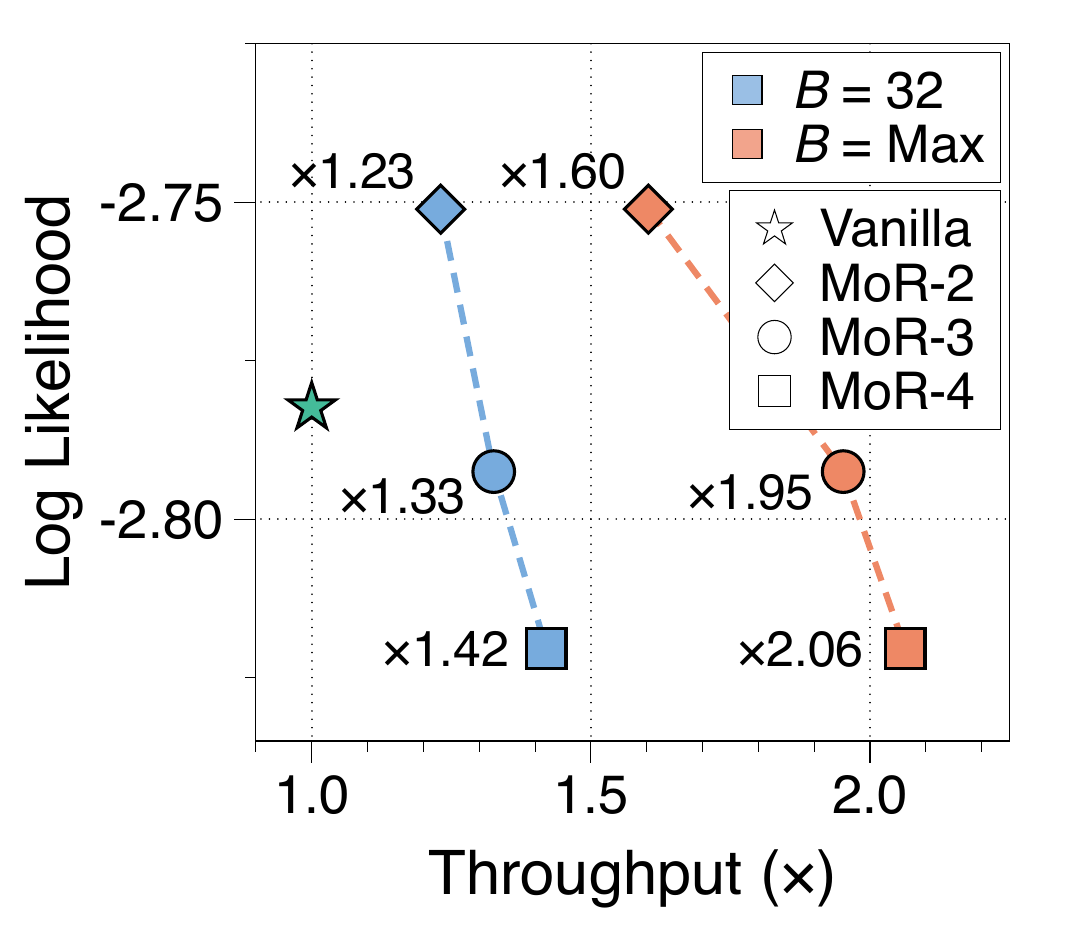}
        \captionsetup{justification=centering}
        \subcaption{Pareto frontier of throughput}
        \label{fig_mor:pareto_frontier_sharing-a}
    \end{subfigure}
    \centering
    \begin{subfigure}[t]{0.325\textwidth}
        \includegraphics[width=\textwidth]{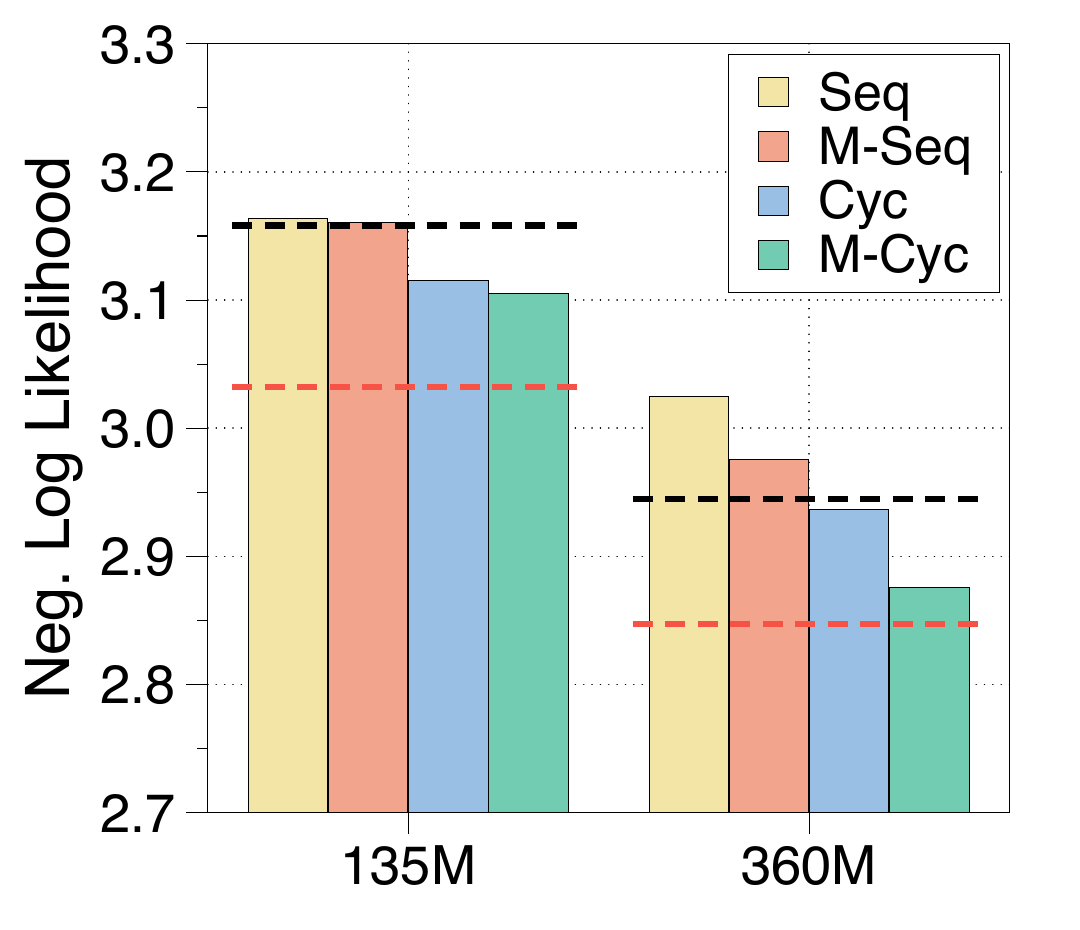}
        \captionsetup{justification=centering}
        \subcaption{Sharing strategy}
        \label{fig_mor:pareto_frontier_sharing-b}
    \end{subfigure}
    \begin{subfigure}[t]{0.325\textwidth}
        \includegraphics[width=\textwidth]{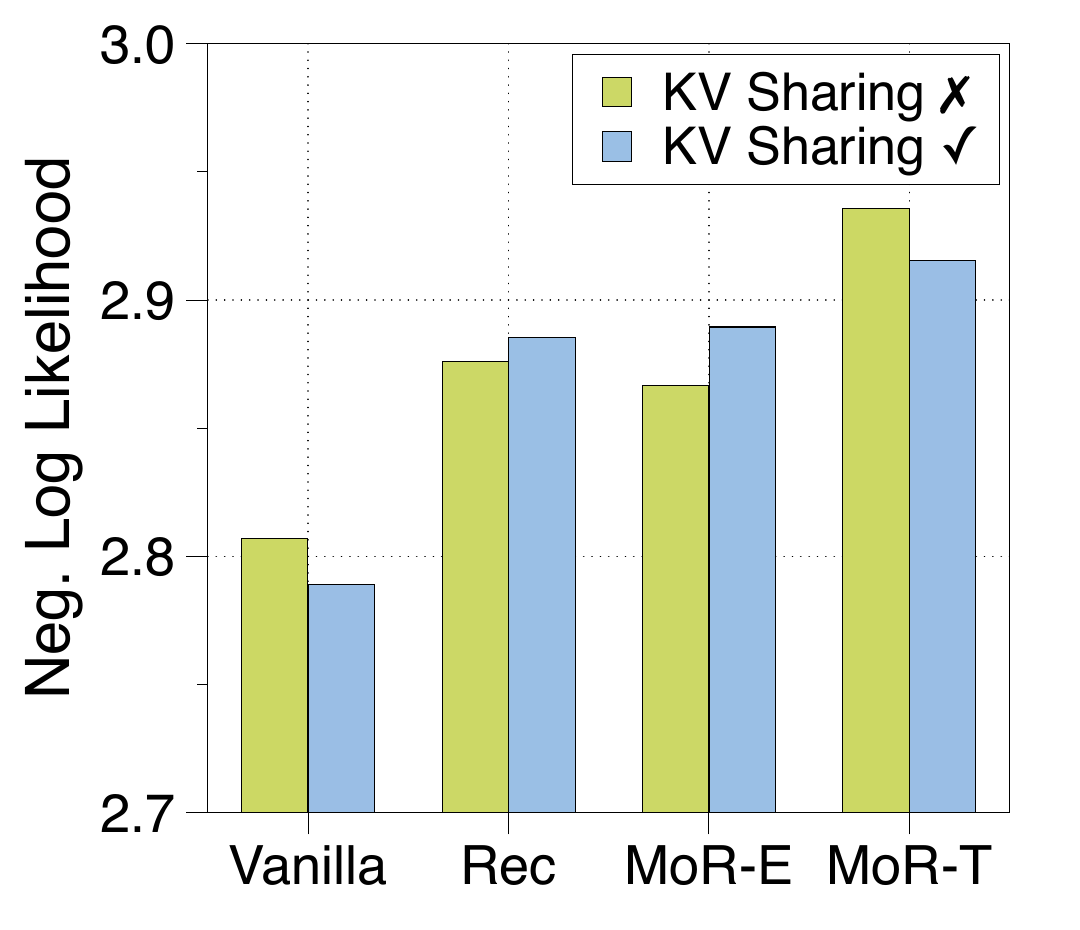}
        \captionsetup{justification=centering}
        \subcaption{KV cache sharing}
        \label{fig_mor:pareto_frontier_sharing-c}
    \end{subfigure}
    \centering
    \caption{
    (a) Pareto frontier of inference throughput and log-likehood for MoR and Vanilla Transformer under fixed and maximum batching scenarios. Setting details are in Appendix\,\ref{app:throughput}.
    (b) Negative log-likelihood (NLL) of Recursive Transformers with $N_r=3$ across four different parameter-sharing strategies. We pretrained the models on 10 billion tokens. The dashed red and black lines denote the full-size Vanilla Transformer and parameter-matched vanilla models (approximately one-third scales), respectively.
    (c) NLL performance comparison across four different architectures with KV sharing. 
    For MoR, green (disabled) and blue (enabled) refer to recursion-wise KV caching and recursive KV sharing strategies. MoR-E and MoR-T denotes expert-choice and token-choice MoR, respectively. All models are based on 360M scale and trained on 10 billion tokens.\looseness=-1
    }    
    \label{fig_mor:pareto_frontier_sharing}
\end{figure}

\vspace{-5pt}
\paragraph{MoR boosts inference throughput with continuous depth-wise batching.}

In Figure\,\ref{fig_mor:pareto_frontier_sharing-a}, across both batch settings, all MoR variants outperform the vanilla baseline, which even leverages continuous sequence-wise batching~\citep{yu2022orca, kwon2023efficient}. Increasing recursion depth leads to more tokens exiting early and a further reduction in KV cache usage. This, in turn, boosts throughput significantly (e.g., MoR-4 achieves up to a 2.06$\times$ speedup with $B=\text{Max}$). While there's a slight performance degradation, it can be a favorable trade-off given the substantial throughput gain. These results support that the integration of the depth-wise batching paradigm with early-exiting can significantly accelerate MoR's actual deployment throughput.\looseness=-1

\section{Ablation Studies}

\subsection{Parameter Sharing Strategies}
\label{subsec:sharing}

\paragraph{Middle-Cycle is the most effective parameter sharing strategy.}

As discussed in \S\ref{subsec:preliminary}, parameter sharing is a key component of Recursive Transformers and MoR. To identify the most effective sharing configuration, we empirically compare four aforementioned strategies: Cycle, Sequence, Middle-Cycle, and Middle-Sequence. We evaluate each strategy on Recursive Transformers based on 135M and 360M model sizes. As shown in Figure\,\ref{fig_mor:pareto_frontier_sharing-b}, ``Middle-Cycle'' consistently achieves the lowest validation loss, and its superiority is further confirmed by the detailed results in Appendix\,\ref{app:parameter_sharing}. Based on these findings, we adopt the ``Middle-Cycle'' configuration for all subsequent MoR and Recursive Transformers presented in this paper.

\subsection{Routing Strategies}
\label{subsec:ablation_routing}

We conduct an extensive ablation study to understand the impact of various design choices within the expert-choice and token-choice routing schemes in our MoR framework. Detailed results are summarized in Appendix\,\ref{app:router_ablation}.

\vspace{-5pt}
\paragraph{In expert-choice routing, auxiliary loss and linear router yield the best performance.}

For the \textit{expert-choice} routing setup (\textit{Left} of Table\,\ref{tab_mor:ablation_study_router}), we evaluate several design aspects: solution to mitigate causality violation (auxiliary router vs. auxiliary loss), normalization functions (\texttt{sigmoid} vs. \texttt{tanh}), router architectures (MLP, Linear, or Wide-MLP), and the impact of an auxiliary z-loss~\citep{zoph2022st}. To assess how well the router performs dynamic allocation, we measure the proportion of ``dead'' tokens—those never selected by the final recursion in a batch—on the validation dataset. Our key findings are as follows: First, using an auxiliary loss is more effective for inference-time behavior than training a separate auxiliary router. Second, a \texttt{sigmoid} normalization function and a simple linear router architecture yield the best performance. Finally, the auxiliary z-loss has a negligible impact on accuracy, though it does slightly reduce the proportion of dead tokens.\looseness=-1

\vspace{-5pt}
\paragraph{In token-choice routing, balancing loss yields stable and accurate routing.}
For \textit{token-choice} routing (\textit{Right} of Table\,\ref{tab_mor:ablation_study_router}), we follow common MoE practices and enable z-loss by default. We compare two balancing strategies: using a balancing loss and training in a loss-free manner using router bias. While both approaches achieve similar log-probability and few-shot accuracy, the explicit balancing loss yields a significantly lower MaxVio~\citep{wang2024auxiliary} in our MoR architectures, making it the preferable choice for stable routing. However, despite this, the model often struggles to balance loads among its heterogeneous experts, even for nearly half of the training steps. Softmax activation with an MLP router performs best, and removing z-loss—though we add back with a very small coefficient in the final design—results in higher performance and routing stability.\looseness=-1

\begin{table*}[t!]
    \caption{
    Ablation results for expert-choice (\textit{Left}) and token-choice (\textit{Right}) routers with various design choices. We use MoR models that apply three recursions to a 360M model with recursion-wise caching. Model performance is measured by NLL and average few-shot accuracy. We evaluate router metrics—dead token ratio (for expert-choice) and MaxVio (for token-choice)—on the validation set. The dead token ratio denotes the proportion of tokens that remain unselected during the last recursion step, measured on 500 samples, each with 2K sequence length. The selected design choice is highlighted in gray.\looseness=-1
    }
    \label{tab_mor:ablation_study_router}
    \small
    \begin{subtable}[t]{0.5\textwidth}
    \label{tab_mor:expert_choice_router}
    \centering
    \resizebox{\linewidth}{!}{
    \setlength{\tabcolsep}{2.5pt}
    \begin{tabular}{lccc|c|c|c}
    \toprule
      \multicolumn{4}{c|}{\textbf{Expert-choice Router}} &  \multicolumn{3}{c}{\textbf{Performance\,($\downarrow$ / $\downarrow$ / $\uparrow$)}} \\
    \cmidrule(l{2pt}r{2pt}){1-4} \cmidrule(l{2pt}r{2pt}){5-7}
    Sampling & Func & Arch & z-loss & Dead & NLL & Few-shot  \\
    \midrule
     Aux\,Rout & $\sigma$ & MLP & \xmark & \,0.0 & 2.8893 & 39.4 \\
     Aux\,Rout & \,\texttt{tanh}\! & MLP & \xmark  & \!66.7 & 2.8720 & 36.2 \\
     \midrule
     Aux\,Loss & $\sigma$ & MLP & \xmark  & \,0.0 & 2.8816 & 40.0 \\
     Aux\,Loss & \,\texttt{tanh}\! & MLP & \xmark  & 0.0 & 2.9933 & 38.8 \\
     \midrule
     \rowcolor[gray]{0.9}
     Aux\,Loss & $\sigma$ & Linear & \xmark  & \,0.1 & \textbf{2.8667} & \textbf{40.1} \\
     Aux\,Loss & $\sigma$ & W-MLP & \xmark  & \,0.4 & 2.8716 & 39.4 \\
     \midrule
     Aux\,Loss & $\sigma$ & Linear & \cmark  & \,0.0 & 2.8824  & 40.0 \\
    \bottomrule
    \end{tabular}
    }
    \end{subtable}
    \hfill
    \begin{subtable}[t]{0.5\textwidth}
    \label{tab_mor:token_choice_router}
    \renewcommand{\arraystretch}{1.04}
    \centering
    \resizebox{\linewidth}{!}{
    \setlength{\tabcolsep}{2.5pt}
    \begin{tabular}{lccc|c|c|c}
    \toprule
      \multicolumn{4}{c|}{\textbf{Token-choice Router}} &  \multicolumn{3}{c}{\textbf{Performance\,($\downarrow$ / $\downarrow$ / $\uparrow$)}} \\
    \cmidrule(l{2pt}r{2pt}){1-4} \cmidrule(l{2pt}r{2pt}){5-7}
    Balancing & Func\!\!\! & Arch & z-loss & M-Vio & NLL & Few-shot  \\
    \midrule
     Loss\,(0.1) & \texttt{soft} & MLP & \cmark & 0.200 & 3.0239 & 38.5  \\
     Loss\,(0.01) & \texttt{soft} & MLP & \cmark & 0.682 & 2.9118 & 39.4 \\
     \midrule
     Loss-free & \texttt{soft} & MLP & \cmark & 0.852 & 2.9081 & 39.4 \\
     Loss-free & $\sigma$ & MLP & \cmark & 1.281  & 3.0188 & 37.6 \\
     \midrule
     Loss\,(0.1) & \texttt{soft} & Linear & \cmark & 0.492 &  2.9974 & 38.4 \\
     Loss\,(0.1) & \texttt{soft} & W-MLP & \cmark & 0.384 & 3.0293 & 38.8 \\
     \midrule
     \rowcolor[gray]{0.9}
     Loss\,(0.1) & \texttt{soft} & Linear & \xmark & 0.266 &  \textbf{2.9358} & \textbf{39.1} \\
    \bottomrule
    \end{tabular}
    }
    \end{subtable}
\end{table*}

\subsection{KV Caching Strategies}
\label{subsec:ablation_kv_cache}

\paragraph{KV sharing robustly works even in parameter-shared architectures.}

In Figure\,\ref{fig_mor:pareto_frontier_sharing-c}, we first investigate the effect of KV sharing in Vanilla and Recursive Transformers. As consistent with prior works~\citep{brandon2024reducing, wu2024layer, sun2024you}, if we pretrain models from the scratch, KV sharing does not often compromise performance due to the greater parameter flexibility. Surprisingly, the Recursive Transformer remains relatively robust to KV sharing, despite its reduced degrees of freedom. We found evidence for this by decomposing the KV pairs at each recursion depth into their magnitude and direction. As detailed in Appendix\,\ref{app:expanded_kv}, depths that share parameters exhibit highly consistent magnitude patterns and high cosine similarity, providing a clear justification for why KV sharing results in only a slight performance drop.

\vspace{-5pt}
\paragraph{KV sharing degrades expert-choice but benefits token-choice routing in MoR.}

We compare recursion-wise KV caching and recursive KV sharing mechanisms in our MoR framework. 
We observe that while recursive KV sharing offers the advantages of reduced memory footprint and overall FLOPs\footnote{Although attention FLOPs increase by $N_{\text{ctx}}/k$ than recursion-wise KV caching, reduced KV projection FLOPs lead to an overall reduction.\looseness=-1}, it leads to quite large performance degradation in expert-choice routing under a fixed token setting. This suggest that exclusively updating and attending to the tokens active in that recursion depth may be more beneficial. Conversely, MoR with token-choice routing could benefit from KV sharing, where its weaker, inaccurate routing decisions can be complemented by the additional contextual information provided by shared KV pairs.

\section{Analysis}

\subsection{Compute-optimal Scaling Analysis}
\label{subsec:scaling_analysis}

\begin{figure}[t!]
    \centering
    \begin{subfigure}[t]{0.32\textwidth}
        \vspace{0pt}
        \includegraphics[width=\textwidth]{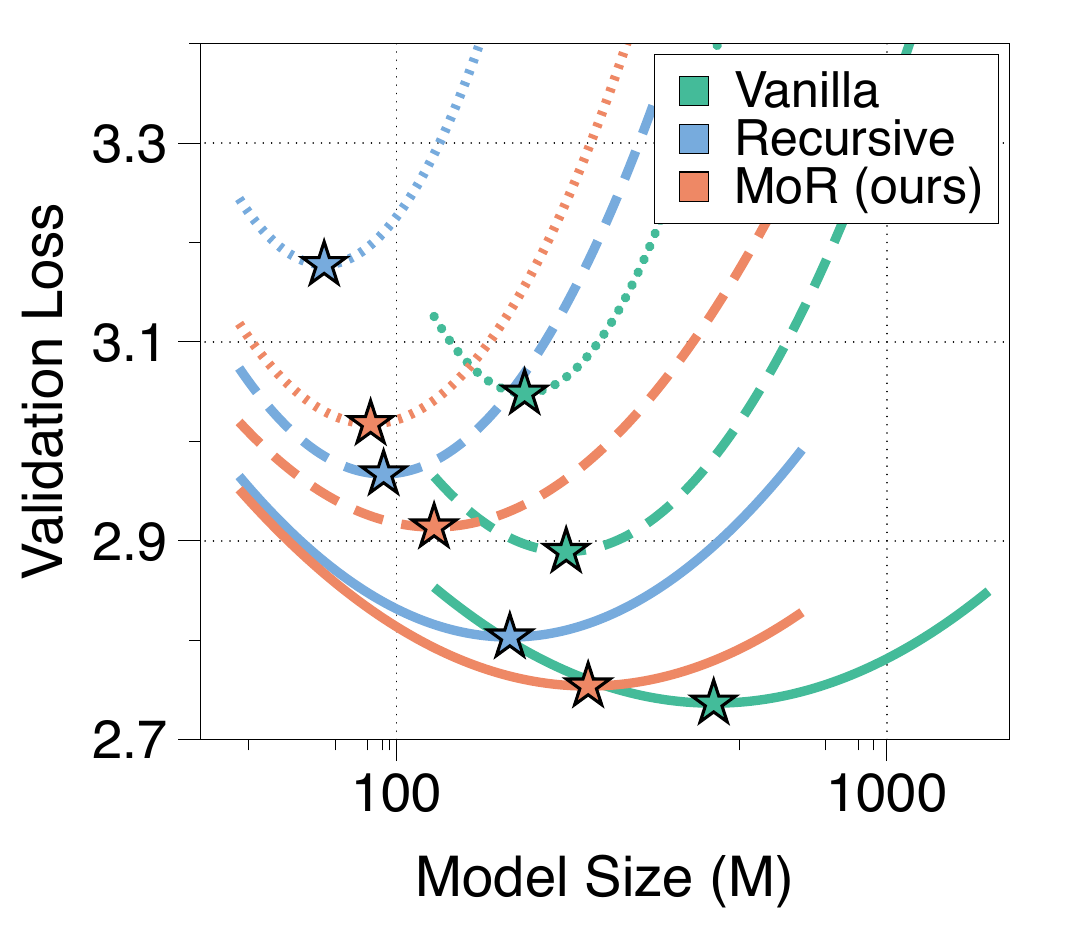}
            \captionsetup{justification=centering}
        \subcaption{Compute-optimal scaling}
        \label{fig_mor:isoflops_qualitative-a}
    \end{subfigure}
    \hfill
    \begin{subfigure}[t]{0.385\textwidth}
        \vspace{-4pt}
        \includegraphics[width=\textwidth]{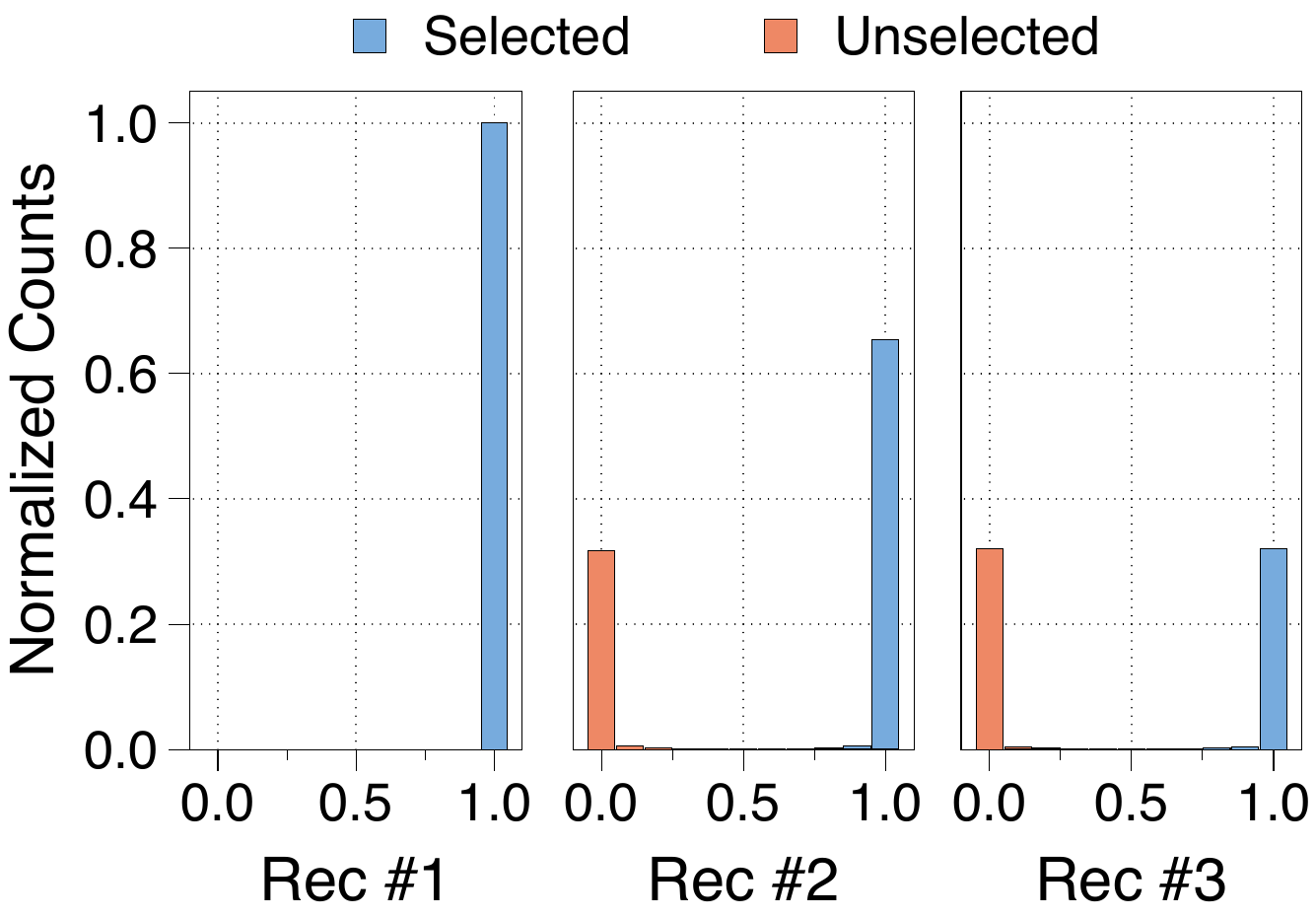}
        \vspace{-4pt}
            \captionsetup{justification=centering}
        \subcaption{Learned router scores}
        \label{fig_mor:isoflops_qualitative-b}
    \end{subfigure}
    \hfill
    \begin{subfigure}[t]{0.275\textwidth}
        \vspace{0pt}
        \includegraphics[width=\textwidth]{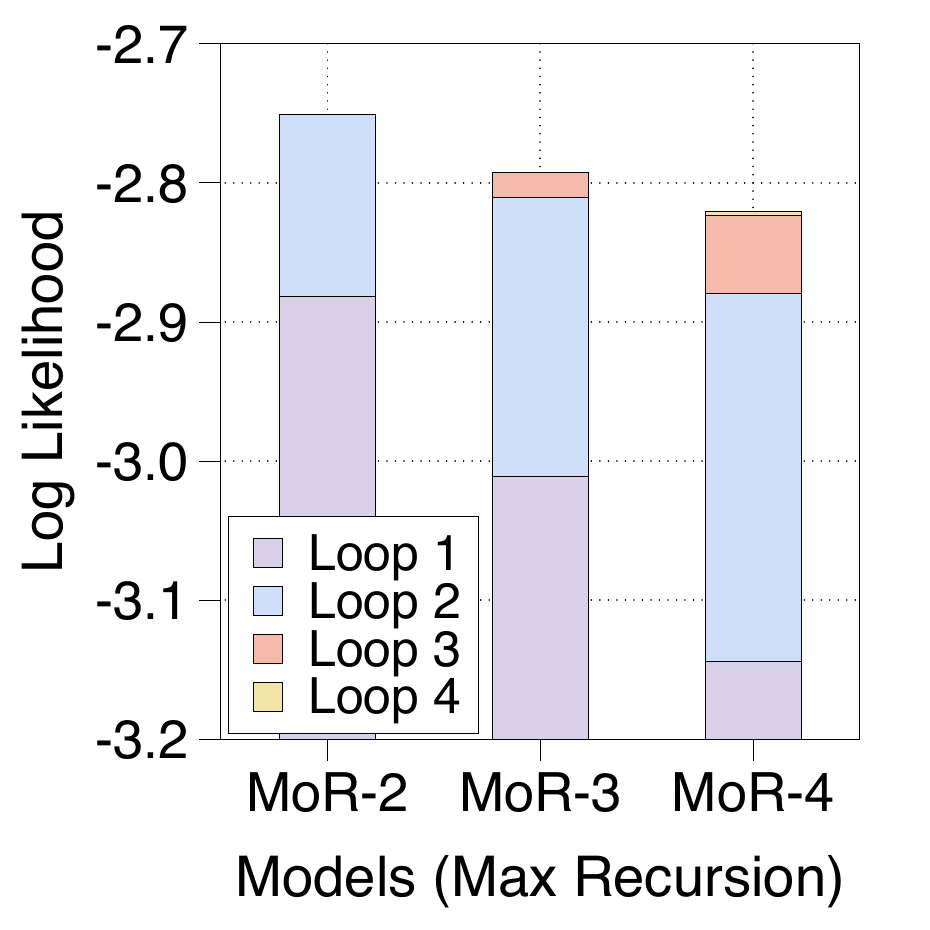}
            \captionsetup{justification=centering}
        \subcaption{Test-time scaling}
        \label{fig_mor:isoflops_qualitative-c}
    \end{subfigure}
    \caption{
    (a) Compute-optimal scaling analysis for three model architectures. Each star indicates the optimal model size for a given compute budget. We visualize the results in \S\ref{subsec:scaling_laws} by fitting polynomial functions for each architecture and FLOPs budget, and derive the optimal points from these fits. (b) Distribution of router outputs for selected and unselected tokens at each recursion step. As an example, a 360M size-based MoR model with $N_r=3$, expert-choice router with auxiliary loss, and recursion-wise caching, is used. (c) Test-time scaling analysis illustrating the cumulative log-likelihood improvement with increasing recursion depth, measured over 500 samples. As we increase $N_r$ based on a 360M model size, the number of unique parameters in MoR decreases, resulting in a gradual decline in overall performance (i.e., a decrease in log-likelihood). All models are trained by an expert-choice router with auxiliary loss and a recursion-wise caching mechanism.\looseness=-1
    }    
    \label{fig_mor:isoflops_qualitative}
\end{figure}

\paragraph{MoR scaling favors model size over training length under isoFLOPs.}

As illustrated in Figure\,\ref{fig_mor:isoflops_qualitative-a}, MoR exhibits a distinct compute-optimal scaling behavior compared to baselines under isoFLOPs constraints. The flatter slope of MoR's optimal path (a line connecting stars) indicates that it benefits more significantly from increases in parameter count (i.e., less data-hungry). This is likely because the performance of the shared parameter block itself becomes important, even more than feeding in additional data. Therefore, the optimal scaling policy for MoR models favors allocating resources to increasing model capacity by using larger models trained for shorter steps.\looseness=-1

\subsection{Routing Analysis}

\paragraph{The allocation of recursion depth reflects contextual predictability of the subsequent token.}

In the \textit{Right} of Figure\,\ref{fig_mor:mor_overview}, we illustrate each token's recursion depth, which directly indicates how easily the next token can be predicted within the given context. For example, the second subword part (e.g., ``-ensively'') of a word is often straightforward to predict, thus needing fewer steps. As also shown in Table\,\ref{tab_mor:qualitative_highlighted_results}, while the initial generation (opening) of function words like ``{-}{-}{-}'', ``('', ``.'', and ``,'' appears easy for models, predicting their closing parts or the first token immediately following their opening might be more challenging.\looseness=-1

\paragraph{Expert-choice router with auxiliary loss perfectly separates selected from unselected tokens.}

Figure\,\ref{fig_mor:isoflops_qualitative-b} visualizes one example of the expert-choice router output distribution at each recursion step in a MoR model with $N_r = 3$. For each recursion step, the normalized counts of routing scores are plotted, distinguishing between tokens selected by the expert (blue) and those not selected (orange). In all steps, auxiliary loss achieves a perfect separation in router outputs, with selected tokens sharply concentrated near a routing score of 1.0 and unselected tokens clustering near 0.0. Router output distributions for the two routing strategies and their associated design choices are detailed in Appendix\,\ref{app:qualitative_results}.\looseness=-1

\subsection{Test-time Scaling Analysis}

\paragraph{MoR enables test-time scaling via deeper recursion.}

We visualize how log-likelihood evolves across recursion steps in MoR models with $N_r = \{2,3,4\}$ in Figure\,\ref{fig_mor:isoflops_qualitative-c}. The overlaid bars illustrate the performance of each model when the maximum thinking (recursion) depth of tokens gradually increases. This suggests that deeper recursion not only provides additional compute but also enables each subsequent step to specialize further in refining the token representation or ``thought process'' at its particular depth, leading to better performance. Thereby, these results support the view that MoR enables test-time scaling: allocating more recursion steps at inference can improve generation quality.

\clearpage
\section{Conclusion}

Mixture-of-Recursions (MoR) presents a unified Transformer architecture that simultaneously leverages parameter sharing, adaptive recursion depth, and efficient KV caching without compromising model quality. 
By dynamically assigning recursion depth to tokens via lightweight routers and selectively caching key-value states for selected tokens, MoR reduces both quadratic attention computation and redundant memory access costs.
Extensive empirical evaluations show that MoR lowers validation perplexity and improves average few-shot accuracy compared to both vanilla and previous recursive baselines, even with higher inference throughput.
These results demonstrate that MoR offers an effective path towards achieving large-model capabilities with significantly reduced computational and memory overhead.\looseness=-1

\subsection{Limitations and Future Works}

\paragraph{Reasoning MoR models.}
Recent studies have highlighted the redundancy within reasoning chains and address it by applying token-level adaptive computation, like early-exit mechanisms~\citep{yang2025dynamic, jiang2025flashthink, dai2025s}. Our MoR framework inherently enables latent reasoning by adaptively determining the necessary recursion depth for individual tokens. Therefore, a crucial future work involves exploring how the router can dynamically learn to adjust to the necessity of chain-of-thought (CoT) chains when post-trained on actual reasoning datasets. Developing advanced routing strategies that explicitly align recursion depth with reasoning complexity may enhance reasoning accuracy, computational efficiency, and even interpretability for deliberative reasoning process.

\vspace{-5pt}
\paragraph{Further scaling model family.}

Our experiments have been limited to models with up to 1.7 billion parameters due to compute constraints. The natural next step is to train MoR models at larger scales (over 3 billion parameters) on substantially larger corpora. To reduce overall pre-training costs, we could also explore continued pre-training (i.e., uptraining), starting from existing pre-trained vanilla LLM checkpoints. As future work, we plan to investigate MoR performance using various initialization strategies for recursive models, as explored in prior work~\citep{bae2024relaxed}. Additionally, to ensure a fair scalability comparison, we need to account for potential performance degradation in Recursive Transformers during post-training for early-exiting~\citep{bae2024relaxed} and incorporate inference throughput constraints for Vanilla Transformers, which further highlights the advantages of MoR.

\vspace{-5pt}
\paragraph{Adaptive capacity control.}
Expert-choice routing offers the significant advantage of guaranteeing perfect load balancing through pre-determined capacity factors~\citep{raposo2024mixture, zhou2022mixture}. However, a limitation arises when we want to allocate different capacities during inference. Specifically, in our MoR models, we observe that when using an auxiliary loss, the router outputs for selected and unselected tokens are almost perfectly separated. This makes it challenging to adjust top-k values after training. Therefore, a more adaptive model design, which can leverage different capacities during both training and inference phases, is needed to address this limitation.

\vspace{-5pt}
\paragraph{Compatibility with sparse algorithms.}

Given MoR's token-level adaptive recursion, we can further optimize computation by integrating structured sparsity. This approach allows for the selective activation of subnetworks or parameters~\citep{liu2023deja}, dynamically pruning unnecessary computations at both the token and layer levels~\citep{raposo2024mixture, elhoushi2024layerskip}. This investigation into sparse model designs promises significant efficiency improvements. We believe many sparsity-based techniques, such as pruning~\citep{han2015learning} or quantization~\citep{jacob2018quantization}, are highly complementary to our MoR framework. This will provide deeper insights into effective sparse architectures within recursive models, offering promising directions for future research.

\vspace{-5pt}
\paragraph{Expansion to multimodal and non‑text domains.} 

MoR's recursion block is inherently modality-agnostic, allowing its adaptive depth mechanism to extend beyond text processing. This crucial property enables MoR to readily integrate into vision, speech, and unified multimodal transformer architectures. Applying token-adaptive recursion to long-context video or audio streams holds the potential for even greater memory efficiencies and substantial throughput gains, crucial for real-world applications. By dynamically adjusting the processing depth for each token or segment, MoR could unlock these significant benefits.\looseness=-1

\clearpage
\section{Appendix}

\subsection{Design Choices for Parameter-sharing Strategy}
\label{app:sharing_strategy}

Table\,\ref{tab_mor:sharing_strategy_middle} shows formulation and visualization of four parameter-sharing strategies: Cycle, Middle-Cycle, Sequence, and Middle-Sequence. These strategies determine how a shared pool of blocks $\Phi'$ are reused across a total of $L$ unrolled layers. The optimal strategy for parameter sharing in recursive models remains an open question.

In the \textbf{Cycle} strategy, a fixed set of parameters is reused cyclically across all recursion steps. By forcing the model to re-engage with the input through the same shared block, it encourages a deeper, iterative refinement process, akin to ``rethinking'' the problem from the ground up at every stage.
However, because the same transformations are applied repeatedly regardless of input variation, it may limit the model’s capacity to learn diverse or highly specialized features.

On the other hand, the \textbf{Sequence} strategy assigns distinct parameters to each recursion block in sequential order. A potential drawback is that simply applying similar transformations twice in a row may lead to redundant features with diminishing returns. Nevertheless, the use of a fixed, sequential order of layers may provide a stable and predictable structure.

Building upon these strategies, the \textbf{Middle} sharing variant further refines parameter reuse by preserving unique parameters at the first and last layers while sharing weights only among the intermediate layers. This approach aims to balance the trade-off between parameter efficiency and representational flexibility, maintaining distinct entry and exit transformations while benefiting from reduced parameter redundancy in the middle layers. In line with recent findings~\citep{kim2023solar, geiping2025scaling}, Middle sharing can capture important input and output nuances more effectively than pure Cycle or Sequence sharing, without significantly increasing model size.

\newcolumntype{M}[1]{>{\centering\arraybackslash}m{#1}}

\begin{table*}[h]
\centering
\caption{
Parameter-sharing strategies in Recursive Transformers. This table shows \textit{Cycle}, \textit{Middle-Cycle}, \textit{Sequence}, and \textit{Middle-Sequence} schemes with layer reuse, where Middle-* retains unique first and last layers.
}
\label{tab_mor:sharing_strategy_middle}
\renewcommand{\arraystretch}{1.}
\resizebox{\textwidth}{!}{%
\setlength{\tabcolsep}{7pt}
\begin{tabular}{
    l
  | M{0.32\textwidth} 
  | M{0.16\textwidth} 
  | M{0.32\textwidth} 
  | M{0.16\textwidth}  
}
\toprule
& \multicolumn{2}{c|}{\textbf{Cycle Strategy}} & \multicolumn{2}{c}{\textbf{Middle-Cycle Strategy}} \\
\cmidrule(l{2pt}r{2pt}){2-3}\cmidrule(l{2pt}r{2pt}){4-5}
\textbf{Layers} & Equation & Figure & Equation & Figure \\
\midrule
Last 
& -- 
& 
& $f\bigl(\mathbf{h}_t^{L-1}; \Phi_{L-1}\bigr)$ 
& \\
[10pt]
Recursion 
& $f\!\Bigl(\mathbf{h}_t^{\ell}; \Phi'_{\ell \bmod (L/N_r)}\Bigr)$ 
& \multirow{-2}{*}{\includegraphics[width=0.91\linewidth]{fig_mor/sharing_strategy/sharing_strategy_cycle.pdf}}
& \!\!\!$f\!\Bigl(\mathbf{h}_t^{\ell}; \Phi'_{(\ell - 1 \bmod ((L-2)/N_r)) + 1}\Bigr)$ 
& \multirow{-3}{*}{\includegraphics[width=0.91\linewidth]{fig_mor/sharing_strategy/sharing_strategy_middle_cycle.pdf}} \\
[18pt]
First 
& -- 
& 
& $f\bigl(\mathbf{h}_t^{0}; \Phi_0\bigr)$ 
& \\ [5pt]
\midrule
& \multicolumn{2}{c|}{\textbf{Sequence Strategy}} & \multicolumn{2}{c}{\textbf{Middle-Sequence Strategy}} \\
\cmidrule(l{2pt}r{2pt}){2-3}\cmidrule(l{2pt}r{2pt}){4-5}
\textbf{Layers} & Equation & Figure & Equation & Figure \\
\midrule
Last 
& -- 
& 
& $f\bigl(\mathbf{h}_t^{L-1}; \Phi_{L-1}\bigr)$ 
& \\
[10pt]
Recursion 
& $f\!\Bigl(\mathbf{h}_t^{\ell}; \Phi'_{ \lfloor{\ell / N_r \rfloor}}\Bigr)$ 
& \multirow{-2}{*}{\includegraphics[width=0.91\linewidth]{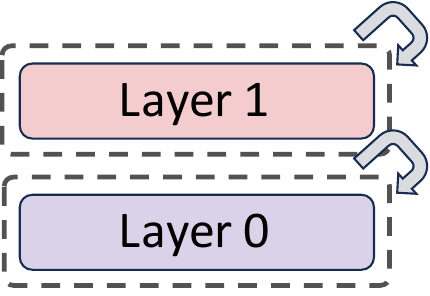}} 
& $f\!\Bigl(\mathbf{h}_t^{\ell}; \Phi'_{\lfloor(\ell - 1) / N_r\rfloor + 1)}\Bigr)$ 
& \multirow{-3}{*}{\includegraphics[width=0.91\linewidth]{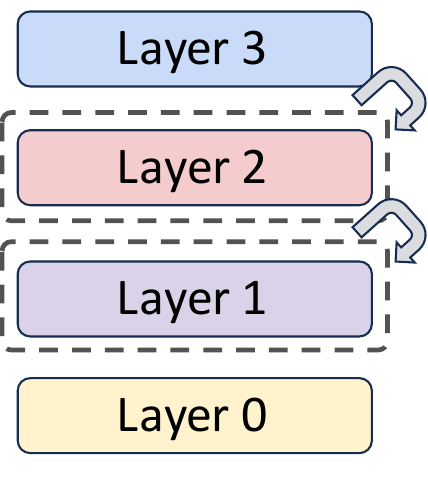}} \\
[20pt]
First 
& -- 
& 
& $f\bigl(\mathbf{h}_t^{0}; \Phi_0\bigr)$ 
& \\[7pt]
\bottomrule
\end{tabular}
}
\end{table*}

\clearpage

\subsection{Design Choices for Routing Strategy}
In this section, we provide an in-depth explanation of the two routing strategies employed in Mixture-of-Recursions: \textit{Expert-choice} and \textit{Token-choice} routers. Each approach has distinct advantages and inherent limitations, which we first outline before discussing the mitigation techniques we utilized.

\paragraph{Expert-choice routing.}
The expert-choice router offers several advantages, including a fixed compute budget that simplifies resource management. However, it suffers from a key issue: the top-$k$ selection operation, which requires information of tokens that appear later in the sequence, violates causality in autoregressive inference. This non-causal dependency (i.e., information leakage) can cause unexpected behavior during inference, potentially reducing model reliability. 

To address these challenges, we explore two approaches: the auxiliary router and the auxiliary loss \citep{raposo2024mixture}. The \textbf{auxiliary router} is an additional lightweight network trained jointly but used only during inference; it predicts whether a token will be among the top-$k$ selection. This additional router is trained with a binary cross-entropy loss, where the top-k selections from the main router are defined as the targets. Importantly, its training is isolated from the main objective through gradient blocking, so it does not affect the primary model training. Meanwhile, the \textbf{auxiliary loss} applies the binary cross-entropy loss to the main router itself, enabling it to simultaneously learn to push top-$k$ tokens towards one and others towards zero during training. This ensures the router can reliably predict which tokens will be selected as top-$k$ during inference.

\paragraph{Token-choice routing.}
In contrast, the token-choice router assigns recursion depths on a per-token basis without enforcing a fixed compute budget, thus avoiding leakage of information across tokens and preserving autoregressive properties. However, this introduces load imbalance across experts, which results in uneven token distribution across experts (or recursion depths), potentially causing inefficient compute allocation and unbalanced training.

To mitigate load imbalance, we employ two solutions from existing literature. \textbf{Balancing Loss}~\citep{lepikhin2020gshard, fedus2022switch} regularizes for a more uniform distribution of tokens across experts. For a sequence of length $T$, a balancing loss for MoR is calculated as follows: 
\begin{equation*}
\begin{aligned}
\mathcal{L}_{\text{Balance}} &= \alpha \sum_{i=1}^{N_r} f_i P_i, \\
f_i &= \frac{N_r}{T} \sum_{t=1}^T \mathbb{I}(\text{Token } t \text{ selects Expert } i), \\
P_i &= \frac{1}{T} \sum_{t=1}^T g_t^i,
\end{aligned}
\end{equation*}
where $N_r$ is the total number of experts (which is also the number of recursion), $g_t^i$ is the routing score of expert $i$ for token $t$, $f_i$ represents the fraction of tokens routed to expert $i$, $P_i$ denotes the average routing scores of expert $i$, and $\lambda$ is a hyperparameter controlling the strength of the auxiliary loss.\looseness=-1

\textbf{Loss-free}~\citep{wang2024auxiliary} utilizes router biasing without explicit regularization loss. Specifically, this method adjusts per-expert bias terms \(b_i\) to balance token assignments across experts. During each training batch, routing scores are computed, and the number of tokens assigned to each expert ($c_i$) is counted. The load violation error is calculated as $e_i = \bar{c}_i - c_i$ where \(\bar{c}_i\) is the average token count for expert \(i\). Biases are then updated via $b_i \leftarrow b_i + u \times \operatorname{sign}(e_i)$, where \(u\) is a bias update rate.
The biased routing scores for selecting top-$k$ expert are calculated as\looseness=-1
\begin{equation*}
g_{i}^t =
\begin{cases}
g_{i}^t, & \text{if } g_{i}^t + b_i \in \texttt{topk}\left(\{g_{j}^t + b_j \mid 1 \leq j \leq N\}, k\right) \\
0, & \text{otherwise}
\end{cases}
\end{equation*}
Note that the expert bias term is only utilized to adjust the routing strategy by influencing the top-$k$ selection.

\subsection{Design Choices for KV Caching Strategy}
This work investigates two principal strategies for key-value (KV) caching to optimize memory usage during Recursive Transformer computations: \textit{recursion-wise caching} and \textit{recursive KV sharing}.

\paragraph{Recursion-wise caching.} This keeps separate KV caches for each recursion step, ensuring tokens attend only to the KV pairs generated in their current recursion block. This prevents distribution mismatches between recursion steps, helping to maintain model accuracy while reducing memory and computational costs.

\paragraph{Recursive KV sharing.} In contrast, recursive sharing reuses KV pairs computed in the first recursion step for all subsequent steps. Although this approach further lowers memory usage and eliminates the need to compute deeper recursion during the prefill phase, it introduces potential mismatches as later recursion steps receive KV representations originally intended for earlier steps. Such mismatch can negatively impact model performance when token routing is precise.
Therefore, recursion-wise caching is generally preferred in settings with selective token routing to avoid performance degradation, while recursive KV sharing may be considered when memory efficiency is prioritized and prefill time is main bottleneck in the system.

\subsection{Experimental Setup}
\label{app:experimental_setup}

\paragraph{Training settings.}
We utilized a Llama-based Transformer architecture~\citep{grattafiori2024llama}, referring to the configurations of the open-source SmolLM models~\citep{allal2024SmolLM}. All models were pretrained on a deduplicated subset of the FineWeb-Edu dataset~\citep{penedo2024the} in SmolLM-Corpus~\citep{benallal2024smollmcorpus}, which comprises 220 billion tokens sourced from educational materials. Pretraining was conducted using four H100 or A100 GPUs. 
In our main and isoFLOPs analysis experiments, we utilized a Trapezoid learning rate scheduler, which consists of warmup (about 5\%), stable, and cooldown (20\%) phases. This approach allows us to efficiently continue pretraining for scaling laws from intermediate checkpoints, eliminating the need to train all models independently. In contrast, for all other experiments, we used a simple cosine annealing scheduler.

\paragraph{Evaluation settings.}

To assess model performance, we evaluated few-shot accuracy on six benchmarks using the Language Model Evaluation Harness: LAMBADA (LD), HellaSwag (HS), PIQA (PQ), WinoGrande (WG), ARC (Easy and Challenge), and MMLU. For all few-shot datasets, excluding LAMBADA, WinoGrande, and MMLU, we normalized accuracy by the byte length of the target string. We adhered to the standard number of shots for each dataset, and used the continuation task specifically for MMLU for simplicity. 
All evaluation performance measurements were conducted using a single H100 or A100 GPU.

\paragraph{Model architecture details}

Table\,\ref{tab_mor:model_arch} summarizes the architectural specifications of the four Vanilla Transformer models used as the base for our recursive models. Each model variant differs in scale, ranging from 135M to 1.7B total parameters (including both non-embedding and embedding components). For consistency and comparability, all models are trained using a vocabulary size of 49K and a maximum input sequence length of 2K tokens.

\begin{table*}[h!]
    \caption{
    Key parameters of four model size variants. A model's size is defined by the total number of its non-embedding and embedding parameters. Three small models utilize Grouped-Query Attention~\citep{DBLP:conf/emnlp/AinslieLJZLS23}, reducing the number of key-value heads. We refer to the base configurations of the open-sourced SmolLM models~\citep{allal2024SmolLM}.
    }
    \label{tab_mor:model_arch}
    \small
    \centering
    \resizebox{\textwidth}{!}{
    \setlength{\tabcolsep}{7.5pt}
    \begin{tabular}{l|cccc|cccc|cc}
    \toprule
      &  \multicolumn{4}{c|}{\textbf{Base Configuration}} & \multicolumn{4}{c|}{\textbf{Attention \& Feed-Forward}} & \multicolumn{2}{c}{\textbf{Input}} \\
    \cmidrule(l{2pt}r{2pt}){2-5} \cmidrule(l{2pt}r{2pt}){6-9} \cmidrule(l{2pt}r{2pt}){10-11}
     \textbf{Models} & N-emb & Emb & $N_L$ & $d_{model}$ & $N_{head}$ & $N_{KV}$ & $d_{head}$ & $d_{inter}$ & Vocab & $L_{ctx}$  \\
    \midrule
    Vanilla 135M & 106M & 28M & 30 & 576 & \,\,\,9 & 3 & 64 & 1536 & 49K & 2K \\[1pt]
    Vanilla 360M & 315M & 47M & 32 & 960 & 15 & 5 & 64 & 2560 & 49K & 2K \\[1pt]
    Vanilla 730M & 654M & 75M & 26 & \!\!\!1536 & 24 & 8 & 64 & 4096 & 49K & 2K \\[1pt]
    Vanilla \,\,\,1.7B & 1.61B & \!\!\!101M & 24 & \!\!\!2048 & 32 & \!\!\!32 & 64 & 8192 & 49K & 2K \\[1pt]
    \bottomrule
    \end{tabular}
    }
\end{table*}

\subsection{Expanded Results of IsoFLOP Analysis}
\label{app:scaling_laws}

In the main paper (\S\ref{subsec:scaling_laws}), we compared Vanilla, Recursive and our Mixture‑of‑Recursions (MoR) models under matched \emph{training compute}. 
Four base model capacities were studied—135M, 360M, 730M and 1.7B parameters. 
For recursive and MoR models, we fix the recursion count to $N_r\!=\!3$, so the number of \emph{unique} parameters is roughly one‑third of the vanilla counterpart.  
Each architecture is trained once for the \emph{largest} compute budget (16.5EB)\footnote{1EB $\,{=}\,10^{18}$ floating‑point operations.} and the resulting checkpoint is re‑used to obtain the 5EB and 2EB variants, as detailed below.

\paragraph{FLOPs approximated calculation of Transformers.}

We follow the approximation for calculating FLOPs as detailed in Kaplan et al. (2020)\,\citep{kaplan2020scaling}. Our analysis solely focuses on forward pass FLOPs, since the FLOPs involved in the backward pass are typically just double those of the forward pass. For most operations within Transformers, which primarily consist of linear projections, the forward pass FLOPs are calculated as two times the number of parameters, excluding the attention mechanism.

Regarding attention, we specifically account for the operations from the dot product between queries and keys and the scaling of values with softmax values. We only calculate FLOPs that contribute to the actual loss, excluding redundant computations in the upper triangular portion due to causality masking. Furthermore, we omit any additional computational costs associated with FlashAttention~\citep{dao2022flashattention}, normalization, and non-linearity operations from our overall FLOPs calculation.

As a result, Vanilla and Recursive Transformers have the same FLOPs. For MoR, the FLOPs calculation varies based on the routing and KV caching strategy. Especially, we calculated FLOPs based on the sequence length at each recursion depth, which is determined by the capacity factor and caching mechanism. In the case of token-choice routing, since the actual token allocation changes at every step, we approximated the FLOPs by assuming perfect balancing.
Furthermore, we add extra layers to a few MoR models to ensure their effective depth is divisible by the recursion number. For example, in a 135M model with 30 layers, setting a base depth of 10 and applying recursion three times (as in the Middle-Cycle strategy) results in a total of 32 layers. These additional layers introduce extra FLOPs, so we reduce the number of training steps accordingly to maintain our predefined FLOP budget.

\paragraph{Trapezoid learning‑rate schedule with checkpoint reuse.}
To avoid retraining every model from scratch for each FLOPs budget, we employ the \textit{trapezoid} schedule~\citep{xing2018walk}. The rule of this scheduler is as follows:\looseness=-1
\begin{equation*}
\eta(t)=
\begin{cases}
\frac{t}{w}\,\eta_{\max}, & 0 \le t < w             \quad\text{(warm‑up)},\\[4pt]
\eta_{\max},                            & w \le t < p             \quad\text{(plateau)},\\[2pt]
\eta_{\max}\!\Bigl(1-\tfrac{t-p}{d}\Bigr), & p\le t < p+d \quad\text{(cool‑down)},
\end{cases}
\end{equation*}

where $w$ denotes the warm‑up interval, $p-w$ is the constant‑LR plateau, and $d$ represents the cool-down segment. This stable phase allows us to efficiently manage experiments by saving intermediate checkpoints and then running additional cool-down steps from those points, according to each budget. For the warm‑up we allocate \(5\,\%\) of the total training steps of the smallest budget (2EB), and we set the cool‑down steps to \(20\,\%\) of the total training steps for each \emph{corresponding} budget.

\paragraph{Results at a glance.}
Table~\ref{tab_mor:scaling_laws} reports NLL on FineWeb-Edu validation set and few‑shot accuracy on six benchmarks. Our findings reveal a clear trend: more compute consistently leads to better models, evidenced by lower NLL and improved accuracy with higher FLOPs. However, weight-sharing alone in recursive models degraded performance compared to Vanilla, a clear trade-off for the reduced parameters. Crucially, token-routed MoR models overcome this shortcoming, catching up to and then surpassing Vanilla models from 360M parameters upward, all while utilizing only one-third of the parameters. This performance advantage persists at 730M and 1.7B parameter scales.\looseness=-1

\begin{table*}[h!]
    \caption{Detailed results of isoFLOP analysis across three compute budgets. We evaluate negative log‑likelihood (NLL) on the FineWeb‑Edu validation set and few‑shot accuracy on six downstream tasks for four base model sizes (135M, 360M, 730M, 1.7B). Each model was initially trained up to 16.5EB and sliced back to 5EB and 2EB via mid‑training checkpoints using a trapezoid learning‑rate schedule. All models used three recursion steps. For MoR models, we use expert-choice routing and recursion-wise caching mechanisms. We highlight the best-performing model in each setting in gray.
    }
    \label{tab_mor:scaling_laws}
    \small
    \centering
    \resizebox{\textwidth}{!}{
    \setlength{\tabcolsep}{4pt}
    \begin{tabular}{lc|cccc|cc|c|cccccc|c}
    \toprule
      & & \multicolumn{4}{c|}{\textbf{Pretrain}} &  \multicolumn{2}{c|}{\textbf{Recursion}} & \textbf{NLL\,$\downarrow$} & \multicolumn{7}{c}{\textbf{Few-shot Accuracy\,$\uparrow$}} \\
    \cmidrule(l{2pt}r{2pt}){3-6} \cmidrule(l{2pt}r{2pt}){7-8} \cmidrule(l{2pt}r{2pt}){9-9} \cmidrule(l{2pt}r{2pt}){10-16}
     \textbf{Models} & \textbf{Base} & N-Emb  & $N_L$ & FLOPs & $N_{tok}$ & Share & Loop & FineWeb & LD & HS & PQ & WG & ARC & \!MMLU\! & Avg  \\
    \midrule
    \rowcolor[gray]{0.9}
    Vanilla & 135M & 106M & 30 & \,\,\,2.0e+18  & 6.5B  & -  & - & 3.0922 & 22.80 & 30.93 & 62.35 & 51.14 & 36.28 & 26.29 & 38.30  \\
    Recursive & 135M & \,\,\,42M & 1+10+1 & \,\,\,2.0e+18  & 6.1B & M-Cyc  & 3 & 3.2058 & 19.79 & 29.32 & 60.17 & 50.59 & 34.83 & 25.40 & 36.68  \\
    MoR & 135M & \,\,\,42M & 1+10+1 & \,\,\,2.0e+18  & 9.2B & M-Cyc  & 3 & 3.1077 & 21.13 & 31.00 & 59.79 & 49.09 & 34.87 & 25.63 & 36.92  \\
    \midrule
    \rowcolor[gray]{0.9}
    Vanilla & 135M & 106M & 30 &  \,\,\,5.0e+18   & \!\!\!16.1B & -  & - & 2.9464 & 26.88 & 33.69 & 63.98 & 51.46 & 37.08 & 27.07 & 40.03  \\
    Recursive & 135M & \,\,\,42M & 1+10+1  &  \,\,\,5.0e+18   & \!\!\!15.1B & M-Cyc  & 3 & 3.0534 & 24.51 & 31.57 & 62.40 & 50.83 & 35.78 & 25.94 & 38.51  \\
    MoR & 135M & \,\,\,42M & 1+10+1  &  \,\,\,5.0e+18   & \!\!\!23.1B & M-Cyc  & 3 & 3.0192 & 22.01 & 32.53 & 61.75 & 49.88 & 35.39 & 26.19 & 37.96  \\
    \midrule
    \rowcolor[gray]{0.9}
    Vanilla & 135M & 106M & 30 & 16.5e+18  & \!\!\!53.3B & -  & - & 2.8432 & 30.16 & 36.51 & 64.80 & 53.43 & 40.17 & 27.82 & 42.15  \\
    Recursive & 135M & \,\,\,42M & 1+10+1 & 16.5e+18  & \!\!\!50.0B & M-Cyc  & 3 & 2.9552 & 25.98 & 33.36 & 63.98 & 51.78 & 36.96 & 26.68 & 39.79  \\
    MoR & 135M & \,\,\,42M & 1+10+1 & 16.5e+18  & \!\!\!76.2B & M-Cyc  & 3 & 2.9490 & 22.61 & 33.99 & 61.92 & 47.83 & 35.95 & 26.36 & 38.11  \\
    \midrule
    \midrule
    Vanilla & 360M & 315M & 32 & \,\,\,2.0e+18   & 2.4B & -  & - & 3.3785 & 17.27 & 27.90 & 59.36 & 51.38 & 32.10 & 25.49 & 35.58  \\
    Recursive & 360M & 118M & 1+10+1 & \,\,\,2.0e+18   & 2.4B & M-Cyc  & 3 & 3.4864 & 10.34 & 26.66 & 58.00 & 51.54 & 30.94 & 24.94 & 33.74  \\
    \rowcolor[gray]{0.9}
    MoR & 360M & 118M & 1+10+1 & \,\,\,2.0e+18   & 3.6B & M-Cyc  & 3 & 3.1026 & 24.14 & 30.53 & 61.86 & 50.99 & 34.74 & 25.50 & 37.96  \\
    \midrule
    Vanilla & 360M & 315M & 32 &  \,\,\,5.0e+18   & 6.0B & -  & - & 3.0097 & 25.17 & 32.10 & 63.22 & 48.62 & 36.01 & 26.69 & 38.63  \\
    Recursive & 360M & 118M & 1+10+1 &  \,\,\,5.0e+18   & 6.0B & M-Cyc  & 3 & 3.0722 & 23.29 & 31.19 & 62.62 & 51.30 & 35.85 & 25.99 & 38.37  \\
    \rowcolor[gray]{0.9}
    MoR & 360M & 118M & 1+10+1 &  \,\,\,5.0e+18   & 9.0B & M-Cyc  & 3 & 2.9161 & 28.33 & 34.53 & 63.22 & 51.07 & 36.70 & 26.98 & 40.14  \\
    \midrule
    Vanilla & 360M & 315M & 32 & 16.5e+18  & \!\!\!19.8B & -  & - & 2.7824 & 31.94 & 37.92 & 66.10 & 51.30 & 39.70 & 27.95 & 42.49  \\
    Recursive & 360M & 118M & 1+10+1 & 16.5e+18  & \!\!\!19.8B & M-Cyc  & 3 & 2.8466 & 29.75 & 35.92 & 64.91 & 51.46 & 39.12 & 27.18 & 41.39  \\
    \rowcolor[gray]{0.9}
    MoR & 360M & 118M & 1+10+1 & 16.5e+18  & \!\!\!29.7B & M-Cyc  & 3 & 2.7924 & 33.15 & 37.94 & 66.97 & 52.09 & 38.46 & 27.49 & 42.68  \\
    \midrule
    \midrule
    Vanilla & 730M & 654M & 26 & \,\,\,2.0e+18   & 1.2B & -  & - & 3.7164 & 07.74 & 26.58 & 57.62 & 51.14 & 29.74 & 24.46 & 32.88  \\
    Recursive & 730M & 252M & 1+8+1 & \,\,\,2.0e+18   & 1.2B & M-Cyc  & 3 & 3.8136 & 05.53 & 26.25 & 55.77 & 50.59 & 29.88 & 24.63 & 32.11  \\
    \rowcolor[gray]{0.9}
    MoR & 730M & 252M & 1+8+1 & \,\,\,2.0e+18   & 1.8B & M-Cyc  & 3 & 3.3300 & 17.93 & 28.74 & 59.30 & 51.46 & 33.14 & 25.37 & 35.99  \\
    \midrule
    Vanilla & 730M & 654M & 26 &  \,\,\,5.0e+18   & 3.1B & -  & - & 3.0821 & 22.05 & 31.99 & 62.68 & 50.67 & 35.88 & 26.12 & 38.23  \\
    Recursive & 730M & 252M & 1+8+1 &  \,\,\,5.0e+18   & 3.1B & M-Cyc  & 3 & 3.1640 & 18.51 & 30.72 & 62.13 & 47.83 & 35.97 & 25.84 & 36.83  \\
    \rowcolor[gray]{0.9}
    MoR & 730M & 252M & 1+8+1 &  \,\,\,5.0e+18   & 4.5B & M-Cyc  & 3 & 3.0067 & 26.18 & 32.76 & 62.46 & 50.91 & 36.93 & 26.37 & 39.27  \\
    \midrule
    Vanilla & 730M & 654M & 26 & 16.5e+18  & \!\!\!10.1B & -  & - & 2.7048 & 34.50 & 40.29 & 66.81 & 49.49 & 40.82 & 28.66 & 43.43  \\
    Recursive & 730M & 252M & 1+8+1 & 16.5e+18  & \!\!\!10.1B & M-Cyc  & 3 & 2.7886 & 30.76 & 37.84 & 65.51 & 52.41 & 39.26 & 27.51 & 42.21  \\
    \rowcolor[gray]{0.9}
    MoR & 730M & 252M & 1+8+1 & 16.5e+18  & \!\!\!14.9B & M-Cyc  & 3 & 2.7438 & 32.93 & 39.55 & 66.32 & 54.38 & 40.00 & 28.09 & 43.55  \\
    \midrule
    \midrule
    Vanilla & \,\,\,1.7B & 1.61B & 24 & \,\,\,2.0e+18   & 0.6B & -  & - & 5.1349 & 00.00 & 24.96 & 51.03 & 51.38 & 25.75 & 23.07 & 29.37  \\
    Recursive & \,\,\,1.7B & 0.67B & 1+8+1 & \,\,\,2.0e+18   & 0.5B & M-Cyc  & 3 & 5.3277 & 00.00 & 25.27 & 51.36 & 48.62 & 26.52 & 22.98 & 29.13  \\
    \rowcolor[gray]{0.9}
    MoR & \,\,\,1.7B & 0.67B & 1+8+1 & \,\,\,2.0e+18   & 0.8B & M-Cyc  & 3 & 4.1175 & 01.44 & 25.80 & 53.97 & 49.64 & 27.56 & 24.08 & 30.42  \\
    \midrule
    Vanilla & \,\,\,1.7B & 1.61B & 24 &  \,\,\,5.0e+18   & 1.5B & -  & - & 3.6926 & 08.33 & 26.84 & 57.29 & 51.30 & 29.72 & 24.51 & 33.00  \\
    Recursive & \,\,\,1.7B & 0.67B & 1+8+1 &  \,\,\,5.0e+18   & 1.3B & M-Cyc  & 3 & 3.8876 & 03.14 & 26.57 & 54.73 & 49.17 & 29.01 & 24.49 & 31.19  \\
    \rowcolor[gray]{0.9}
    MoR & \,\,\,1.7B & 0.67B & 1+8+1 &  \,\,\,5.0e+18   & 2.0B & M-Cyc  & 3 & 3.2905 & 17.62 & 28.32 & 59.03 & 49.80 & 32.14 & 25.28 & 35.37  \\
    \midrule
    Vanilla & \,\,\,1.7B & 1.61B & 24 & 16.5e+18  & 4.8B & -  & - & 2.8658 & 26.94 & 35.61 & 64.74 & 50.59 & 38.55 & 26.81 & 40.54  \\
    Recursive & \,\,\,1.7B & 0.67B & 1+8+1 & 16.5e+18  & 4.5B & M-Cyc  & 3 & 3.0042 & 23.25 & 32.09 & 62.95 & 50.75 & 37.64 & 26.53 & 38.87  \\
    \rowcolor[gray]{0.9}
    MoR & \,\,\,1.7B & 0.67B & 1+8+1 & 16.5e+18  & 6.5B & M-Cyc  & 3 & 2.8316 & 28.10 & 36.18 & 64.64 & 50.99 & 38.68 & 27.25 & 40.97  \\
    \bottomrule
    \end{tabular}
    }
\end{table*}

\clearpage
\subsection{Details of Experimental Settings for Throughput Measurement}
\label{app:throughput}

We implement a continuous depth-wise batching inference system~\citep{bae2024relaxed, hooper2023speed} to evaluate decoding throughput of MoR models. Queries are enqueued and scheduled dynamically during decoding using 1K samples from the FineWeb-Edu validation set. In particular, for MoR, when some queries exit early, the vacant slots in the batch are immediately filled with new queries waiting in the queue, maintaining a fully utilized batch at all times.\looseness=-1

We compare the throughput of Vanilla and MoR models (at a 360M parameter scale) for generating a certain length of tokens per sample, where the number is sampled from a normal distribution with a mean of 256, starting without any input prefix. The speeds of the MoR models are normalized against the speed of the Vanilla Transformer. For pretraining MoR-4 models, we add two additional layers (34 layers in total) before applying recursion. This ensures the total effective depth is divisible by the recursion number (specifically for the Middle-Cycle strategy). Consequently, the speed comparison for MoR-4 is made against a modified vanilla model that includes these two extra layers, resulting in a total of 34 layers (32 original layers + 2 added layers).\looseness=-1

We use two batching settings: (1) a \textit{fixed batch size of 32} and (2) a \textit{relative maximum batch size}, derived by multiplying 32 by the ratio of the maximum batch sizes of vanilla and MoR models. Specifically, based on the H100 GPU's VRAM size, we calculated the maximum batch sizes by considering model parameters and their KV cache memory. For simplicity, we omit the memory size from the hidden states at the current position. Under these adaptive conditions, MoR-2 supports a batch size of 42, MoR-3 supports 48, and MoR-4 supports up to 51. By employing recursion-wise KV caching, MoR allows a substantial increase in batch size stemming from its reduced parameter and KV cache memory footprint.

For implementation, we use a queue to enable continuous depth-wise batching and employ FlashAttention 2~\citep{dao2023flashattention} to support variable-length KV caches within a batch. We adopt a \textit{static}-sized cache where each position is updated over time, since this is compatible with \texttt{torch.compile}~\citep{paszke2019pytorch} to further optimize inference speeds. Furthermore, mimicking real-world deployment scenarios~\citep{kwon2023efficient, zhong2024distserve}, we decouple the transformer block phase from the rest of the computation by pre-processing the input embeddings or the first non-shared layer before passing into the transformer blocks. Then, we measured the actual time taken during the forward pass. Note that we include a warmup stage by running the model for 100 iterations before actual measurement, in order to obtain stable timing results. For further optimization, tokens that exited early were accumulated up to the maximum batch size before being processed by the last non-shared layer, classifier, and embedding layers (including the non-shared first layer in the case of MoR models). After this, we queue them for sequential batching by following a FIFO (First-In, First-Out) strategy. For implementation convenience, we exclude the time spent on caching and updating for KV pairs, as these aspects can be significantly optimized through various engineering techniques~\citep{kwon2023efficient}. We leave a more precise speed comparison, which accounts for these considerations, as future work.

\clearpage
\subsection{Expanded Results of Parameter Sharing Strategy}
\label{app:parameter_sharing}

This section complements the ablation in \S\ref{subsec:sharing} by providing the full quantitative panorama behind Figure\,\ref{fig_mor:pareto_frontier_sharing-b}. We revisit the four weight‑tying schemes (\textit{Cycle}, \textit{Sequence}, \textit{Middle‑Cycle}, and \textit{Middle‑Sequence}) on two base model scales (135M and 360M non‑embedding parameters) and two different recursion depths ($N_{r}=2$ and~3). All models were trained from scratch for 10B tokens under identical optimization hyperparameters. Validation NLL on FineWeb‑Edu and averaged few‑shot accuracy over six benchmarks are summarized in Table\,\ref{tab_mor:revist_sharing_strategy}.

\begin{table*}[h!]
  \caption{
    Comparison of parameter-sharing strategies (Cycle, Sequence, Middle-Cycle, Middle-Sequence) across two model scales (135M and 360M) and two recursion depths ($N_R = 2$ and $N_R = 3$). All models are pretrained from scratch on 10B tokens. We report validation negative log-likelihood (NLL) on FineWeb-Edu and few-shot accuracy across six tasks. Middle-Cycle consistently outperforms other strategies in both NLL and average task accuracy, especially at higher recursion depth. We highlight the optimal strategy for each setting in gray.\looseness=-1
    }

  \label{tab_mor:revist_sharing_strategy}
  \small
  \centering
  \resizebox{\textwidth}{!}{
  \setlength{\tabcolsep}{3pt}
  \begin{tabular}{l|ccc|cc|c|cccccc|c}
  \toprule
   & \multicolumn{3}{c|}{\textbf{Pretrain}} & \multicolumn{2}{c|}{\textbf{Recursion}} & \textbf{NLL\,$\downarrow$} & \multicolumn{7}{c}{\textbf{Few-shot Accuracy\,$\uparrow$}} \\
  \cmidrule(l{2pt}r{2pt}){2-4} \cmidrule(l{2pt}r{2pt}){5-6} \cmidrule(l{2pt}r{2pt}){7-7} \cmidrule(l{2pt}r{2pt}){8-14} 
    \textbf{Base Model} & N-Emb & $N_L$ & $N_{tok}$ & Share & Loop & FineWeb & LD & HS & PQ & WG & ARC & \!MMLU\! & Avg \\
  \midrule
  Vanilla\,135M & 106M & 30 & 10B & - & - & 3.0323 & 24.14 & 31.12 & 61.15 & 52.01 & 34.74 & 25.95 & 38.19 \\
  Vanilla\,135M & \,\,\,53M & 15 & 10B & - & - & 3.0818 & 23.64 & 30.10 & 60.94 & 50.99 & 35.38 & 25.93 & 37.83 \\
  Vanilla\,135M & \,\,\,35M & 10 & 10B & - & - & 3.1582 & 21.46 & 29.30 & 60.01 & 52.01 & 34.40 & 25.53 & 37.12 \\
  \midrule
  \rowcolor[gray]{0.9}
  Vanilla\,135M & \,\,\,53M & 15 & 10B & Cyc & 2 & 3.0071 & 25.52 & 31.25 & 61.10 & 50.99 & 36.08 & 26.11 & 38.51 \\
  Vanilla\,135M & \,\,\,53M & 15 & 10B & Seq & 2 & 3.1093 & 22.39 & 29.60 & 61.10 & 50.12 & 34.46 & 25.72 & 37.23 \\
   Vanilla\,135M & \,\,\,57M & 1+14+1 & 10B & M-Cyc & 2 & 3.0330 & 23.40 & 31.20 & 61.59 & 50.59 & 35.44 & 25.54 & 37.96 \\
    Vanilla\,135M & \,\,\,57M & 1+14+1 & 10B & M-Seq & 2 & 3.0991 & 21.70 & 30.06 & 60.45 & 49.41 & 35.20 & 25.74 & 37.09 \\
    \midrule
    Vanilla\,135M & \,\,\,35M & 10 & 10B & Cyc & 3 & 3.1154 & 21.42 & 30.14 & 60.61 & 49.72 & 34.15 & 25.57 & 36.94 \\
    Vanilla\,135M & \,\,\,35M & 10 & 10B & Seq & 3 & 3.1637 & 19.99 & 29.39 & 59.25 & 51.62 & 33.79 & 25.32 & 36.56 \\
    \rowcolor[gray]{0.9}
    Vanilla\,135M & \,\,\,39M & 1+9+1 & 10B & M-Cyc & 3 & 3.1048 & 22.41 & 30.35 & 61.04 & 49.01 & 34.80 & 25.91 & 37.26 \\
    Vanilla\,135M & \,\,\,39M & 1+9+1 & 10B & M-Seq & 3 & 3.1602 & 20.69 & 29.35 & 61.43 & 51.30 & 34.40 & 25.51 & 37.11 \\
    \midrule
    \midrule
    Vanilla\,360M & 315M & 32 & 10B & - & - & 2.8471 & 27.27 & 34.78 & 64.20 & 52.80 & 38.29 & 26.72 & 40.68 \\
    Vanilla\,360M & 157M & 16 & 10B & - & - & 2.8908 & 27.01 & 33.49 & 64.42 & 52.09 & 37.40 & 26.54 & 40.16 \\
    Vanilla\,360M & \,\,\,98M & 10 & 10B & - & - & 2.9449 & 26.41 & 32.93 & 63.38 & 50.36 & 37.15 & 26.48 & 39.45 \\
    \midrule
    Vanilla\,360M & 157M & 16 & 10B & Cyc & 2 & 2.8487 & 28.47 & 34.79 & 63.06 & 49.96 & 37.38 & 26.81 & 40.08 \\
    Vanilla\,360M & 157M & 16 & 10B & Seq & 2 & 2.9467 & 26.33 & 32.49 & 62.89 & 52.41 & 36.37 & 26.24 & 39.46 \\
    \rowcolor[gray]{0.9}
    Vanilla\,360M & 167M & 1+15+1 & 10B & M-Cyc & 2 & 2.8295 & 28.59 & 34.98 & 64.53 & 50.51 & 39.68 & 27.20 & 40.91 \\
    Vanilla\,360M & 167M & 1+15+1 & 10B & M-Seq & 2 & 2.9303 & 26.14 & 32.71 & 62.79 & 51.38 & 36.31 & 25.73 & 39.18 \\
    \midrule
    Vanilla\,360M & \,\,\,98M & 10 & 10B & Cyc & 3 & 2.9363 & 25.87 & 32.98 & 62.89 & 50.28 & 36.35 & 26.54 & 39.15 \\
    Vanilla\,360M & \,\,\,98M & 10 & 10B & Seq & 3 & 3.0245 & 24.55 & 31.48 & 63.11 & 49.25 & 35.65 & 25.73 & 38.30 \\
    \rowcolor[gray]{0.9}
    Vanilla\,360M & 118M & 1+10+1 & 10B & M-Cyc & 3 & 2.8760 & 28.51 & 34.89 & 64.31 & 50.51 & 39.51 & 27.20 & 40.82 \\
    Vanilla\,360M & 118M & 1+10+1 & 10B & M-Seq & 3 & 2.9753 & 24.18 & 31.89 & 62.08 & 49.72 & 36.47 & 26.27 & 38.44 \\
    \bottomrule
  \end{tabular}
  }
  
\end{table*}

\begin{figure}[h]
    \centering
    \begin{subfigure}[t]{0.35\textwidth}
    \captionsetup{justification=centering}
        \includegraphics[width=\textwidth]{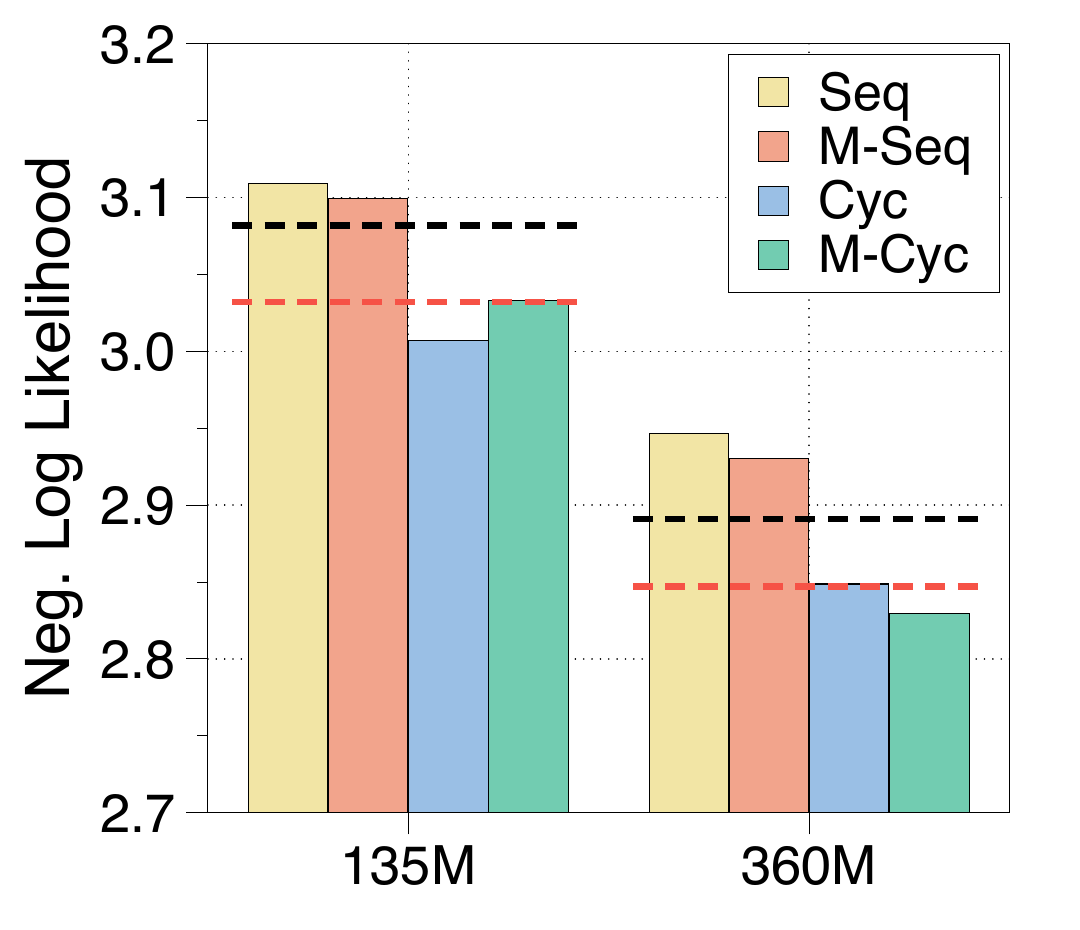}
        \subcaption{Recursion of two\,($N_R$ = 2)}
        \label{fig_mor:revisit_sharing_rec2_app}
    \end{subfigure}
    \hspace{10pt}
    \centering
    \begin{subfigure}[t]{0.35\textwidth}
    \captionsetup{justification=centering}
        \includegraphics[width=\textwidth]{fig_mor/revisit_sharing/revisit_sharing_rec3.pdf}
        \subcaption{Recursion of three\,($N_R$ = 3)}
        \label{fig_mor:revisit_sharing_rec3_app}
    \end{subfigure}
    \caption{
    Validation negative log‑likelihood (lower is better) on \emph{FineWeb‑Edu} for four parameter‑sharing strategies.
    Bars are grouped by model capacity (135M vs.\ 360M parameters). 
    Middle‑Cycle consistently attains the lowest NLL, with its margin widening as either model size or depth increases.
    The horizontal dashed lines mark the \emph{untied} (non-sharing) baselines: the lower red line represents the full capacity model, while the upper black line represents a parameter-matched reduced model with a footprint equal to the unique trainable parameter sizes of the recursive model.
    }
    \label{fig_mor:revisit_sharing_app}
\end{figure}

\clearpage
\paragraph{Middle‑Cycle is consistently the safest choice.}
For the 360M models, Middle-Cycle achieves the lowest NLL at both depths ($N_{r}=2$ and $N_{r}=3$) and also shows the largest improvement in average accuracy compared to vanilla reduced models. For the 135M models, while Cycle is slightly ahead at two recursion setting (3.0071 vs. 3.0330), Middle-Cycle overtakes when recursion depth rises (3.1048 vs. 3.1154) and shows a steadier accuracy profile. Meanwhile, pure Sequence sharing records the worst NLL in all four settings, and its accuracy gap widens with recursion depth. The Middle strategy slightly improves the performance of the Sequence, but it still performs worse than the Cycle-based methodology. We visualized the results in Figure\,\ref{fig_mor:revisit_sharing_app}.\looseness=-1

\paragraph{Behavior under continued pre‑training (up‑training).}
Table\,\ref{tab_mor:revist_sharing_strategy_uptrain} extends the study by ``up‑training'' models---continuing from open-sourced SmolLM~\citep{allal2024SmolLM} checkpoints for an additional 5B tokens.
Both Middle strategies demonstrate superior performance across all settings, and notably, they significantly outperform the reduced baseline models that are initialized in the same manner but without recursion.
The other strategies reach a performance plateau earlier, suggesting that they have limited room for further improvement in capacity.

\begin{table*}[h!]
    \caption{
    Uptraining results across four parameter sharing strategies. Models are trained on 5B tokens from FineWeb-Edu and evaluated by train NLL and few-shot accuracy across six benchmarks. ARC denotes average of ARC-Easy and ARC-Challenge tasks, MMLU denotes the MMLU-Cont task. We highlight the optimal strategy for each setting in gray.\looseness=-1
    }
    \label{tab_mor:revist_sharing_strategy_uptrain}
    \small
    \centering
    \resizebox{\textwidth}{!}{
    \setlength{\tabcolsep}{3pt}
    \begin{tabular}{l|ccc|ccc|c|cccccc|c}
    \toprule
      &  \multicolumn{3}{c|}{\textbf{Pretrain}} &  \multicolumn{3}{c|}{\textbf{Recursion}} & \textbf{NLL\,$\downarrow$} & \multicolumn{7}{c}{\textbf{Few-shot Accuracy\,$\uparrow$}} \\
    \cmidrule(l{2pt}r{2pt}){2-4} \cmidrule(l{2pt}r{2pt}){5-7} \cmidrule(l{2pt}r{2pt}){8-8} \cmidrule(l{2pt}r{2pt}){9-15} 
     \textbf{Base Model} & N-Emb & $N_L$ & $N_{tok}$ & Share & Init & Loop & FineWeb & LD & HS & PQ & WG & ARC & \!MMLU\! & Avg  \\
    \midrule
    Vanilla\,360M & 315M & 32 & 5B & -  & - &   & 2.4825 & 41.67 & 50.63 & 70.35 & 55.09 & 46.99 & 30.82 & 49.26  \\
    Vanilla\,360M & 157M & 16 & 5B & - & Step  & 1   & 2.7168 & 31.85 & 37.59 & 64.74 & 53.20 & 41.06 & 27.34 & 42.63  \\
    Vanilla\,360M & 157M & 16 & 5B & Cyc  & Avg & 1  & 2.8603 & 22.14 & 30.36 & 60.07 & 48.22 & 34.99 & 25.56 & 36.89  \\
    Vanilla\,360M & 157M & 16 & 5B & Seq  & Avg & 1  & 2.7919 & 25.40 & 32.35 & 62.30 & 50.12 & 35.88 & 26.19 & 38.71  \\
    Vanilla\,360M & \,\,\,98M & 10 & 5B & -  & Step & 1  & 2.8915 & 26.63 & 35.03 & 64.42 & 52.09 & 38.75 & 26.86 & 40.63    \\
    Vanilla\,360M & \,\,\,98M & 10 & 5B & Cyc  & Avg & 1  & 3.0512 & 23.19 & 31.27 & 62.30 & 51.22 & 36.71 & 26.29 & 38.50   \\
    Vanilla\,360M & \,\,\,98M & 10 & 5B & Seq  & Avg & 1  & 2.9915 & 25.67 & 32.21 & 62.30 & 51.38 & 36.68 & 26.56 & 39.14  \\
    \midrule
     \rowcolor[gray]{0.9}
    Vanilla\,360M & 157M & 16 & 5B & Cyc  & Step & 2  & 2.7165 & 31.30 & 37.68 & 64.91 & 52.17 & 39.29 & 27.53 & 42.15  \\
    Vanilla\,360M & 157M & 16 & 5B & Cyc  & Avg & 2  & 2.8263 & 23.21 & 30.52 & 60.55 & 50.28 & 36.01 & 25.50 & 37.68  \\
    Vanilla\,360M & 157M & 16 & 5B & Cyc  & Lower & 2  & 2.8024 & 27.67 & 34.71 & 63.49 & 49.88 & 38.12 & 26.87 & 40.13  \\
    Vanilla\,360M & 157M & 16 & 5B & Cyc  & Upper & 2  & 2.7915 & 18.26 & 34.88 & 63.06 & 51.85 & 39.27 & 26.88 & 39.03  \\
    Vanilla\,360M & 157M & 16 & 5B & Cyc  & Rand & 2  & 2.7575 & 25.29 & 34.78 & 61.64 & 52.01 & 38.09 & 26.62 & 39.74   \\
    \midrule
     \rowcolor[gray]{0.9}
    Vanilla\,360M & 157M & 16 & 5B & Seq  & Step & 2  & 2.6862 & 34.14 & 42.49 & 67.90 & 53.35 & 43.24 & 28.80 & 44.99  \\
    Vanilla\,360M & 157M & 16 & 5B & Seq  & Avg & 2  & 2.7508 & 29.01 & 34.13 & 63.60 & 52.09 & 36.31 & 26.58 & 40.29  \\
    Vanilla\,360M & 157M & 16 & 5B & Seq  & Lower & 2  & 2.8300 & 27.50 & 33.28 & 63.38 & 51.38 & 37.44 & 26.32 & 39.88  \\
    Vanilla\,360M & 157M & 16 & 5B & Seq  & Upper & 2  & 2.7498 & 30.49 & 40.07 & 65.61 & 52.25 & 40.28 & 28.17 & 42.81  \\
    Vanilla\,360M & 157M & 16 & 5B & Seq  & Rand & 2  & 2.7153 & 32.00 & 41.31 & 66.10 & 53.35 & 42.13 & 28.52 & 43.90   \\
    \midrule
    Vanilla\,360M & 167M & 1+15+1 & 5B & M-Cyc  & Step & 2  & 2.6800 & 35.47 & 42.39 & 67.19 & 50.99 & 42.54 & 28.79 & 44.56  \\
    Vanilla\,360M & 167M & 1+15+1 & 5B & M-Cyc  & Avg & 2  & 2.7314 & 33.81 & 40.42 & 66.87 & 51.78 & 41.68 & 28.17 & 43.79  \\
    Vanilla\,360M & 167M & 1+15+1 & 5B & M-Cyc  & Lower & 2  & 2.7449 & 30.76 & 39.50 & 66.16 & 50.99 & 41.12 & 28.07 & 42.77  \\
     \rowcolor[gray]{0.9}
    Vanilla\,360M & 167M & 1+15+1 & 5B & M-Cyc  & Upper & 2  & 2.6605 & 34.41 & 43.74 & 67.46 & 53.20 & 43.75 & 28.93 & 45.25  \\
    Vanilla\,360M & 167M & 1+15+1 & 5B & M-Cyc  & Rand & 2  & 2.6730 & 35.65 & 43.04 & 67.74 & 52.17 & 42.60 & 28.62 & 44.97  \\
    \midrule
     \rowcolor[gray]{0.9}
    Vanilla\,360M & 167M & 1+15+1 & 5B & M-Seq  & Step & 2  & 2.6627 & 35.09 & 43.34 & 67.57 & 51.22 & 43.66 & 28.91 & 44.97  \\
    Vanilla\,360M & 167M & 1+15+1 & 5B & M-Seq  & Avg & 2  & 2.7143 & 33.92 & 40.93 & 66.49 & 51.70 & 40.72 & 28.24 & 43.66   \\
    Vanilla\,360M & 167M & 1+15+1 & 5B & M-Seq & Lower & 2  & 2.7696 & 30.74 & 38.36 & 65.94 & 51.78 & 41.27 & 27.73 & 42.64  \\
    Vanilla\,360M & 167M & 1+15+1 & 5B & M-Seq  & Upper & 2  & 2.6931 & 32.66 & 42.35 & 67.14 & 52.80 & 42.83 & 28.49 & 44.38  \\
    Vanilla\,360M & 167M & 1+15+1 & 5B & M-Seq  & Rand & 2  & 2.6908 & 35.07 & 42.03 & 66.32 & 53.75 & 43.11 & 28.42 & 44.78  \\
    \midrule
    Vanilla\,360M & \,\,\,98M & 10 & 5B & Cyc  & Step & 3  & 2.8901 & 27.46 & 35.26 & 63.82 & 51.54 & 39.35 & 27.44 & 40.81  \\
    Vanilla\,360M & \,\,\,98M & 10 & 5B & Seq & Step & 3  & 2.8258 & 30.43 & 37.37 & 63.76 & 52.33 & 40.55 & 27.65 & 42.02  \\
     \rowcolor[gray]{0.9}
    Vanilla\,360M & 118M & 1+10+1 & 5B & M-Cyc  & Step & 3  & 2.7735 & 31.38 & 39.31 & 65.51 & 50.51 & 40.70 & 27.65 & 42.51  \\
     \rowcolor[gray]{0.9}
    Vanilla\,360M & 118M & 1+10+1 & 5B & M-Seq  & Step & 3  & 2.7678 & 31.67 & 39.23 & 65.89 & 52.09 & 40.65 & 27.90 & 42.91  \\
    Vanilla\,360M & 118M & 1+10+1 & 5B & M-Cyc  & Upper & 3  & 2.8100 & 28.86 & 37.61 & 65.51 & 52.57 & 41.44 & 28.17 & 42.36  \\
    \bottomrule
    \end{tabular}
    }
\end{table*}

\clearpage

\subsection{Details of Router Design Configurations}
\label{appx:router_details}

We investigate various router design choices to optimize performance and stability. Specifically, we tune the coefficient values controlling the strength of auxiliary or balancing loss terms (\textit{Coeff}), and adjust the scaling factor applied after the router function ($\alpha$) to modulate routing weights. Moreover, we test different activation functions (\textit{Func}), such as sigmoid or softmax, are evaluated, with architectural variations (\textit{Arch}) of the router network, including linear layer, 2-layer MLP with GELU activation, Wide-MLP that expands the hidden layer size by a factor of four.\looseness=-1

We also incorporate several techniques to stabilize training. To improve training stability, we utilize the \textit{router z-loss}~\citep{zoph2022st}, which penalizes large logits produced by the gating network. Large logits can cause numerical instability and hinder effective training of the router. The z-loss is computed as follows:
\begin{equation*}
L_z(x) = \frac{1}{B} \sum_{i=1}^B \left( \log \sum_{j=1}^{N_r} e^{x_j^{(i)}} \right)^2,
\end{equation*}
where $B$ is the number of tokens in the batch, $N_r$ is the number of experts, and $x \in \mathbb{R}^{B \times N_r}$ denotes the logits input to the router. This regularization encourages the gating network to produce smaller logits, promoting more stable and reliable routing decisions.

\subsection{Router Performance Evaluation Metrics}

\paragraph{Expert-choice routing.}
We evaluate dead token ratio and sampling accuracy to assess the router’s selection behavior. The \textit{dead token ratio} measures the proportion of tokens at specific positions within the batch that are consistently unselected during the final recursion step, indicating a positional bias where certain token positions are systematically neglected by the router. The \textit{sampling accuracy} how well the router used during inference predicts whether a token belongs to the top-$k$ tokens identified during training, reflecting the router’s ability to consistently select the most relevant tokens.
Ideally, high sampling accuracy with a low dead token ratio indicates a router that both identifies important tokens accurately and maintains diversity in token selection.

\paragraph{Token-choice routing.}
We evaluate the router’s ability to balance token assignments across experts using MaxVio (maximum violation) and entropy metrics. \textit{MaxVio}~\citep{wang2024auxiliary} measures the load imbalance across experts:\looseness=-1
\begin{equation*}
\mathrm{MaxVio} = \frac{\max_i \mathrm{Load}_i - \overline{\mathrm{Load}}_i}{\overline{\mathrm{Load}}_i},
\label{eq:maxvio}
\end{equation*}
where $\mathrm{Load}_i$ denotes the actual number of tokens assigned to the $i$-th expert, and $\overline{\mathrm{Load}}_i$ represents the expected load per expert assuming perfect balance.

To measure the diversity of token assignments across experts, we also compute the \textit{Entropy} of the average selection probabilities for each expert:
\begin{equation*}
H = - \sum_{i=1}^{N_r} \overline{p}_i \log \overline{p}_i,
\label{eq:entropy}
\end{equation*}
where $\overline{p}_i$ is the average probability of selecting the $i$-th expert over all tokens in the evaluation batch, and $N_r$ is the total number of experts. A higher entropy indicates a more uniform distribution of tokens among experts, reflecting balanced and diverse routing decisions.\looseness=-1

\subsection{Extended Evaluation Results of Router Designs}
\label{app:router_ablation}

The results presented in Table\,\ref{tab_mor:ablation_study_expert_choice_app} indicate that although both the auxiliary router and auxiliary loss methods enhance sampling accuracy, they are also associated with high dead token ratios. In particular, some auxiliary router variants exhibit dead token ratios as high as 66.7\%, suggesting that the router always selects tokens from the same positions across inputs, reflecting a positional bias. Notably, employing a linear router architecture in conjunction with auxiliary loss effectively reduces the dead token ratio without compromising sampling accuracy.\looseness=-1

Results from Table\,\ref{tab_mor:ablation_study_token_choice_app} reveal that applying an explicit balancing loss significantly reduces MaxVio and increases entropy, leading to improved load balance without sacrificing overall model performance. Loss-free approaches, while simpler, tend to show higher MaxVio and lower entropy, indicating less balanced token routing. Architectures such as MLP and Linear routers perform comparably under balancing loss, with z-loss often contributing to improved routing stability and model accuracy. Nevertheless, it still struggles to achieve balance during quite long initial stage. The heterogeneity among the experts, stemming from the use of computation blocks with varying recursion depths as experts, likely complicates load balancing.\looseness=-1

\begin{table*}[h!]
    \caption{
    Ablation results on using expert-choice router with different routing configurations. We use the recursion-wise KV caching strategy by default. Coeff denotes coefficient values for auxiliary loss term, and $\alpha$ denotes scaling term after router function. Dead token ratio are measured within evaluation batch size of 500. Warmup refers to gradually decreasing the capacity from 1.0 to desired value over warmup steps. The last highlighted row represents a chosen final strategy, and intermediate best-performing designs (based on performance and routing metrics) are highlighted to illustrate how it was derived.\looseness=-1
    }
    \label{tab_mor:ablation_study_expert_choice_app}
    \small
    \centering
    \resizebox{\linewidth}{!}{
    \setlength{\tabcolsep}{2pt}
    \begin{tabular}{lcccccc|cc|c|ccccccc}
    \toprule
      \multicolumn{7}{c|}{\textbf{Expert-choice Configurations}} &  \multicolumn{2}{c|}{\textbf{Router\,Metrics}} & \textbf{NLL}\,$\downarrow$ & \multicolumn{7}{c}{\textbf{Few-shot Accuracy\,$\uparrow$}} \\
    \cmidrule(l{2pt}r{2pt}){1-7} \cmidrule(l{2pt}r{2pt}){8-9} \cmidrule(l{2pt}r{2pt}){10-10} \cmidrule(l{2pt}r{2pt}){11-17}
    Sampling & Coeff & \!Func\! & $\alpha$ & Arch & Warmup & z-loss & Dead\,$\downarrow$ & Samp-Acc\,$\uparrow$ & FineWeb & LD & HS & PQ & WG & ARC & \!MMLU\! & Avg  \\ 
    \midrule
     - & - & \texttt{rand} & - & - & \xmark & \xmark & \,\,\,0.0 & - & 2.9335 & 26.0 & 33.1 & 61.6 & 52.3 & 35.8 & 26.2 & 39.1  \\
     \midrule
     Aux\,Router & - & - & - & MLP & \xmark & \xmark & 66.7 & 50.0 & \texttt{NaN} & 0.0 & 25.04 & 49.5 & 49.6 & 23.9 & 23.0 & 28.5  \\
     Aux\,Router & - & $\sigma$ & 0.1 & MLP & \xmark & \xmark & \,\,\,0.0 & 89.2 & 2.8893 & 26.1 & 33.8 & 62.0 & 51.5 & 36.6 & 26.4 & 39.4 \\
     \rowcolor[gray]{0.9}
     Aux\,Router & - & $\sigma$ & 1.0 & MLP & \xmark & \xmark & 66.7 & 50.0 & 2.8867 & 26.4 & 33.6 & 63.0 & 52.4 & 37.0 & 24.1 & 39.8 \\
     Aux\,Router & - & \texttt{tanh} & 0.1 & MLP & \xmark & \xmark & 66.7 & 98.6 & 2.8720 & 13.9 & 31.8 & 60.7 & 49.3 & 35.8 & 25.8 & 36.2  \\
     Aux\,Router & - & \texttt{tanh} & 1.0 & MLP & \xmark & \xmark & 66.7 & 97.0 & 3.0624 & 18.26 & 29.7 & 60.1 & 50.9 & 34.6 & 25.5 & 36.5 \\
     \midrule
     Aux\,Loss & 0.01 & - & - & MLP & \xmark & \xmark & 66.7 & 50.0 & \texttt{NaN} & 0.0 & 25.04 & 49.5 & 49.6 & 23.9 & 23.0 & 28.5 \\
     \rowcolor[gray]{0.9}
     Aux\,Loss & 0.01 & $\sigma$ & 0.1 & MLP & \xmark & \xmark & 0.0 & 99.6 & 2.8967 & 24.8 & 33.6 & 63.3 & 50.3 & 36.6 & 26.6 & 39.2\\
     Aux\,Loss & 0.01 & $\sigma$ & 1.0 & MLP & \xmark & \xmark & 65.9 & \!\!\!100.0 & 2.9189 & 12.0 & 31.6 & 59.4 & 51.5 & 33.2 & 25.3 & 35.5\\
     Aux\,Loss & 0.01 & \texttt{tanh} & 0.1 & MLP & \xmark & \xmark & 32.8 & 99.7 & 2.9426 & 23.5 & 32.4 & 62.4 & 49.8 & 35.6 & 26.0 & 38.3 \\
     Aux\,Loss & 0.01 & \texttt{tanh} & 1.0 & MLP & \xmark & \xmark & 0.0 & 98.8 & 3.2743 & 16.4 & 28.14 & 58.8 & 52.2 & 31.6 & 24.8 & 35.3 \\
     \midrule
     Aux\,Loss & 0.1 & $\sigma$ & 0.1 & MLP & \xmark & \xmark & 0.0 & 99.8 & 3.0416 & 21.5 & 31.0 & 61.8 & 50.3 & 35.0 & 26.0 & 37.6 \\
     \rowcolor[gray]{0.9}
     Aux\,Loss & \!\!0.001\!\! & $\sigma$ & 0.1 & MLP & \xmark & \xmark & 0.0 & 99.1 & 2.8816 & 27.6 & 34.3 & 63.0 & 51.6 & 36.7 & 26.5 & 40.0 \\
     Aux\,Loss & \!\!0.001\!\! & \texttt{tanh} & 0.1 & MLP & \xmark & \xmark & 0.0  & 56.4 & 2.9933 & 25.0 & 32.3 & 61.5 & 51.5 & 36.6 & 26.0 & 38.8 \\
     \midrule
     \rowcolor[gray]{0.9}
     Aux\,Loss & \!\!0.001\!\! & $\sigma$ & 0.1 & Linear & \xmark & \xmark & 0.1 & 99.2 & 2.8667 & 27.4 & 34.6 & 63.2 & 51.5 & 37.2 & 26.5 & 40.1  \\
     Aux\,Loss & \!\!0.001\!\! & $\sigma$ & 0.1 & W-MLP & \xmark & \xmark & 0.4 & 99.2 & 2.8716 & 27.8 & 33.9 & 62.4 & 49.9 & 36.3 & 26.3 & 39.4\\
     \midrule
     Aux\,Loss & \!\!0.001\!\! & $\sigma$ & 0.1 & Linear & \cmark & \xmark & 4.9 & 99.1 & 2.8744 & 26.0 & 33.9 & 62.0 & 51.2 & 36.1 & 26.1 & 39.2 \\
     Aux\,Loss & \!\!0.001\!\! & $\sigma$ & 0.1 & Linear & \xmark & \cmark & 0.0 & 99.3 & 2.8824 & 26.9 & 34.0 & 63.8 & 52.3 & 36.8 & 26.4 & 40.0\\
    \bottomrule
    \end{tabular}
    }
\end{table*}

\begin{table*}[h!]
    \caption{
    Ablation results on token-choice router under different routing configurations. We use the recursion-wise KV caching strategy by default. Coeff denotes coefficient for balancing loss term, or updating coefficient ($u$) for loss-free algorithm. The last highlighted row represents a chosen final strategy, but we added z-loss back in with a small coefficient of 1e-3 since it often stabilizes load balancing. The intermediate best-performing designs (based on performance and routing metrics) are highlighted to illustrate how it was derived.\looseness=-1
    }
    \label{tab_mor:ablation_study_token_choice_app}
    \small
    \centering
    \resizebox{\linewidth}{!}{
    \setlength{\tabcolsep}{3.5pt}
    \begin{tabular}{lccccc|cc|c|ccccccc}
    \toprule
      \multicolumn{6}{c|}{\textbf{Token-choice Configurations}} &  \multicolumn{2}{c|}{\textbf{Router\,Metrics}} & \textbf{NLL}\,$\downarrow$ & \multicolumn{7}{c}{\textbf{Few-shot Accuracy\,$\uparrow$}} \\
    \cmidrule(l{2pt}r{2pt}){1-6} \cmidrule(l{2pt}r{2pt}){7-8} \cmidrule(l{2pt}r{2pt}){9-9} \cmidrule(l{2pt}r{2pt}){10-16}
    Balancing & Coeff & \!Func\! & $\alpha$ & Arch & Z-loss & MaxVio\,$\downarrow$ & Entropy\,$\uparrow$ & FineWeb & LD & HS & PQ & WG & ARC & \!MMLU\! & Avg  \\
    \midrule
     - & - & \texttt{rand} & - & - &  \cmark & 0.007 & 1.099 & 3.0268 & 24.8 & 32.0 & 61.4 & 52.2 & 35.5 & 26.1 & 38.7 \\
     \midrule
     Loss & 0.1 & \texttt{soft} & 1.0 & MLP &  \cmark & 0.200 & 1.076 & 3.0239 & 24.2 & 31.9 & 61.4 & 51.5 & 35.7 & 26.2 & 38.5 \\
     \rowcolor[gray]{0.9}
     Loss & 0.01 & \texttt{soft} & 1.0 & MLP &  \cmark & 0.682 & 0.921 & 2.9118 & 28.0 & 33.3 & 62.8 & 49.7 & 36.4 & 26.2 & 39.4 \\
     \midrule
     Loss-free & 0.01 & \texttt{soft} & 1.0 & MLP &  \cmark & 1.788 & 0.297 & 2.9078 & 25.5 & 32.5 & 61.3 & 52.3  & 36.1 & 26.0 & 38.9 \\
     Loss-free & 0.01 & $\sigma$ & 0.1 & MLP &  \cmark & 0.956 & 0.646 & 3.1144 &21.8 & 29.8 & 60.3 &51.6 & 34.0 & 25.7 & 37.2 \\
     Loss-free & 0.01 & $\sigma$ & 1.0 & MLP &  \cmark & 0.918 & 0.749 & 3.0188 & 23.4 & 31.3 & 59.9 & 50.0 & 35.2 & 25.8 & 37.6 \\
     \rowcolor[gray]{0.9}
     Loss-free & 0.001 & \texttt{soft} & 1.0 & MLP &  \cmark & 0.852 & 0.915 & 2.9081 & 25.8 & 33.6 & 62.8 &50.6 & 37.5 & 26.7 & 39.5  \\
     Loss-free & 0.001 & $\sigma$ & 0.1 & MLP &  \cmark & 1.281 & 0.551 & 2.9165 &23.9 & 33.1 & 61.2 & 51.6 & 37.3 & 26.2 &38.9 \\
     Loss-free & 0.001 & $\sigma$ & 1.0 & MLP &  \cmark & 0.542 & 0.941 & 3.0188 &24.9 & 32.0 & 61.9 & 51.4 & 35.5 & 25.9 & 38.6 \\
     \midrule
     \rowcolor[gray]{0.9}
     Loss & 0.1 & \texttt{soft} & 1.0 & Linear &  \cmark & 0.492 & 0.960 & 2.9974 &23.7 & 31.3 &  62.2 & 50.3 & 36.7 & 26.0 & 38.4 \\
     Loss & 0.1 & \texttt{soft} & 1.0 & W-MLP &  \cmark & 0.384  & 1.037 & 3.0293 & 25.3 & 31.5 & 62.2 &51.2 & 36.4 & 26.3 & 38.8 \\
     \midrule
     \rowcolor[gray]{0.9}
     Loss & 0.1 & \texttt{soft} & 1.0 & Linear &  \xmark & 0.266 & 1.056 & 2.9358 & 25.7 & 32.6 & 61.9 & 51.7 & 36.4 & 26.5 & 39.1 \\
    \bottomrule
    \end{tabular}
    }
\end{table*}

\subsection{Key Value Representation Trends in Recursive Transformers}

Sharing KV caches across model depths has emerged as a promising approach to improve inference throughput in Vanilla Transformers~\citep{brandon2024reducing}. This technique can reduce the memory footprint required for KV caches, enabling larger inference batch sizes. Significant speedups can be also achieved by skipping the KV projection and even prefill operations at shared depths, especially with Cycle strategy~\citep{sun2024you}. Due to the high degree of freedom in Vanilla models—where trainable parameters can be well optimized for shared caches—these models exhibit only marginal performance drops when KV caches are shared between adjacent layers.
In contrast, Recursive Transformers have far fewer parameters available for being optimized to tied KV states. Nevertheless, we hypothesize that similar patterns may emerge between shared blocks. To investigate this, we decomposed the KV states from pretrained Recursive Transformers into magnitude and directional components.\looseness=-1

\begin{figure}[h]
    \centering
    \begin{subfigure}[t]{0.3334\textwidth}
    \captionsetup{justification=centering}
        \includegraphics[width=\textwidth]{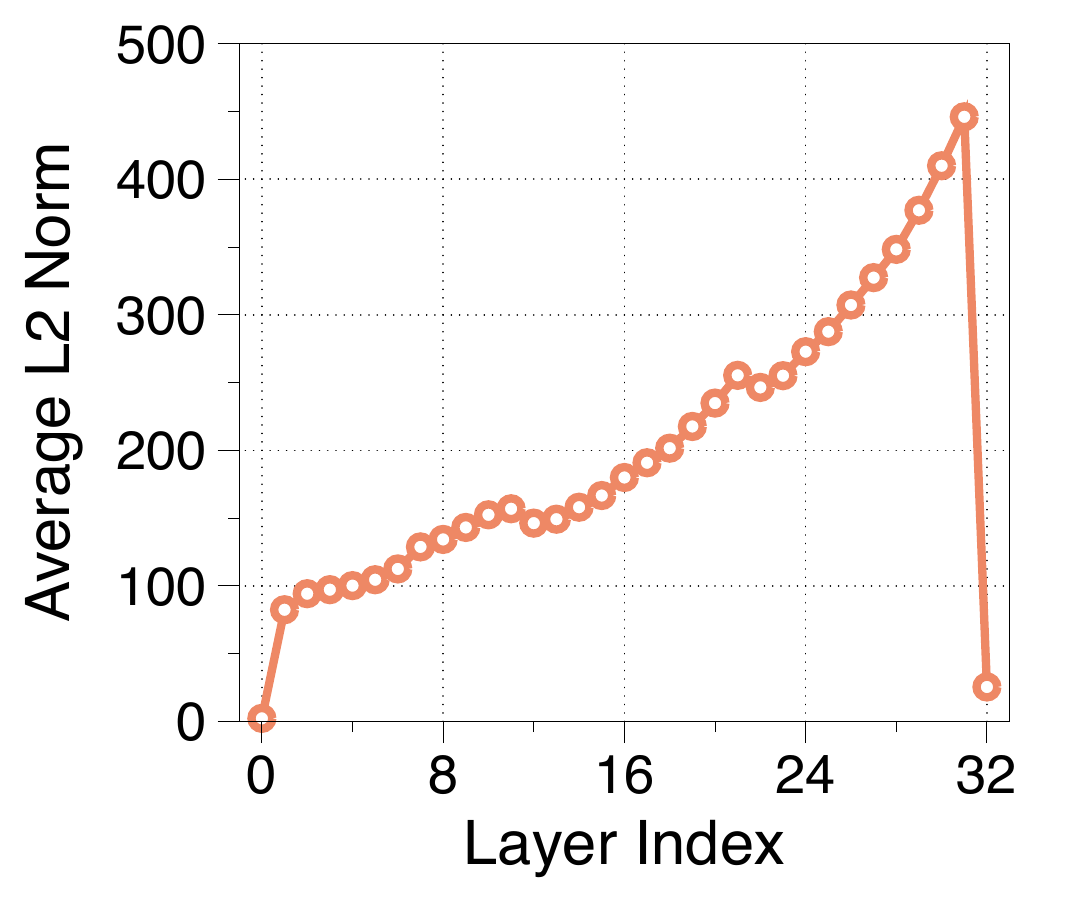}
        \subcaption{Hidden states}
    \end{subfigure}
    \centering
    \begin{subfigure}[t]{0.303\textwidth}
    \captionsetup{justification=centering}
        \includegraphics[width=\textwidth]{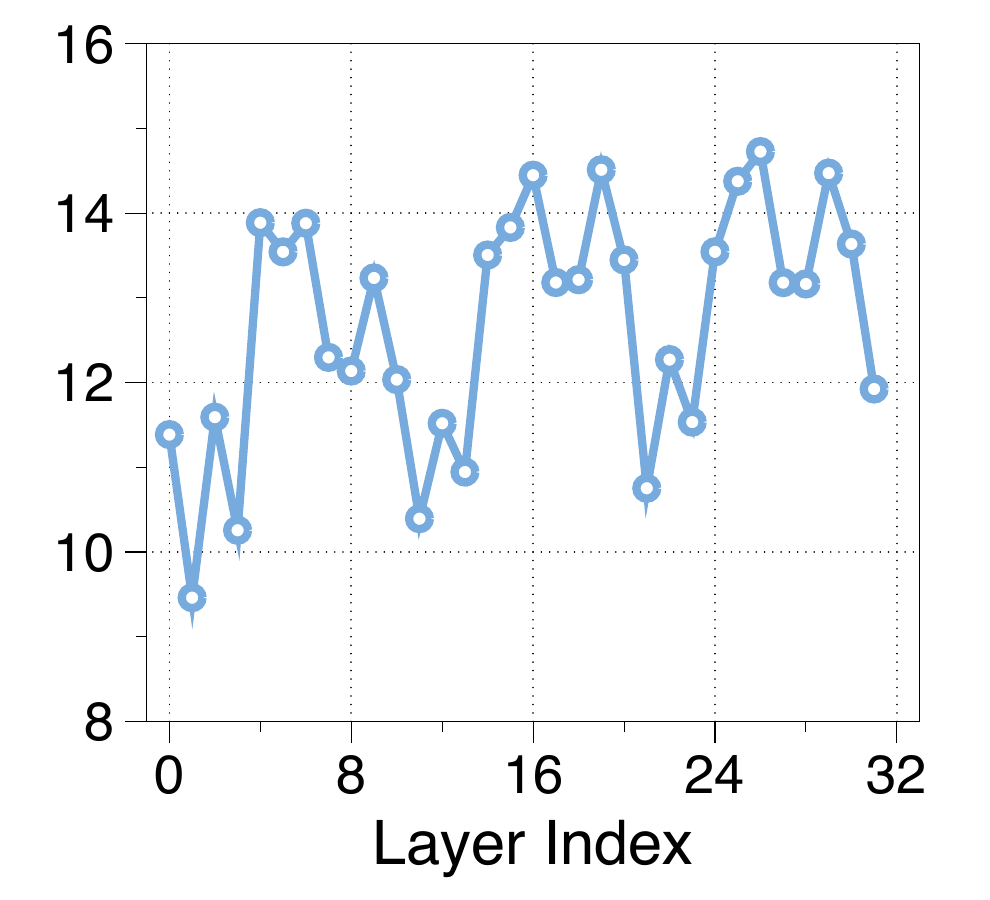}
        \subcaption{Key states}
    \end{subfigure}
    \centering
    \begin{subfigure}[t]{0.303\textwidth}
    \captionsetup{justification=centering}
        \includegraphics[width=\textwidth]{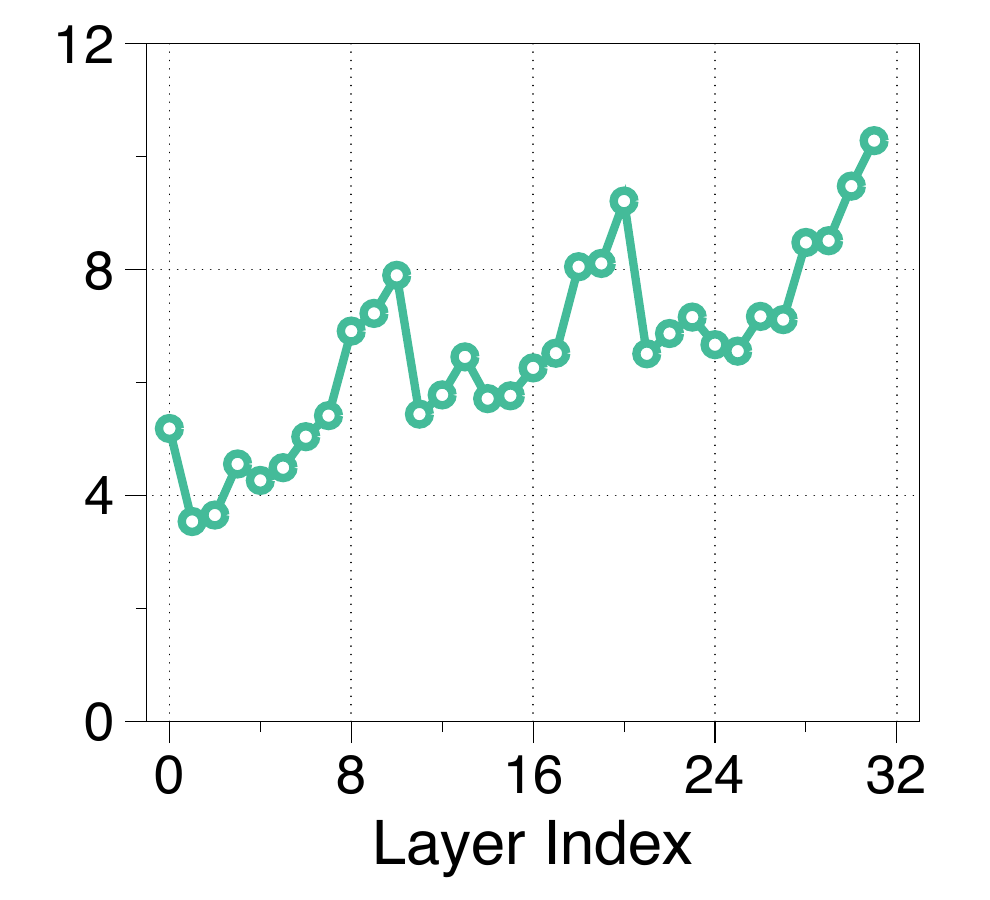}
        \subcaption{Value states}
    \end{subfigure}
    \caption{ Average L2 norm magnitude of (a) hidden states, (b) key states, and (c) value states across layers in a Middle-Cycled Recursive Transformer 360M with 3 recursion steps. 
    Note that the last hidden states correspond to the final hidden states after the last layer normalization. 
    }
    \label{fig_mor:kv_sharing_mag_app}
\end{figure}

As shown in Figure\,\ref{fig_mor:kv_sharing_mag_app}, the sharing of key and value projection layers across recursion depths leads to clear recursive patterns in the \textit{magnitude} values. Although the magnitudes of hidden states tend to increase, the projection layers appear to be trained to produce similar signal sizes at corresponding depths within each recursion.\looseness=-1

\begin{figure}[h!]
    \centering
    \begin{subfigure}[t]{0.33\textwidth}
    \captionsetup{justification=centering}
        \includegraphics[width=\textwidth]{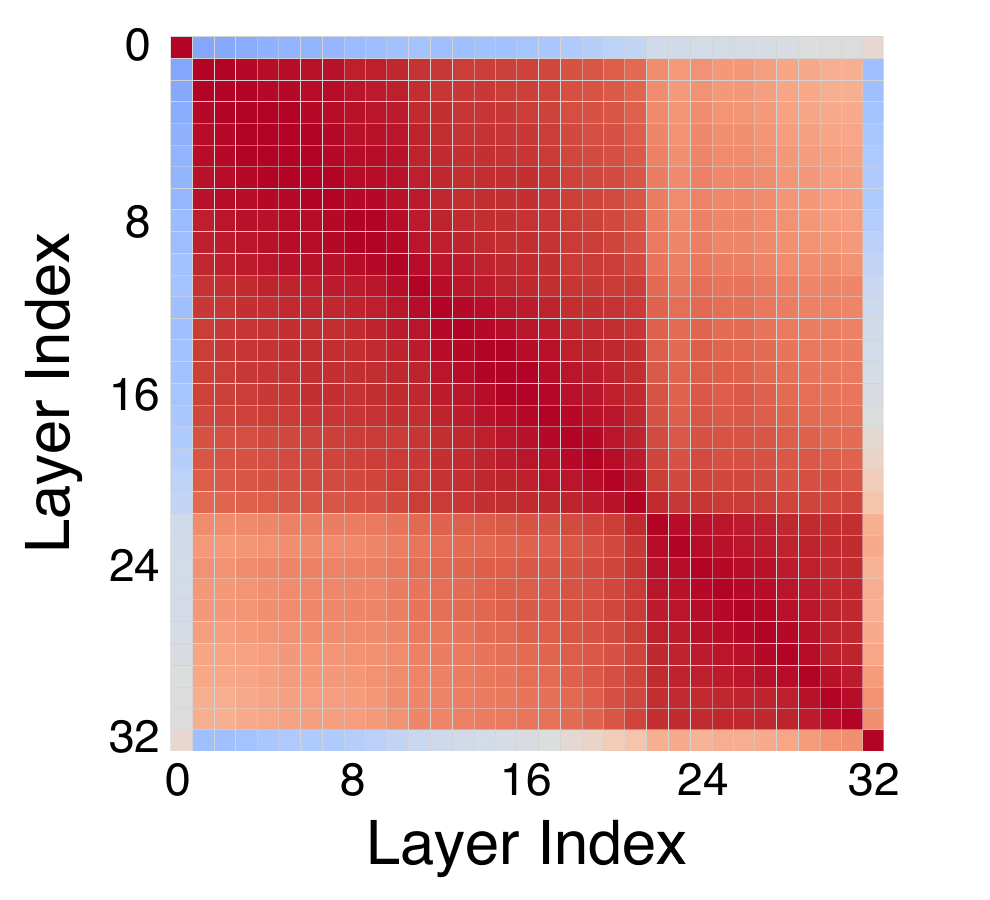}
        \subcaption{Hidden states}
    \end{subfigure}
    \centering
    \begin{subfigure}[t]{0.30\textwidth}
    \captionsetup{justification=centering}
        \includegraphics[width=\textwidth]{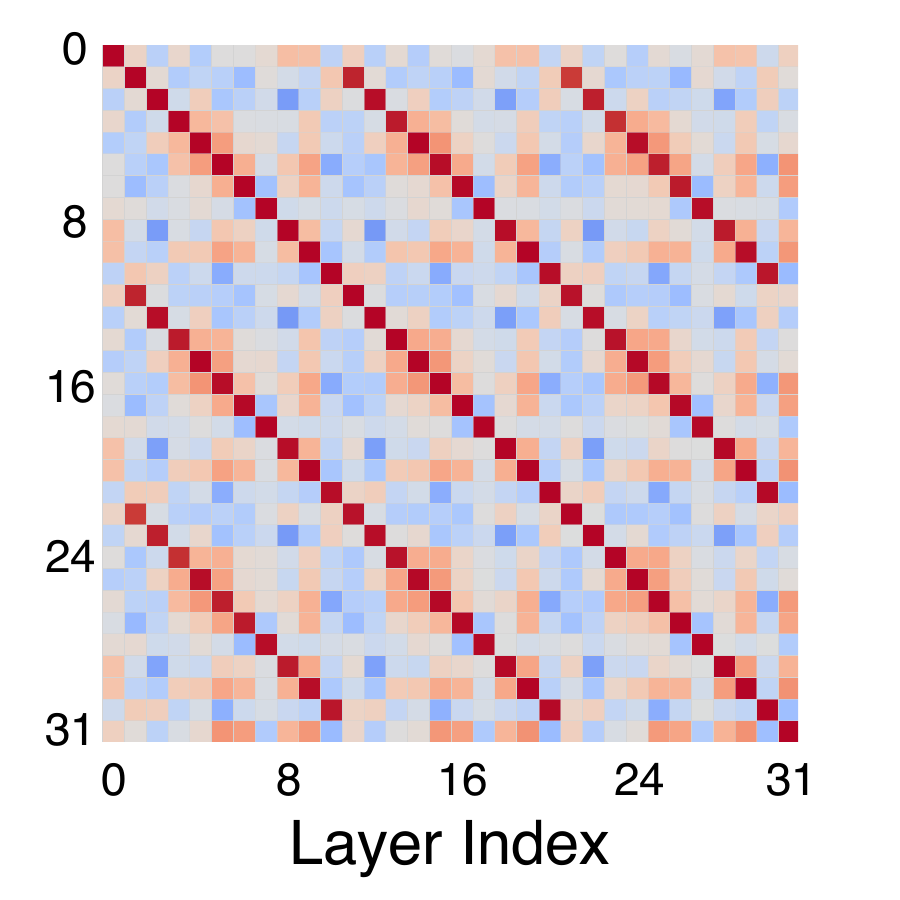}
        \subcaption{Key states}
    \end{subfigure}
    \centering
    \begin{subfigure}[t]{0.33\textwidth}
    \captionsetup{justification=centering}
        \includegraphics[width=\textwidth]{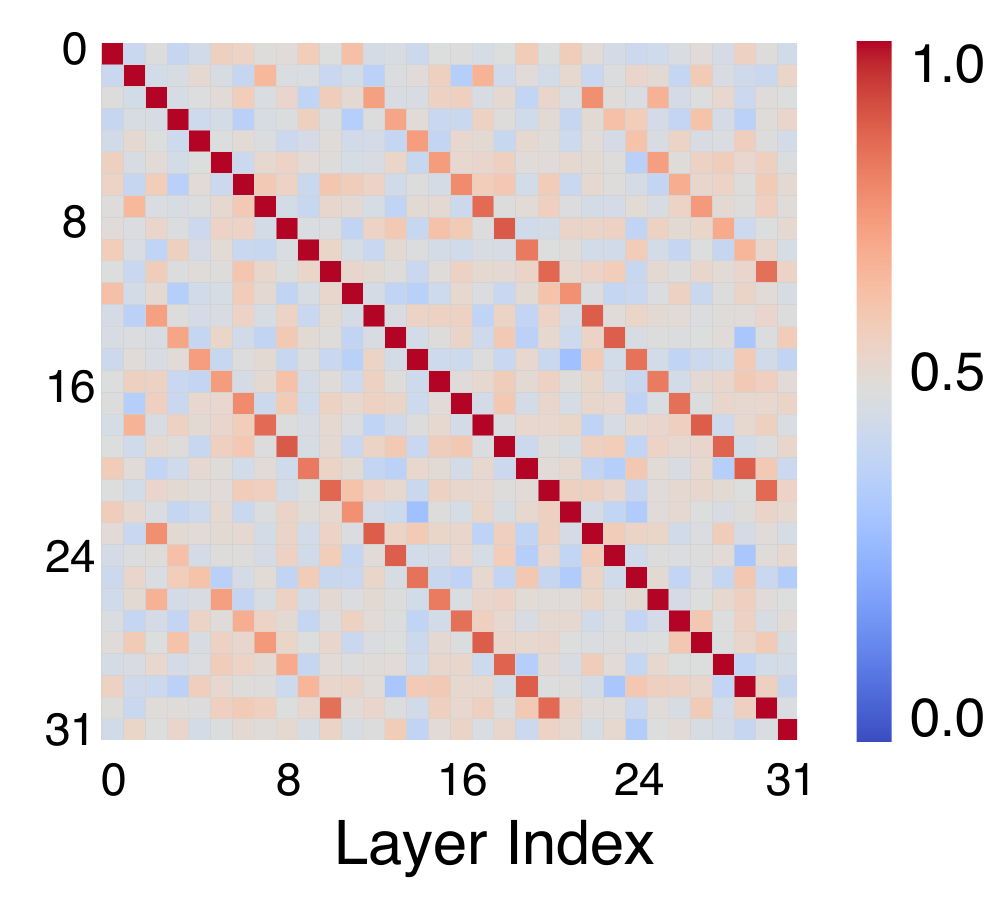}
        \subcaption{Value states}
    \end{subfigure}
    \caption{ Cosine similarity matrices showing the layer-wise similarity of (a) hidden states, (b) key states, and (c) value states in Recursive Transformer with Middle-Cycle strategy and recursion depth 3. Results are from a 360M parameter model with 32 layers. The hidden states matrix includes the final hidden states after the last layer normalization.
    }
    \label{fig_mor:kv_sharing_dir_app}
\end{figure}

When we measured the cosine similarity in Figure\,\ref{fig_mor:kv_sharing_dir_app}, distinct diagonal patterns emerge, suggesting that shared projection layers generate highly similar key and value representations. While sharing value states across recursions appears to be more challenging than sharing key states, these findings suggest that the performance drop from KV cache sharing can be marginal even in Recursive Transformers.\looseness=-1

\subsection{Performance Comparison of KV Sharing Strategy}
\label{app:expanded_kv}

\paragraph{Experimental results of key-value cache sharing.}
In Table\,\ref{tab_mor:key_value_sharing_results}, we present the performance results when applying KV cache sharing to Vanilla, Recursive, and MoR models. Especially, we tested various strategies for KV caches, including Cycle or Sequence strategies that share the same concepts as parameter sharing (see \S\ref{app:sharing_strategy} for details). Interestingly, KV cache sharing even improves the performance of vanilla models, where sharing acts as a regularization technique.
In case of recursive models, we align the sharing strategy for parameters and KV caches. Despite some variations in the results after applying KV sharing, the Middle-Cycle strategy (the best parameter sharing strategy) showed a slight perplexity drop, albeit not substantial.

When moving to MoR models, they still introduced a small amount of degradation in our best settings (expert-choice router). However, considering the reduced parameter sizes and cache sizes, we believe this minor drop is acceptable. Furthermore, we explored an alternative sharing strategy (indicated by $\dagger$) that utilized shared caches for inactive (unselected) positions while updating active positions through actual computation. This method is analogous to a recursive caching scheme but initializes inactive positions with key-value pairs from the first recursive iteration. Although it did not provide additional benefits, it is still worth exploring combinations of KV sharing and actual updates.\looseness=-1

\begin{table*}[h]
    \caption{
    Comparison of KV cache sharing strategies across Vanilla, Recursive, and MoR Transformers. Models are pretrained on 10B tokens of FineWeb-Edu, and evaluated using negative log-likelihood (NLL) on train set and few-shot accuracy across benchmarks. KV sharing denotes use of recursive KV sharing mechanism. If MoR is mentioned without further specification of KV sharing strategy, it implies the use of the recursion-wise caching strategy. $\text{}^\dagger$It indicates training with hybrid KV sharing that leverages shared caches for inactive positions while updating active ones through actual computation.\looseness=-1
    }
    \label{tab_mor:key_value_sharing_results}
    \small
    \centering
    \resizebox{\textwidth}{!}{
    \setlength{\tabcolsep}{3.5pt}
    \begin{tabular}{l|cc|cc|c|cc|c|cccccc|c}
    \toprule
      &  \multicolumn{2}{c|}{\textbf{Pretrain}} &  \multicolumn{2}{c|}{\textbf{Recursion}} & \textbf{MoR} & \multicolumn{2}{c|}{\textbf{KV Sharing}} & \textbf{NLL\,$\downarrow$} & \multicolumn{7}{c}{\textbf{Few-shot Accuracy\,$\uparrow$}} \\
    \cmidrule(l{2pt}r{2pt}){2-3} \cmidrule(l{2pt}r{2pt}){4-5} \cmidrule(l{2pt}r{2pt}){6-6} \cmidrule(l{2pt}r{2pt}){7-8} \cmidrule(l{2pt}r{2pt}){9-9} \cmidrule(l{2pt}r{2pt}){10-16} 
     \textbf{Models} & N-Emb & $N_L$ & Share & Loop & Type & Share & Loop & FineWeb & LD & HS & PQ & WG & ARC & \!MMLU\! & Avg  \\
    \midrule
    Vanilla & 315M & 32 & - & - & - & - & - & 2.8471 & 27.3 & 34.8 & 64.2 & 52.8 & 38.3 & 26.7 & 40.7\\
    Vanilla & 315M & 32 & - & - & - & Seq & 2 & 2.7848 & 30.0 & 36.5 & 64.6 & 50.7 & 39.4 & 26.9 & 41.3 \\
    \rowcolor[gray]{0.9}
    Vanilla & 315M & 32 & - & - & - & Cyc & 2 & 2.7650 & 30.0 &	36.7 & 65.2 & 51.1  & 39.6 & 27.5 & 41.7\\
    \midrule
    Vanilla & 295M & 30 & - & - & - & - & - & 2.8069 & 29.1 & 35.6 & 65.1 & 50.4 & 38.5 & 27.3 & 41.0 \\
    \rowcolor[gray]{0.9}
    Vanilla & 295M & 30 & - & - & - & Seq & 3 & 2.7879 & 28.3 & 36.4 & 64.3 & 52.7 & 39.4 & 27.3 & 41.4 \\
    Vanilla & 295M & 30 & - & - & - & Cyc & 3 & 2.7890 & 28.9 & 36.5 & 64.6 & 51.4  & 39.0 & 27.6 & 41.3 \\
    \midrule
    Recursive & 157M & 16 & Seq & 2 & - & - & - & 2.9467 & 26.3 & 32.5 & 62.9 & 52.4 & 36.4 & 26.2 & 39.5\\
    Recursive &157M  & 16 & Seq & 2 & - & Seq & 2 & 2.8904 & 26.4 & 33.4 & 64.0 & 51.0 & 37.0 & 26.9 & 39.8\\
    Recursive & 157M & 16 & Cyc & 2 & - & - & - & 2.8487 & 28.5 & 34.8 & 63.1 & 50.0 & 37.4 & 28.8 & 40.1\\
    Recursive & 157M & 16 & Cyc & 2 & - & Cyc & 2 & 2.8577 & 26.2 & 34.5 & 64.2 & 51.4 & 37.3 & 26.9 & 40.1\\
    \rowcolor[gray]{0.9}
    Recursive & 167M & 1+15+1 & M-Cyc & 2 & - & - & - & 2.8295 & 28.6 & 35.0 & 64.5 & 50.5 & 39.7 & 27.2 & 40.9\\
    Recursive & 167M & 1+15+1 & M-Cyc & 2 & - & M-Cyc & 2 & 2.8451 & 27.3 &	34.7 & 63.7 & 50.5 & 37.8 & 27.0 & 40.2 \\
    \midrule
    Recursive & \,\,\,98M & 10 & Seq & 3 & - & - & - & 3.0245 & 24.6 & 31.5 & 63.1 & 49.3 & 35.7 & 25.7 & 38.3\\
    Recursive & \,\,\,98M & 10 & Seq & 3 & - & Seq & 3 & 2.9554 & 24.2 & 32.3 & 62.5 & 52.7 & 36.6 & 26.2 & 39.1\\
    Recursive & \,\,\,98M & 10 & Cyc & 3 & - & - & - & 2.9363 & 25.9 & 33.0 & 62.9 & 50.3 & 36.4 & 26.5 & 39.2\\
    Recursive & \,\,\,98M & 10 & Cyc & 3 & - & Cyc & 3 & 2.9155 & 24.1 & 32.9 & 62.4 & 51.2 & 37.4 & 26.7 & 39.1 \\
    \rowcolor[gray]{0.9}
    Recursive & 118M & 1+10+1 & M-Cyc & 3 & - & - & - & 2.8760 & 28.5 & 34.9 & 64.3 & 50.5 & 39.5 & 27.2 & 40.8\\
    Recursive & 118M & 1+10+1 & M-Cyc & 3 & - & M-Cyc & 3 & 2.8854 & 27.3 &	33.8 & 63.3 & 52.3 & 37.5 & 26.8 &	40.2\\
    \midrule
    \rowcolor[gray]{0.9}
    MoR & 118M & 1+10+1 & M-Cyc & 3 & Expert & - & - & 2.8667 & 27.4 & 34.6 & 63.2 & 51.5 & 37.2 & 26.5 & 40.1\\
    MoR & 118M & 1+10+1 & M-Cyc & 3 & Expert & M-Cyc & 3 & 2.8895 & 34.0 & 	61.6 & 50.2 & 26.0 & 36.5 & 27.0 & 39.2 \\
    MoR & 118M & 1+10+1 & M-Cyc & 3 & Expert & \,\,$\text{M-Cyc}^{\dagger}$ & 3 & 2.8653 & 24.8 & 34.3 & 62.0 & 50.1 & 36.7 & 26.7 & 39.1\\
    \midrule
    \rowcolor[gray]{0.9}
    MoR & 118M & 1+10+1 & M-Cyc & 3 & Token & - & - & 2.9358 & 25.7 & 32.6 & 61.9 & 51.7 & 36.4 & 26.5 & 39.1\\
    MoR & 118M & 1+10+1 & M-Cyc & 3 & Token & M-Cyc & 3 & 2.9155& 25.7 & 32.6 &	61.8 & 49.4 & 36.2 & 26.0 & 38.6\\
    \bottomrule
    \end{tabular}
    }
\end{table*}

\vspace{-10pt}
\paragraph{Relaxation for key-value sharing constraints.}
We also investigated relaxing the constraints on KV sharing in Table\,\ref{tab_mor:key_value_sharing_relaxation}, similar to the relaxation approach in Bae et al. (2024) \,\citep{bae2024relaxed} for parameter sharing constraints. Specifically, we first re-examined four relaxation techniques for standard Recursive Transformers with very small ranks~\citep{hu2022lora, liu2024dora} or prefix lengths~\citep{liu2021p}. We also experimented with the position encoding~\citep{chen2025inner}, where trainable embeddings are element-wise multiplied with the output of each recursion block. 

Our results show that these techniques do not provide substantial performance improvements when pretraining relaxed models from scratch, consistent with prior studies, as they introduce only a limited number of additional parameters. Although we hypothesize that incorporating prefix-based approaches (such as adding trainable prefixes to attention) into KV sharing might lead to greater benefits, our experiments did not reveal substantial differences in this regard. Further exploration of more sophisticated techniques for efficiently relaxing KV cache sharing constraints remains an open direction for future research.\looseness=-1

\begin{table*}[h]
    \caption{
    Experimental results of relaxing parameter sharing and KV cache sharing constraints in Recursive Transformers. All models are trained on FineWeb-Edu with 10B tokens, and we apply the Middle-Cycle parameter sharing for 360M models with 3 recursion depths. We evaluate them based on training NLL and few-shot accuracy across six benchmarks. Relaxation types include encoding trainable embeddings on recursion outputs via element-wise multiplication (Enc), applying LoRA and DoRA to query and value weight matrices, and adaptation prompt tuning (Adapt-P). 
    }
    \label{tab_mor:key_value_sharing_relaxation}
    \small
    \centering
    \resizebox{\textwidth}{!}{
    \setlength{\tabcolsep}{3.5pt}
    \begin{tabular}{l|cc|ccc|cc|c|cccccc|c}
    \toprule
      &  \multicolumn{2}{c|}{\textbf{Pretrain}} &  \multicolumn{3}{c|}{\textbf{Relaxation}} &  \multicolumn{2}{c|}{\textbf{KV Sharing}} & \textbf{NLL\,$\downarrow$} & \multicolumn{7}{c}{\textbf{Few-shot Accuracy\,$\uparrow$}} \\
    \cmidrule(l{2pt}r{2pt}){2-3} \cmidrule(l{2pt}r{2pt}){4-6} \cmidrule(l{2pt}r{2pt}){7-8} \cmidrule(l{2pt}r{2pt}){9-9} \cmidrule(l{2pt}r{2pt}){10-16} 
     \textbf{Models} & N-Emb & $N_L$ & Type & Rank & Len & Share & Loop & FineWeb & LD & HS & PQ & WG & ARC & \!MMLU\! & Avg  \\
    \midrule
    Recursive & 118M & 1+10+1 & - & - & - & - & - & 2.8854 & 27.3 &	33.8 & 63.3 & 52.3 & 37.5 & 26.8 &	40.2\\
    Recursive & 118M & 1+10+1 & Enc & - & - & - & - & 2.8604 & 27.3 & 34.6 & 63.9 & 53.4 & 38.6 & 26.7 & 40.2\\
    Recursive & 124M & 1+10+1 & LoRA & 64 & - & - & - & 2.8599 & 27.3 & 34.6 & 64.3 & 50.9 & 38.0	& 26.9 & 39.7 \\
    Recursive & 124M & 1+10+1 & DoRA & 64 & - & - & - & 2.8945 & 26.4 &	33.6 & 64.4 & 50.6 & 37.4 &	26.5 & 39.2 \\
    Recursive & 126M & 1+10+1 & Adapt-P & - & 256 & - & - & 2.8626 & 27.1 & 34.7 & 64.0 & 51.9	& 37.6 & 26.8 & 39.7\\
    \midrule
    Recursive & 118M & 1+10+1 & - & - & - & M-Cyc & 3 & 2.8854 & 27.3 &	33.8 & 63.3 & 52.3 & 37.5 & 26.8 & 40.2\\
    Recursive & 126M & 1+10+1 & Adapt-P & - & 256 & M-Cyc & 3 & 2.9030 & 24.5 & 33.1 & 63.0 & 52.2 & 26.7 & 37.6 &	39.5\\
    \bottomrule
    \end{tabular}
    }
\end{table*}

\subsection{Analysis on Adaptive Computation Paths}

Table\,\ref{tab_mor:qualitative_highlighted_results} illustrates a qualitative analysis of the recursion depth assigned to each subword token. This visualization provides a detailed insight into how tokens within each sample exhibit varying levels of recursive processing, showcasing the adaptive computation mechanism within the MoR framework. Notably, some tokens exit early (purple), while others require deeper processing (blue and red), reflecting the model’s ability to focus more compute on challenging parts of the input.

\begin{table*}[ht]
\centering
\footnotesize
\caption{Visualization of the recursion depth for each subword token, with colors representing the number of recursion steps: \raisebox{0pt}[0.5em][0.1em]{\colorbox{DarkOrchid!25}{1}}, \raisebox{0pt}[0.5em][0.1em]{\colorbox{SkyBlue!55}{2}}, and \raisebox{0pt}[0.5em][0.1em]{\colorbox{Salmon!60}{3}}. This depth specifically indicates how many times recursion is applied at that token's position to predict the subsequent token. Each row corresponds to a single sample, offering a clear illustration of the token-level recursion distribution in practice. We use an MoR model with $N_r = 3$, auxiliary loss, and recursion-wise KV caching. This model is built on a 360M parameter base and trained on 30B tokens.
}
\begin{tabular}{@{}l p{13cm}@{}}
\toprule
\textbf{Sample} & \textbf{Text} \\
\bottomrule
\end{tabular}
\label{tab_mor:qualitative_highlighted_results}
\end{table*}

\clearpage

\subsection{Analysis on Router Weights}
\label{app:qualitative_results}

\begin{figure}[h]
    \centering
    \begin{subfigure}[t]{0.4\textwidth}
    \captionsetup{justification=centering}
        \includegraphics[width=\textwidth]{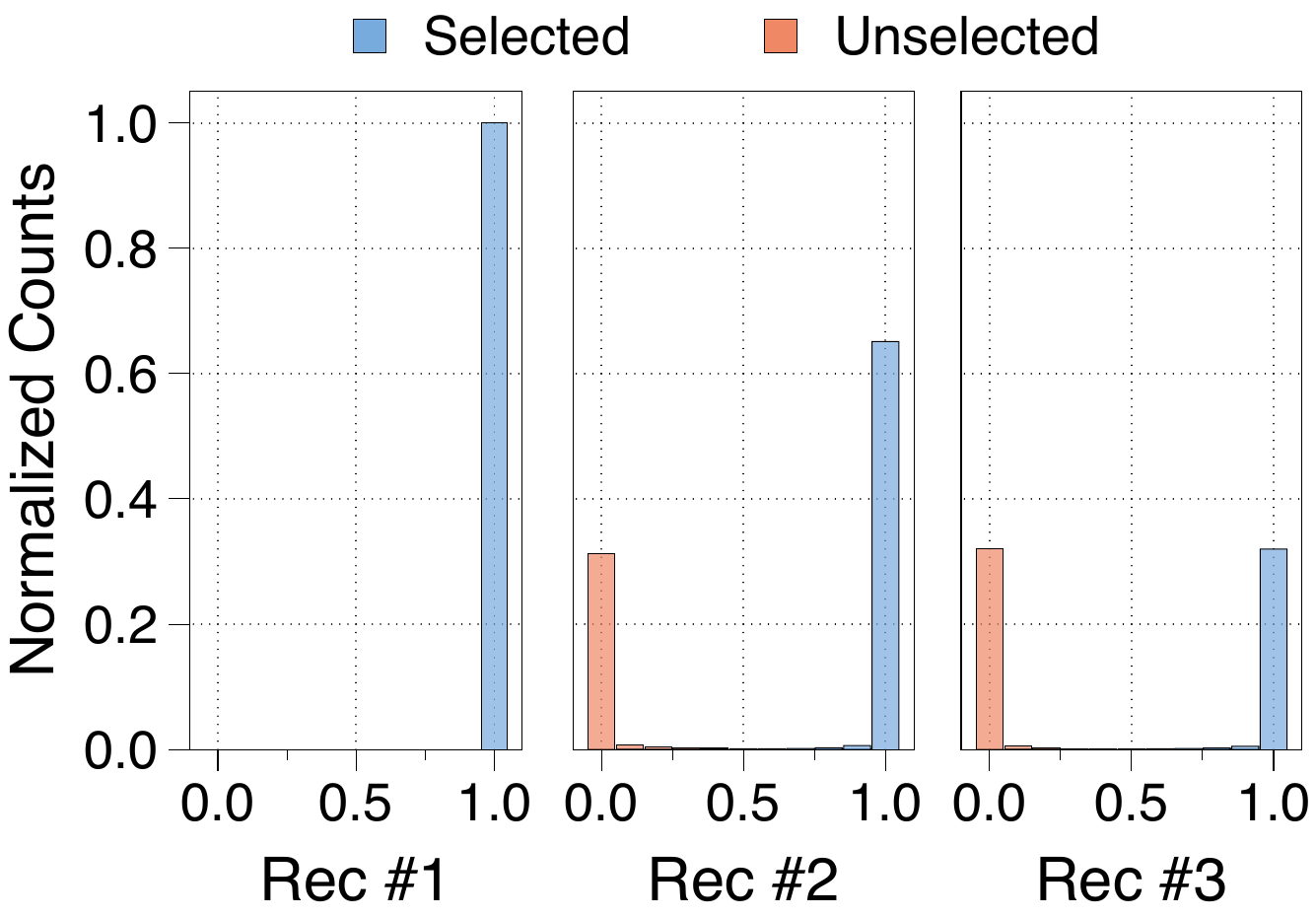}
        \subcaption{Expert-choice (Auxiliary loss)}
    \end{subfigure}
    \centering
    \hspace{25pt}
    \begin{subfigure}[t]{0.4\textwidth}
    \captionsetup{justification=centering}
        \includegraphics[width=\textwidth]{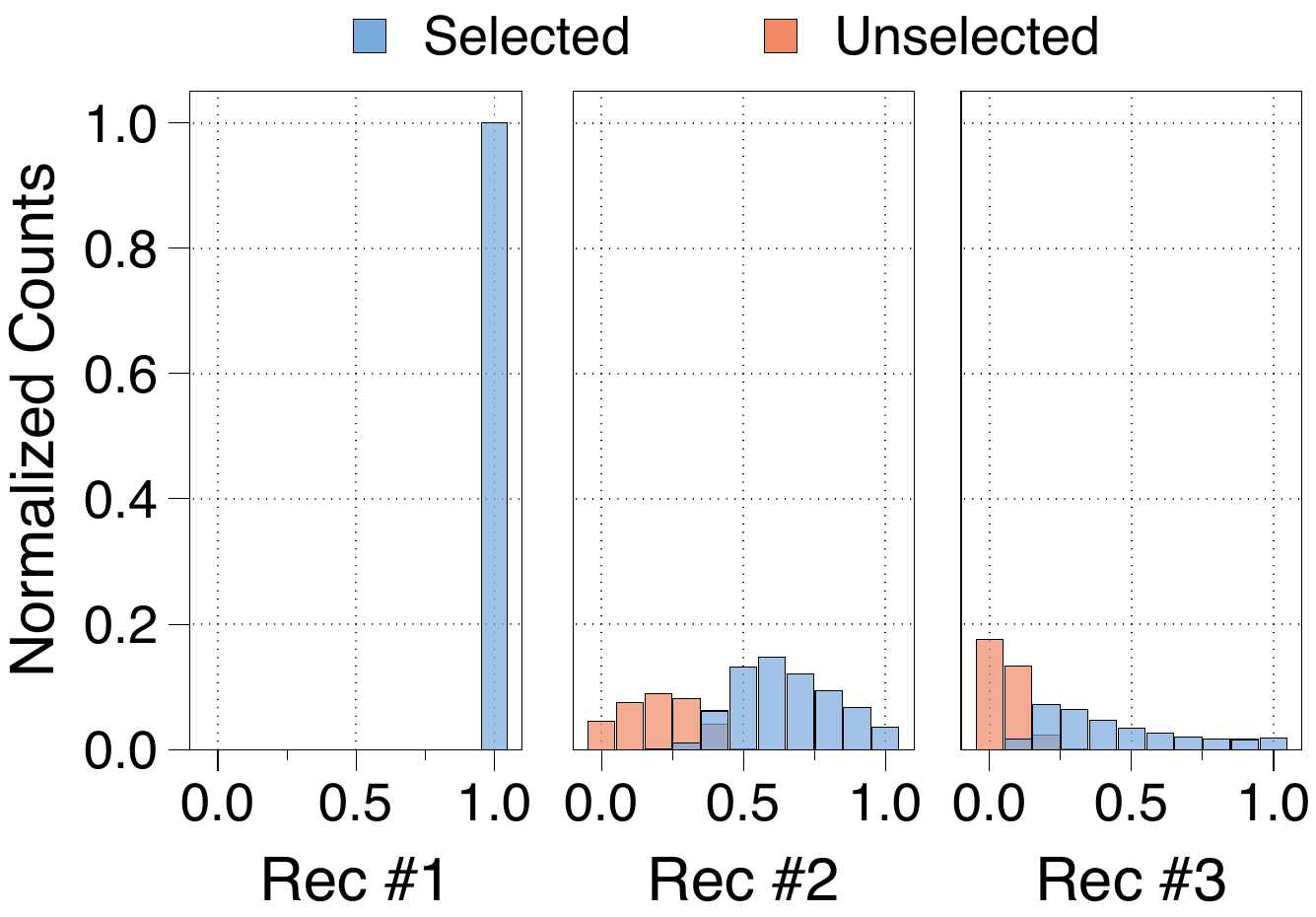}
        \subcaption{Expert-choice (Auxiliary router)}
    \end{subfigure}
    \centering
    \begin{subfigure}[t]{0.4\textwidth}
    \captionsetup{justification=centering}
        \includegraphics[width=\textwidth]{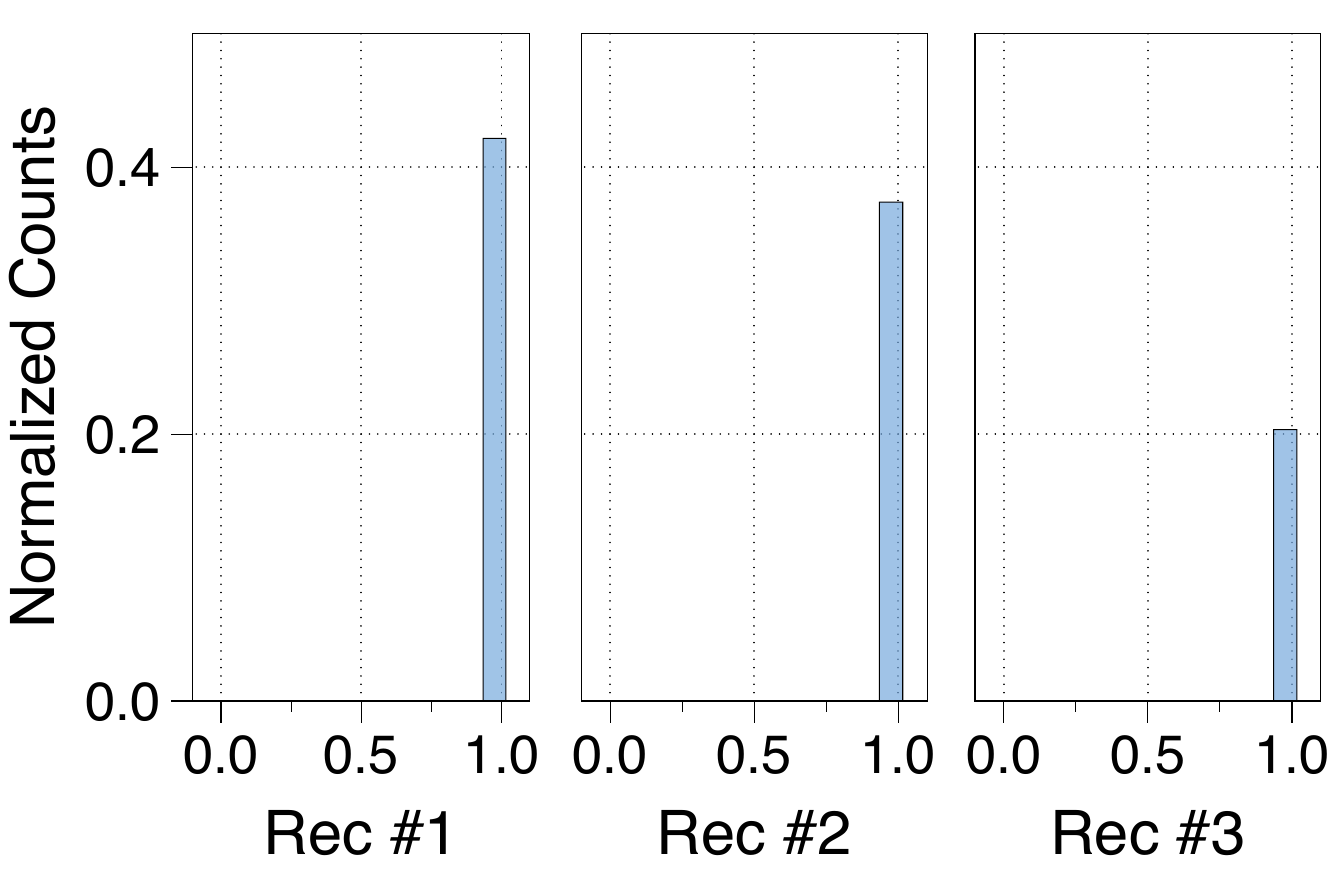}
        \subcaption{Token-choice (Balancing loss)}
    \end{subfigure}
    \centering
    \hspace{25pt}
    \begin{subfigure}[t]{0.4\textwidth}
    \captionsetup{justification=centering}
        \includegraphics[width=\textwidth]{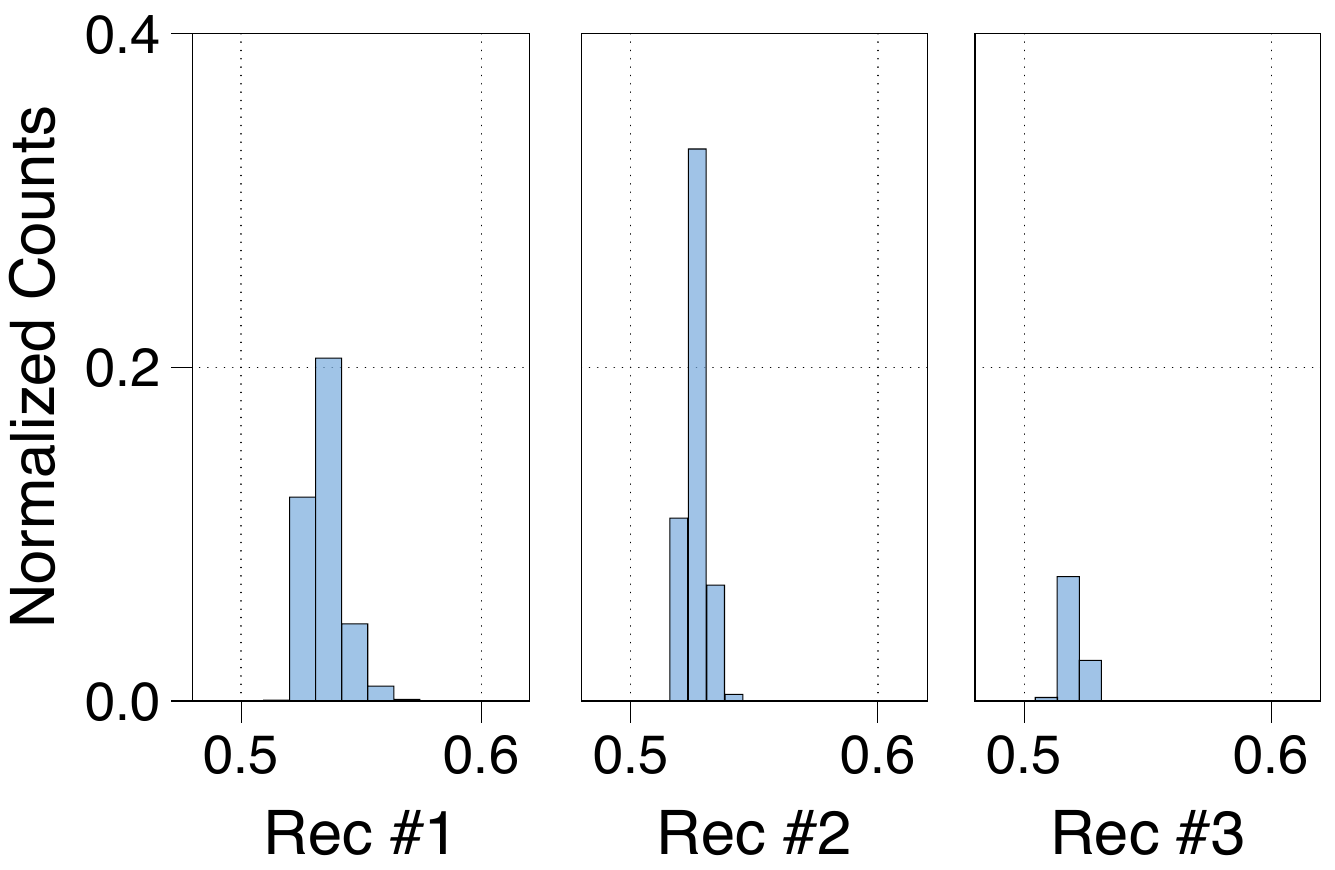}
        \subcaption{Token-choice (Loss-free)}
    \end{subfigure}
    \caption{
    Distribution of router weights for selected and unselected tokens at each recursion step in expert-choice and token-choice MoR~($N_r$=3). All models use the recursion-wise KV caching strategy. The subplots show results for (a) expert-choice routing with auxiliary loss, (b) auxiliary router, (c) token-choice routing with balancing loss, and (d) loss-free algorithm. Each subplot uses the best hyperparameter settings identified in Table\,\ref{tab_mor:ablation_study_router}.
    }    
    \label{fig_mor:weight_values_app}
\end{figure}

To gain insight into the router output distributions, we visualized the results in Figure\,\ref{fig_mor:weight_values_app}. Our analysis reveals that various routing mechanisms are optimized to balance expert loads according to the desired capacity. Notably, expert-choice routers achieved nearly perfect load balancing with the auxiliary loss, resulting in almost binary values (1 or 0) for selected and unselected tokens, respectively. 
For the auxiliary router, it was able to distinguish between selected and unselected tokens to some extent, but still showed overlapping. Since this strategy allows for different capacity factors during training and inference\footnote{The auxiliary router learns to capture intra-token differences within selected or unselected groups, rather than being biased towards extreme points.}, further research into methodologies that can more distinctly separate inter-cluster variations seems necessary.

Other token-choice strategies also exhibited good balancing properties with reasonable router values, which are used to refine the outputs of the corresponding recursion computation blocks. However, most cases failed to converge to optimal load balancing (i.e., these were edge cases where they achieved their own optimal load balancing), highlighting the challenges of achieving consistent performance in heterogeneous expert settings.

\chapter{Concluding Remark}

This dissertation addresses the critical inference inefficiency of LLMs stemming from their significant computational costs. While early-exiting frameworks offer a promising solution, their practical deployment is hindered by fundamental bottlenecks inherent to their dynamic nature—most notably, the problem of missing key-value caches and batch synchronization issues. To resolve these obstacles, this work proposes a series of methodologies that systematically co-design adaptive algorithms and model architectures, enabling the practical, high-throughput inference required for real-world applications.

In Chapter \ref{chapter:free}, we introduce the Fast and Robust Early-Exiting (FREE) framework to address critical limitations in conventional early-exiting. Specifically, we tackle the performance degradation from missing KV caches of exited tokens and the sensitivity to model sizes and sequence lengths. FREE incorporates a simplified shallow-deep module architecture and a novel synchronized parallel decoding mechanism. This parallel decoding strategy explicitly computes the KV caches even for the remaining layers of exited tokens, while maintaining significantly enhanced throughput. Concurrently, we introduce a Beta mixture model that adaptively estimates the optimal confidence threshold during decoding. Experimental results demonstrated that FREE significantly improves inference speed by up to 2.16$\times$ and enhances robustness across various generation tasks, providing a more practical foundation for adaptive computation.\looseness=-1

Chapter \ref{chapter:rrt} tackles the synchronization issue in batched inference, a critical bottleneck where exited tokens are forced to idle. This occurs because they cannot be batched with tokens at different model depths, severely hindering throughput. To resolve this, we introduce the Recursive Transformers, which leverage parameter sharing across layers. We maintain high performance within a significantly more compact model through smart initialization for looped layers and depth-wise LoRA modules to relax the strict parameter sharing constraints. Crucially, this architecture enables Continuous Depth-wise Batching, a new inference paradigm where tokens at different depths can be processed together in a single batch as they utilize a shared parameter block. This approach, especially when combined with early exiting, mitigates the synchronization bottleneck, thereby substantially boosts overall throughput.

Chapter \ref{chapter:mor} introduces Mixture-of-Recursions (MoR), a unified framework that addresses key challenges in batched inference and the missing KV cache problem. Unlike previous Recursive Transformers that require two sequential training procedures for parameter sharing and early-exiting—often leading to performance degradation—MoR uses a single, cohesive training process. It employs lightweight routers that dynamically learn to assign the optimal recursion depth for each token, allocating computation only where it is needed. MoR further enhances efficiency with a novel recursion-wise caching mechanism. This feature strategically resolves the missing KV cache issue by selectively storing entries only for tokens routed to a specific recursion step, simultaneously reducing KV memory sizes. Scaling law experiments confirm that MoR establishes a new Pareto frontier, achieving lower perplexity than both vanilla and recursive baselines with significantly higher throughput.

In summary, this dissertation has presented methodologies designed to overcome the fundamental bottlenecks that prevent the real-world application of early exiting in LLMs. We believe the pathways established by this research not only create more sustainable and efficient models today, but also serve as a foundational contribution to the critical challenge of ensuring AGI's long-term sustainability.

\bibliography{main}
\bibliographystyle{plain}

\curriculumvitae[4]

    \begin{personaldata}
        \name       {배 상 민}
        \dateofbirth{1996}{2}{23}
    \end{personaldata}

    \begin{education}
        \item[2011. 3.\ --\ 2014. 2.] 공주한일고등학교
        \item[2014. 3.\ --\ 2019. 2.] 한국과학기술원 산업및시스템공학과 학사과정
        \item[2019. 3.\ --\ 2021. 2.] 한국과학기술원 산업및시스템공학과 석사과정
        \item[2021. 3.\ --\ 2025. 8.] 한국과학기술원 김재철 AI대학원 박사과정
    \end{education}

    \begin{career}
        \item[2017. 12.\ --\ 2018. 6.] Research Internship at Human Factors and Ergonomics Lab, KAIST
        \item[2018. 7.\ --\ 2018. 8.] Research Internship at Optimization and Statistical Inference Lab, KAIST
        \item[2018. 9.\ --\ 2019. 2.] Research Internship at Kakao Recommendation Team
        \item[2024. 5.\ --\ 2024. 8.] Research Internship at Google DeepMind
        \item[2025. 5.\ --\ 2025. 9.] Research Internship at Meta FAIR
    \end{career}


    \begin{publication}
        \item Gihun Lee*, \textbf{\underline{Sangmin Bae}*}, Jaehoon Oh, $\text{Se-Young Yun}^\dagger$. \textit{``SIPA: A Simple Framework for Efficient Networks"}. \textbf{ICDM Workshop} on Big Data Analysis for Smart Energy. 2020.


        \item Sungnyun Kim*, Gihun Lee*, \textbf{\underline{Sangmin Bae}*}, $\text{Se-Young Yun}^\dagger$. \textit{``MixCo: Mix-up Contrastive Learning for Visual Representation"}. \textbf{NeurIPS Workshop} on Self-Supervised Learning: Theory and Practice. 2020.

        \item Gihun Lee*, Minchan Jeong*, Yongjin Shin, \textbf{\underline{Sangmin Bae}}, $\text{Se-Young Yun}^\dagger$. \textit{``Preservation of Global Knowledge by Not-True Distillation in Federated Learning"}. Thirty-Sixth Conference on Neural Information Processing Systems \textbf{(NeurIPS)}. 2022.
        
        \item \textbf{\underline{Sangmin Bae}}*, Sungnyun Kim*, Jongwoo Ko, Gihun Lee, Seungjong Noh, $\text{Se-Young Yun}^\dagger$, {\it ``Self-Contrastive Learning: Single-viewed Supervised Contrastive Framework using Sub-network"}, Thirty-Seventh AAAI Conference on Artificial Intelligence {\bf (AAAI)}. 2023.

        \item Sangmook Kim*, \textbf{\underline{Sangmin Bae}*}, $\text{Hwanjun Song}^\dagger$, $\text{Se-Young Yun}^\dagger$. \textit{``Re-thinking Federated Active Learning based on Inter-class Diversity"}. The IEEE/CVF Conference on Computer Vision and Pattern Recognition \textbf{(CVPR)}. 2023.

        \item Sungnyun Kim*, \textbf{\underline{Sangmin Bae}}*, $\text{Se-Young Yun}^\dagger$. {\it ``Coreset Sampling from Open-Set for Fine-Grained Self-Supervised Learning"}. The IEEE/CVF Conference on Computer Vision and Pattern Recognition \textbf{(CVPR)}. 2023.

        \item \textbf{\underline{Sangmin Bae}*}, June-Woo Kim*, Won-Yang Cho, Hyerim Baek, Soyoun Son, Byungjo Lee, Changwan Ha, Kyongpil Tae,  $\text{Sungnyun Kim}^\dagger$, $\text{Se-Young Yun}^\dagger$. {\it ``Patch-Mix Contrastive Learning with Audio Spectrogram Transformer on Respiratory Sound Classification"}. Conference of the International Speech Communication Association \textbf{(INTERSPEECH)}. 2023. 

        \item \textbf{\underline{Sangmin Bae}*}, Jongwoo Ko*, $\text{Hwanjun Song}^\dagger$, $\text{Se-Young Yun}^\dagger$. {\it ``Fast and Robust Early-Exiting Framework for Autoregressive Language Models with Synchronized Parallel Decoding"}. Conference on Empirical Methods in Natural Language Processing \textbf{(EMNLP)}. 2023. 

        \item Felix den Breejen, \textbf{\underline{Sangmin Bae}}, Stephen Cha, Tae-Young Kim, Seoung-Hyun Koh, $\text{Se-Young Yun}^\dagger$. {\it ``Fine-Tuning the Retrieval Mechanism for Tabular Deep Learning"}. \textbf{NeurIPS Workshop} on Table Representation Learning. 2023. 

        \item June-Woo Kim, Chihyeon Yoon, Miika Toikkanen, \textbf{\underline{Sangmin Bae}}, $\text{Ho-Young Jung}^\dagger$. {\it ``Adversarial Fine-tuning using Generated Respiratory Sound to Address Class Imbalance"}. \textbf{NeurIPS Workshop} on Deep Generative Models for Health. 2023.

        \item June-Woo Kim, \textbf{\underline{Sangmin Bae}}, Won-Yang Cho, Byungjo Lee, $\text{Ho-Young Jung}^\dagger$. {\it ``Stethoscope-guided Supervised Contrastive Learning for Cross-domain Adaptation on Respiratory Sound Classification"}. IEEE International Conference on Acoustics, Speech and Signal Processing \textbf{(ICASSP)}. 2024.

        \item Yujin Kim, Jaehong Yoon, Seonghyeon Ye, \textbf{\underline{Sangmin Bae}}, Namgyu Ho, $\text{Sung Ju Hwang}^\dagger$, $\text{Se-Young Yun}^\dagger$. {\it ``Carpe diem: On the Evaluation of World Knowledge in Lifelong Language Models"}. Conference of the North American Chapter of the Association for Computational Linguistics \textbf{(NAACL)}. 2024.

        \item June-Woo Kim, Miika Toikkanen, \textbf{\underline{Sangmin Bae}}, $\text{Minseok Kim}^\dagger$, $\text{Ho-Young Jung}^\dagger$. {\it ``RepAugment: Input-Agnostic Representation-Level Augmentation for Respiratory Sound Classification"}. International Conference of the IEEE Engineering in Medicine and Biology Society \textbf{(EMBC)}. 2024.

        \item Yunseon Choi, \textbf{\underline{Sangmin Bae}}, Seonghyun Ban, Minchan Jeong, Chuheng Zhang, Lei Song, Li Zhao, Jiang Bian, $\text{Kee-Eung Kim}^\dagger$. {\it ``Hard Prompts Made Interpretable: Sparse Entropy Regularization for Prompt Tuning with RL"}. The Association for Computational Linguistics \textbf{(ACL)}. 2024.

        \item Sungnyun Kim*, Kangwook Jang*, \textbf{\underline{Sangmin Bae}}, $\text{Hoirin Kim}^\dagger$, $\text{Se-Young Yun}^\dagger$. {\it ``Learning Video Temporal Dynamics with Asymmetric Cross-Modal Attention for Robust Audio-Visual Speech Recognition"}. IEEE Spoken Language Technology Workshop \textbf{(SLT)}. 2024.

        \item Namgyu Ho*, \textbf{\underline{Sangmin Bae}*}, Taehyeon Kim, Hyunjik Jo, Yireun Kim, Tal Schuster, Adam Fisch, $\text{James Thorne}^\dagger$, $\text{Se-Young Yun}^\dagger$. {\it ``Block Transformer: Global-to-Local Language Modeling for Fast Inference"}. Conference on Neural Information Processing Systems \textbf{(NeurIPS)}. 2024. 

        \item Felix den Greejen, \textbf{\underline{Sangmin Bae}}, Stephen Cha, $\text{Se-Young Yun}^\dagger$. {\it ``Fine-tuned In-Context Learning Transformers are Excellent Tabular Data Classifiers"}. Preprint. 2024.
        
        \item Yongjin Yang*, Sihyeon Kim*, Hojung Jung, \textbf{\underline{Sangmin Bae}}, SangMook Kim, $\text{Se-Young Yun}^\dagger$, $\text{Kimin Lee}^\dagger$. {\it ``Automated Filtering of Human Feedback Data for Aligning Text-to-Image Diffusion Models"}. International Conference on Learning Representations \textbf{(ICLR)}. 2025.
        
        \item Sungnyun Kim, Sungwoo Cho, \textbf{\underline{Sangmin Bae}}, Kangwook Jang, $\text{Se-Young Yun}^\dagger$. {\it ``Multi-Task Corrupted Prediction for Learning Robust Audio-Visual Speech Representation"}. International Conference on Learning Representations \textbf{(ICLR)}. 2025.
        
        \item \textbf{\underline{Sangmin Bae}}, Adam Fisch, Hrayr Harutyunyan, Ziwei Ji, Seungyeon Kim, $\text{Tal Schuster}^\dagger$. {\it ``Relaxed Recursive Transformers: Effective Parameter Sharing with Layer-wise LoRA".} International Conference on Learning Representations \textbf{(ICLR)}. 2025.

        \item Sungnyun Kim, Kangwook Jang, \textbf{\underline{Sangmin Bae}}, Sungwoo Cho, $\text{Se-Young Yun}^\dagger$. {\it ``MoHAVE: Mixture of Hierarchical Audio-Visual Experts for Robust Speech Recognition"}. International Conference on Machine Learning \textbf{(ICML)}. 2025.


        \item Sihyeon Kim*, Boryeong Cho*, \textbf{\underline{Sangmin Bae}}, $\text{Sumyeong Ahn}^\dagger$, $\text{Se-Young Yun}^\dagger$. {\it ``VSCoDe: Visual-Augmentation Selection for Contrastive Decoding"}. Transactions on Machine Learning Research \textbf{(TMLR)}. 2025.

        \item \textbf{\underline{Sangmin Bae}*}, Yujin Kim*, Reza Bayat*, Sungnyun Kim, Jiyoun Ha, Tal Schuster, Adam Fisch, Hrayr Harutyunyan, Ziwei Ji, $\text{Aaron Courville}^\dagger$, $\text{Se-Young Yun}^\dagger$. {\it ``Mixture-of-Recursions: Learning Dynamic Recursive Depths for Efficient Language Modeling"}. Conference on Neural Information Processing Systems \textbf{(NeurIPS)}. 2025.
        
    \end{publication}

  \label{paperlastpagelabel}     
\end{document}